\documentclass[lettersize,journal]{IEEEtran}

\usepackage{newtxtext}
\usepackage{graphicx}
\usepackage{multirow}
\usepackage[table]{xcolor}
\usepackage{textcomp}
\usepackage{manyfoot}
\usepackage{booktabs}
\usepackage{algorithm}
\usepackage{algorithmicx}
\usepackage{algpseudocode}
\usepackage{listings}
\usepackage{lmodern}
\usepackage{hyperref}
\usepackage{chngcntr}
\usepackage{silence}
\usepackage{wrapfig}
\usepackage{tcolorbox}
\tcbuselibrary{breakable,skins,minted}
\usepackage{minted}
\usepackage{amsmath,amsfonts}
\usepackage{array}
\usepackage[caption=false,font=normalsize,labelfont=sf,textfont=sf]{subfig}
\usepackage{stfloats}
\usepackage{url}
\usepackage{verbatim}
\usepackage{cite}

\setminted{autogobble,breaklines,fontsize=\footnotesize}

\newtcolorbox{promptbox}[2][]{%
  colback=gray!2,
  colframe=gray!55,
  boxrule=0.6pt,
  arc=2mm,
  left=1.2mm,right=1.2mm,top=1mm,bottom=1mm,
  breakable,
  title={\textbf{#2}},
  enhanced,
  #1
}

\begin{document}

\title{TacticGen: Grounding Adaptable and Scalable Generation of Football Tactics}

\author{Sheng Xu,
Guiliang Liu,
Tarak Kharrat,
Yudong Luo,
Mohamed Aloulou,\\
Javier L\'opez Pe\~na,
Konstantin Sofeikov,
Adam Reid,
Paul Roberts,
Steven Spencer,\\
\qquad Joe Carnall,
Ian McHale,
Oliver Schulte, 
Hongyuan Zha,
and Wei-Shi Zheng
\thanks{Corresponding author: Guiliang Liu. Email: liuguiliang@cuhk.edu.cn.}
\thanks{Sheng Xu, Guiliang Liu, Yudong Luo, and Hongyuan Zha are with the School of Data Science, The Chinese University of Hong Kong, Shenzhen, China.}
\thanks{Tarak Kharrat, Mohamed Aloulou, Javier L\'opez Pe\~na, Konstantin Sofeikov, Adam Reid, and Paul Roberts are with Real Analytics, London, UK.}
\thanks{Steven Spencer and Joe Carnall are with Birmingham City Football Club, Birmingham, UK.}
\thanks{Ian McHale is with the Management School, University of Liverpool, Liverpool, UK.}
\thanks{Oliver Schulte is with the School of Computing Science, Simon Fraser University, Vancouver, Canada.}
\thanks{Wei-Shi Zheng is with the School of Computer Science and Engineering, Sun Yat-sen University, Guangzhou, China.}
}



\maketitle

\begin{abstract}
Success in association football relies on both individual skill and coordinated tactics. While recent advancements in spatio-temporal data and deep learning have enabled predictive analyses like trajectory forecasting, the development of tactical design remains limited. Bridging this gap is essential, as prediction reveals what is likely to occur, whereas tactic generation determines what should occur to achieve strategic objectives. In this work, we present TacticGen, a generative model for adaptable and scalable tactic generation. 
TacticGen formulates tactics as sequences of multi-agent movements and interactions conditioned on the game context. It employs a multi-agent diffusion transformer with agent-wise self-attention and context-aware cross-attention to capture cooperative and competitive dynamics among players and the ball. 
Trained with over 3.3 million events and 100 million tracking frames from top-tier leagues, TacticGen achieves state-of-the-art precision in predicting player trajectories. Building on it, TacticGen enables adaptable tactic generation tailored to diverse inference-time objectives through classifier guidance mechanism, specified via rules, natural language, or neural models.
Its modeling performance is also inherently scalable. A case study with football experts confirms that TacticGen generates realistic, strategically valuable tactics, demonstrating its practical utility for tactical planning in professional football. The project page is available at: \url{https://shengxu.net/TacticGen/}.
\end{abstract}


\section{Introduction}

Association football, commonly known as soccer, is one of the world’s most popular sports~\cite{worldatlas2025}.
As a highly complex and strategic sport, the success of football depends not only on individual technical skill and physical ability, but also on the execution of sophisticated tactics that coordinate the movements of players across an expansive pitch under continuous play~\cite{sarmento2014match}. The combination of spatial scale, temporal continuity, and multi-agent interaction distinguishes football from sports such as basketball or ice hockey, and makes tactics the true currency of competitive advantage.

\begin{figure*}[htbp]
    \centering
    \includegraphics[width=0.9\textwidth]{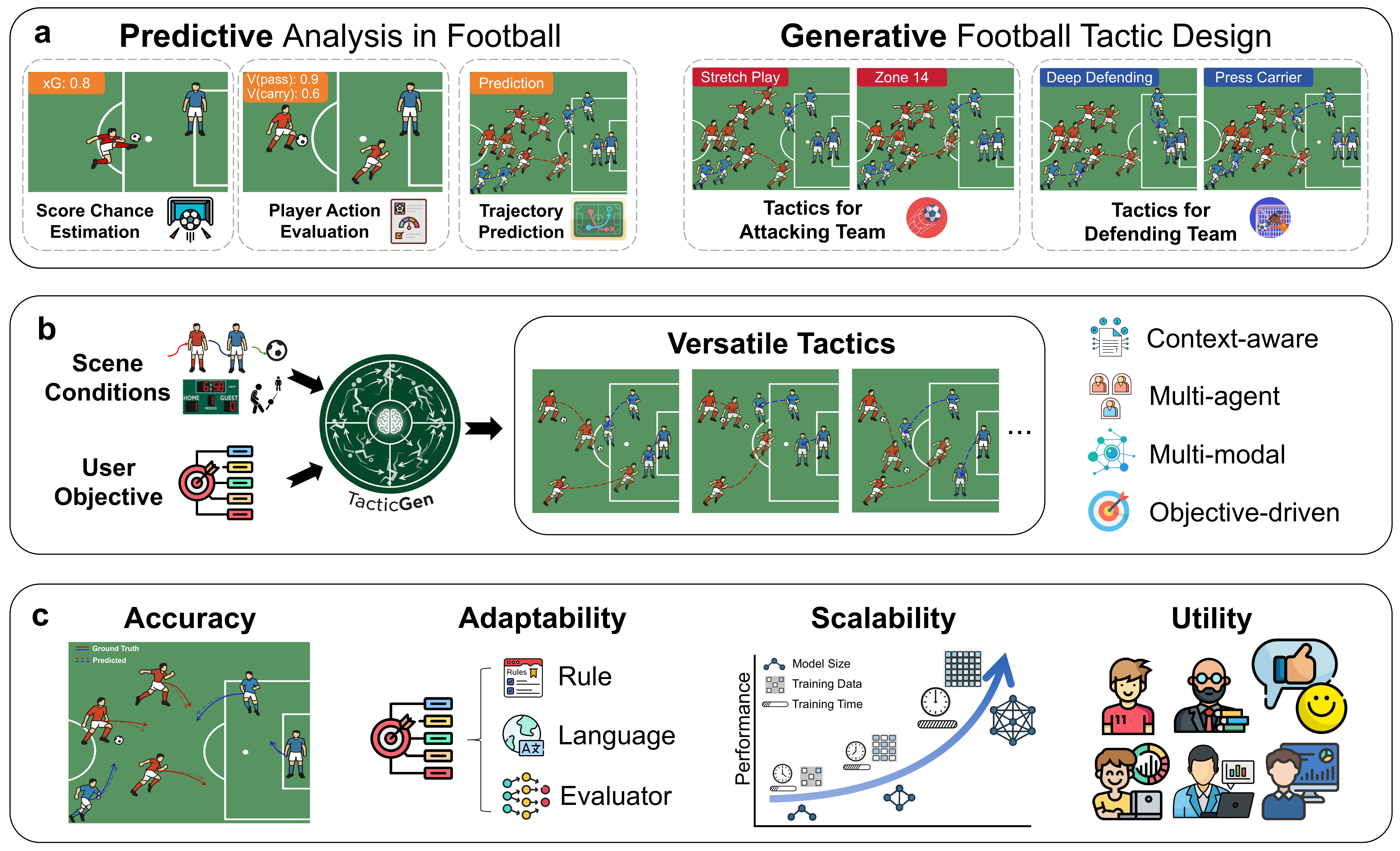}
    \caption{\textbf{Generating football tactics by modeling coordinated player movements.} \textbf{a} This paper aims to move the field of football analytics from predictive modeling to generative tactical design, requiring models that adapt to diverse objectives. \textbf{b} Conditioned on the football scene context and diverse user objectives, TacticGen generates versatile tactics represented through subsequent player movements, characterized as context-aware, multi-agent, multi-modal, and objective-driven. \textbf{c} TacticGen generates player movements with high accuracy that capture the underlying patterns of football play. Building on this foundation, it demonstrates adaptability by guiding trajectories toward diverse user-specified tactical objectives, scalability in tactic modeling consistent with scaling laws, and, most importantly, strong practical utility as validated by five football experts.}\label{fig:motivation}
\end{figure*}

Fueled by the growing availability of spatiotemporal and multimodal sports data~\cite{mendes2024towards, beal2019artificial}, AI-driven methods have reshaped football analytics and enabled advances in \textit{predictive analysis}, including scoring chance estimation~\cite{anzer2021goal,cefis2025accuracy}, player action evaluation~\cite{hassani2025dynamic,nakahara2023action,xarles2025action}, next-event prediction~\cite{alves2025score,hong2026modeling}, and motion trajectory prediction~\cite{omidshafiei2022multiagent,yuan2021agentformer,xu2025sports,yang2025smgdiff}. Although some of these methods employ generative architectures, they remain fundamentally predictive, addressing \textit{what might happen} rather than \textit{what should happen} to realize a tactical objective. From a technical perspective, prior approaches mainly model future trajectories under observed context or predefined conditioning signals, but lack a mechanism for flexibly steering generation toward diverse objectives at inference time. This gap is particularly critical in football, where the challenge is not merely to predict plausible future movements, but to generate coordinated player interactions that satisfy diverse tactical objectives.
Figure~\ref{fig:motivation}a depicts the shift from predictive analysis to \textit{generative tactical design}, which represents the central unsolved problem of football analytics. Solving this challenge would allow analysts and coaches to systematically simulate, evaluate, and design strategies before they unfold on the pitch.

Initial progress has been made in AI-driven tactical design. For example, TacticAI~\cite{wang2024tacticai} demonstrates the potential of generating coordinated player positioning in corner-kick scenarios under the objective of maximizing shot probability. However, it is evaluated in highly structured set-piece contexts, leaving its applicability to the continuous dynamics of open play unverified. More recently, TacEleven~\cite{zhao2025taceleven} and GenTac~\cite{rao2026gentac} have extended this line of research to open-play tactic generation. Nevertheless, both rely on curated text-to-trajectory datasets during training, where trajectory-relevant descriptions are provided as conditioning inputs for generation. As a result, at inference time, they remain tied to this finite set of predefined conditions, which may limit their adaptability to accommodate broader tactical objectives and diverse user-defined intentions. Moreover, their evaluations are conducted on relatively small datasets, providing little evidence of scalability to larger datasets. These limitations highlight the need for a general framework that can model complex multi-agent interactions, accommodate diverse objectives during inference time, and scale effectively with both data and model size.



In this paper, we formulate tactical design as \textit{the generation of coordinated player movement sequences conditioned on the game context and guided by diverse tactical objectives}. In our setting, {\em adaptability} denotes the capacity of a single pretrained model to produce tactics that align with diverse objectives during inference, such as enhancing pitch control, creating attacking overloads, and preserving defensive compactness~\cite{bommasani2021opportunities}. {\em Scalability} refers to consistent improvements as data and model capacity grow, following established scaling laws~\cite{bahri2024explaining}. In football terms, scalability ensures that as more match data becomes available, the model can generate more reliable tactics across teams, leagues, and playing styles. Together, these capabilities are essential for moving beyond scenario-specific models toward practical tools for tactical design and decision support in professional football.

To address this, we propose TacticGen, a framework for grounding adaptable and scalable generation of football tactics (see Figure~\ref{fig:motivation}b).
TacticGen leverages a multi-agent Diffusion Transformer backbone~\cite{peebles2023scalable}, complemented by a specially designed attention mechanism that captures the dynamics of both competitive and cooperative relationships among players, as well as the surrounding game context across a temporal sequence. Trained on more than 3.3 million annotated events and 100 million tracking frames from top-tier leagues, TacticGen generates player trajectories with high accuracy and realism, consistently outperforming state-of-the-art methods. Beyond prediction, it demonstrates strong adaptability during inference by guiding the generation toward diverse tactical goals using a differentiable classifier mechanism, whether through rule-based functions, natural language descriptions, or learned evaluators. Systematic experiments show that TacticGen scales well with model size, training duration, and data volume, adhering to established scaling laws, which highlights its potential as a high-capacity model for football tactics. Finally, case studies with experts from Birmingham City Football Club\footnote{\url{https://www.bcfc.com/}} demonstrate its practical utility: the generated tactics were often indistinguishable from real trajectories and were preferred by the experts over ground-truth plays in 80\% of scenarios due to their greater strategic effectiveness, underscoring its real-world impact.


As shown in Figure~\ref{fig:motivation}c, TacticGen unifies accuracy, adaptability, scalability, and utility, representing a foundational step toward generative tactical design in football. This framework has the potential to extend beyond football, transforming sports analytics from descriptive and predictive analysis to generative decision support, and opening new opportunities for preparation in professional sports.

\section{Related Works}
\subsection{Trajectory Prediction in Sports}
Most prior work on trajectory modeling in sports focuses on \emph{predictive analysis}, where models forecast future agent movements from observed context~\cite{zhao2025survey}. Broadly, these methods fall into two categories: deterministic architectures, such as LSTMs~\cite{hochreiter1997lstm} and transformers~\cite{vaswani2017attention}, with representative examples including~\cite{monti2021dag,capellera2024transportmer,zhang2025football,capellera2024footbots}; and multimodal formulations built on generative architectures, such as conditional VAEs~\cite{doersch2016tutorial} and diffusion models~\cite{ho2020denoising}, with representative examples including~\cite{omidshafiei2022multiagent,chen2024playbest,xu2025sports,capellera2025unified}. Generative models, in particular, typically achieve stronger performance by better capturing multi-modality and multi-agent coordination, leading to more accurate and diverse forecasts of plausible future motions~\cite{huang2022survey,zhao2025survey}. However, despite these architectural differences, their objective remains fundamentally \emph{predictive}: to estimate what is likely to happen next given observed history. As a result, they are well-suited for anticipation and analysis, but are not designed for controllable tactic generation or objective-driven strategic planning in football.

\subsection{Trajectory Generation in Sports}
In this paper, \emph{generation} refers not merely to sampling diverse plausible futures from observed history, but to synthesizing trajectories explicitly steered toward user-specified tactical objectives. Compared to prediction, \emph{trajectory generation} in sports remains relatively underexplored. TacticAI~\cite{wang2024tacticai} demonstrates this potential in corner-kick scenarios, but is restricted to structured settings and a single objective of maximizing shot probability. TacEleven~\cite{zhao2025taceleven} extends generative football analysis to open-play tactic discovery by autoregressively composing multi-step trajectories from successive single-step generations. While effective, this sequential generation paradigm can be computationally expensive and may compromise long-horizon coordination. GenTac~\cite{rao2026gentac} further adopts diffusion models with predefined tactical objectives, enabling one-stage generation of continuous trajectories. Yet, despite using richer conditions than TacticAI, both of them rely on curated conditional training, where generation is driven by conditioning signals specified in the dataset. Consequently, their apparent generative capability at inference time is largely realized by varying a finite set of predefined conditions, placing them closer to \emph{conditional prediction} than to truly flexible objective-guided generation. This reliance limits adaptability and prevents the seamless incorporation of user-specified objectives during inference, particularly open-ended goals expressed in natural language. In contrast, TacticGen adopts a classifier-guided diffusion framework that models continuous multi-agent trajectories within a single trained model. At inference time, it enables adaptable tactic generation under diverse user-specified objectives through classifiers constructed from rule-based functions, natural language inputs, or learned value models, rather than confining generation to a fixed set of predefined conditions.
\section{Methods}\label{sec:method}
In this section, we first present the problem formulation and essential background on diffusion models (Section~\ref{sec:problem-formulation}), followed by a detailed description of the proposed TacticGen model (Section~\ref{sec:tacticgen}). We then introduce three guidance mechanisms for guided trajectory generation (Section~\ref{sec:tacticgen-guided-generation}).

\subsection{Problem Formulation}\label{sec:problem-formulation}

\textbf{Definition of Tactic.} We define a \textit{tactic} as a coordinated sequence of player movements, along with interactions with the ball, executed by a team to achieve a strategic objective within a specific game context, such as creating scoring opportunities, maintaining possession, or enhancing pitch control.

\textbf{Diffusion Model.}
Diffusion models learn a data distribution by gradually perturbing clean samples with Gaussian noise and training a model to reverse this process~\cite{sohl2015deep,ho2020denoising}. In this work, we model the trajectory distribution $p(\tau)$, where $\tau=(s_t,s_{t+1},\ldots,s_{t+H-1})$ is a trajectory of length $H$ and $s_t\in\mathbb{R}^2$ denotes the 2D position at time $t$. We omit explicit time indices and use $\tau$ to denote the full trajectory for simplicity.

Given diffusion steps $K$ and variance schedule $\{\beta_k\}_{k=1}^K$, let $\alpha_k=1-\beta_k$ and $\bar{\alpha}_k=\prod_{i=1}^k \alpha_i$. The forward process is:
\begin{equation}
q(\tau_k \mid \tau_{k-1})
=
\mathcal{N}\!\left(\tau_k;\sqrt{\alpha_k}\,\tau_{k-1},\, (1-\alpha_k)\mathbf{I}\right),
\end{equation}
which implies the closed-form sampling equation:
\begin{equation}
\tau_k
=
\sqrt{\bar{\alpha}_k}\,\tau_0
+
\sqrt{1-\bar{\alpha}_k}\,\epsilon,
\quad
\epsilon\sim\mathcal{N}(0,\mathbf{I}).
\end{equation}

The reverse process is then modeled as:
\begin{equation}
p_\theta(\tau_{k-1}\mid \tau_k)
=
\mathcal{N}\!\left(\tau_{k-1};\,\mu_\theta(\tau_k,k),\,\sigma_k^2\mathbf{I}\right),
\end{equation}
where $\sigma_k^2$ is a fixed timestep-dependent variance. Following~\cite{ho2020denoising}, learning the reverse process is equivalent to learning a noise predictor $\epsilon_\theta(\tau_k,k)$, optimized with:
\begin{equation}
\label{eq:diff-loss}
\mathcal{L}(\theta)
=
\mathbb{E}_{k \sim \{1,\ldots,K\},\, \tau_0,\, \epsilon \sim \mathcal{N}(0,\mathbf{I})}
\left[\left\lVert \epsilon - \epsilon_\theta(\tau_k,k) \right\rVert^2\right].
\end{equation}

\textbf{Conditional Multi-agent Diffusion.} We extend the diffusion model to a conditional setting by introducing an auxiliary variable $C$, which may encode current states, historical context, or event-level information such as action type. The forward process remains $q(\boldsymbol{\tau}_k \mid \boldsymbol{\tau}_{k-1})$, while the reverse process becomes $p_\theta(\boldsymbol{\tau}_{k-1} \mid \boldsymbol{\tau}_k, C)$, modeling the conditional distribution of the clean trajectory given $C$.

We formulate football as a multi-agent trajectory generation task, where each of the $A$ agents corresponds to either one of the 22 players or the ball. Let $\boldsymbol{\tau} = (\tau^1,\ldots,\tau^A)$ denote the joint trajectory of all agents, where $\tau^i$ is the trajectory of agent $i$. The training objective is:
\begin{equation}
\label{eq:cond-madiff-loss}
    \mathcal{L}(\theta) 
    = \mathbb{E}_{k \sim \{1,\ldots,K\},\, \boldsymbol{\tau}_0,\, \epsilon \sim \mathcal{N}(0,\mathbf{I})}
    \big[ \lVert \epsilon - \epsilon_\theta(\boldsymbol{\tau}_k, k, C) \rVert^2 \big].
\end{equation}

Notably, in this paper, we use \textbf{bold} symbols to denote a matrix (e.g., $\boldsymbol{\tau}$ denotes a multi-agent trajectory matrix).

\begin{figure*}[htbp]
    \centering
    \includegraphics[width=0.9\textwidth]{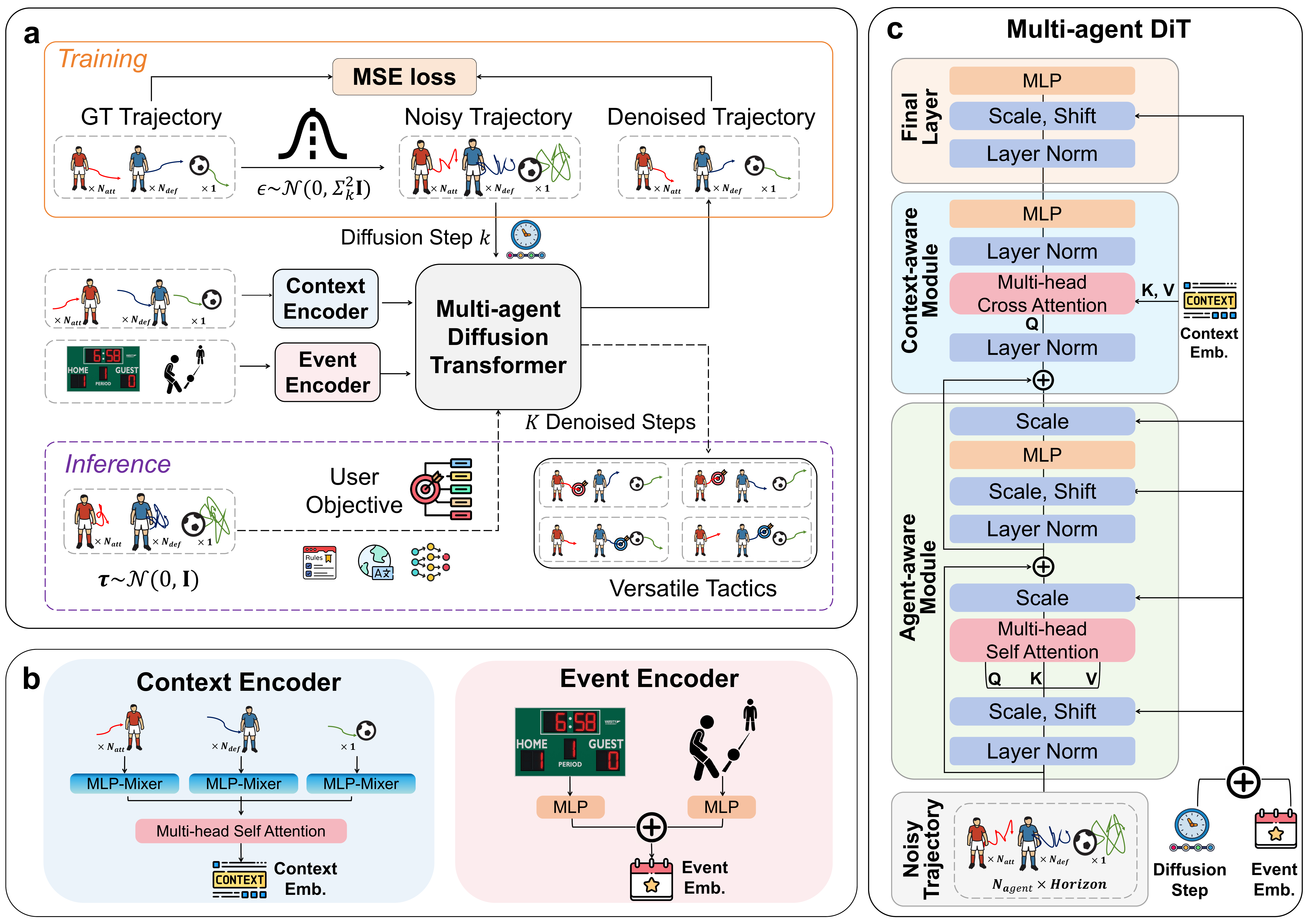}
    \caption{\textbf{The proposed TacticGen framework.} \textbf{a} Overview of the training and inference processes. During training, the ground-truth trajectories of $N_{\text{att}}$ attackers, $N_{\text{def}}$ defenders, and the ball are altered by adding noise, and TacticGen learns to recover the corresponding denoised trajectories, conditioned on the context (past trajectories), the event type (e.g., pass, block, clearance), and the diffusion step. During inference, trajectories are generated by denoising samples from Gaussian noise, where users can specify different objectives to guide TacticGen in producing versatile tactics aligned with their goals. \textbf{b} Architectures of the context and event encoders. The context encoder processes player and ball trajectories through MLP-Mixers, followed by self-attention to fuse the representations. The event encoder extracts global features via MLPs and concatenates them into a unified embedding. \textbf{c} Architecture of the multi-agent diffusion transformer backbone. It employs self-attention on different agents to encode noisy trajectories and cross-attention with the context to enable context-aware modeling. Event and time embeddings are concatenated and injected into the diffusion process.}\label{fig:TacticGen-framework}
\end{figure*}

\textbf{Classifier Guidance.} Classifier guidance~\cite{dhariwal2021diffusion} steers diffusion sampling toward trajectories that satisfy a desired criterion. It introduces an auxiliary classifier $p(y \mid \boldsymbol{\tau}_k, k, C)$, which estimates whether the noisy trajectory $\boldsymbol{\tau}_k$ at diffusion step $k$, under condition $C$, satisfies guidance signal $y$. The reverse process is then modified as:
\begin{align}
\label{eq:guide-diffusion}
\tilde{\mu}_\theta(\boldsymbol\tau_k, k, C, y)
&\approx \mu_\theta(\boldsymbol\tau_k, k, C) \nonumber\\
&\quad + \omega \,\Sigma_k \,\nabla_{\boldsymbol{\tau}_k} \log p(y \mid \boldsymbol\tau_k, k, C),
\end{align}
where $\omega \ge 0$ is the guidance scale and $\Sigma_k$ is the covariance at step $k$. The gradient term biases sampling toward trajectories favored by the classifier, enabling generation aligned with user-specified semantic or task-level objectives.

Importantly, classifier guidance is applied only at sampling time. Thus, once the diffusion model $\mu_\theta$ is trained, different objectives can be incorporated through the classifier without retraining the generative model.

\subsection{TacticGen Modules}\label{sec:tacticgen}
Building on the diffusion transformer (DiT)~\cite{peebles2023scalable}, we propose TacticGen, which extends the DiT backbone to a multi-agent DiT (MADiT), and introduces three key designs tailored for football: (i) a multi-agent self-attention mechanism to capture agent-wise interactions, (ii) a context encoder with self-attention to embed contextual information, and (iii) a multi-agent cross-attention mechanism between the context encoding and the generated trajectories. Figure~\ref{fig:TacticGen-framework} presents the overall framework of TacticGen, with each component described in detail as follows.

\subsubsection{Multi-Agent Diffusion Transformer Backbone}
DiT is originally designed for single-agent sequential modeling, where attention is applied along the temporal dimension of a sequence. Specifically, given the noised trajectories of all the agents (i.e., $\boldsymbol{\tau}_k \in \mathbb{R}^{A \times H \times D}$, where $A$ denotes the number of agents, $H$ the prediction horizon, and $D$ the x-y coordinates), DiT reshapes the input into $\mathbb{R}^{H \times (A \times D)}$, treating the horizon $H$ as the attention axis.

In the multi-agent setting, we argue that capturing the spatial dependencies among \textit{agents} is more critical than modeling the \textit{temporal} structure of their individual trajectories~\cite{zheng2025diffusion}. Thus, we rearrange the input as $\boldsymbol{\tau}_k \in \mathbb{R}^{A \times (H \times D)}$ and apply attention across the agent dimension. This design is driven by two insights. First, interdependence among agents shapes collective outcomes in multi-agent systems (e.g., players must coordinate with teammates). Second, the large football field means an agent’s position typically follows simple dynamics, making capturing temporal structure less crucial.

Consequently, MADiT employs a multilayer perceptron (MLP) to encode temporal features, while reserving transformer layers for modeling agent-level interactions. This yields the initial embedding of the noised multi-agent trajectories:
\begin{equation}
\label{eq:extract-emb}
e(\boldsymbol{\tau}_k) = \mathrm{MLP}(\boldsymbol{\tau}_k), \boldsymbol{\tau}_k \in \mathbb{R}^{A \times (H \times D)}
\end{equation}
where $e(\boldsymbol{\tau}_k)\in \mathbb{R}^{A \times D_e}$ denotes the trajectory embedding and $D_e$ the embedding dimension.

To further capture inter-agent relationships, each agent $i$ is assigned a role label $\ell^i \in \{a, d, b\}$, distinguishing between attacking players ($a$), defending players ($d$), and the ball ($b$). Collectively, these labels form an identity matrix $\ell(\boldsymbol{\tau}_k)$. A learnable embedding layer $\mathrm{Emb}(\cdot)$ then maps $\ell(\boldsymbol{\tau}_k)$ into a role embedding $r(\boldsymbol{\tau}_k)$:
\begin{equation}
r(\boldsymbol{\tau}_k) = \mathrm{Emb}(\ell(\boldsymbol{\tau}_k) )\in \mathbb{R}^{A \times D_e},
\end{equation}

The final agent embedding is then obtained by incorporating both trajectory and role information:
\begin{equation}
e(\boldsymbol{\tau}_k) \leftarrow e(\boldsymbol{\tau}_k) + r(\boldsymbol{\tau}_k).
\end{equation}

After obtaining the agent embedding, we then apply multi-head dot-product attention \cite{vaswani2017attention} at the agent level. For each agent $i \in [A]$ at diffusion step $k$, its hidden representation $e(\tau_k^i)$ is projected into query, key, and value vectors using head-specific learnable parameters $W_Q^h$, $W_K^h$, and $W_V^h$:
\begin{equation}
Q_i^h = W_Q^h e(\tau_k^i),  K_i^h = W_K^h e(\tau_k^i),  V_i^h = W_V^h e(\tau_k^i).
\end{equation}

The attention weight between agent $i$ and agent $j$ under the $h$-th head is then computed as:
\begin{equation}
\mu_{ij}^h = \mathrm{softmax}_j \!\left(\frac{Q_i^h (K_j^h)^\top}{\sqrt{d_k}} \right),
\label{eq:self-atten1}
\end{equation}
where $d_k$ is the key dimension, used as a scaling factor to stabilize training~\cite{vaswani2017attention}.
Finally, the attended representation of agent $i$ is obtained as:
\begin{equation}
\bar{e}(\tau^i_k) = W_O \Bigg(\mathrm{concat}_{h=1}^H \sum_{j=1}^{A} \mu_{ij}^h V_j^h \Bigg),
\label{eq:self-atten2}
\end{equation}
where $W_O$ maps the concatenated outputs of all heads back to the hidden dimension. Concatenating all agents yields the final embedding $\bar{e}(\boldsymbol{\tau}_k)$.

The architecture of TacticGen is shown in Figure~\ref{fig:TacticGen-framework}c. This design enables MADiT to efficiently capture inter-agent dependencies while preserving role information, yielding a more faithful representation of the multi-agent dynamics underlying complex football tactics.

\subsubsection{Context Encoding and Information Fusion}\label{sec:context-encoder}
Our task aims to generate future steps of trajectories based on a given observed trajectories, which we refer to as the context. As part of the condition $C$, the context is defined as $c(\boldsymbol{\tau}) = \boldsymbol{\tau}_{[H_c]}$, where the subscript $[H_c]$ denotes the first $H_c$ steps of $\boldsymbol{\tau}$ with $H_c < H$. In this work, we set the context length to $H_c=10$. 

In football, the ball’s movement is a decisive factor in decision-making, as both teams continuously track it to create scoring opportunities or to prevent attacks. To this end, this paper considers two ball-related context settings:
(i) Predictive ball modeling. In the common case where ball positions are available only within the context window ($H_c = 10$), TacticGen exploits this partial information to jointly generate the future trajectories of both players and the ball.
(ii) Conditional ball modeling. When the full ball trajectory across the prediction horizon ($H = 64$) is available or predefined (e.g., by a coach), TacticGen treats the ball as a fixed reference and focuses solely on generating coherent player trajectories. Formally, let the trajectories of the attacking players be denoted by $\boldsymbol{\tau}^a = \{\tau^{a_1}, \ldots, \tau^{a_{11}}\}$, those of the defending players by $\boldsymbol{\tau}^d = \{\tau^{d_1}, \ldots, \tau^{d_{11}}\}$, and the trajectory of the ball by $\tau^b$. In the context ball input setting, the context consists of the first $H_c$ steps of both players and the ball, i.e., $c(\boldsymbol{\tau})=\{(\boldsymbol{\tau}^a,\boldsymbol{\tau}^d,\tau^b)_{[H_c]}\}$. In the second scenario, the context consists of the first $H_c$ steps of all players and the complete $H$ steps of the ball trajectory, i.e., $c(\boldsymbol{\tau})=\{(\boldsymbol{\tau}^a, \boldsymbol{\tau}^d)_{[H_c]},\tau^b\}$. These two setups enhance the model’s flexibility and are well aligned with practical use cases in sports~\cite{capellera2024footbots,omidshafiei2022multiagent}.

Traditional diffusion-based approaches treat $c(\boldsymbol{\tau})$ as a fixed constraint~\cite{janner2022planning,zheng2025diffusion}. During denoising, $c(\boldsymbol{\tau})$ stays fixed, with generation applied only to the remaining parts. However, these hard-coded constraints can cause discontinuities (e.g., sudden jerks) between fixed and denoised segments, as the denoising process lacks context awareness. In multi-agent settings, it’s crucial to model agent interactions within the context, ensuring temporally coherent trajectories while preserving relational structures. To address this problem, TacticGen introduces a context encoder that leverages self-attention to capture contextual information while preserving agent-wise interactions, and integrates the resulting embeddings into the MADiT backbone for context-aware trajectory generation.

\textbf{Context Encoding.} For each agent $i$, the context is represented as $c(\tau^i) \in \mathbb{R}^{H_{c^i} \times D}$, where $H_{c^i}$ denotes the context horizon and $D$ the feature dimension (i.e., x-y coordinates). Following~\cite{zheng2025diffusion}, we adopt the MLP-Mixer architecture~\cite{tolstikhin2021mlp} to extract representations for agents. This is achieved by iteratively passing $c(\tau^i)$ through mixing layers that operate on both the temporal (horizon) and feature dimensions~\cite{zheng2025diffusion}.

After the mixing, we apply pooling along the horizon dimension and concatenate all agents’ representations to form a matrix $c(\boldsymbol{\tau}) \in \mathbb{R}^{A \times D_e}$, where $A$ is the number of agents and $D_e$ is the embedding dimension (consistent with Eq. (\ref{eq:extract-emb})). Finally, multi-head self-attention (Eq. (\ref{eq:self-atten2})) is applied for the context embedding to obtain the final context encoding $\bar{c}(\boldsymbol{\tau})$.

This process enables the context encoder to distill temporal information into compact agent-level embeddings, while attention further captures dependencies across agents. As a result, $\bar{c}(\boldsymbol{\tau})$ provides a richer representation that facilitates more coordinated and context-aware trajectory generation.

\textbf{Context Fusion.} We integrate the extracted context information $\bar{c}(\boldsymbol{\tau})$ into MADiT by introducing a cross-attention mechanism. Recall that at diffusion step $k$, MADiT produces an embedding $\bar{e}(\boldsymbol{\tau}_k)$. To enable context-aware denoising, we compute cross-attention between $\bar{e}(\boldsymbol{\tau}_k)$ and $\bar{c}(\boldsymbol{\tau})$. Unlike self-attention, where queries, keys, and values are derived from the same input, cross-attention uses $\bar{e}(\boldsymbol{\tau}_k)$ as the query and $\bar{c}(\boldsymbol{\tau})$ as the source of keys and values. This design allows $\bar{e}(\boldsymbol{\tau}_k)$ to retrieve the most relevant contextual signals through query-key similarity. The fused output provides a refined representation for the noised trajectory $\boldsymbol{\tau}_k$, denoted as $e^*(\boldsymbol{\tau}_k)$.

\textbf{Event Information Fusion.} In addition to context trajectories, the condition $C$ also includes event-related information, such as the action type (e.g., pass, block, clearance), global features (e.g., goal difference, event outcome), and the event timestamp. These features provide crucial contextual signals for trajectory generation and should therefore be explicitly modeled. Specifically, the action type is encoded using an embedding layer followed by an MLP, while the global features and event timestamp are concatenated and processed through another MLP. The resulting representations are then concatenated to form the event-level embedding. This event-level embedding is further integrated with the diffusion timestep embedding derived from $k$, and the fused representation is applied through an adaptive layer normalization block~\cite{peebles2023scalable} to condition the trajectory generation process. The architectures of the context and event encoders are shown in Figure~\ref{fig:TacticGen-framework}b, with their integration into the backbone in Figure~\ref{fig:TacticGen-framework}c.

By integrating all aforementioned components and trainable parameters into a unified function $\epsilon_\theta(\boldsymbol{\tau}_k, k, C)$, TacticGen is trained to predict trajectories conditioned on $C$, following the conditional diffusion objective in Eq.~(\ref{eq:cond-madiff-loss}). The overall training and inference paradigms of TacticGen are shown in Figure~\ref{fig:TacticGen-framework}a.

\subsection{Guided Generation through Classifier Guidance}\label{sec:tacticgen-guided-generation}
Effective tactic generation demands the flexibility to align with user-specified objectives. Thus, we move beyond passive \textit{prediction} to actively \textit{guide the generation} in pursuit of desired tactical outcomes.
In contrast to prior conditional-generation approaches~\cite{wang2024tacticai, ho2022classifier}, which typically necessitate training distinct models for different tactical intents, TacticGen employs a classifier-guidance mechanism~\cite{dhariwal2021diffusion, janner2022planning} that can adaptively incorporate diverse guidance objectives into a single diffusion-based trajectory generator. Specifically, this approach allows us to train a single, high-capacity prediction model trained by Eq.~(\ref{eq:cond-madiff-loss}), which can then be flexibly guided to generate diverse tactics during inference Eq.~(\ref{eq:guide-diffusion}), without the need for retraining. The guidance objectives can be specified through pre-defined rules (Section \ref{sec:method-rule-function}), natural language descriptions (Section \ref{sec:method-llm-function}), or learned value models (Section \ref{sec:method-value-function}).

\subsubsection{Rule-based Function as Classifier Guidance}\label{sec:method-rule-function}
Due to their close connection with energy-based models~\cite{lu2023contrastive}, diffusion models naturally support the integration of user-defined objectives formulated as differentiable rule-based functions through classifier guidance~\cite{zheng2025diffusion}. During sampling, these functions act as classifiers $p(y \mid \boldsymbol{\tau}_k, k, C)$, providing gradient signals that steer trajectory generation toward tactically consistent behaviors via Eq.~(\ref{eq:guide-diffusion}).

A key challenge, however, is that most rule-based functions are defined on clean trajectories $\boldsymbol{\tau}$ rather than their noisy counterparts $\boldsymbol{\tau}_k$, making direct evaluation during the diffusion process intractable. To address this issue, we follow~\cite{zheng2025diffusion} and adopt diffusion posterior sampling (DPS)~\cite{chung2022diffusion, xu2025rethinking} as a training-free solution. DPS first leverages the trained diffusion model $\mu_\theta$ to approximate the clean trajectory $\boldsymbol{\tau}$ at each step given the noisy trajectory $\boldsymbol{\tau}_k$, which is then passed through differentiable functions to obtain numerical values. Since both the functions and the neural network $\mu_\theta$ are differentiable, gradients can be propagated through the computed values and subsequently used to facilitate guided generation. Consequently, users can predefine rule-based functions grounded in domain knowledge to enable flexible and adaptable generation of versatile football tactics, as long as these functions are differentiable. Notably, these functions can be composed with adjustable weights to form multi-factor objectives, which allows TacticGen to balance different tactical considerations.

We outline several rule-based guidance functions applicable for football tactic generation in Appendix~\ref{sec:rule-example}, and provide some code snippets in Appendix~\ref{sec:llm-func-detail} (see the code examples in the LLM prompt).

\subsubsection{Generating Guidance Function with LLM}\label{sec:method-llm-function} Recent advancements in Large Language Models (LLMs) have demonstrated their strong ability to understand instructions, reason over structured domains, and generate executable code across a variety of applications~\cite{achiam2023gpt,jaech2024openai}. Their capacity to translate natural language specifications into structured functions makes them particularly well-suited for automatically generating differentiable guidance terms for trajectory generation.

To enable natural language-driven guidance, we design an automated pipeline that queries OpenAI’s GPT-5~\cite{singh2025openai} with structured prompts specifying the tactical objective, background information, and pitch geometry, while enforcing strict output constraints that require executable code. The input-output interface is standardized such that the functions accept trajectories as inputs and return a scalar guidance score. This pipeline ensures that LLM-generated functions can be seamlessly integrated into the trajectory generation process, allowing natural language descriptions of tactical intent to be reliably translated into differentiable, programmatic guidance functions. This feature is particularly valuable for fans or users without professional football expertise or programming skills, as it enables them to specify high-level objectives easily. More details about the prompts can be found in Appendix~\ref{sec:llm-func-detail}.

\subsubsection{Value-based Function as Classifier Guidance}\label{sec:method-value-function}
While combining the diffusion posterior process~\cite{chung2022diffusion,xu2025rethinking} with differentiable guidance functions, either pre-defined or generated by LLMs, offers computational efficiency, it depends on explicitly specified rules that may fail to generalize across diverse scenarios. To address this limitation, we propose an alternative approach that trains an auxiliary classifier by using the reward signals as defined in reinforcement learning~\cite{sutton1998reinforcement,janner2022planning}. In our setting, the reward $r(\boldsymbol{\tau})$ quantifies the outcome of an event associated with the trajectory $\boldsymbol{\tau}$. Positive rewards are assigned for advantageous outcomes induced by the trajectory, such as scoring goals, creating high-quality chances, or earning penalties, while negative rewards are given for corresponding outcomes that benefit the opposing team. More details of reward signals can be found in Appendix~\ref{sec:data-reward-detail}.

Reward-guided diffusion has been widely applied in planning to generate higher-reward trajectories during sampling~\cite{zhu2023diffusion}. 
Let $y$ be a binary random variable denoting the optimality of a trajectory $\boldsymbol{\tau}$, with 
$p(y=1)=\exp(r(\boldsymbol{\tau}))$, a common approach is to train a reward predictor $r_\phi(\boldsymbol{\tau}_k,k,C)$ that estimates the reward of a noisy trajectory $\boldsymbol{\tau}_k$, with $\phi$ the model parameters. 
During sampling, its gradient can be used for classifier guidance (Eq.~(\ref{eq:guide-diffusion}))~\cite{janner2022planning} through:
\begin{equation}\label{eq:reward-guide}
\nabla_{\boldsymbol{\tau}_k}\log p_\phi(y\mid \boldsymbol{\tau}_k,k,C) =\nabla_{\boldsymbol{\tau}_k} r_\phi(\boldsymbol{\tau}_k,k,C),
\end{equation}  
where $r_\phi(\boldsymbol{\tau}_k,k,C)$ serves as a surrogate energy function that promotes the trajectory towards higher predicted reward.

A key limitation of Eq.(~\ref{eq:reward-guide}) is that it only evaluates the immediate reward of the current trajectory, without considering future returns. In football, coaches focus on how a present action contributes to future scoring opportunities or goals. Therefore, instead of relying on the immediate reward $r(\boldsymbol{\tau})$, it is more effective to guide trajectory generation using the future return $R(\boldsymbol{\tau})$, which represents the discounted cumulative rewards from the current trajectory until the episode ends (e.g., when a goal is scored; see Appendix~\ref{sec:data-detail} for details). This approach aligns decisions with long-term strategic objectives for more practical tactic planning.

To this end, we employ the value model $V$ from reinforcement learning~\cite{sutton1998reinforcement,janner2022planning}, which estimates the expected discounted return as $V_\phi(\boldsymbol{\tau}) = \mathbb{E}[R(\boldsymbol{\tau})]$. Specifically, for each trajectory, we compute the future return $R(\boldsymbol{\tau})$ from its reward label and episode information. The value model $V_\phi$ is then trained via the Monte Carlo method~\cite{sutton1998reinforcement,arulkumaran2017deep} to estimate $R(\boldsymbol{\tau})$, chosen for its simplicity and its ability to yield unbiased estimates from precomputed returns. Since the rewards of the attacking and defending teams are strictly complementary, we train $V_\phi$ using only the attacking team’s rewards and define the defending value as its negative. The gradient of $V_\phi$ is then used to guide the trajectory generation toward outcomes with higher expected returns. The value model $V_\phi$ can be trained using nearly the same architecture as the diffusion backbone, with the only difference being that its output is a scalar return rather than a denoised trajectory.
\section{Empirical Evaluations}
We first provide the necessary background in Section~\ref{sec:background} to clarify the model and experimental setup. To thoroughly evaluate TacticGen’s capacity for generating adaptable and scalable football tactics, we organize our experiments around the following key questions:

1) Can TacticGen generate accurate and realistic multi-player motion trajectories? (Section~\ref{sec:exp-prediction})

2) Can TacticGen generate adaptable tactics for diverse objectives? (Section~\ref{sec:exp-guidance})

3) Does TacticGen adhere to scaling laws and exhibit scalable generalization? (Section~\ref{sec:scaling-law})

4) Does TacticGen provide practical utility in real-world applications? (Section~\ref{sec:case-study})

\subsection{Experimental Setup}\label{sec:background}
\textbf{Dataset.} The football dataset integrates event data and tracking data. Event data provide time-stamped annotations of in-game actions (e.g., shots, passes, tackles) together with event-level features such as goal difference and outcome. Tracking data captures the positions of all players and the ball. Positional coordinates are flipped so that the attacking team always scores on the right target. Event-tracking alignment is performed using the Needleman-Wunsch algorithm~\cite{marekblog2020}. The resulting dataset comprises 1,432 matches from top-tier leagues spanning the 2018-2025 seasons, including over 3.3 million events and nearly 100 million frames. The dataset is randomly shuffled and split into training (80\%) and test (20\%) sets, with consistent partitioning across all experiments to ensure no test events are included in the training set. More details are provided in Appendix~\ref{sec:dataset}.

\textbf{Model Setup.} TacticGen conditions on the past 10 frames (1 second) of agent trajectories, enriched with event-level information (e.g., action type, goal difference), to generate the next 54 frames (5.4 seconds) of coordinated agent movements. We consider two ball-related settings: (i) \textbf{Predictive Ball Modeling}: When ball positions are observed only within the context window (10 frames), TacticGen uses this context to generate future player and ball trajectories, referred to as TacticGen-Predictive (\textbf{TacticGen-P}). (ii) \textbf{Conditional Ball Modeling}: When the full ball trajectory over the prediction horizon (64 frames) is available or predefined, TacticGen conditions on the ball's movement to generate player trajectories, referred to as TacticGen-Conditional (\textbf{TacticGen-C}). The first setting is standard for trajectory prediction tasks, while the second is relevant to real-world football, where coaches anticipate ball progression and coordinate player positioning. The two models differ in the context encoder length for ball encoding, with other components remaining the same.

\textbf{Evaluation metrics.} For prediction tasks, we evaluate model performance using three primary metrics: Average Displacement Error (ADE), measuring mean positional error over time; Final Displacement Error (FDE), assessing error at the final prediction step; and Miss Rate (MR), the percentage of final-step predictions deviating more than 2 meters from the ground truth. For multi-modal approaches, we use the best-of-$N$ strategy with $N=20$, as in prior work~\cite{hauri2021multi, gu2022stochastic, mao2023leapfrog, xu2025sports, qi2024learning, zhangmulti2024}. Marginal metrics (e.g., ADE, FDE, MR) evaluate each player independently, while joint metrics (e.g., JADE, JFDE, JMR) assess team-level coordination and interaction dynamics~\cite{mao2023leapfrog,xu2025sports,weng2023joint}.
For adaptive generation, the Guidance Score (GS) measures the value determined by specific tactical objective functions. Additionally, two case studies with football experts from Birmingham City Football Club assess the realism and practical applicability of TacticGen. Further metric details are provided in Appendix~\ref{sec:metric}.

\subsection{Accurate Generation of Multi-Player Motion Trajectories}\label{sec:exp-prediction}

\subsubsection{Quantitative Results}
We conduct comprehensive evaluations on our large-scale football dataset to benchmark the performance of TacticGen against state-of-the-art methods. For fairness and consistency with prior baselines, we adopt TacticGen-P as the representative version of TacticGen for direct comparison. All models are trained for 600K steps with a batch size of 512 on the training set and evaluated on the test set, predicting the next 54 frames of both the ball and players from the past 10 frames. The corresponding results are presented in Table~\ref{tab:prediction-results}.

\begin{table}[htbp]
\vspace{-0.1in}
\centering
\caption{{Performance of different methods on our football dataset.} Models marked with $^*$ are deterministic. Best-of-$N$ results are reported where applicable ($N=20$). \textbf{Bolded} values denote the best performance (lower is better). All diffusion-based methods are equipped with the event encoder. All distance-based metrics are measured in meters.}\label{tab:prediction-results}
\resizebox{0.5\textwidth}{!}{%
\begin{tabular}{@{}lcccccc@{}}
\toprule
\multirow{2}{*}{\bf Method} 
  & \multicolumn{3}{c}{\bf Marginal} 
  & \multicolumn{3}{c}{\bf Joint} \\
\cmidrule(lr){2-4} \cmidrule(lr){5-7}
  & ADE & FDE & MR (\%) & JADE & JFDE & JMR (\%) \\
\midrule
LSTM$^*$~\cite{hochreiter1997lstm} & $-$ &$-$ &$-$ &1.65 &3.52 &45.32 \\
Social LSTM$^*$~\cite{alahi2016social} &$-$ &$-$ &$-$ &1.11 &2.33 &31.50 \\
Transformer$^*$~\cite{vaswani2017attention} &$-$ &$-$ &$-$ &1.31 &2.54 &37.47 \\
STGAT$^*$~\cite{huang2019stgat} &$-$ &$-$ &$-$ &1.09 &2.32 &31.13 \\
GRNN$^*$~\cite{dick2021rating} &$-$ &$-$ &$-$ &1.25 &2.40 &32.14 \\
DAG-Net$^*$~\cite{monti2021dag} &$-$ &$-$ &$-$ &1.01 &1.83 &25.90 \\
FootBots$^*$~\cite{capellera2024footbots} &$-$ &$-$ &$-$ &0.85 &1.47 &19.14 \\
Trajectron++~\cite{salzmann2020trajectron++} &0.61 &0.99 &11.23 &0.98 &1.52 &20.10 \\
mmTransformer~\cite{liu2021multimodal} &0.43 &0.68 &7.64 &0.69 &1.25 &15.50 \\
Scene Transformer~\cite{ngiam2021scene} &0.38 &0.64 &7.38 &0.67 &1.24 &14.98 \\
GVRNN~\cite{yeh2019diverse} &0.53 &0.84 &9.23 &0.79 &1.42 &19.20 \\
Graph Imputer~\cite{omidshafiei2022multiagent} &0.49 &0.77 &8.50 &0.73 &1.35 &18.16 \\
Sports-Traj~\cite{xu2025sports} &0.42 &0.65 &7.69 &0.62 &1.18 &14.30 \\
Diffuser~\cite{janner2022planning} &0.51 &0.79 &8.78 &0.80 &1.41 &18.55 \\
DiT~\cite{peebles2023scalable} &0.57 &0.64 &7.02 &0.87 &1.33 &17.76 \\
MID~\cite{gu2022stochastic} &0.36 &0.60 &6.34 &0.68 &1.31 &16.80 \\
LED~\cite{mao2023leapfrog} &0.36 &0.55 &5.70 &0.65 &1.27 &15.01 \\
PlayBest~\cite{chen2024playbest} &0.48 &0.75 &8.03 &0.75 &1.34 &18.02 \\
MADiff~\cite{zhu2023madiff} &0.33 &0.57 &5.73 &0.58 &1.14 &14.19 \\
\midrule
\textbf{TacticGen}  &\textbf{0.29} &\textbf{0.52} &\textbf{4.73} &\textbf{0.45} &\textbf{0.92} &\textbf{10.66} \\
\bottomrule
\end{tabular}%
}
\end{table}

The results demonstrate that TacticGen, composed of the multi-agent diffusion transformer, self-attention mechanism~\cite{vaswani2017attention}, and cross-attention module (check Section~\ref{sec:tacticgen} for detailed architectures), consistently outperforms existing trajectory prediction methods across multiple evaluation metrics. This highlights the effectiveness of TacticGen in modeling complex inter-agent interactions and leveraging contextual information for future trajectory prediction. 

To further validate this finding, we conduct a comprehensive ablation study that systematically examines the contribution of each component within TacticGen. The results underscore the effectiveness of both the context encoder and the event encoder, highlighting the critical role of the model design. Detailed results and analyses are provided in Appendix~\ref{sec:ablation}. 

Moreover, following TacticAI~\cite{wang2024tacticai}, we evaluated how TacticGen performs under temporal drift by re-running experiments and comparing it with the top-5 models from Table~\ref{tab:prediction-results} in a temporal setting. In this setup, we use the most recent 20\% of events from our dataset for testing, with the remaining 80\% for training. The results of the temporal split experiments are presented in Appendix~\ref{sec:temporal-exp}.

In addition, to demonstrate the generality and transferability of TacticGen beyond the football domain, we evaluate its performance on a widely used public basketball trajectory dataset, enabling direct comparison with established baselines~\cite{gupta2018socialgan,huang2019stgat,mohamed2020socialstgcnn,mangalam2020pecnet,yu2020star,salzmann2020trajectron++,xu2022memonet,bae2022npsn,xu2022groupnet,gu2022stochastic,mao2023leapfrog,zhu2023madiff}. The results demonstrate that TacticGen still achieves competitive performance, highlighting its potential as a general framework for multi-agent trajectory generation across diverse domains. Appendix~\ref{sec:nba-results} provides detailed results.

\subsubsection{Qualitative Results}
To gain deeper insights into TacticGen’s generating behaviors beyond numerical metrics, we present visualizations that highlight the predictive patterns of different models. In addition to TacticGen-P, we also visualize TacticGen-C, which enhances the model’s ability by incorporating a complete ball trajectory to better predict player trajectories, and it reflects scenarios where the ball’s path is assumed or externally specified during tactical planning.

\begin{figure}[htbp]
\vspace{-0.05in}
  \centering

  \begin{minipage}[t]{0.16\textwidth}
    \centering
    \small \textbf{(a)} Ground Truth\\
    \includegraphics[width=\linewidth]{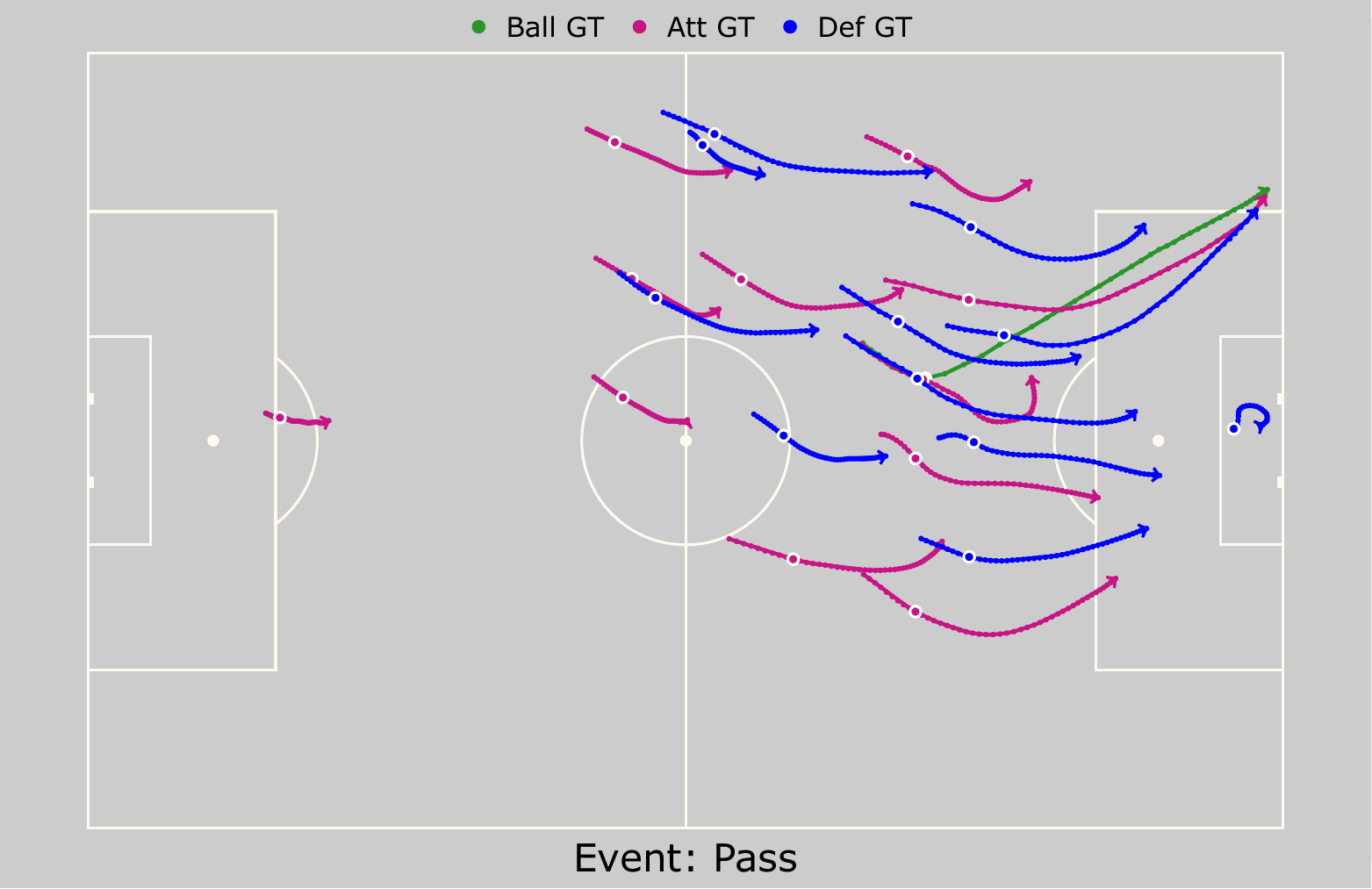}
  \end{minipage}\hfill
  \begin{minipage}[t]{0.16\textwidth}
    \centering
    \small \textbf{(b)} Diffuser\\
    \includegraphics[width=\linewidth]{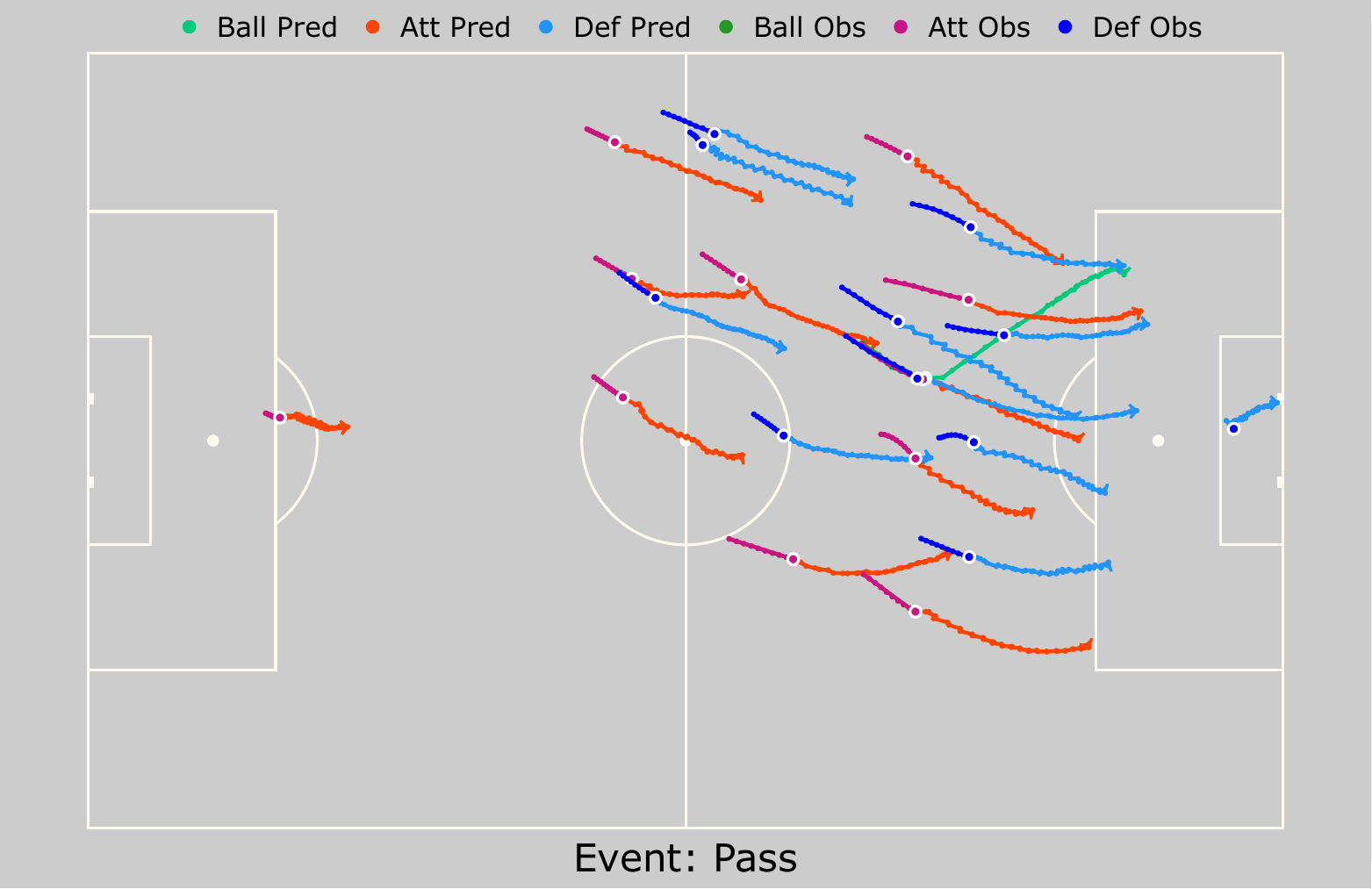}
  \end{minipage}\hfill
  \begin{minipage}[t]{0.16\textwidth}
    \centering
    \small \textbf{(c)} MID\\
    \includegraphics[width=\linewidth]{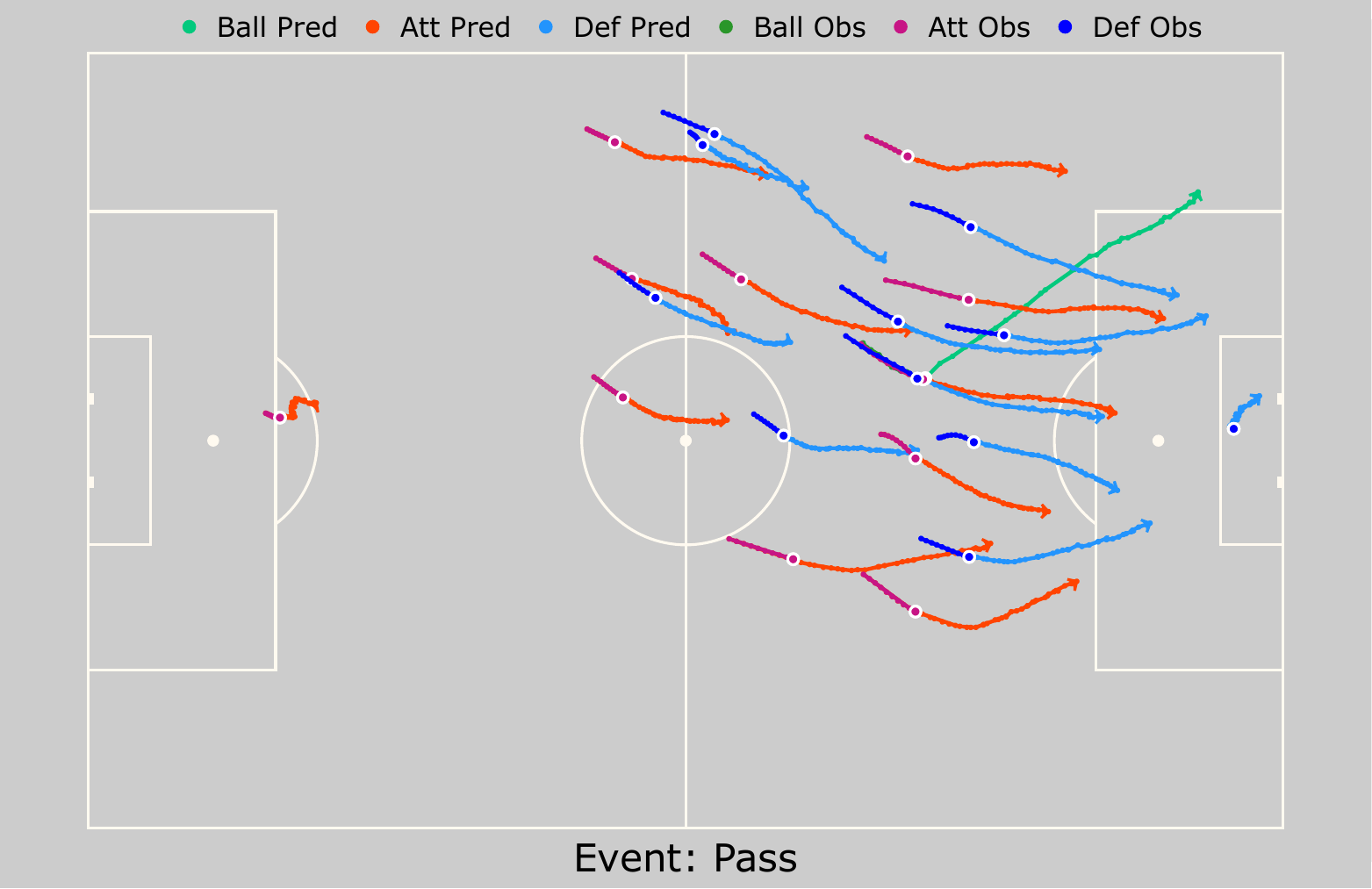}
  \end{minipage}

  \vspace{0.5em} 

  \begin{minipage}[t]{0.16\textwidth}
    \centering
    \small \textbf{(d)} MADiff\\
    \includegraphics[width=\linewidth]{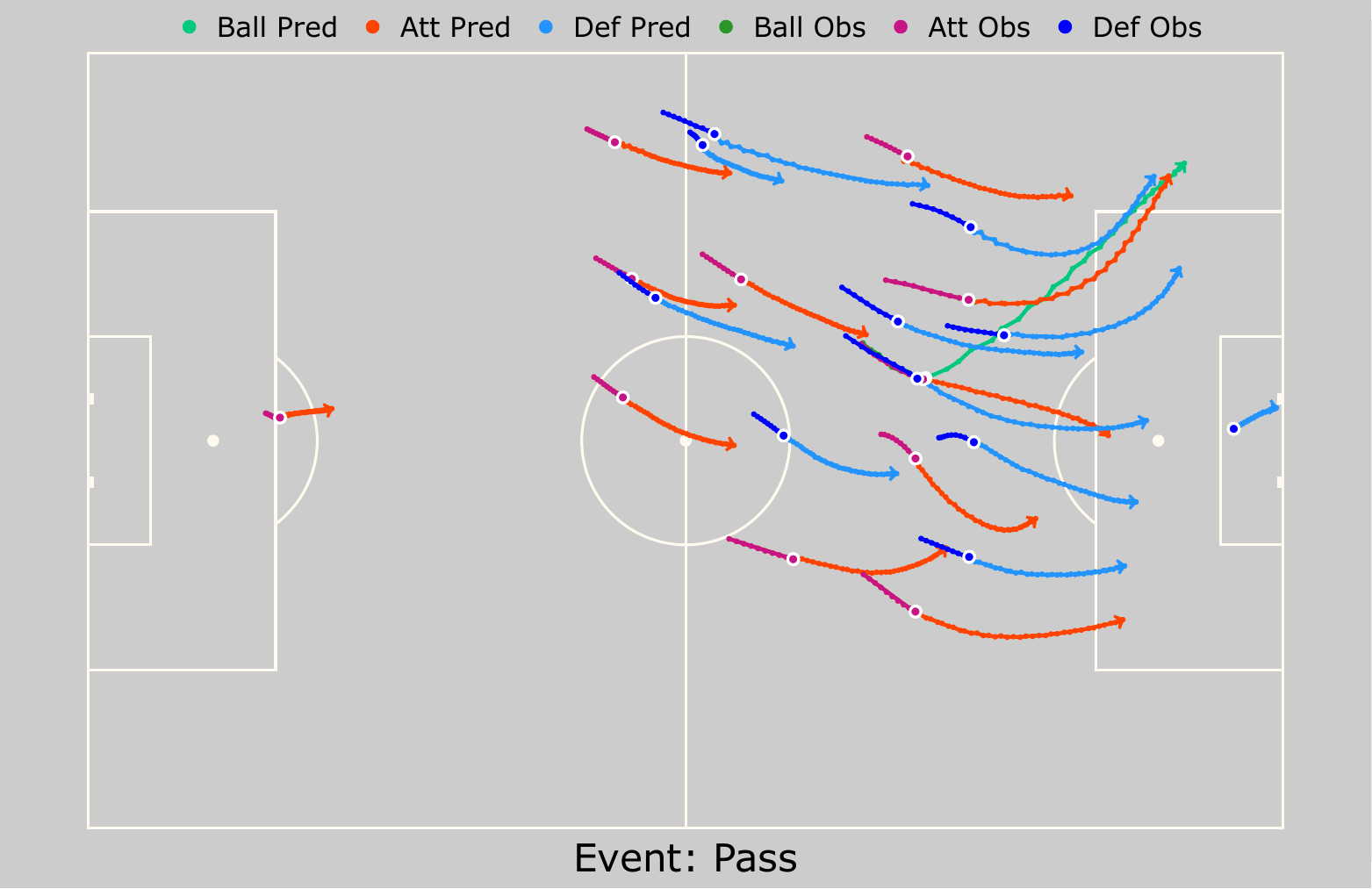}
  \end{minipage}\hfill
  \begin{minipage}[t]{0.16\textwidth}
    \centering
    \small \textbf{(e)} TacticGen-P\\
    \includegraphics[width=\linewidth]{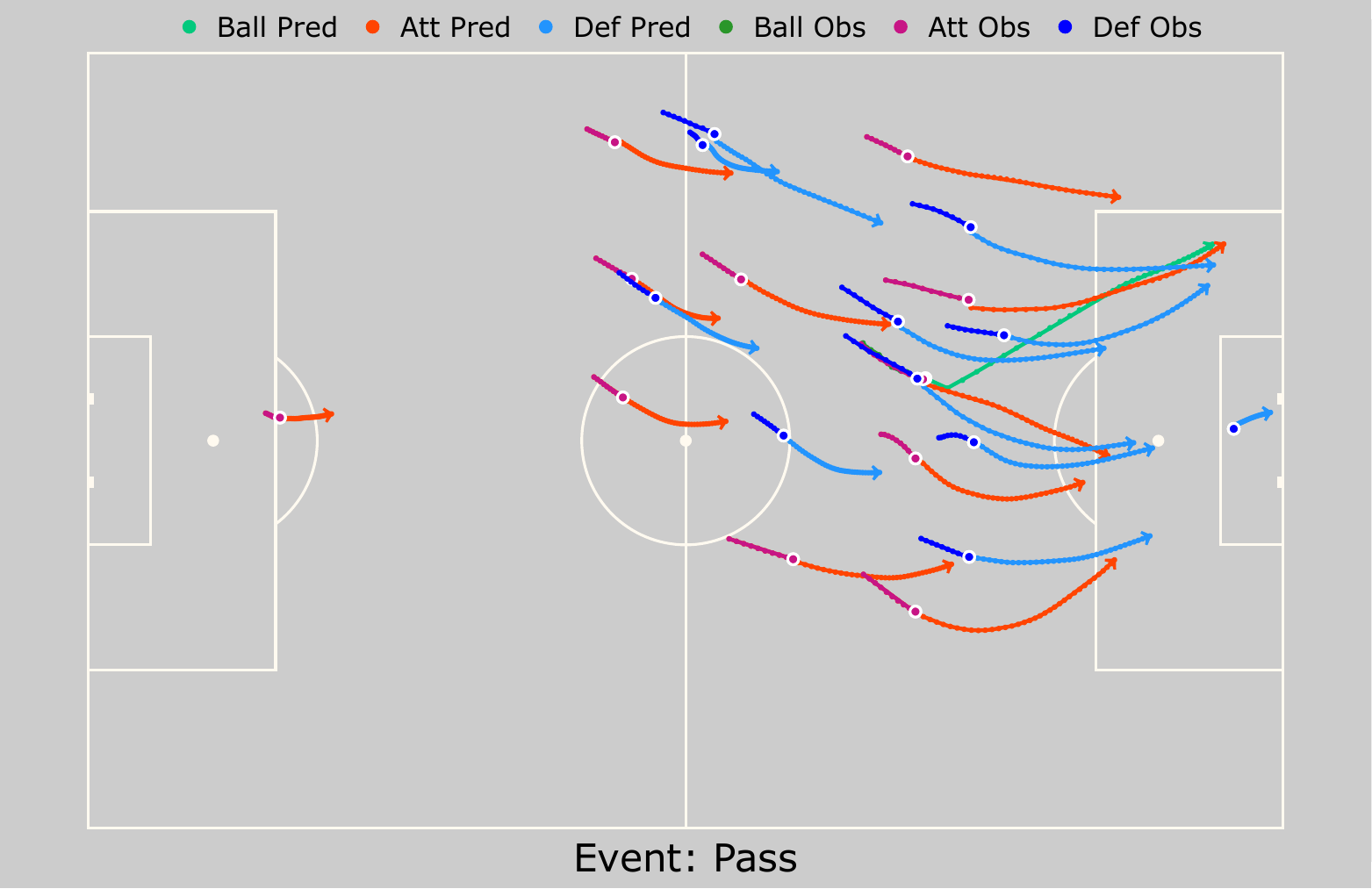}
  \end{minipage}\hfill
  \begin{minipage}[t]{0.16\textwidth}
    \centering
    \small \textbf{(f)} TacticGen-C\\
    \includegraphics[width=\linewidth]{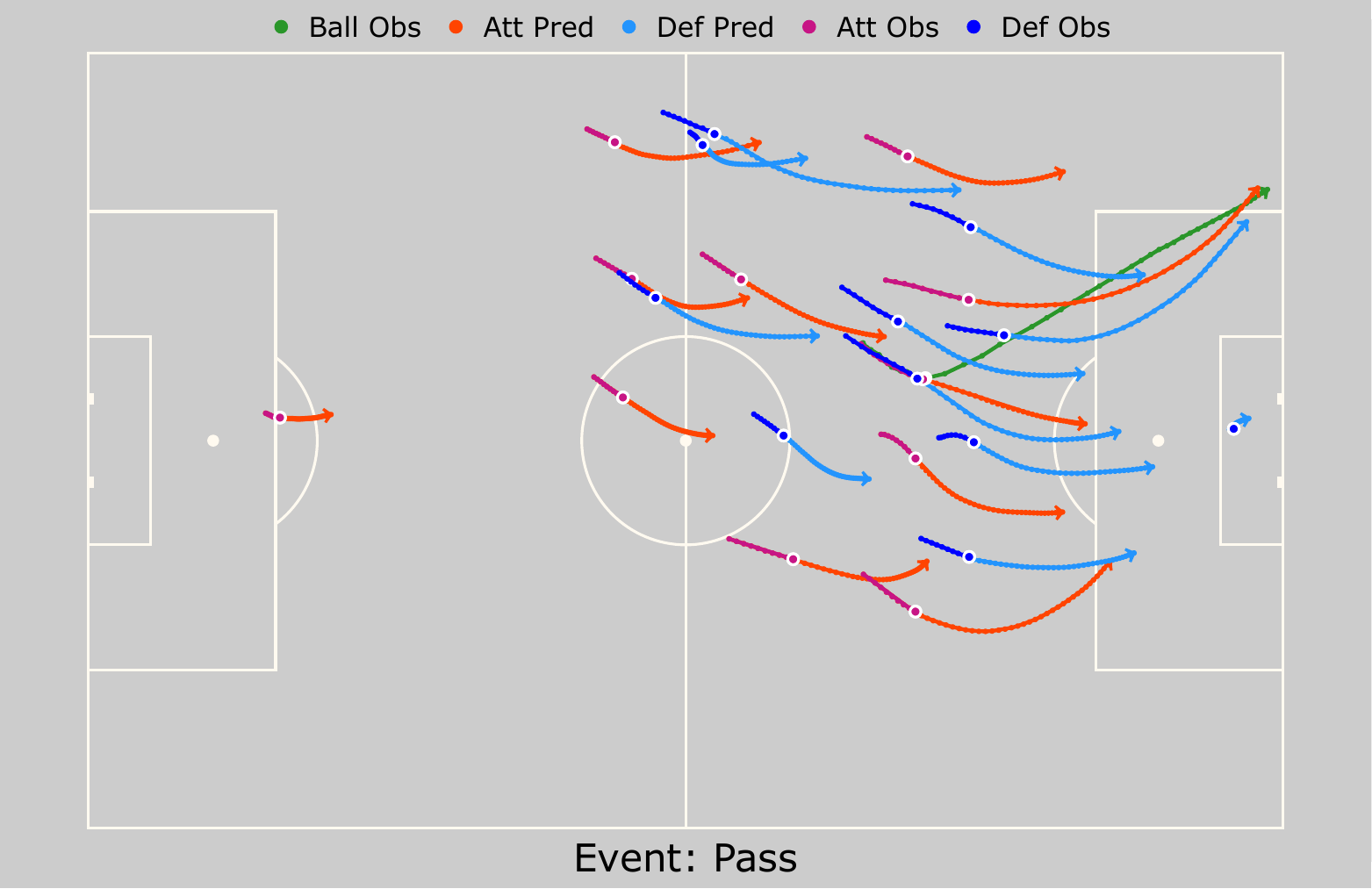}
  \end{minipage}

  \caption{{Ground truth trajectories for a pass event and the corresponding best-of-20 predictions from five methods.} GT denotes ground truth, Obs denotes observed conditions, and Pred denotes predicted trajectories.}  \label{fig:pred-pass-46}
  \vspace{-0.05in}
\end{figure}

Figure~\ref{fig:pred-pass-46} illustrates a representative pass event by comparing the ground truth trajectories with the best-of-20 predictions from five different methods. The trajectories generated by TacticGen visually align more closely with the ground truth (also with lower JADE, JFDE, and JMR), which demonstrates its strong ability to capture the underlying dynamics and structured movement patterns inherent in multi-agent football scenarios. More importantly, in terms of realism, the predictions from Diffuser\cite{janner2022planning}, MID\cite{gu2022stochastic}, and MADiff~\cite{zhu2023madiff} often exhibit noticeable inconsistencies, such as abrupt directional changes or flickering motions among certain players. Instead, both variants of TacticGen consistently produce smooth, coherent, and visually realistic trajectories that not only outperform the baselines in visual fidelity but also better reflect plausible team behaviors. The player movements generated by TacticGen show clear attention to the ball’s motion, whether the ball trajectory is predicted by the model (TacticGen-P) or specified as an observed condition (TacticGen-C).

\begin{figure}[htbp]
\vspace{-0.1in}
  \centering
  \resizebox{0.5\textwidth}{!}{%
    \begin{minipage}[t]{0.33\textwidth}
      \centering
      \small \textbf{(a)} TacticGen-P\\
      \includegraphics[width=\linewidth]{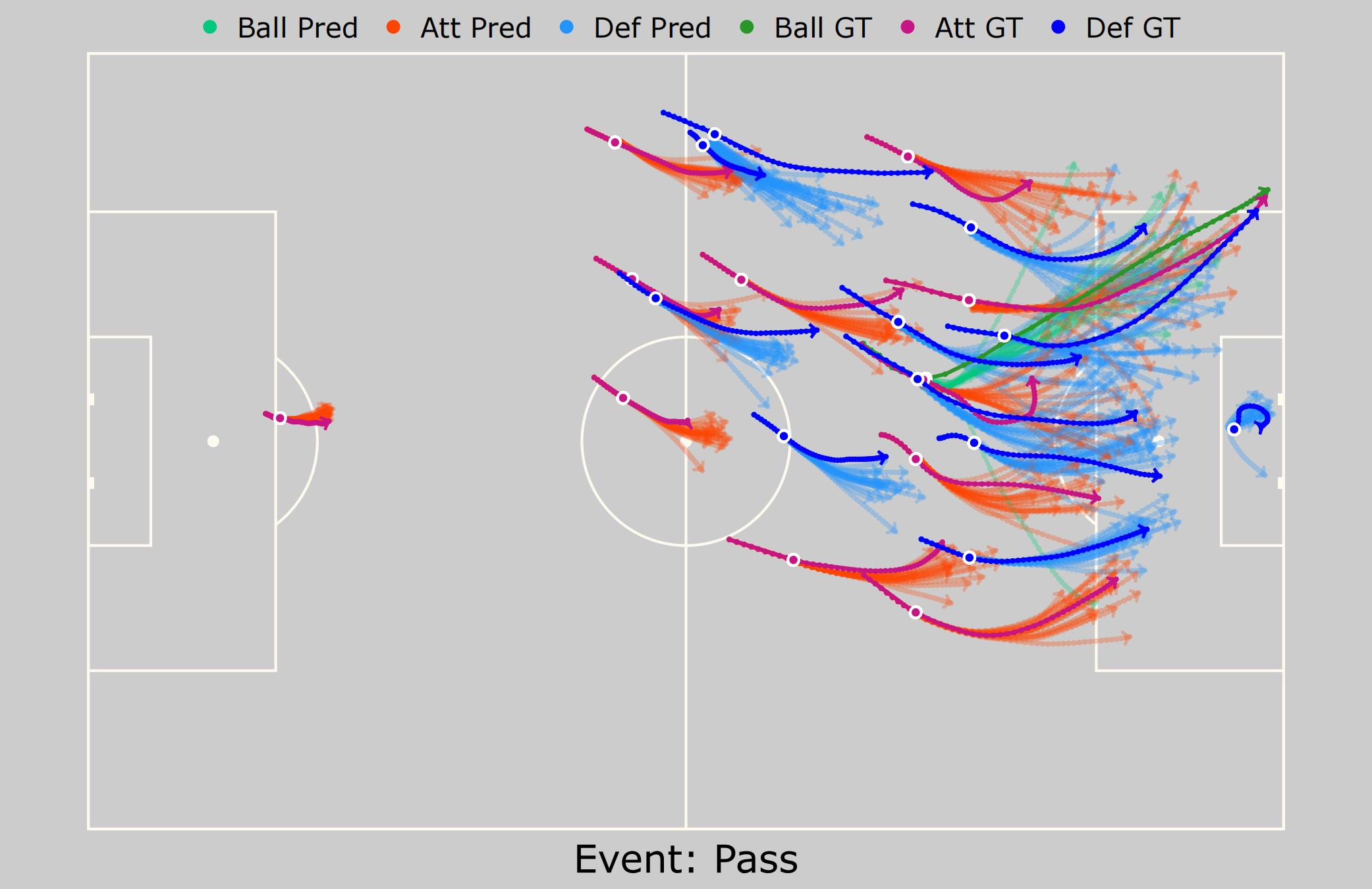}
    \end{minipage}
    \hspace{0.01\textwidth}
    \begin{minipage}[t]{0.33\textwidth}
      \centering
      \small \textbf{(b)} TacticGen-C\\
      \includegraphics[width=\linewidth]{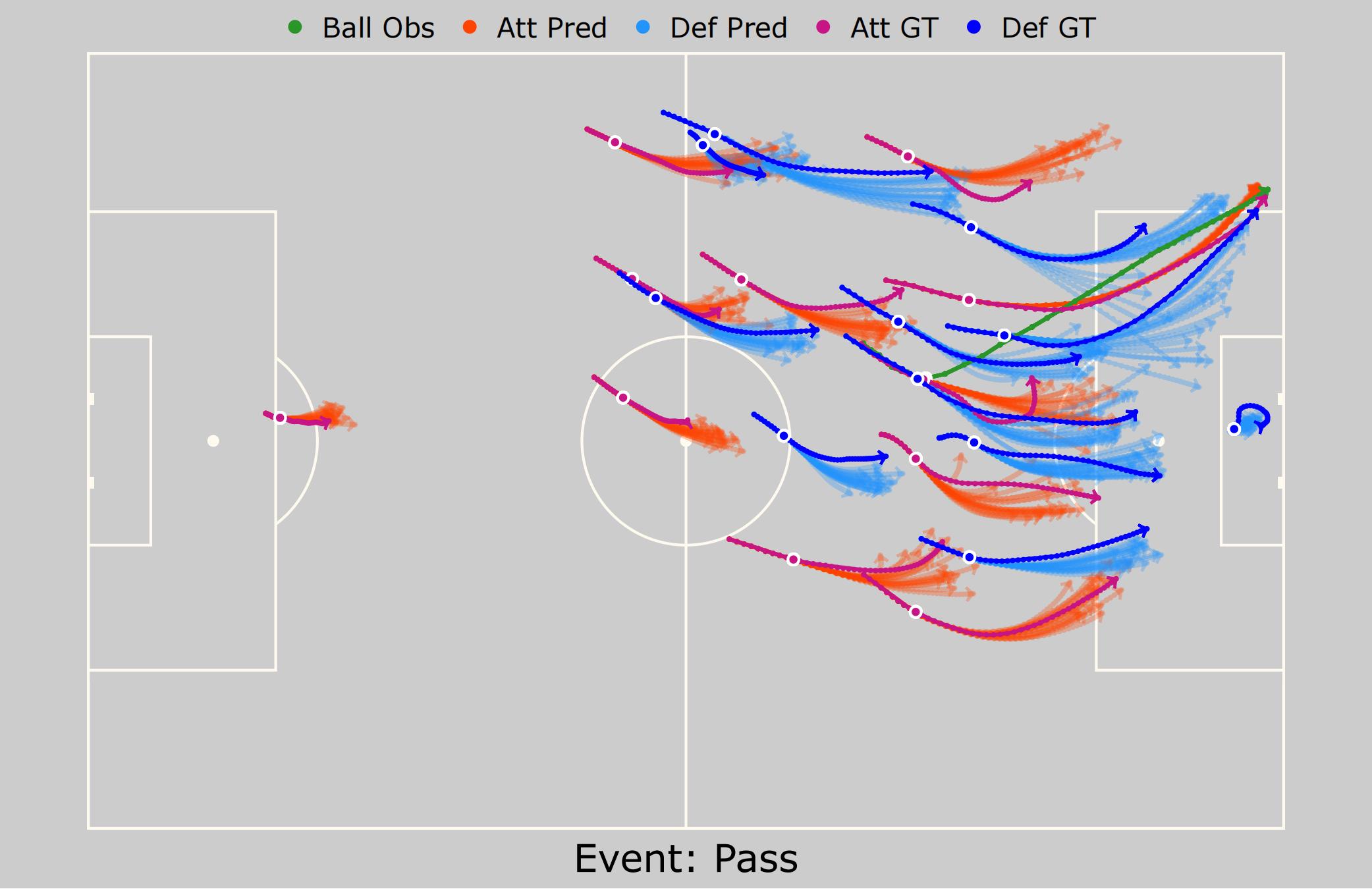}
    \end{minipage}%
  }
  \caption{{20 predicted trajectory samples by TacticGen variants for a pass event.}}
  \label{fig:footdiff-diversity-pass46}
    \vspace{-0.05in}
\end{figure}

Figure~\ref{fig:footdiff-diversity-pass46} presents 20 trajectory samples generated by TacticGen-P and TacticGen-C, demonstrating their ability to produce diverse and multi-modal predictions. It is worth noting that TacticGen-C exhibits slightly less diversity than TacticGen-P, as the ball trajectory is fixed as a condition, and players are inclined to run accordingly. This capability stems from its diffusion-based generative framework, which allows sampling from a wide distribution of possible future trajectories. The sampled trajectories demonstrate coherent team level coordination, plausible motion dynamics, and tactically meaningful variations, such as alternative supporting runs, passing lanes, or defensive adjustments. Crucially, this diversity reflects structured differences in collective behavior rather than stochastic perturbations, indicating that the model learns a distribution over plausible trajectories instead of generating unrealistic samples. Such controlled diversity forms a principled basis for adaptable tactic generation, enabling the production of varied tactical solutions aligned with different objectives, as shown in Section~\ref{sec:exp-guidance}. We also provide more prediction visualizations in Appendix~\ref{sec:more-exp-pred}.

The accurate prediction of multi-player trajectories shows that TacticGen can capture underlying movement patterns and interactions of both players and the ball. This provides a solid foundation for adaptable tactic generation as follows.

\subsection{Adaptable Tactic Generation for Diverse Objectives}\label{sec:exp-guidance}
A key prerequisite for tactic generation is the ability to adapt to user-specified objectives~\cite{wang2024tacticai}. To this end, we move beyond passive \textit{prediction} and actively \textit{guide generation} toward desired tactical outcomes. TacticGen leverages a classifier-guidance mechanism~\cite{dhariwal2021diffusion, janner2022planning} to incorporate diverse objectives into a unified diffusion-based trajectory generator, enabling a single high-capacity prediction model to be trained once and then flexibly guided to generate diverse tactics without retraining (see Section~\ref{sec:tacticgen-guided-generation} for more details). These objectives can be specified through pre-defined rules, natural language descriptions, or learned value models.

In practice, coaches typically design tactics by anticipating ball movement and then adjusting player positioning accordingly. Therefore, we use TacticGen-C as the representative version of TacticGen for tactic generation when no specific descriptions are provided. A key advantage of TacticGen’s guidance mechanism is its flexibility in jointly generating trajectories for both attacking and defending teams while selectively applying guidance. When guiding one team, the other can either (i) follow \textit{recorded trajectories from the dataset}, (ii) replay \textit{the model’s original predictions}, or (iii) produce \textit{reactive trajectories in response}. In the following, we present results for the first setting, with additional results for the other two settings provided in Appendix~\ref{sec:replay-reactive-results}. We also report results for an additional event in Appendix~\ref{sec:more-event-result}.

\subsubsection{Rule-based Guidance with Domain Knowledge}
In football, tactics can be expressed as rule-based functions over player movements, capturing high-level objectives such as maintaining team shape or pressing opponents. In TacticGen, users can define rule-based functions grounded in domain knowledge to enable the generation of versatile football tactics, as long as these functions are differentiable. Notably, these functions can be flexibly composed with adjustable weights at inference time, providing a principled solution for tactic generation under complex and multifaceted objectives. It is worth noting that the experiments with predefined objectives in this paper serve only as illustrative examples. In practice, users can define additional rules beyond those provided, as long as the corresponding functions represent the intended tactics and are differentiable.

\begin{figure}[htbp]
\vspace{-0.05in}
  \centering

  \begin{minipage}[t]{0.16\textwidth}
    \centering
    \small \textbf{(a)} Ground Truth
    \includegraphics[width=\linewidth]{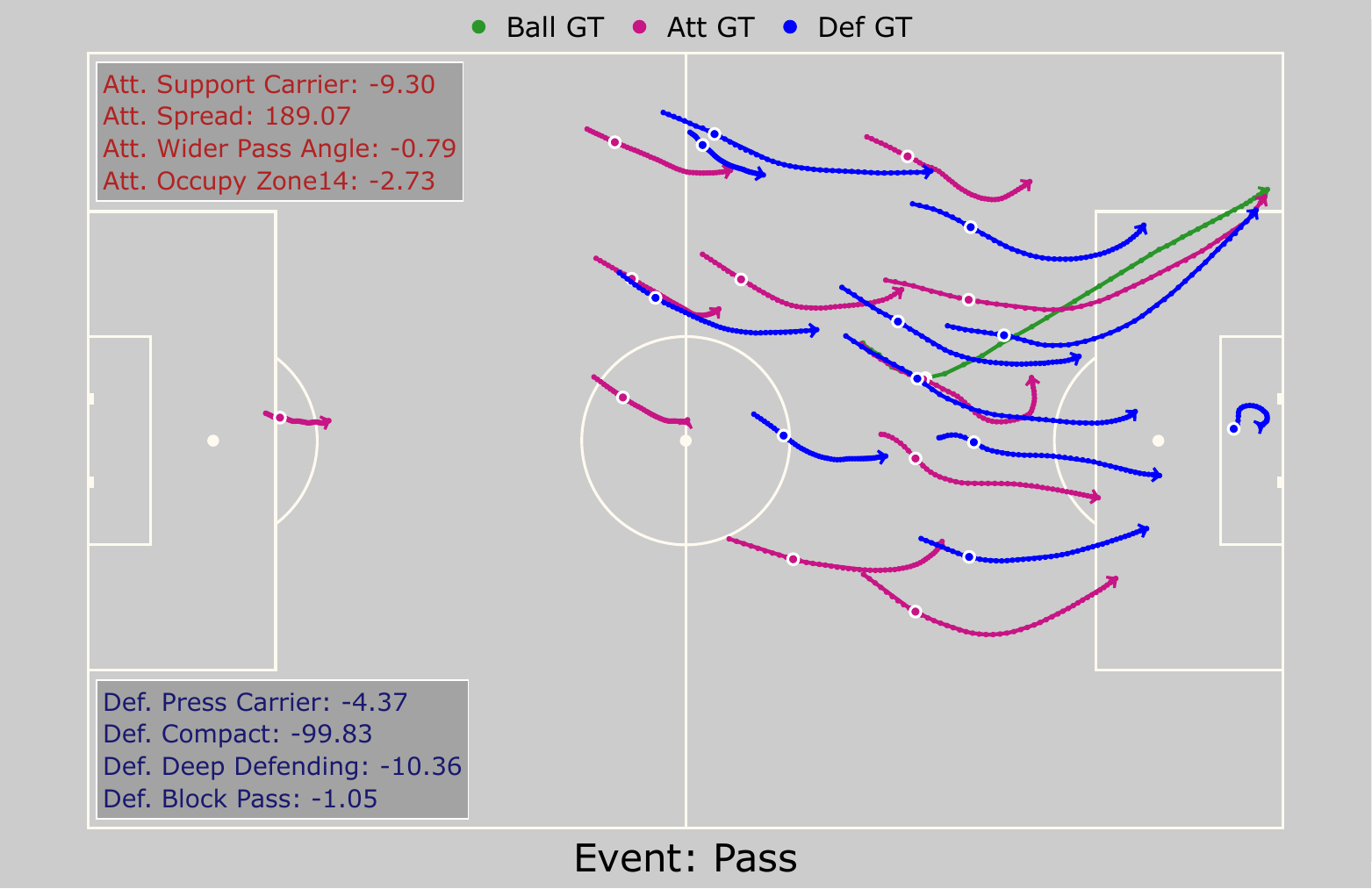}
  \end{minipage}\hfill
  \begin{minipage}[t]{0.16\textwidth}
    \centering
    \small \textbf{(b)} Att. Rule Guid.
    \includegraphics[width=\linewidth]{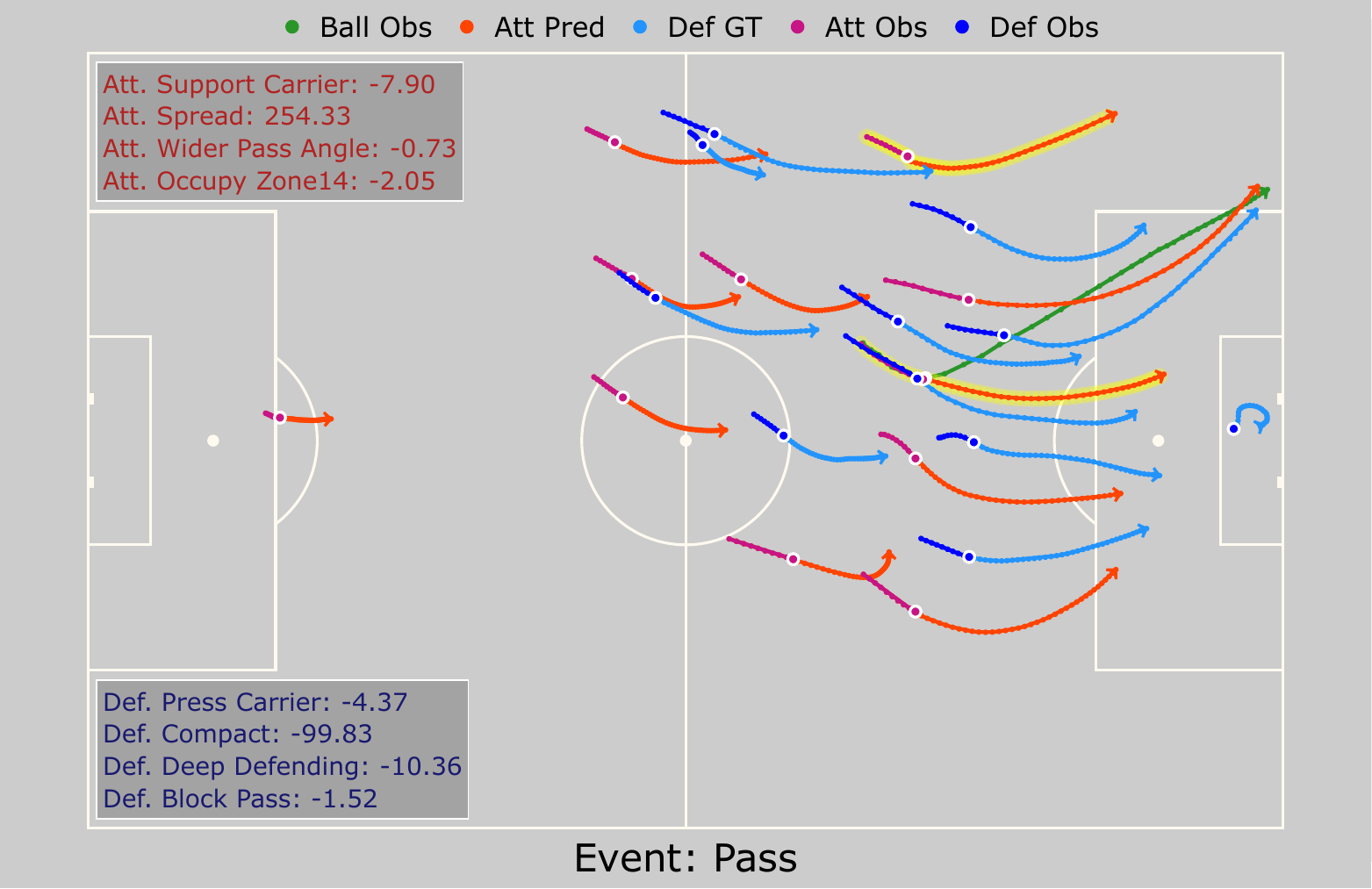}
  \end{minipage}\hfill
  \begin{minipage}[t]{0.16\textwidth}
    \centering
    \small \textbf{(c)} Def. Rule Guid.
    \includegraphics[width=\linewidth]{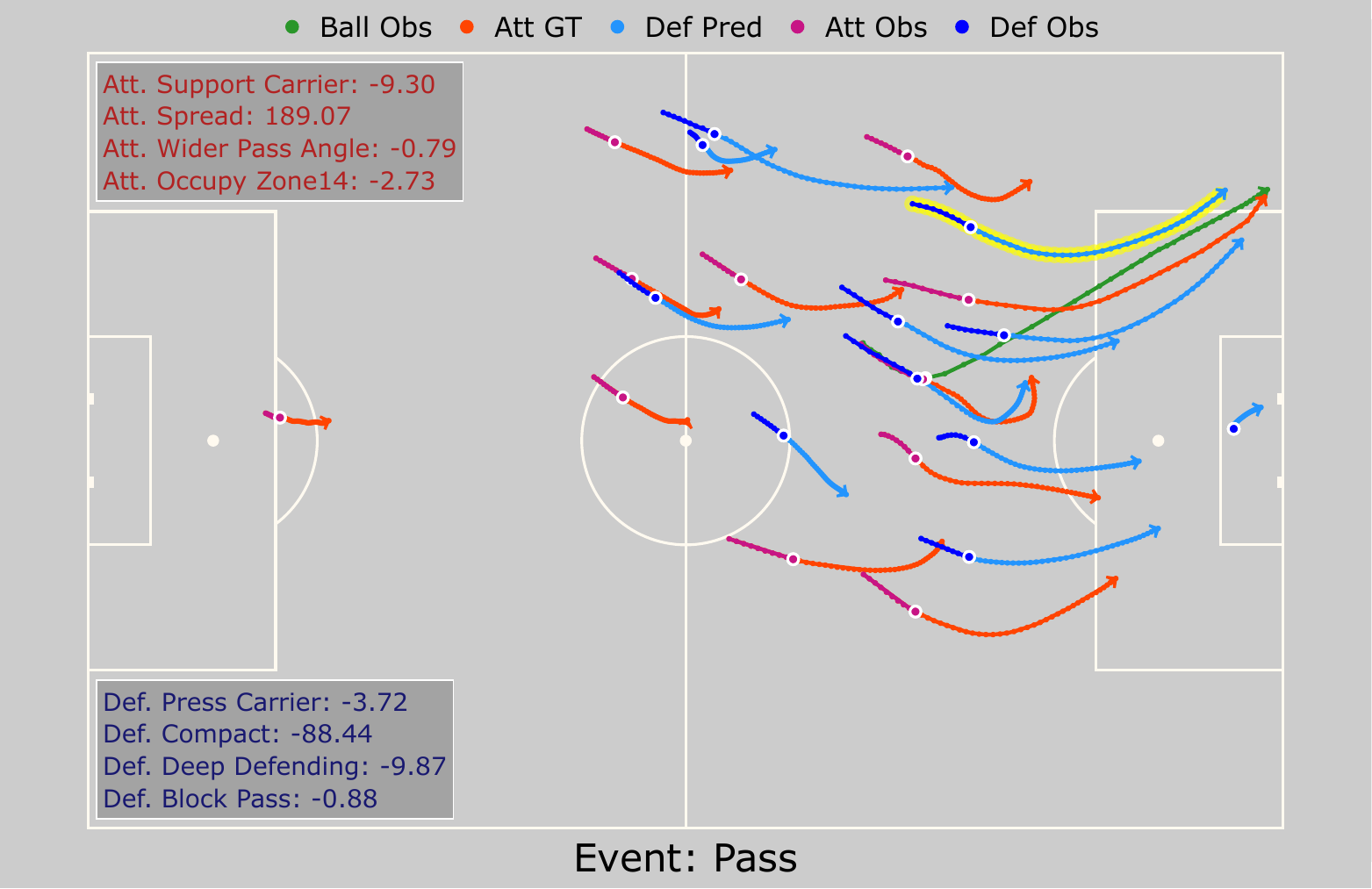}
  \end{minipage}

  \caption{{Trajectories generated by TacticGen for a pass event.} Guidance scores are displayed in the corner boxes as evaluation metrics. \textbf{(a)} Ground Truth. \textbf{(b)} Guidance for the attacking team with composed rules. Notably, the top-right player shifts toward the corner to help maintain team width, while the central player moves to occupy Zone 14 instead of retreating. \textbf{(c)} Guidance for the defending team with composed rules. It is evident that, instead of a single defender pressing the ball carrier as in the ground truth, TacticGen generates a scenario where two defenders move to apply pressure.}\label{fig:fce-rule-guide-pass46}
  \vspace{-0.05in}
\end{figure}

Figure~\ref{fig:fce-rule-guide-pass46} illustrates the generated trajectories guided by a combination of rule-based functions of tactic objectives for both the attacking and defending teams. For the attacking side, the guidance includes objectives such as supporting the ball carrier, spreading team shape, creating wider passing angles, and occupying zone 14 (the central area outside the penalty box)\footnote{\url{https://the-footballanalyst.com/zone-14-the-most-dangerous-area-in-football/}}. For the defending side, the functions encourage behaviors like pressing the ball carrier, collapsing to the ball, deepening defending, and blocking the passing paths. The corresponding guidance scores for each team are shown in the corner boxes. Upon applying guidance, the corresponding guidance values alternate as expected, and the generated trajectories display more structured and purposeful movements aligned with tactical intentions. These results demonstrate that rule-based guidance functions can effectively steer multi-agent trajectory generation toward diverse strategic objectives.

\begin{figure}[htbp]
\vspace{-0.05in}
  \centering

  \begin{minipage}[t]{0.16\textwidth}
    \centering
    \small \textbf{(a)} Ground Truth\\
    \includegraphics[width=\linewidth]{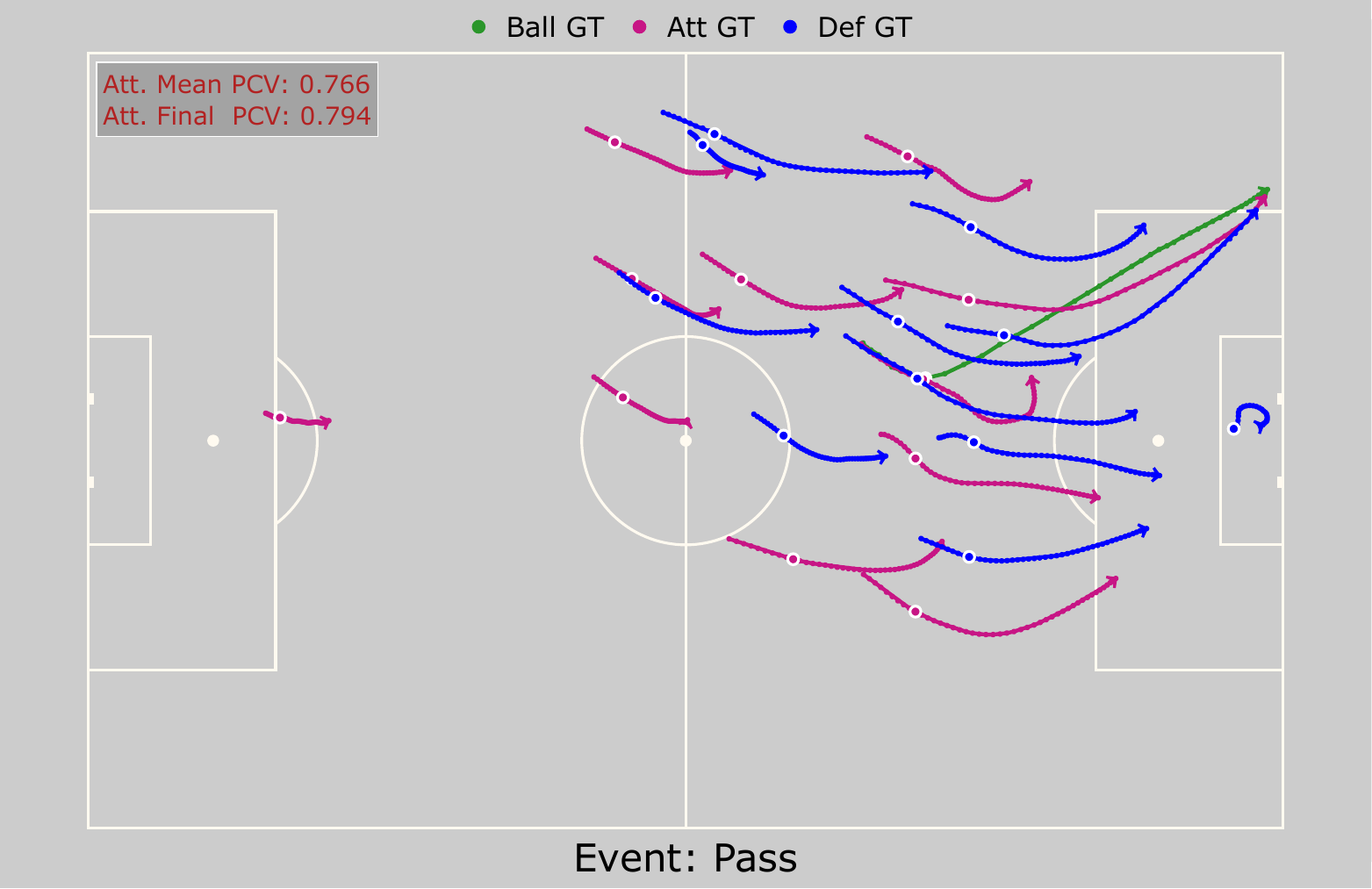}
  \end{minipage}\hfill
  \begin{minipage}[t]{0.16\textwidth}
    \centering
    \small \textbf{(b)} Att. High PCV\\
    \includegraphics[width=\linewidth]{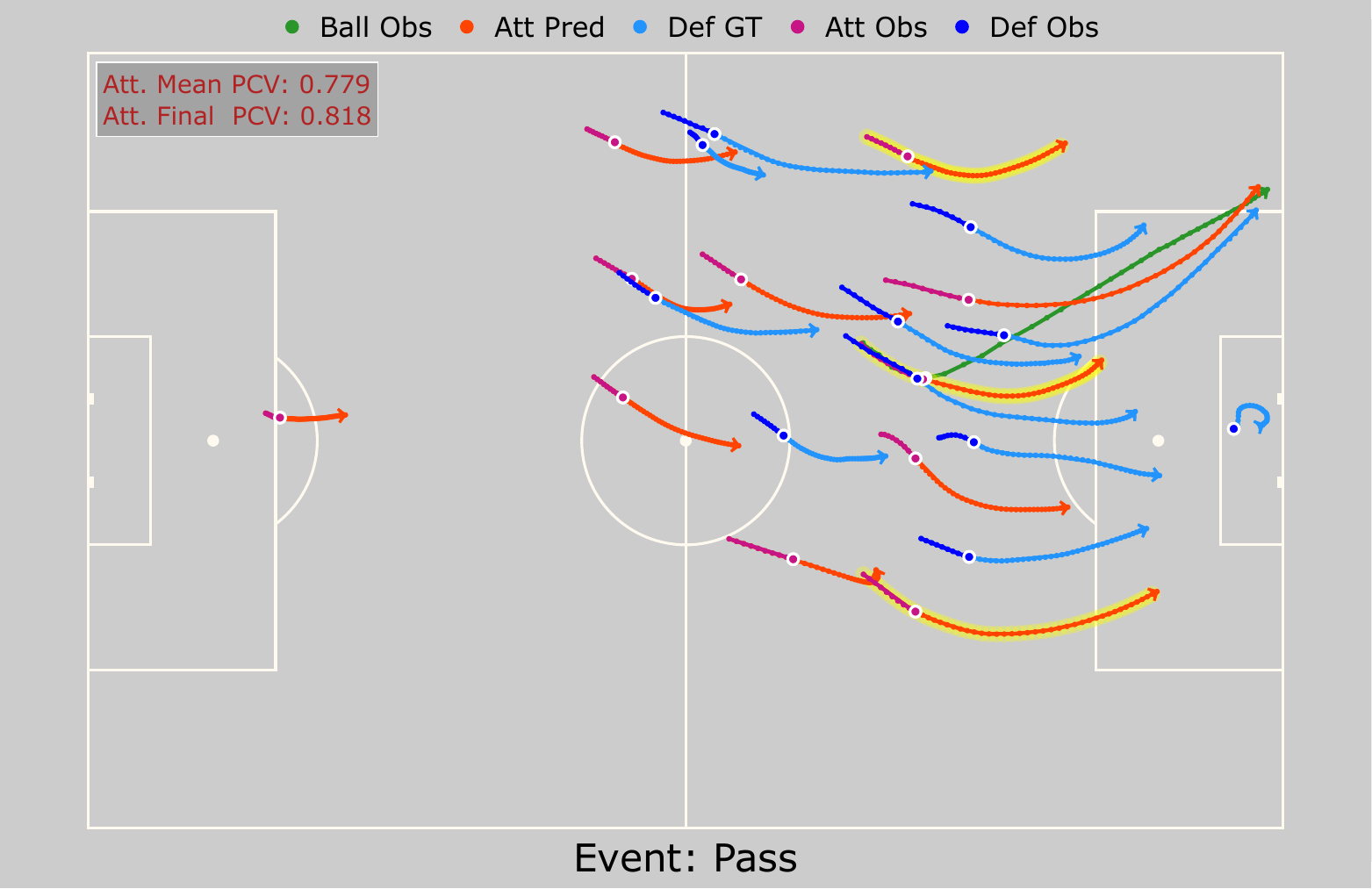}
  \end{minipage}\hfill
  \begin{minipage}[t]{0.16\textwidth}
    \centering
    \small \textbf{(c)} Def. High PCV\\
    \includegraphics[width=\linewidth]{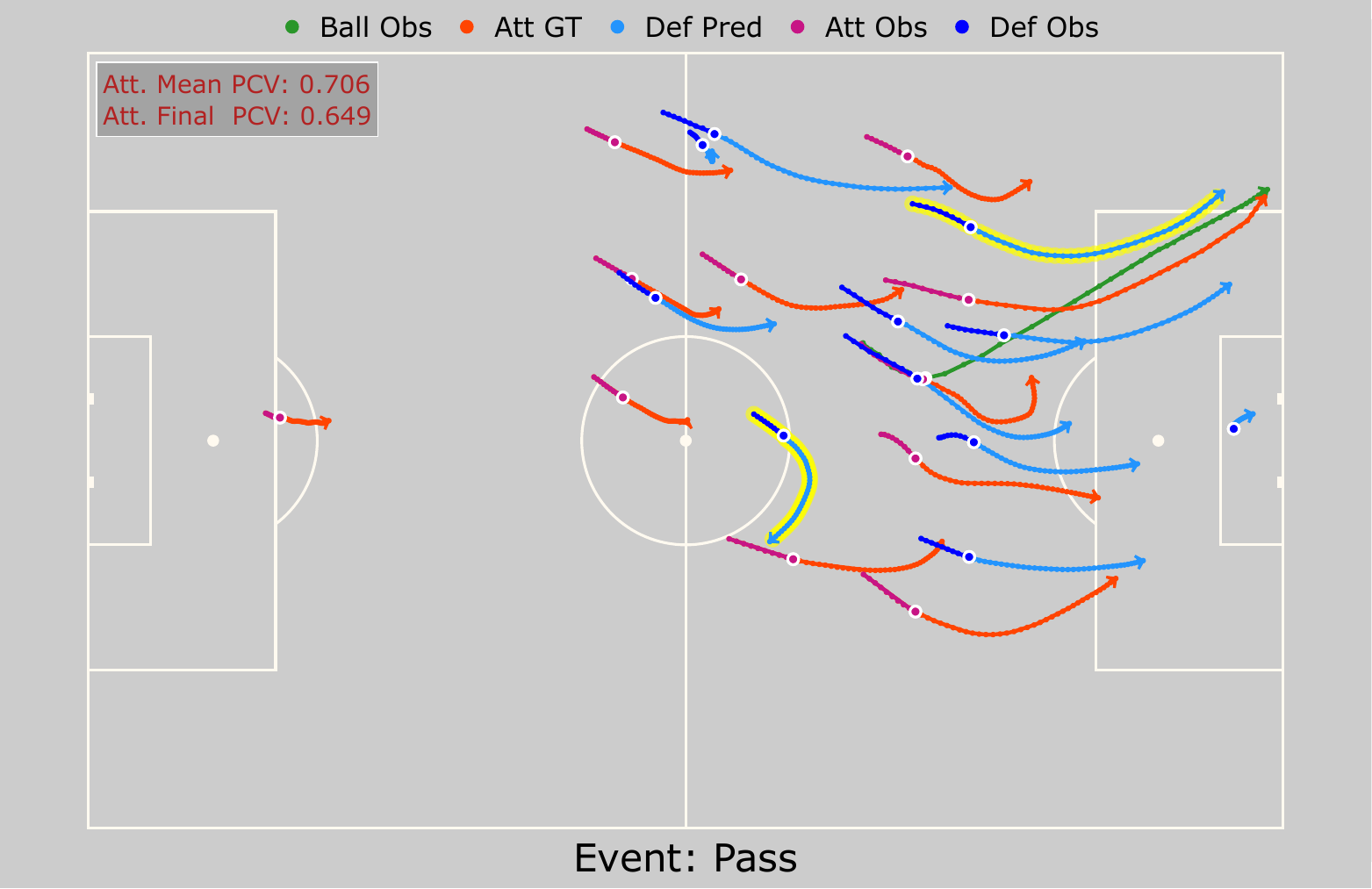}
  \end{minipage}

  \vspace{0.4em}

  \begin{minipage}[t]{0.16\textwidth}
    \centering
    \includegraphics[width=\linewidth]{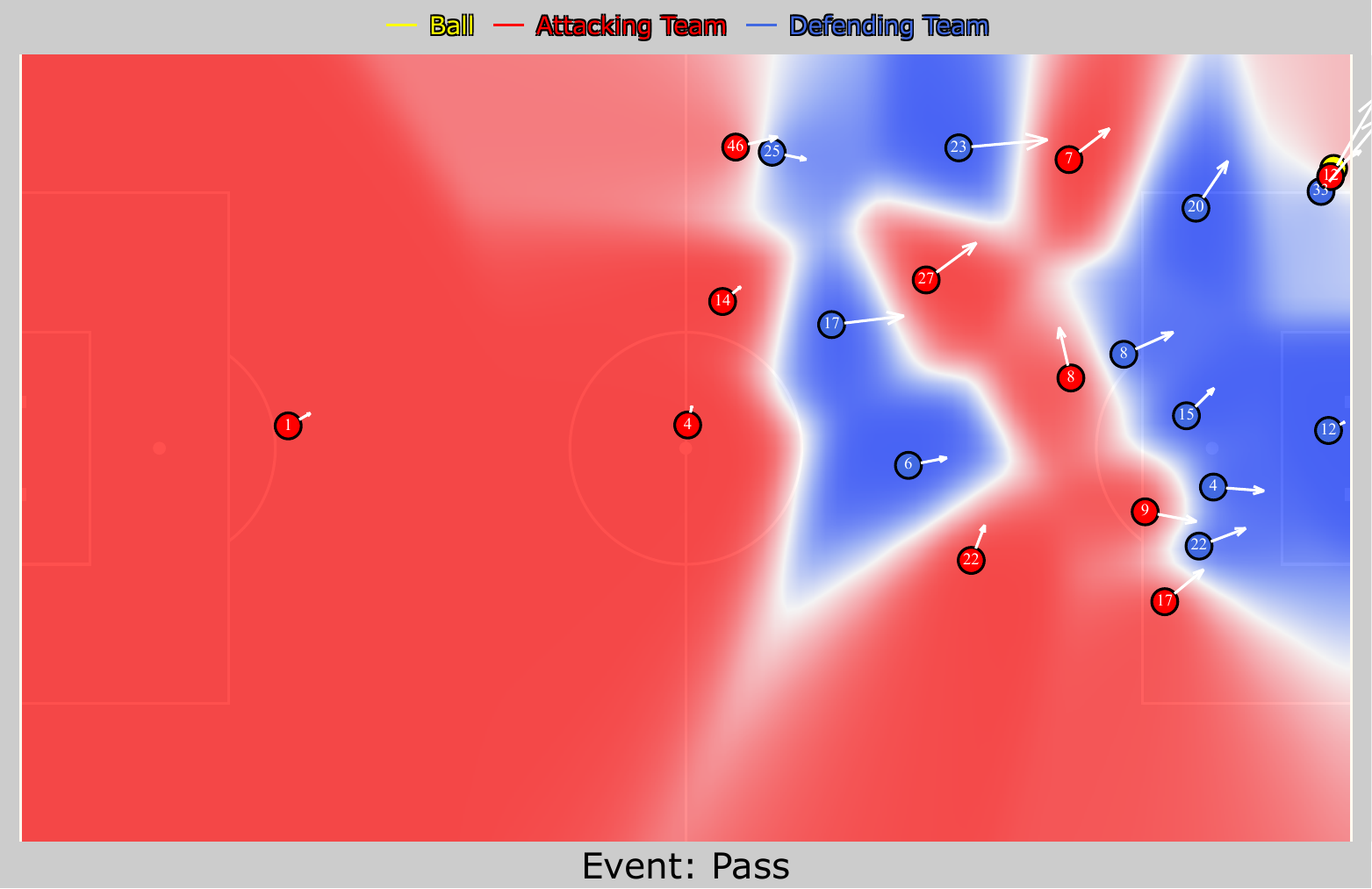}
  \end{minipage}\hfill
  \begin{minipage}[t]{0.16\textwidth}
    \centering
    \includegraphics[width=\linewidth]{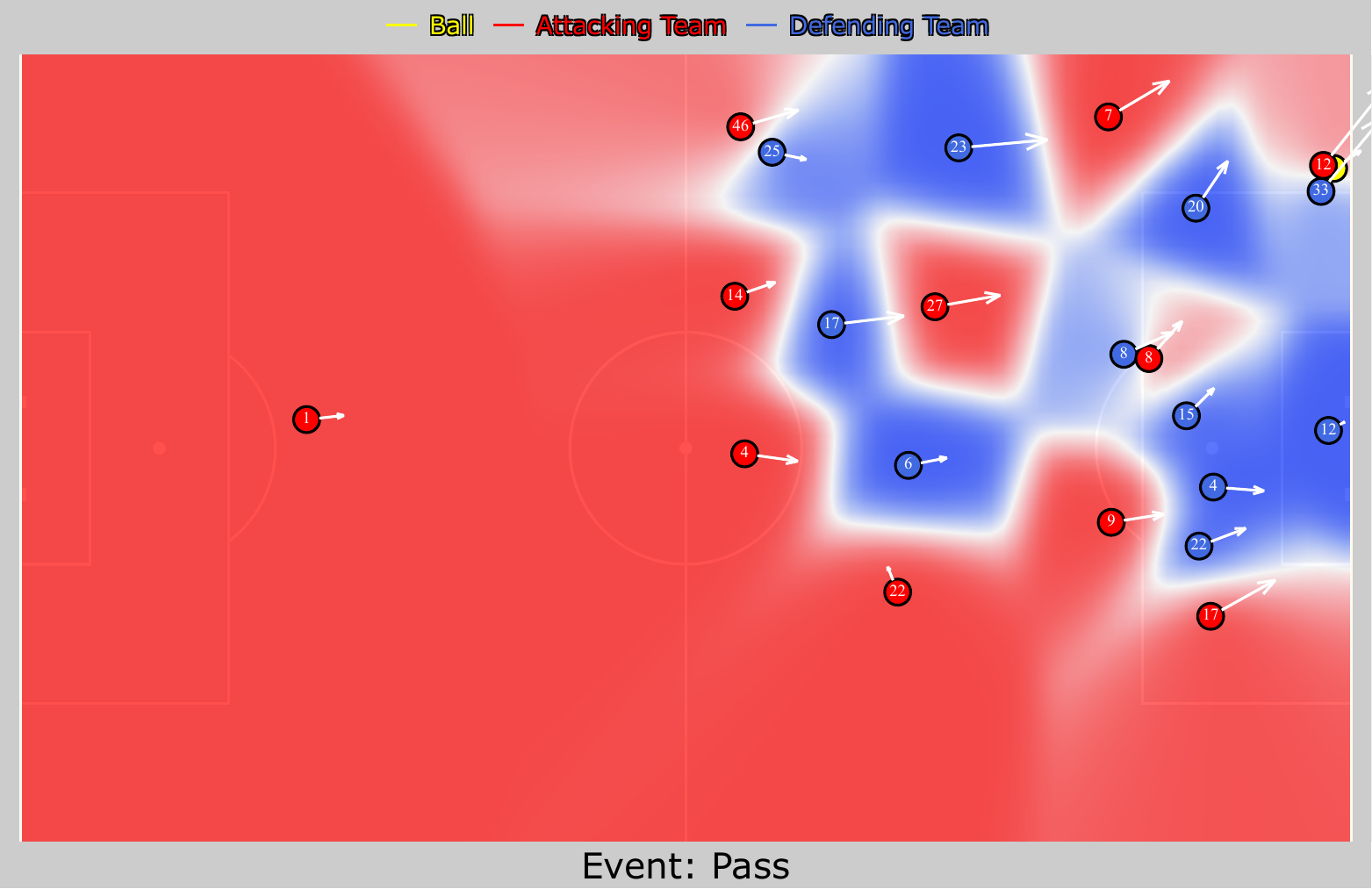}
  \end{minipage}\hfill
  \begin{minipage}[t]{0.16\textwidth}
    \centering
    \includegraphics[width=\linewidth]{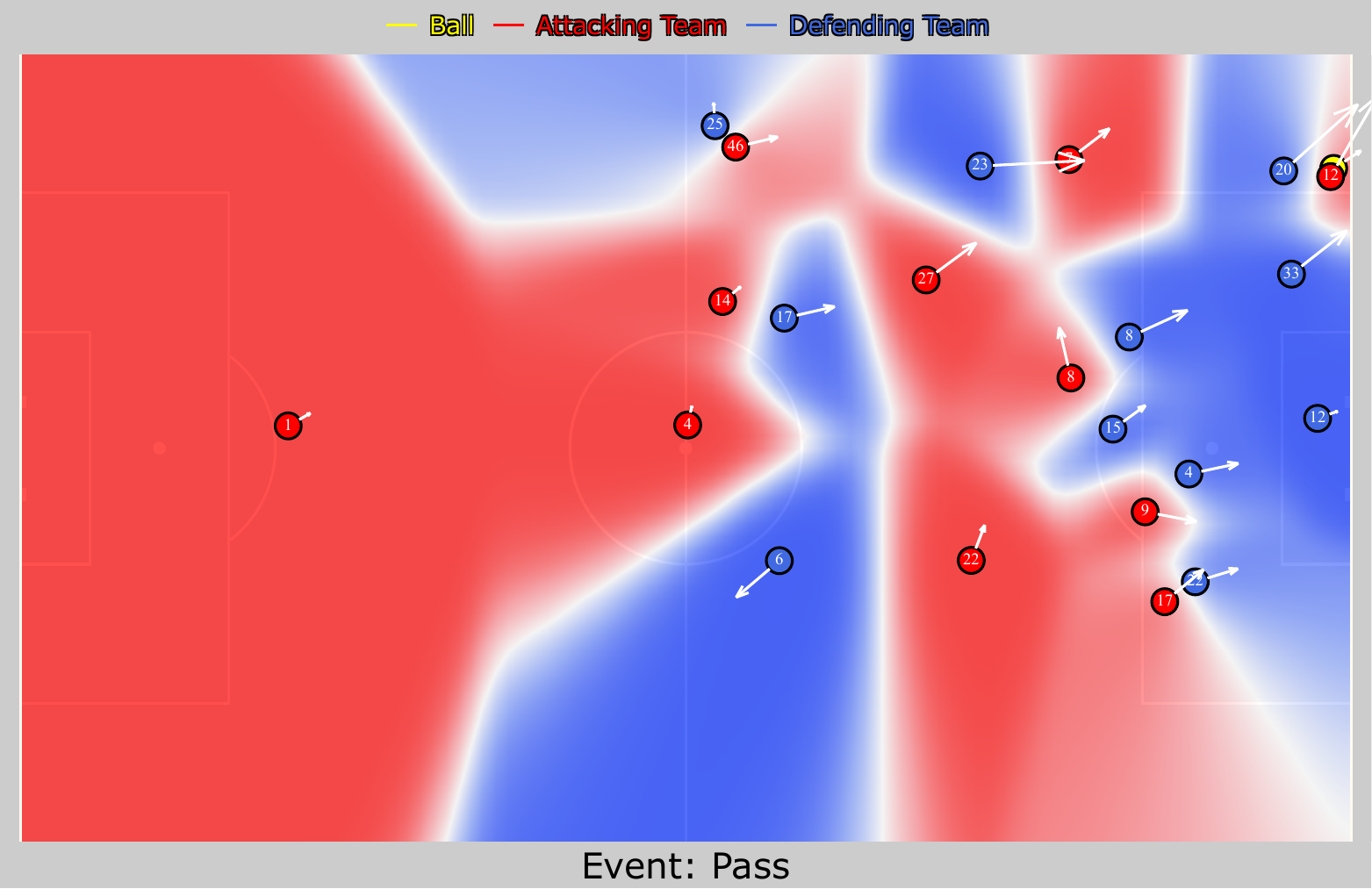}
  \end{minipage}

  \vspace{-0.1in}
  \caption{{Visualizations of trajectories (top) and pitch control values (PCV) at the final frame (bottom) generated by TacticGen for a pass event.} \textbf{(a)} Ground Truth. \textbf{(b)} Guided trajectories aimed at maximizing PCV for the attacking team. Notably, the two wingers and the central player move more rapidly to create additional space. \textbf{(c)} Guided trajectories aimed at maximizing PCV for the defending team. Notably, the blue control area expands, with the top-right defender pressing the ball carrier and the central defender turning to prepare for a possible counterattack.}\label{fig:fce-rule-guide-pcm-pass46}
  \vspace{-0.1in}
\end{figure}

In addition to the previously discussed functions, pitch control \cite{spearman2017pcm, spearman2018beyond, fernandez2018wide, martens2021space, higgins2023measuring} is a foundational tactical metric in football analytics, as it quantifies each team's spatial dominance across different regions of the pitch. In this context, we design guidance functions that steer player trajectories toward maximizing their own team’s pitch control value (PCV), computed following~\cite{fernandez2018wide}. Figure \ref{fig:fce-rule-guide-pcm-pass46} shows the generated trajectories alongside their corresponding PCV heatmaps. The results clearly demonstrate how the pitch control guidance influences spatial control movements, encouraging the attacking team to occupy strategically advantageous offensive zones while guiding the defending team to cover and reclaim key areas of the pitch.

\subsubsection{Natural Language-based Guidance with LLM}\label{sec:exp-llm}
In addition to manually designed rule-based guidance, TacticGen also supports natural language objectives. This feature is especially useful for users without professional knowledge of football tactics or programming skills. By leveraging Large Language Models (LLMs)~\cite{achiam2023gpt, jaech2024openai, singh2025openai}, TacticGen can automatically generate differentiable guidance functions from natural language inputs. This enables more flexible and diverse tactical behaviors that align closely with user intent expressed in natural language under diverse scenarios.

\begin{figure}[htbp]
\vspace{-0.05in}
  \centering

  \begin{minipage}[t]{0.16\textwidth}
    \centering
    \includegraphics[width=\linewidth]{figures/prediction/Pass_986528_1816044_46/gt/GroundTruth.pdf}
  \end{minipage}\hfill
  \begin{minipage}[t]{0.16\textwidth}
    \centering
    \includegraphics[width=\linewidth]{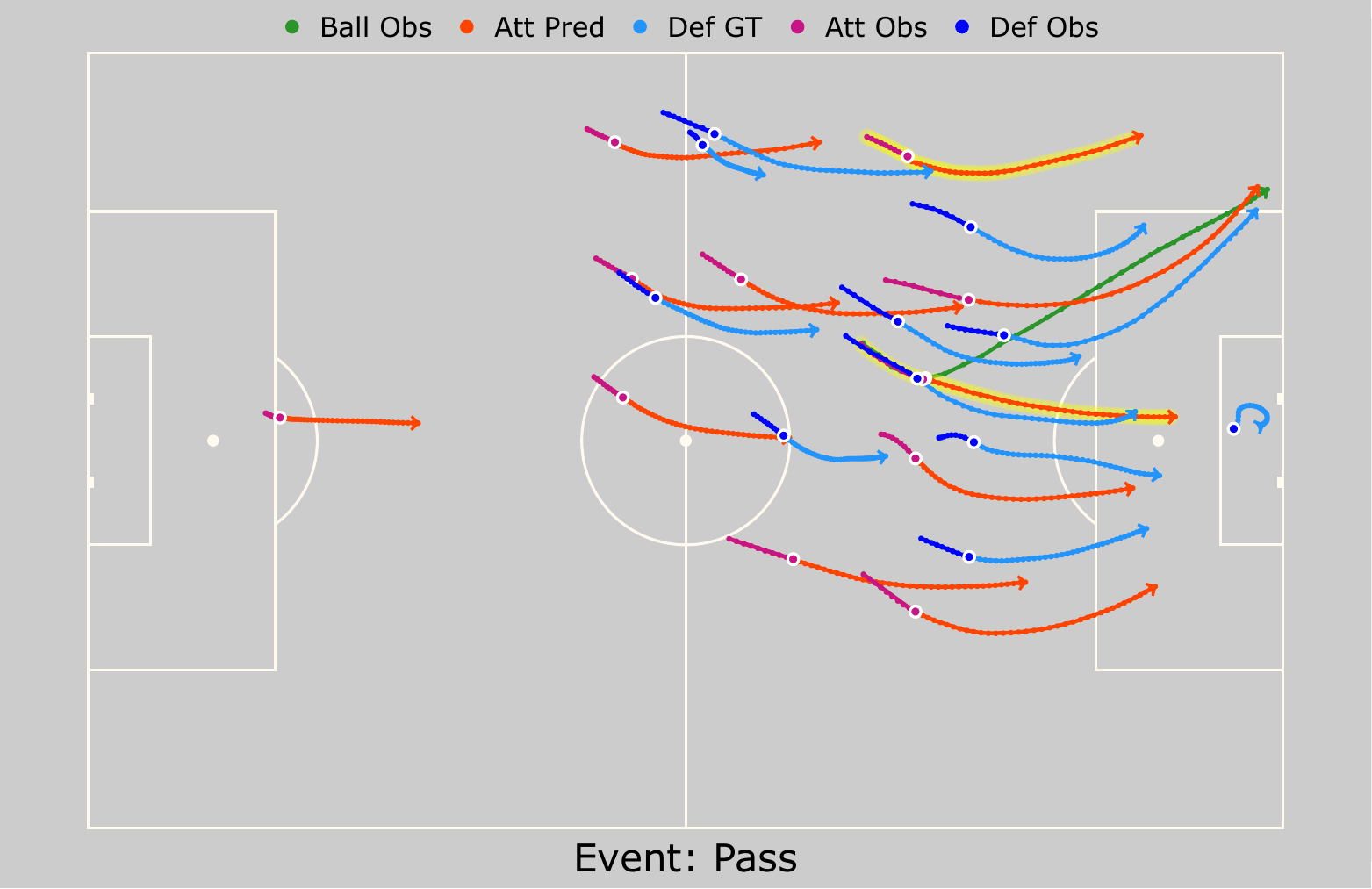}
  \end{minipage}\hfill
  \begin{minipage}[t]{0.16\textwidth}
    \centering
    \includegraphics[width=\linewidth]{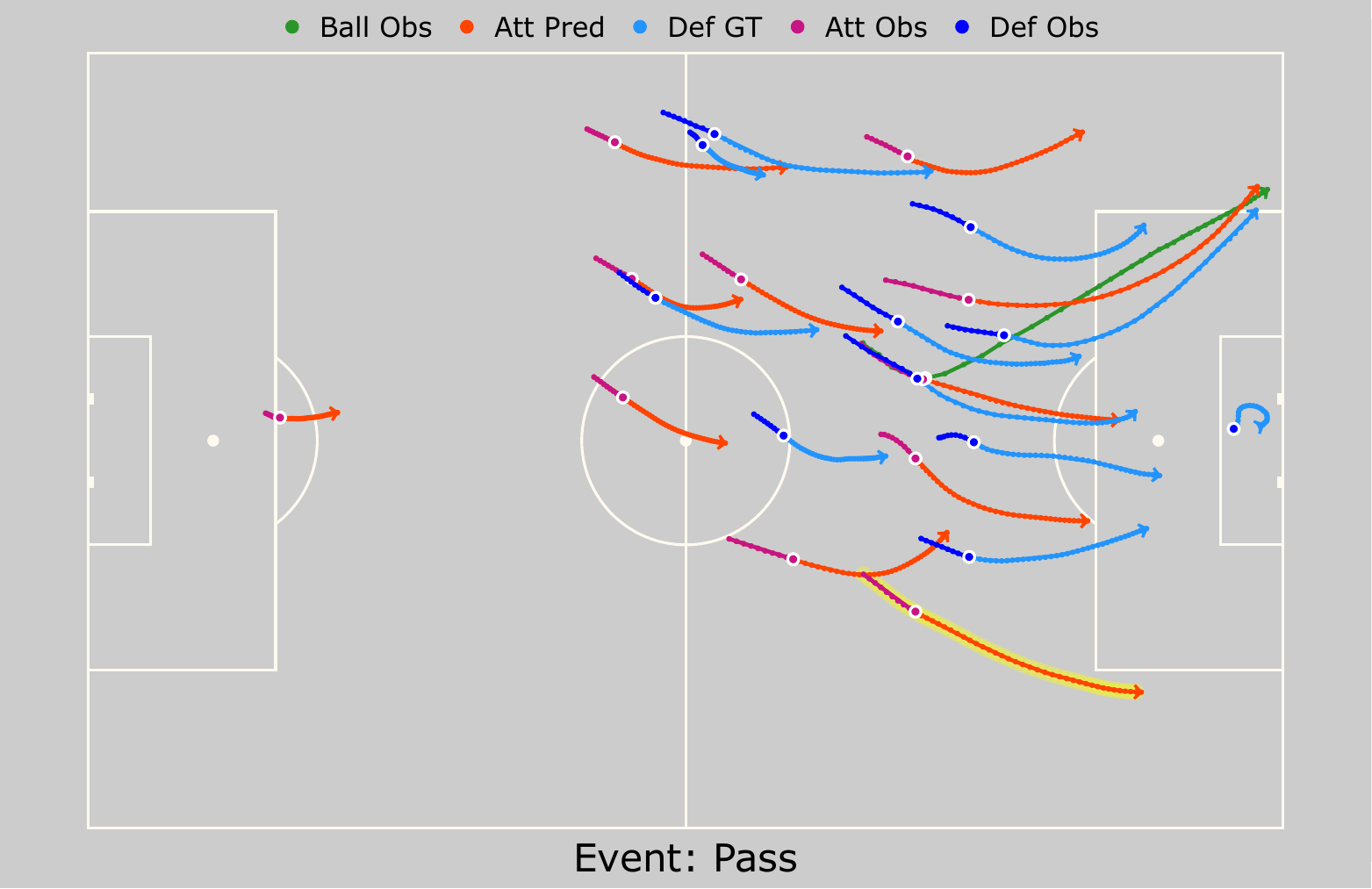}
  \end{minipage}

  \vspace{-0.1in}
  \caption{{Trajectories generated by TacticGen for a pass event under different guidance functions generated by LLM.} \textbf{Left} Ground Truth. \textbf{Middle} Guided generation with the prompt, “Make the attacking team move forward more aggressively.” Notably, the attacking players respond by increasing their speed and covering greater distances toward the defending goal. \textbf{Right} Guided generation with the prompt, “Make the right winger drift into the corner to stretch the defense and open up more space.” Clearly, the right-bottom wing follows the objective by running towards the corner.}
  \label{fig:fce-rule-guide-llm-pass46}
  \vspace{-0.05in}
\end{figure}

Figure~\ref{fig:fce-rule-guide-llm-pass46} showcases two examples where natural language objectives are provided to the LLM for generating guidance functions, which are then used for guided trajectory generation. These objectives can be specified at either the team level (middle fig.) or the player level (right fig.). The results show that the LLM accurately interprets language objectives and generates guidance functions that steer the trajectory generation toward the intended outcomes. The corresponding prompts and the LLM-generated functions are provided in Appendix~\ref{sec:llm-func-detail}.

\subsubsection{Learning-based Guidance with Value Model}
Although using differentiable guidance functions, e.g., pre-defined or generated by LLMs, is computationally efficient, they rely on explicitly defined rules or instructions that may not generalize well across different scenarios. To address this limitation, we train an auxiliary value model $V$ as defined in reinforcement learning~\cite{sutton1998reinforcement}, which estimates the expected future returns, i.e., the overall gains accumulated starting from the current trajectory. We model the value model as a neural network and train it using Monte Carlo estimates. Its gradient is then leveraged by TacticGen to guide trajectory generation toward outcomes with higher expected returns. 

\begin{wrapfigure}{r}{0.15\textwidth}
    \centering
    \vspace{-0.18in}
    \includegraphics[width=0.15\textwidth]{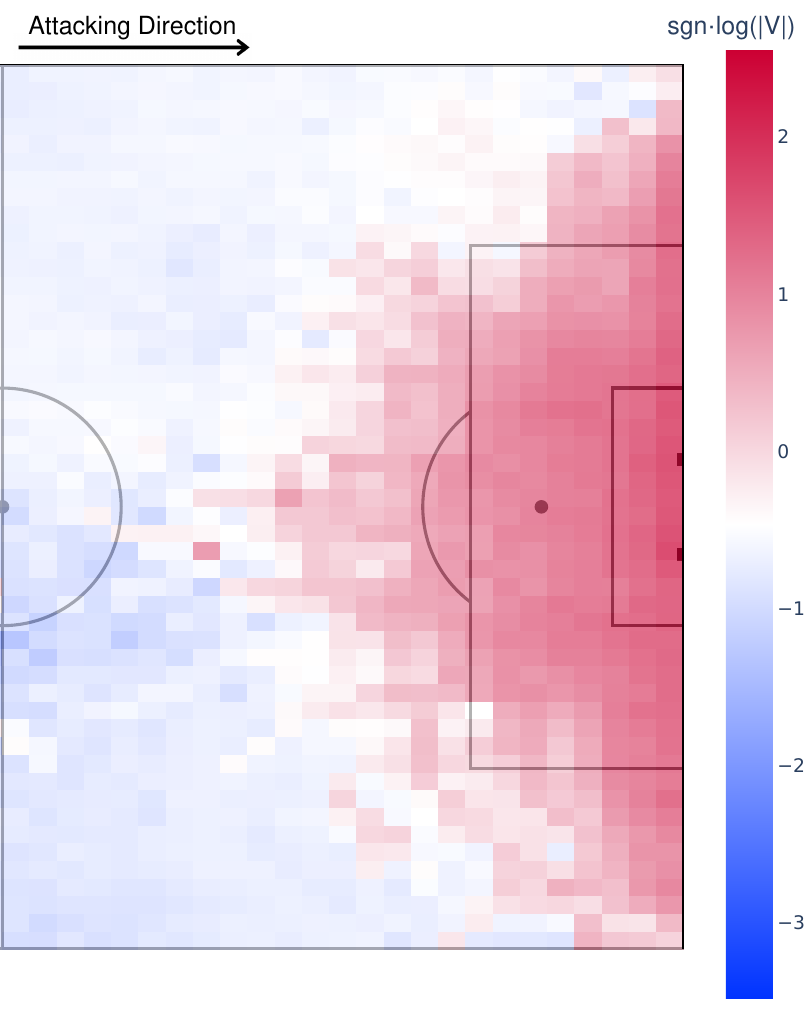}
    \vspace{-0.3in}
    \caption{{Log-scale value heatmap.}}\label{fig:value_heat_map}
    \vspace{-0.2in}
\end{wrapfigure}
Figure~\ref{fig:value_heat_map} shows a log-scale heatmap of the learned value model evaluated on the test set, focusing on the right half of the pitch. Warmer regions indicate higher value, highlighting the penalty-box area as the most critical region. Cooler regions that are relatively far from the penalty box yield lower long-term payoff. This visualization underscores how the value model has learned to prioritize locations most likely to lead to scoring opportunities.

\begin{figure}[htbp]
  \centering

  \begin{minipage}[t]{0.16\textwidth}
    \centering
    \small \textbf{(a)} Ground Truth\\
    \includegraphics[width=\linewidth]{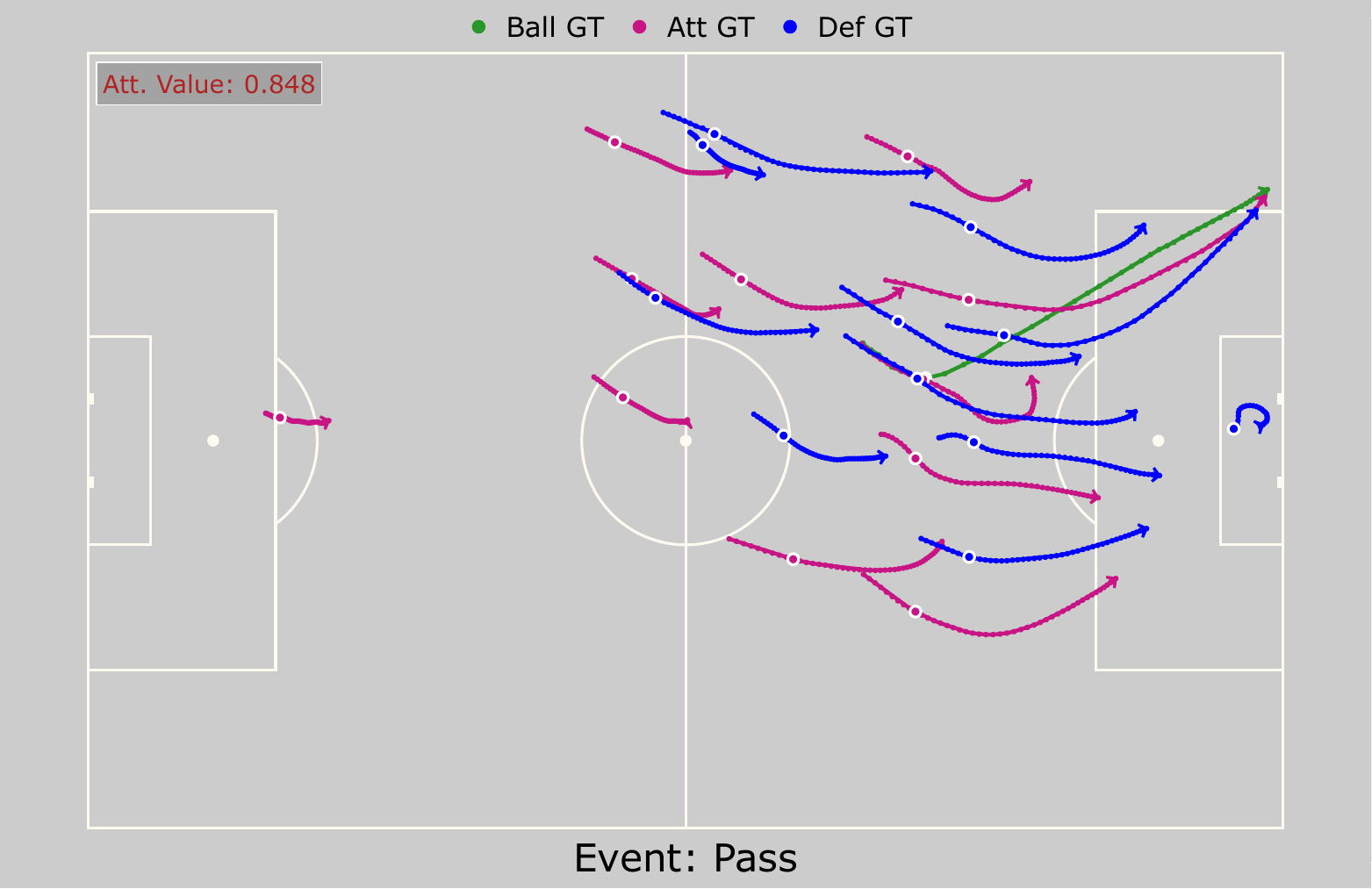}
  \end{minipage}\hfill
  \begin{minipage}[t]{0.16\textwidth}
    \centering
    \small \textbf{(b)} Att. High $V$\\
    \includegraphics[width=\linewidth]{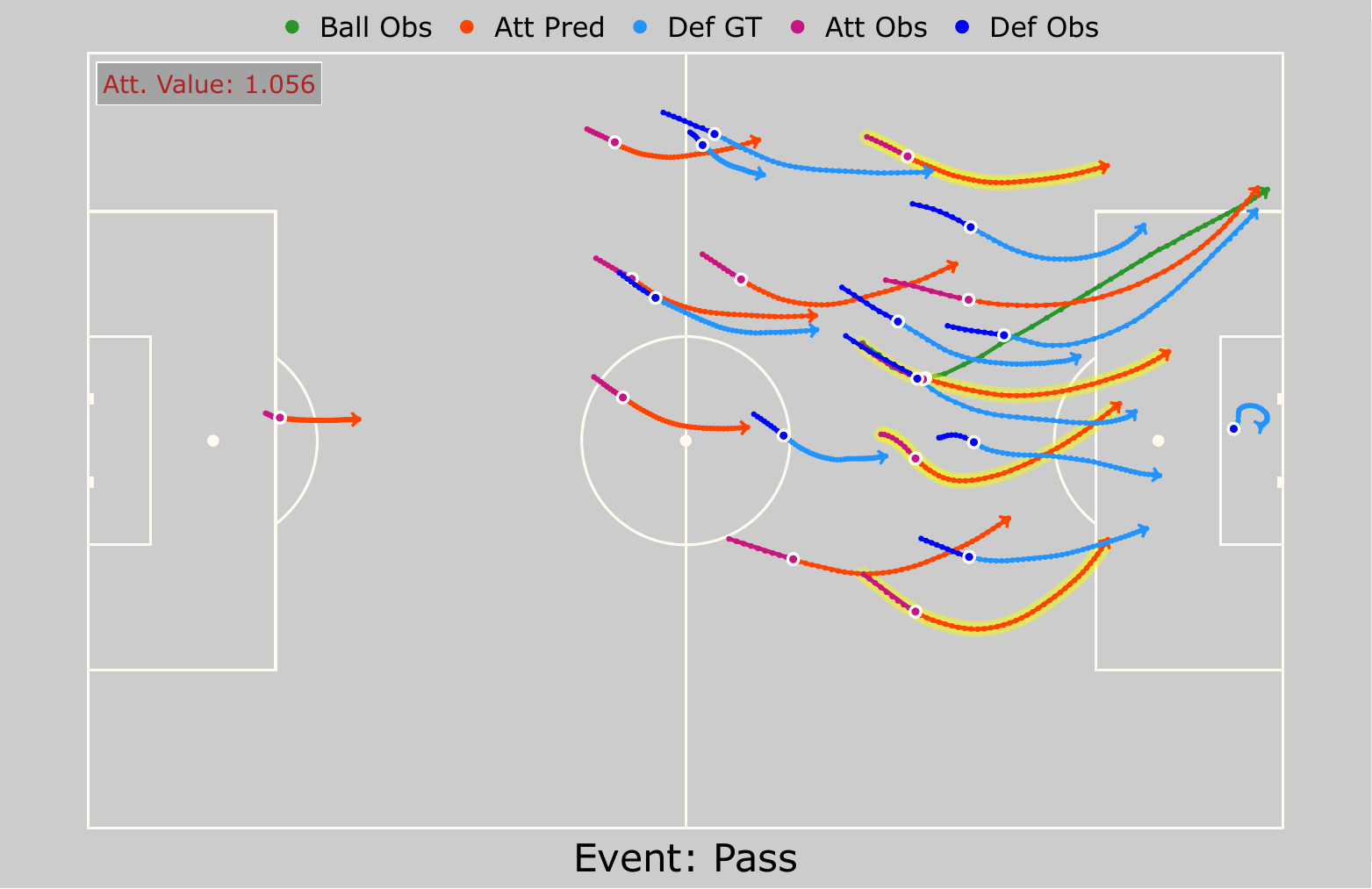}
  \end{minipage}\hfill
  \begin{minipage}[t]{0.16\textwidth}
    \centering
    \small \textbf{(c)} Def. High $V$\\
    \includegraphics[width=\linewidth]{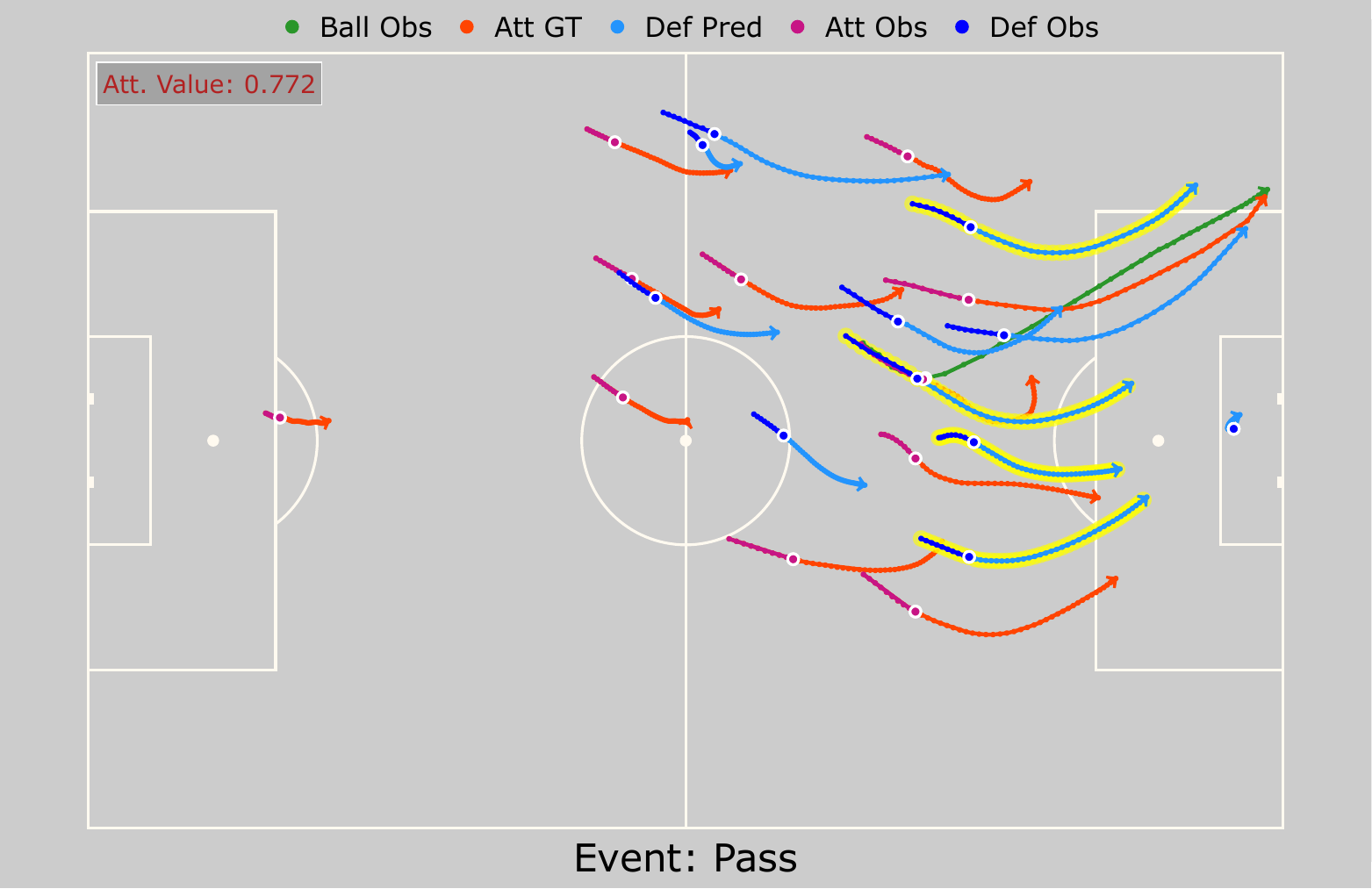}
  \end{minipage}
  \caption{{Visualizations of trajectories generated by TacticGen for a pass event.} \textbf{(a)} Ground Truth. \textbf{(b)} Guided trajectories aimed at maximizing the $V$ value for the attacking team. Notably, the attacking players increase their speed to push the team forward, bringing them closer to the defending area and potentially increasing the probability of scoring. \textbf{(c)} Guided trajectories aimed at maximizing the $V$ value for the defending team (i.e., minimizing the attacking value). Notably, the top-right defender accelerates toward the ball to press the carrier, yielding a higher estimated value than the ground-truth.}
  \label{fig:fce-rule-guide-value-pass46}
  \vspace{-0.05in}
\end{figure}

The middle and right panels of Figure~\ref{fig:fce-rule-guide-value-pass46} depict trajectories generated by TacticGen for a pass event, guided by the learned value model for the attacking and defending teams, respectively. When guiding the attacking team, the objective is to maximize the attacking value, whereas for the defending team, the objective is to minimize it. The results demonstrate the effectiveness of the learned value model in producing gradients that guide tactic generation toward higher-value outcomes.

\subsection{Scalable Generalization to Diverse Model Configurations}\label{sec:scaling-law}
To assess the scalability of TacticGen, we analyze its behavior by systematically scaling key factors, including \textit{model size}, \textit{training steps}, and \textit{training data capacity}, following established methods~\cite{peebles2023scalable, bahri2024explaining}. Such analysis is crucial for developing a high-capacity tactical generation model tailored to football. We adopt TacticGen-P and exclude its context encoder, as this variant best represents the model’s core architecture. We evaluate its predictive ability since it forms the foundation of tactic generation, with stronger prediction indicating a better capture of agent movement patterns.

The backbone of TacticGen consists of a sequence of multi-agent DiT blocks. Each block operates with a hidden dimension and a specified number of attention heads for the multi-head attention mechanism. Following the design principles established in DiT~\cite{peebles2023scalable}, we adopt five model configurations: Small (S), Base (B), Large (L), XLarge (XL), and XXLarge (XXL). These configurations jointly scale the embedding and hidden dimensions, number of layers, and attention heads. They span a wide range of model capacities, from 1.74 million to 311.50 million parameters, enabling a comprehensive evaluation of TacticGen's scaling behavior with respect to model size. Appendix~\ref{sec:hyperparameters} and~\ref{sec:scale-config-detail} give more details of the hyperparameters and model configurations.

To investigate the effects of training data size and training steps, we fix the test set as before (around 20 million examples) and divide the training set (around 78 million examples) into six levels, increasing in steps of 13 million examples. Each level is constructed by incrementally sampling from the full training pool. For each training data volume, we train TacticGen for up to 600K steps, evaluating its performance every 100K steps using the same test set. This setup enables a systematic evaluation of performance as the training data and steps increase. We train all models using the AdamW optimizer~\cite{loshchilov2017decoupled}, with a learning rate of $3 \times 10^{-5}$ and a batch size of 512. Denoising steps are set to 20. Scaling performance is evaluated using Joint Average Displacement Error (Joint ADE) and Joint Final Displacement Error (Joint FDE), which are widely considered the key metrics for multi-agent trajectory prediction~\cite{weng2023joint, ngiam2021scene, cheng2023gatraj, sun2022m2i}.

\begin{figure}[htbp]
    \centering
    \begin{minipage}{0.25\textwidth}
        \centering
        \includegraphics[width=\linewidth]{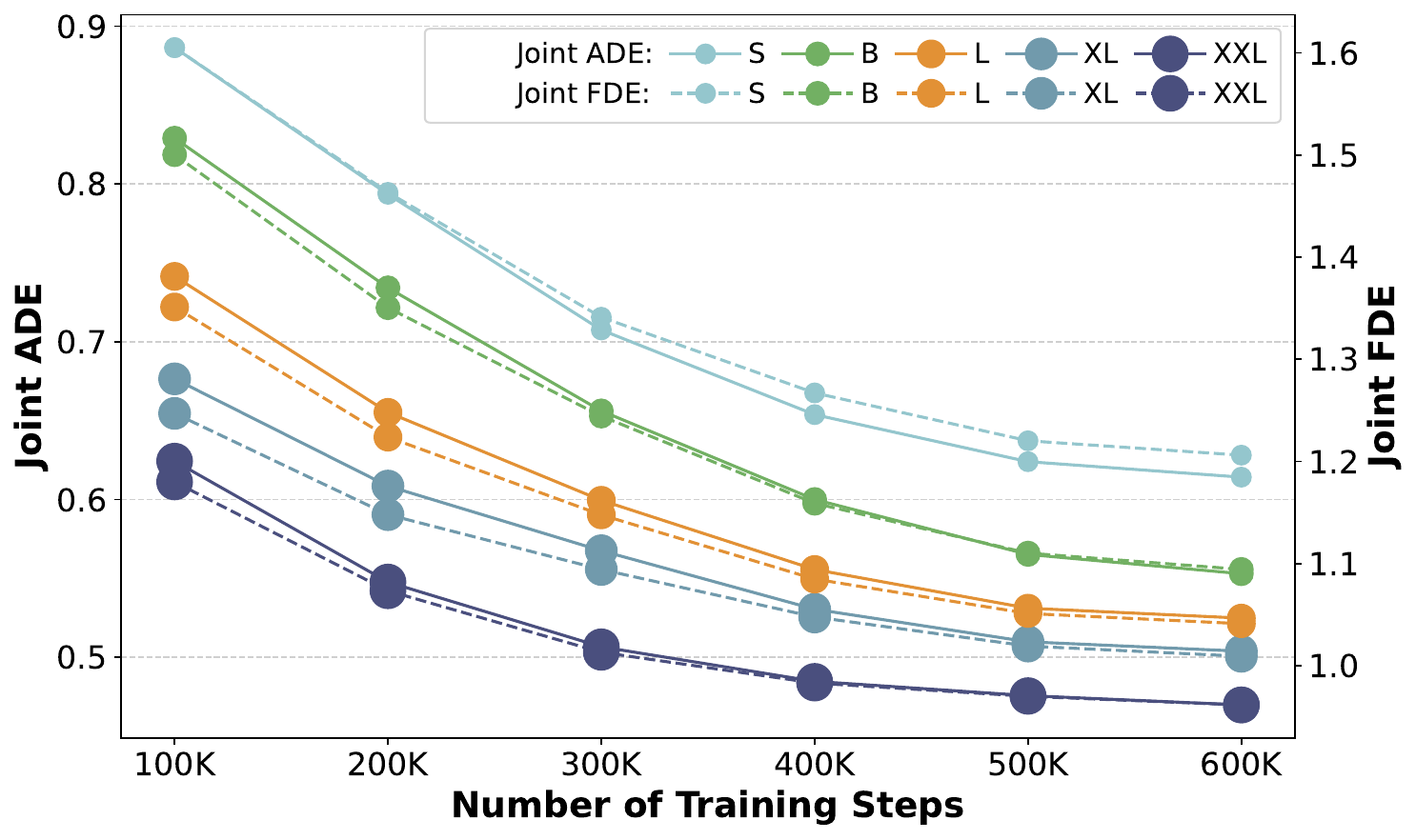}
    \end{minipage}%
    \begin{minipage}{0.25\textwidth}
        \centering
        \includegraphics[width=\linewidth]{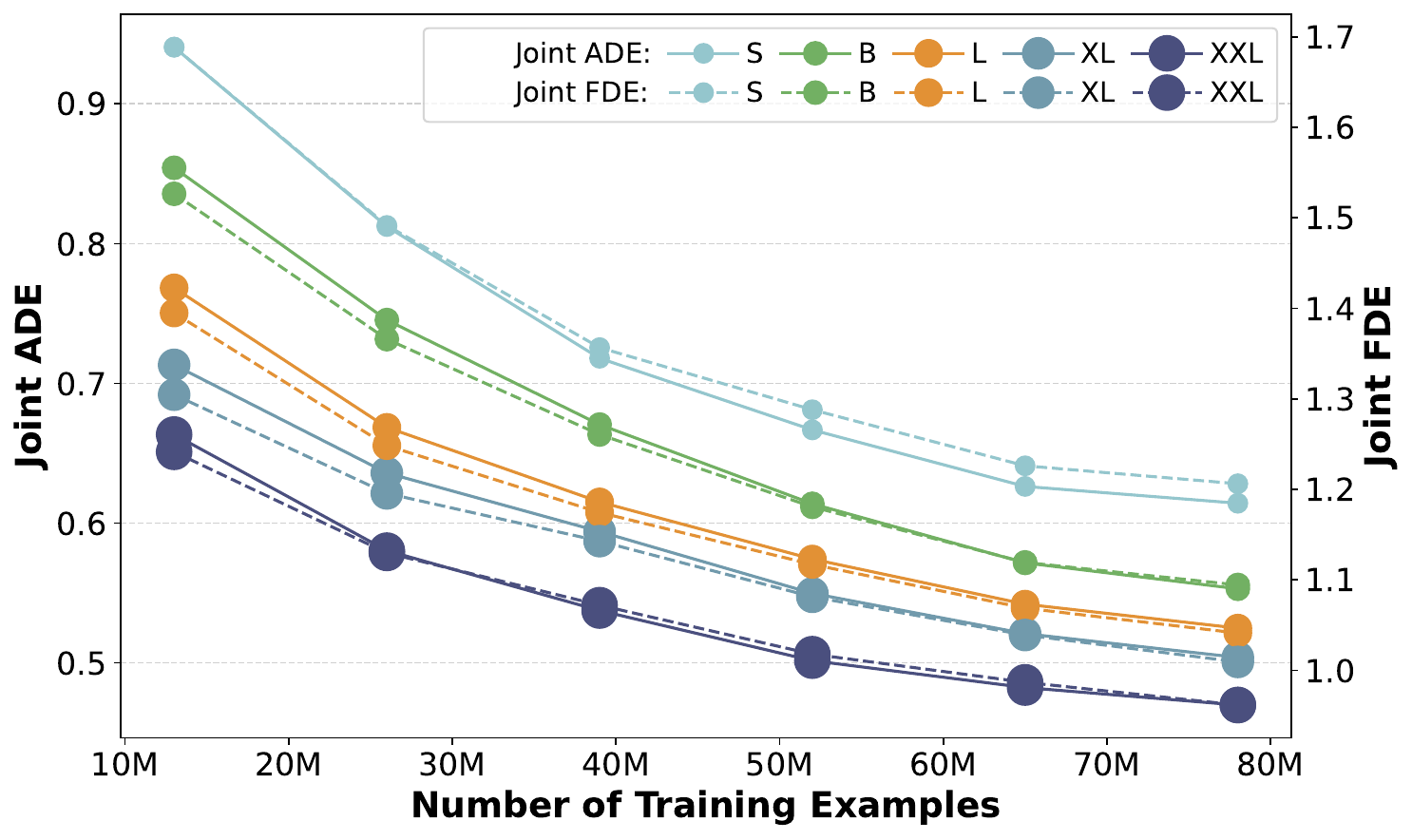}
    \end{minipage}
    
    \caption{{Scaling performance of TacticGen across different model sizes.} \textbf{Left} Performance trends over training steps using the full training dataset. \textbf{Right} Performance across varying amounts of training data, with each model trained for 600K steps. Joint ADE (solid lines) and Joint FDE (dashed lines) are used to assess the quality of multi-agent trajectory prediction.}\label{fig:scaling-law}
    \vspace{-0.05in}
\end{figure}

\textbf{Scalability to Model Size.} As shown in Figure~\ref{fig:scaling-law}, we present the evaluation performance for five model configurations with respect to training steps (Left) and number of training examples (Right). The left panel shows that, when trained on the full dataset, larger models consistently outperform smaller ones across the entire training process, maintaining a clear performance gap in both JADE (solid lines) and JFDE (dashed lines). The right panel highlights a similar scaling trend when evaluating the final performance of different models with respect to the amount of training data, with larger-capacity models achieving lower errors. The consistent performance improvements observed with larger model sizes suggest that there remains some space for scaling. Further gains can be expected by increasing model capacity in proportion to available training data and computing resources. This scaling potential underscores the strength of TacticGen’s architecture, with the scalable generation capacity to advance the frontier of multi-agent trajectory generation in football.

\textbf{Scalability to Training Steps.} Figure~\ref{fig:scaling-law} (Left) presents evaluation performance across models of varying parameter sizes, trained on the full dataset for up to 600K steps. For all sizes, both Joint ADE and Joint FDE decrease steadily, demonstrating consistent performance gains. Substantial improvements emerge during the early rounds (before 300K steps), indicating rapid learning of core predictive capabilities. Beyond this point, progress slows, and by around 500K steps, the curves plateau, indicating diminishing returns as models approach convergence. This trend highlights that while extended training yields some benefits, marginal gains taper in later stages, underscoring the need to balance training duration against performance improvements.

\textbf{Scalability to Data Capacity.} Figure~\ref{fig:scaling-law} (Right) shows evaluation performance for models of varying sizes, each trained for 600K steps on different scale datasets. Across all models, both Joint ADE and Joint FDE decrease as the number of training examples increases, indicating that more data enables better modeling of player movement and subsequent motion prediction. For the smallest model, TacticGen-S (1.74M parameters), gains plateau when expanding the dataset from 65M to 78M examples, suggesting limited capacity to exploit additional data. Larger models, however, continue to benefit, achieving measurable improvements even with the full dataset. This pattern underscores the superior data efficiency of larger models and their capacity to absorb richer training signals when sufficient data is available. It further suggests that TacticGen’s performance could be enhanced by scaling up model size alongside access to more extensive football data.

\subsection{Case Study with Experts}\label{sec:case-study}
While the preceding experiments demonstrate the strong capabilities of TacticGen in generating football tactics, the ultimate goal is to ensure that its practical utility is recognized by professionals within the football domain and industry. To achieve this, we conducted a case study in collaboration with our partners from the football sector. We invited five football experts to participate in the case study: three data analysts, one former professional player, and one professor specializing in football analytics. Each expert has over 10 years of experience. To facilitate evaluation, we presented the trajectories as simulated video clips and asked the experts to complete two tasks assessing the \textit{realism} and \textit{utility} of the trajectories generated by TacticGen. All samples are drawn from the test set, and results are reported as mean values with corresponding standard deviations. TacticGen-C is adopted as the model for this experiment. Please refer to Appendix~\ref{sec:case-study-detail} for more details about the case study.

\subsubsection{Case Study on Realism}
We evaluated the \textit{realism} of the trajectories generated by TacticGen. Following the evaluation protocol in~\cite{wang2024tacticai}, we collected 50 realistic trajectories and 50 generated by TacticGen, resulting in a total of 100 samples. Experts were then asked to decide whether a given sample was real or generated. The results showed that the average $F_1$ score among the raters for distinguishing between real and generated trajectories was only $0.50 \pm 0.07$. Individual $F_1$ scores were as follows: $F_1^A=0.42$, $F_1^B=0.62$, $F_1^C=0.49$, $F_1^D=0.46$, $F_1^E=0.51$. These results indicate that TacticGen-generated trajectories are highly realistic and often indistinguishable from real ones, even by domain experts, demonstrating the model’s ability to capture authentic patterns of player movements.

\begin{figure}[htbp]
\vspace{-0.05in}
    \centering
    \begin{minipage}{0.25\textwidth}
        \centering
        \includegraphics[width=\linewidth]{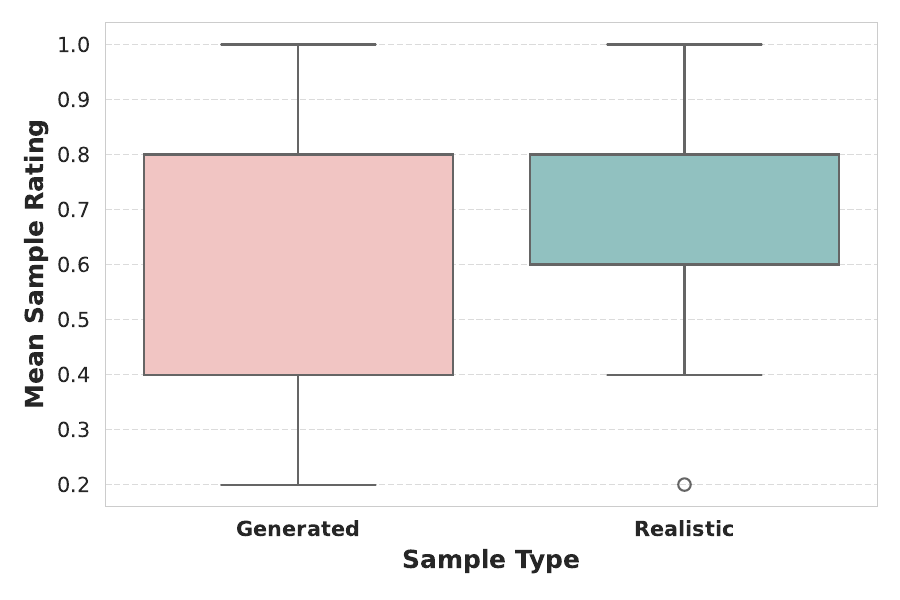}
    \end{minipage}%
    \begin{minipage}{0.25\textwidth}
        \centering
        \includegraphics[width=\linewidth]{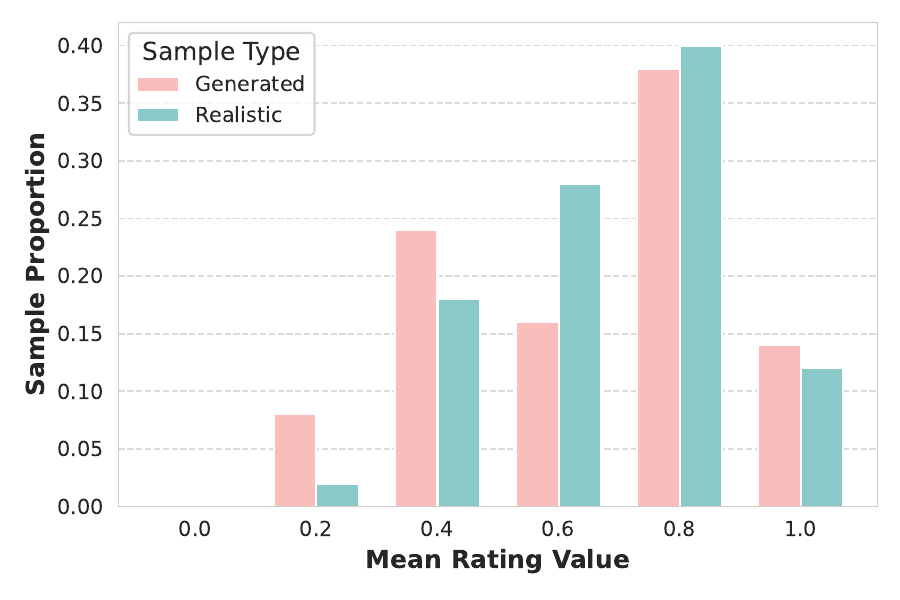}
    \end{minipage}
    
    \caption{{Results of the case study on realism assessment.} \textbf{Left} Distribution of ratings assigned to generated and realistic samples. \textbf{Right} Corresponding histograms showing the distribution of rating values. No statistical difference in the mean was observed in the case ($z=-0.49$, $p>0.05$).}\label{fig:feedback-realism}
    \vspace{-0.05in}
\end{figure}

Following~\cite{wang2024tacticai}, we further conducted an analysis on the realism score. Specifically, each sample was assigned a score of $+1$ if a human rater identified it as real, and $0$ otherwise. The average rating for each sample was then computed across all five raters. The results are presented in Figure~\ref{fig:feedback-realism}. The scores for generated and realistic samples were $0.65 \pm 0.22$ and $0.68 \pm 0.20$, respectively, with both having a median score of $0.8$. Notably, statistical analysis of the rating distributions revealed no significant difference between the average ratings assigned to generated and realistic samples ($z = -0.49$, $p > 0.05$). This suggests that, on average, human raters perceived the generated and realistic trajectories as statistically indistinguishable in terms of realism.

Additionally, we gathered feedback from the experts after their evaluations. All five experts reported that the task was challenging and that distinguishing between the generated and realistic samples was difficult, even for the expert with the highest $F_1$ score of $0.62$. Interestingly, this data analyst commented, “The task was challenging; I had to watch most clips multiple times to make a decision,” and the ex-professional footballer remarked, “It's challenging; I'm trying to imagine myself as the passer in the game, assessing the movement in front of me to determine if it looks realistic!”

\subsubsection{Case Study on Utility}

We evaluated the \textit{utility} of the trajectories generated by TacticGen. Specifically, we collected 25 pairs of trajectories, each consisting of ground-truth player movements and a corresponding version generated by TacticGen with guidance. The true labels were hidden, and the display order of the two trajectories in each pair was randomly shuffled to prevent bias. To quantify expert preferences, we adopt a win odds metric\footnote{\url{https://en.wikipedia.org/wiki/Winning_percentage}}. Specifically, for each trajectory pair, experts indicate whether the generated trajectory, the ground truth, or neither demonstrates superior tactical quality. A score of $1$ is assigned if the generated trajectory is preferred, $0$ if the ground truth is preferred, and $0.5$ in the case of equal preference. The final score is computed by averaging across all pairs, producing a value in the range $[0, 1]$ that quantifies the probability of TacticGen-generated trajectories being preferred to the ground truth.

\begin{figure}[htbp]
\vspace{-0.05in}
    \centering
    \begin{minipage}{0.25\textwidth}
        \centering
        \includegraphics[width=\linewidth]{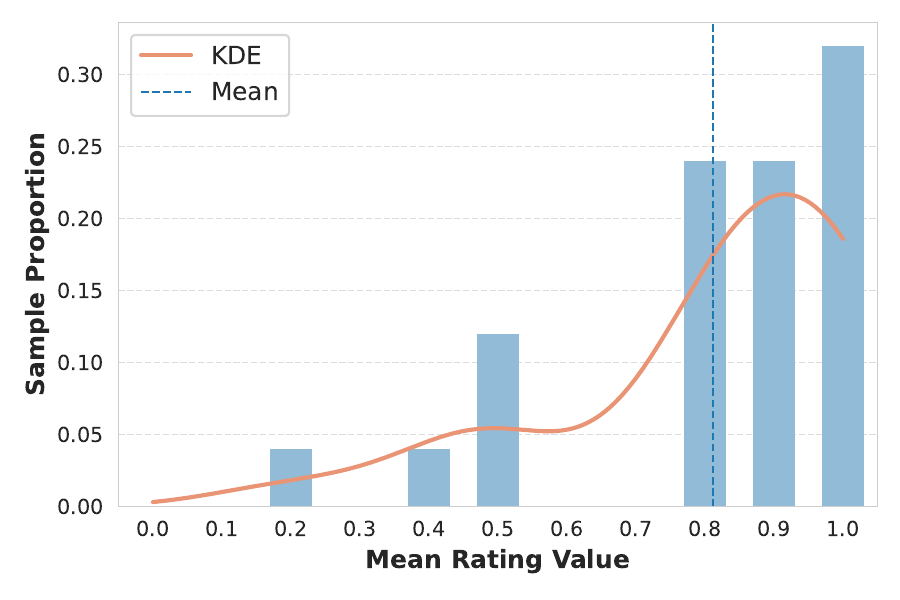}
    \end{minipage}%
    \begin{minipage}{0.25\textwidth}
        \centering
        \includegraphics[width=\linewidth]{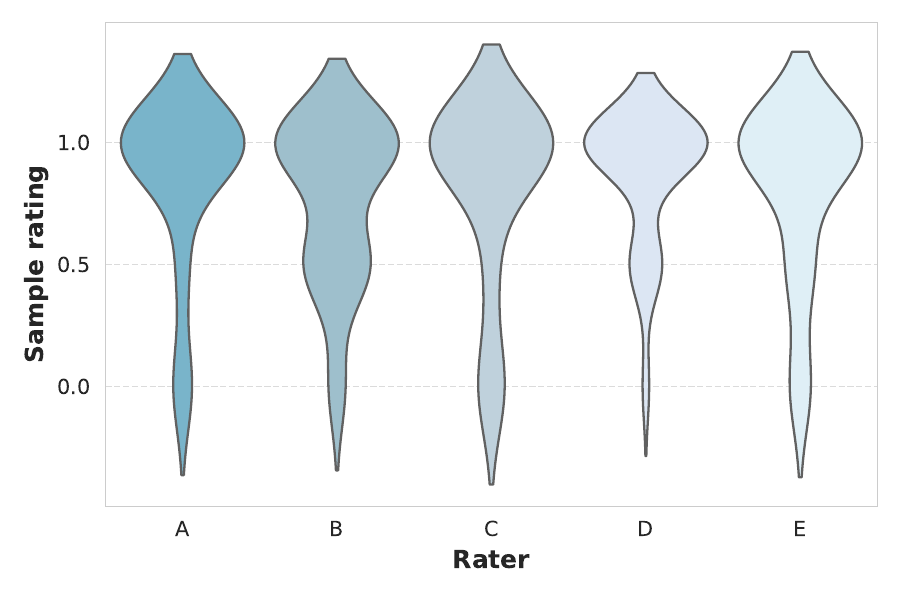}
    \end{minipage}
    
    \caption{{Results of the case study on utility assessment.} 
    \textbf{Left} Histogram of the distribution of mean rating values for the generated samples with kernel density estimate (KDE). 
    \textbf{Right} Ratings of the generated samples across individual human raters. 
    The raters showed general agreement on the effectiveness of the tactics generated by TacticGen ($F_{4,96}=0.47, p>0.05$). 
    Note that a continuous probability distribution is used in the plot, which assigns small, nonzero probabilities to values outside the discrete rating range. 
    For clarity, only the valid rating values are shown on the y-axis.}
    \label{fig:feedback-utility}
\end{figure}

We visualize the average rating of generated samples across the five raters in Figure~\ref{fig:feedback-utility} (Left). The overall mean score was $0.81 \pm 0.04$, with individual scores as follows: $S_1^A = 0.84$, $S_1^B = 0.76$, $S_1^C = 0.80$, $S_1^D = 0.86$, and $S_1^E = 0.80$. These results reflect a strong and consistent preference for the trajectories generated by TacticGen over the ground truth. Furthermore, TacticGen achieved an average score strictly greater than 0.5 in 20 of the 25 pairs (80\%), indicating that its trajectories were tactically favored over the ground truth. In addition, in 80\% of the pairs, a majority of raters (at least three out of five) preferred the generated trajectories. This finding underscores TacticGen’s ability to generate strategic principles in football game-play, highlighting its practical value for tactical recommendation and decision support.

To further validate the results, we conducted statistical significance tests on the observed ratings greater than 0.5 (as good as real tactics). For each of the 25 scenarios, ratings from the five experts were averaged and subjected to a one-sample $t$-test to verify whether the
mean rating was significantly larger than 0.5. The results revealed that TacticGen’s generated player movements were overall constructive, with $t^{\text{avg}}_{24}=7.08, p<0.001$. We then examined each rater individually by applying the $t$-test to their ratings. Figure~\ref{fig:feedback-utility} (Right) visualizes the ratings. All five raters reported mean scores significantly greater than 0.5: $t_{24}^A=4.92, p^A<0.001$; $t_{24}^B=3.98, p^B<0.001$; $t_{24}^C=3.93, p^C<0.001$; $t_{24}^D=6.65, p^D<0.001$; and $t_{24}^E=4.24, p^E=0.001$. These results confirm the constructive effect of TacticGen’s recommendations. Furthermore, the raters’ evaluations exhibited strong consistency, as indicated by repeated-measures ANOVA ($F_{4,96}=0.47, p>0.05$). This suggests that, despite their diverse professional backgrounds, the experts largely agreed on the practical utility of TacticGen’s generated tactics. In Appendix~\ref{sec:vis-tactic}, we provide two illustrative cases in which all five experts agreed that the tactics generated by TacticGen outperformed the ground truth.

\section{Conclusion}
In this work, we have demonstrated the effectiveness of TacticGen, the first generative model for football tactics that is accurate, adaptable, and scalable. Beyond outperforming state-of-the-art trajectory prediction methods, TacticGen addresses the central challenge of tactic generation, which is to move from merely predicting player movements to actively designing coordinated strategies that achieve specific objectives. Our results show that TacticGen not only produces realistic and tactically meaningful behaviors across diverse goals but also adheres to established scaling laws and proves its practical value through expert validation. By enabling objective-driven tactical design, TacticGen marks a fundamental shift in football analytics, advancing from passive prediction toward active data-driven support for strategic planning and decision-making in professional football.

Looking forward, several extensions could further enhance the capabilities of TacticGen. Benefiting from the scalability of TacticGen, future developments could broaden its scope by incorporating larger and more diverse datasets from additional football leagues and other sports domains. Such expansion would enable the model to generalize across a wider range of strategic contexts. Another promising direction is the integration of multi-modal signals, such as broadcast video, together with richer player metadata (e.g., height, weight, fatigue) and more detailed semantic roles (e.g., goalkeeper, fullback, striker). These additions would provide a deeper understanding of player dynamics and tactical behavior, allowing the model to better capture player representations.

\section{Practical Impact}
TacticGen provides substantial practical value for both in-match decision-making and post-match analysis. During inference, the model predicts the subsequent 5.4 seconds of player trajectories from the preceding 1 second of input in approximately 0.5 seconds on an RTX 5090 GPU, leveraging FP8 precision and tensor core acceleration. In practice, the inference time can be further reduced to around 0.2 seconds by using fast
sampling techniques~\cite{nichol2021improved,song2020denoising,lu2022dpm}. This efficiency makes TacticGen-P suitable for in-match applications, as it can generate the movements of both teams efficiently and assist coaches in making tactical adjustments during critical moments of the game. By producing coordinated player movement sequences aligned with specific tactical objectives, TacticGen enables efficient decision support and facilitates the dynamic refinement of strategies during matches. In post-match analysis, TacticGen serves as a powerful “what-if” tactical simulator for professional football, offering in-depth strategic planning and evaluation. The system is flexible, allowing coaches and analysts to choose between TacticGen-P or TacticGen-C depending on data availability or usage requirements. This flexibility enables the isolation of critical moments and the generation of optimized sequences of coordinated movements aligned with specific tactical goals. More importantly, TacticGen transforms traditional post-match review into structured counterfactual analysis, allowing practitioners to explore not just what happened but also what could have happened under alternative strategic intentions. Bridging descriptive analytics with prescriptive decision support, TacticGen empowers clubs to simulate, compare, and refine tactical solutions before implementing them in future competitions.

\bibliography{references}
\bibliographystyle{IEEEtran}

 




\clearpage
\appendices

\renewcommand{\thefigure}{S\arabic{figure}}
\renewcommand{\thetable}{S\arabic{table}}
\renewcommand{\theequation}{S\arabic{equation}}
\setcounter{figure}{0}
\setcounter{table}{0}
\setcounter{equation}{0}

\renewcommand{\thesubsectiondis}{\thesection.\arabic{subsection}}
\renewcommand{\thesubsubsectiondis}{\thesubsectiondis.\arabic{subsubsection}}

\renewcommand{\thesubsection}{\thesection.\arabic{subsection}}
\renewcommand{\thesubsubsection}{\thesubsection.\arabic{subsubsection}}

\section{Football Dataset}\label{sec:dataset}
\subsection{Data Composition}\label{sec:data-detail}
We utilize two primary types of data: \textbf{(1) Event data}, or play-by-play data, which are manually annotated by professional Opta analysts\footnote{\url{https://www.statsperform.com/opta/}} and provide time-stamped logs of in-game actions. Each log entry records the action type (e.g., shot, pass, tackle, etc.), a time stamp, $(x,y)$ coordinates of the ball at the time of the event, and some further descriptors that give further context to the on-ball event, such as the type of pass or the body part used to execute the action. \textbf{(2) Tracking data}, obtained through an optical tracking system and collected by the official provider of each league, which records the $(x, y)$ positions of all players and the ball 25 times per second. Positional data are mapped onto a standardized pitch measuring $105 \times 68$ meters, with the origin $(0,0)$ located on the flag of the bottom left corner and down-sampled to 10 frames per second. For consistency, agent coordinates are flipped so that the attacking team always scores on the right target (center of the goal located at $(105,34)$). Player coordinates are normalized by centering them at the pitch midpoint and scaling them with predefined factors, mapping positions to $[-1, 1]$. This normalization stabilizes training and improves spatial interaction learning. 

\begin{table}[htbp]
  \centering
  \caption{{Summary of the information in each event.}}\label{tab:dataset}
  \renewcommand{\arraystretch}{1.0}
  \begin{tabular}{p{0.15\textwidth} p{0.3\textwidth}}
    \toprule
    \textbf{Field} & \textbf{Description} \\
    \midrule
    \texttt{event\_metadata} & General event metadata, including game ID, event ID, episode ID, etc. \\
    \rowcolor{gray!20}
    \texttt{global\_feature} & Global features at event level, including goal difference, outcome, possession length, and whether the team controlling the ball. \\
    \texttt{time\_to\_event} & Time in seconds until the next event. \\
    \rowcolor{gray!20}
    \texttt{action} & Unified action name, chosen from a predefined set of 30 labels. \\
    \texttt{action\_destination} & Coordinates of the ball at the time of the next event. \\
    \rowcolor{gray!20}
    \texttt{is\_home\_action} & Binary indicator of whether the action was performed by the home team (1) or the away team (0). \\
    \texttt{is\_attacking\_action} & Binary indicator of whether the action belongs to an attacking action (1) or a defending action (0). \\
    \rowcolor{gray!20}
    \texttt{home\_reward} & Immediate reward assigned to the home team for taking the current action. \\
    \texttt{away\_reward} & Immediate reward assigned to the away team for taking the current action. \\
    \rowcolor{gray!20}
    \texttt{done} & Boolean indicating whether the current episode has terminated. \\
    \texttt{context\_positions} & Agent x/y positions over the fixed past timesteps up to the current event. \\
    \rowcolor{gray!20}
    \texttt{context\_features} & Agent context features including side, jersey, visibility, and involvement. \\
    \texttt{trajectory\_positions} & Agent x/y positions over a variable-length trajectory from the current event to the subsequent event. \\
    \rowcolor{gray!20}
    \texttt{trajectory\_features} & Agent trajectory features (same as \texttt{context\_features}). \\
    \bottomrule
  \end{tabular}
\end{table}

The events and tracking data were aligned using an adaptation of the Needleman-Wunsch algorithm \cite{marekblog2020}. After processing, each event includes the organized information summarized in Table~\ref{tab:dataset}.

\begin{figure}[htbp]
    \centering
    \includegraphics[width=0.5\textwidth]{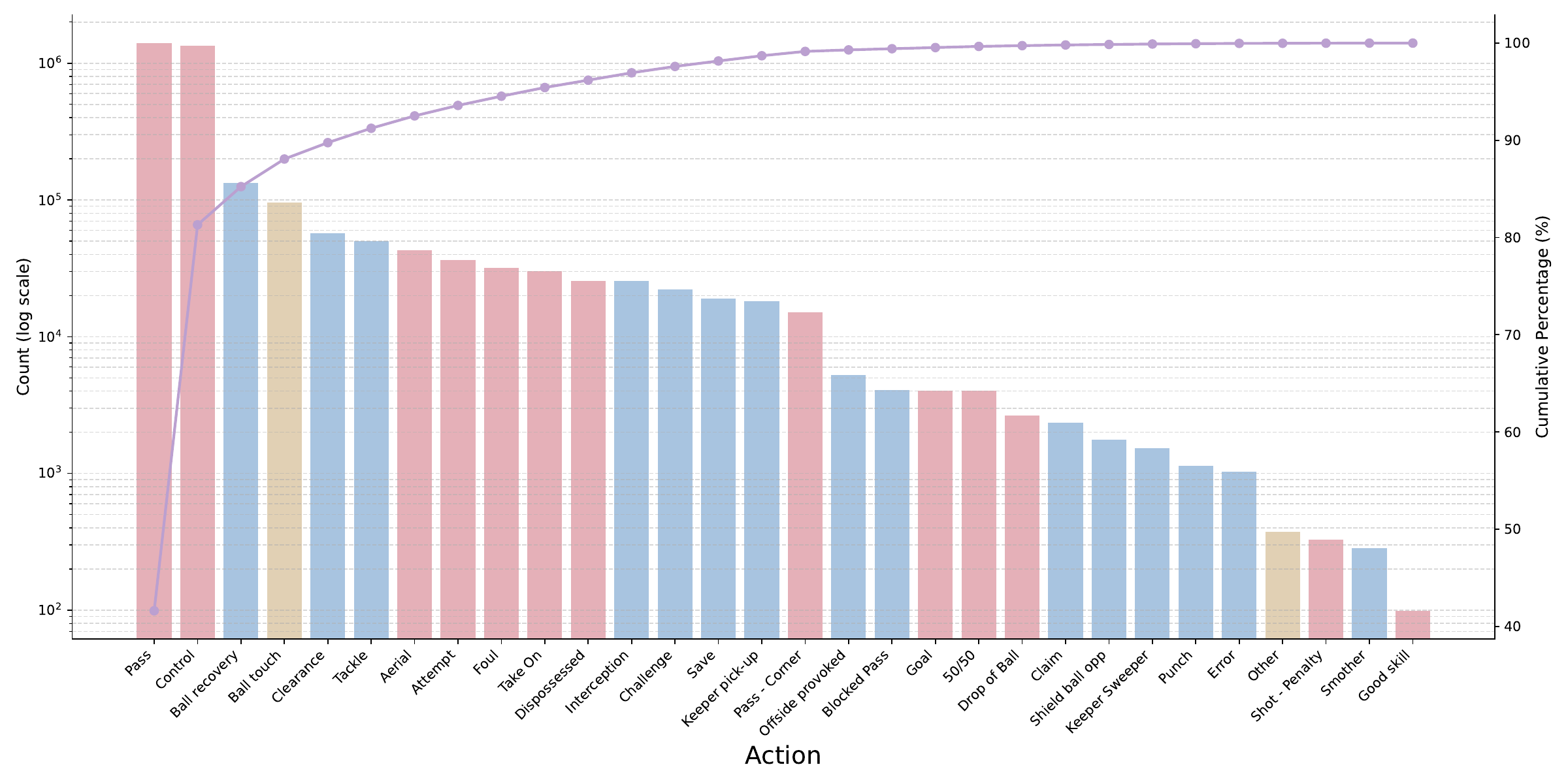}
    \vspace{-0.3in}
    \caption{{Log-scaled Pareto chart of event actions grouped by type: attacking (pale pink), defending (light blue), and neutral (light beige).}}
    \label{fig:action_type}
\end{figure}

Finally, our dataset consists of 1,432 games collected from a range of top-tier football leagues, including 829 from the EFL Championship\footnote{\url{https://en.wikipedia.org/wiki/EFL_Championship}}, 489 from the English Premier League\footnote{\url{https://en.wikipedia.org/wiki/Premier_League}}, 50 from the Major League Soccer\footnote{\url{https://en.wikipedia.org/wiki/Major_League_Soccer}}, 48 from the Dutch Eredivisie\footnote{\url{https://en.wikipedia.org/wiki/Eredivisie}}, and 16 games from other top European leagues, including the German Bundesliga\footnote{\url{https://en.wikipedia.org/wiki/Bundesliga}}, French Ligue 1\footnote{\url{https://en.wikipedia.org/wiki/Ligue_1}}, and Belgian Pro League\footnote{\url{https://en.wikipedia.org/wiki/Belgian_Pro_League}}, which cover the 2018-2025 seasons. In total, the dataset comprises 3,374,599 events and 97,760,895 frames (timesteps). Each frame records the $(x,y)$ positions of the ball and all the players from both teams (usually 22 players). Figure \ref{fig:action_type} presents a Pareto chart of event actions, grouped by action type.

The game was further segmented into discrete \textit{playing episodes}. An episode was terminated under any of the following conditions:
\begin{itemize}
    \item A goal is scored;
    \item The referee signals the end of a half;
    \item The ball remains out of play for more than 30 seconds, a threshold identified by football experts as significantly disrupting game dynamics.
\end{itemize}

\begin{figure}[htbp]
    \centering
    \includegraphics[width=0.5\textwidth]{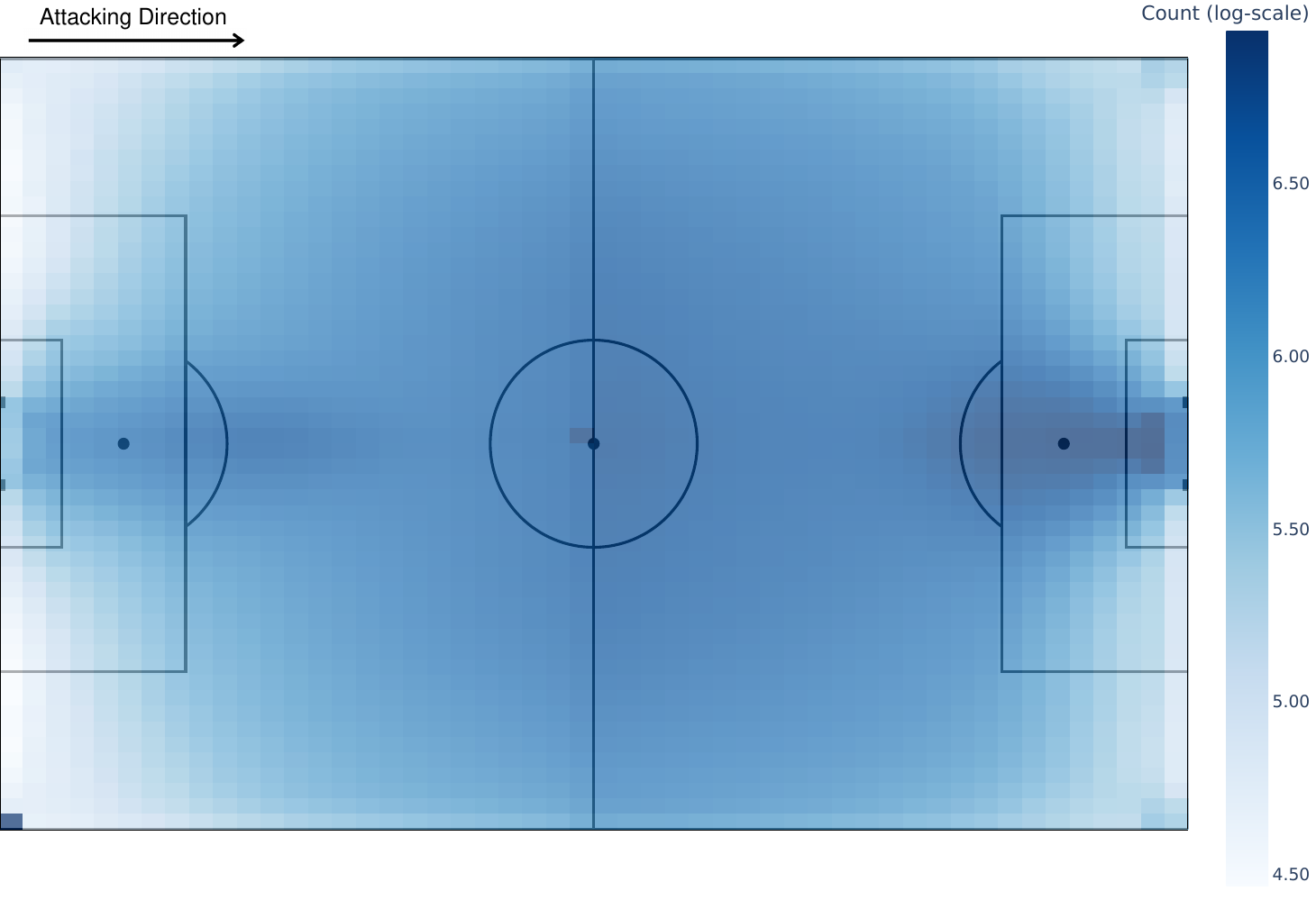}
    \vspace{-0.3in}
    \caption{{Log‐scale spatial density heatmap of processed tracking data.}}\label{fig:heat_map}
\end{figure}

As diffusion models require a fixed-length output, we standardize the variable-length trajectory data by either truncating or padding to a fixed horizon of 64. Compared to the context length of 10, this extended length allows for meaningful trajectory forecasting and strategic recommendations. Figure~\ref{fig:heat_map} shows the heatmap of the processed tracking data, which reveals a diverse range of movement across the pitch. 

\subsection{Reward Function}\label{sec:data-reward-detail}
A reward function was defined to quantify the outcomes of actions, with a team receiving a positive reward under the following conditions:
\begin{itemize}
    \item \textbf{Scoring a goal}: A base reward of 1 is assigned, with a bonus modifier to reflect goal importance (e.g., a decisive 90th-minute goal in a tied game receives more credit than a goal in a lopsided scoreline);
    \item \textbf{Creating a high-quality scoring opportunity}, as annotated by Opta analysts as a ``big chance,'' earns a reward of 0.75;
    \item \textbf{Earning a penalty}: The fouled team receives +0.75, which is roughly the empirical probability of scoring a penalty in our dataset.
    
\end{itemize}

In all cases, the opposing team receives a negative reward equal in magnitude to the positive reward. These reward rules were carefully developed in consultation with football experts, guided by three key principles: reducing reward sparsity, minimizing subjectivity, and penalizing defensive errors in a symmetric manner.

\section{Evaluation Metrics}\label{sec:metric}
For the trajectory prediction task, we follow prior works~\cite{hauri2021multi,mao2023leapfrog,xu2025sports,capellera2025unified} and adopt the following metrics to evaluate the predictive performance of each model: 
\begin{enumerate}
\item \textbf{Average Displacement Error (ADE)}, the mean Euclidean distance between predicted and ground-truth positions over all time steps
\item \textbf{Final Displacement Error (FDE)}, the Euclidean distance between the predicted and true positions at the final time step
\item \textbf{Miss Rate (MR)}: the proportion of predicted trajectories whose final displacement error exceeds 2 meters.
\end{enumerate}

To better capture the stochastic nature of future trajectories, we adopt a best-of-$N$ evaluation strategy~\cite{gu2022stochastic,xu2025sports,capellera2025unified}, where $N$ diverse predictions are generated and the best results among them are reported. Empirically, the generated trajectories show meaningful tactical variations, such as alternative passing lanes, defensive adjustments, and different player positioning, while still maintaining realistic spatial coordination among players. Using multiple predictions (e.g., $N=20$) is a common practice in trajectory forecasting to approximate the distribution of possible futures and increase the likelihood that one prediction aligns closely with the ground truth. In practical applications, users can adjust the number of generated samples according to available computational resources and task requirements, allowing them to explore different possible tactical outcomes.

In addition to reporting \textit{marginal metrics}, which evaluate each agent independently across $N$ samples, we also report \textit{joint metrics} in consideration of the multi-agent nature of football. Following~\cite{cheng2023gatraj,sun2022m2i,weng2023joint,ngiam2021scene}, joint evaluation is performed at the group level by first averaging the metrics across all agents within each sample, and then selecting the sample with the best overall performance based on the aggregated result. This approach ensures that agents are not mixed across different samples, preserving the coherence of multi-agent interactions.

In the guided trajectory generation task, the objective extends beyond generating realistic samples that resemble the ground truth. In addition to realism, a crucial requirement is that the generated trajectories adhere to specific guidance objectives. Following prior works~\cite{chen2024playbest, wang2024optimizing}, we evaluate performance using the \textbf{Guidance Score (GS)}, which measures the value determined by specific tactical objective functions, quantifying the degree to which the generated trajectories align with the intended guidance signals.

Although quantitative evaluation of the above components is essential for TacticGen’s technical development, its ultimate measure of success lies in practical utility, particularly as assessed by professionals in the football industry. To this end, we also conducted a case study in collaboration with our football domain partners.
\section{Ablation Study}\label{sec:ablation}
To assess the effectiveness of individual components in TacticGen, we perform a comprehensive ablation study that evaluates the contributions of the context encoder, the event encoder, and the impact of conditional ball modeling (i.e., encoding the complete ball trajectory). The results are reported in Table~\ref{tab:ablation}.

\begin{table}[htbp]
\centering
\caption{\textbf{Ablation study on the key components of TacticGen.} PB denotes ball prediction, CE the context encoder, and EE the event encoder. $^*$ indicates replacing the proposed context encoder with a graph neural network encoder~\cite{omidshafiei2022multiagent}. For CE, $\triangle$ denotes predictive ball modeling, whereas $\checkmark$ denotes conditional ball modeling. For EE, $\triangle$ denotes exclusion of the ball destination from global features, while $\checkmark$ denotes its inclusion. The fifth and final variants correspond to TacticGen-P and TacticGen-C, respectively.}
\label{tab:ablation}
\resizebox{0.5\textwidth}{!}{%
\begin{tabular}{lcccccccccc}
\toprule
\multirow{2}{*}{\bf Method} 
  & \multicolumn{3}{c}{\bf Modules} 
  & \multicolumn{3}{c}{\bf Marginal Metrics} 
  & \multicolumn{3}{c}{\bf Joint Metrics} \\
\cmidrule(lr){2-4} \cmidrule(lr){5-7} \cmidrule(lr){8-10}
  & PB & CE & EE 
  & ADE & FDE & MR (\%) 
  & JADE & JFDE & JMR (\%) \\
\midrule
\multirow{7}{*}{\textbf{TacticGen}} 
  & $\checkmark$ & $-$ & $-$ & 0.41 & 0.84 & 10.59 & 0.63 & 1.30 & 16.69 \\
  & $\checkmark$ & $\triangle$ & $-$ & 0.38 & 0.75 & 9.02 & 0.53 & 1.10 & 13.57 \\
  & $\checkmark$ & $-$ & $\triangle$ & 0.32 & 0.54 & 5.34 & 0.55 & 1.09 & 13.45 \\
  & $\checkmark$ & $\triangle^*$ & $\triangle$ & 0.30 & 0.53 & 4.98 & 0.46 & 0.94 & 11.03 \\
  & $\checkmark$ & $\triangle$ & $\triangle$ & 0.29 & 0.52 & 4.73 & 0.45 & 0.92 & 10.66 \\
  & $\checkmark$ & $-$ & $\checkmark$ & 0.26 & 0.42 & 3.00 & 0.46 & 0.87 & 9.45 \\
  & $\checkmark$ & $\triangle$ & $\checkmark$ & 0.24 & 0.40 & 2.75 & 0.39 & 0.76 & 7.78 \\
  & $-$ & $\checkmark$ & $\checkmark$ & 0.21 & 0.38 & 2.59 & 0.35 & 0.74 & 7.59 \\
\bottomrule
\end{tabular}%
}
\end{table}

Comparing models with and without the event encoder, we find that it provides valuable information by capturing both event type and global trajectory features, thereby improving prediction accuracy. Moreover, even the baseline variant of TacticGen, which uses only the MADiT backbone (third variant in Table~\ref{tab:ablation}) with an event encoder, already outperforms existing state-of-the-art trajectory prediction methods across multiple metrics (Table~\ref{tab:prediction-results}), underscoring the effectiveness of our architecture in modeling complex inter-agent interactions. Building on this foundation, TacticGen-P (fifth variant) incorporates a context encoder and a cross-attention mechanism, further enhancing performance by more effectively leveraging contextual information for future trajectory generation.

It is also instructive to examine the fourth variant in Table~\ref{tab:prediction-results}, where the context encoder in TacticGen-P is replaced by a graph-based architecture~\cite{omidshafiei2022multiagent}. The results show that our context encoder, which combines an MLP-mixer with an attention mechanism, achieves superior performance.

Finally, TacticGen-C (last variant) achieves the highest overall performance by encoding the complete ball trajectory into the context encoder. Although this setting is not a strictly fair comparison baseline since it leverages information unavailable to other models, it remains highly practical in scenarios where coaches can anticipate or assume the ball’s movement when designing tactics or adjusting player positioning.

\section{Temporal Split Experiments}\label{sec:temporal-exp}

As discussed in the main paper, our training and test datasets are randomly sampled from all events. While this approach provides a fair evaluation of TacticGen’s ability, we recognize that tactics may evolve over time~\cite{wang2024tacticai}. To assess the model’s robustness to temporal shifts and naturally evolving tactics, we re-ran our experiments using a temporal split. In this setting, the 20\% of events from the most recent games in our dataset were used for testing. We then compared TacticGen with the top-5 methods (mesured in JADE) listed in Table~\ref{tab:prediction-results} from the main experiments. The results are summarized in Table~\ref{tab:temporal-results}.

\begin{table}[htbp]
\centering
\vspace{-0.1in}
\caption{{Performance of different methods in the temporal split setting.}}\label{tab:temporal-results}

\resizebox{0.49\textwidth}{!}{%
\begin{tabular}{@{}lcccccc@{}}
\toprule
\multirow{2}{*}{\bf Method} 
  & \multicolumn{3}{c}{\bf Marginal} 
  & \multicolumn{3}{c}{\bf Joint} \\
\cmidrule(lr){2-4} \cmidrule(lr){5-7}
  & ADE & FDE & MR (\%) & JADE & JFDE & JMR (\%) \\
\midrule
Scene Transformer~\cite{ngiam2021scene} &0.44 &0.73 &8.62 &0.78 &1.35 &16.23 \\
Sports-Traj~\cite{xu2025sports} &0.47 &0.73 &8.70 &0.72 &1.29 &15.74 \\
MID~\cite{gu2022stochastic} &0.42 &0.69 &7.53 &0.77 &1.40 &18.13 \\
LED~\cite{mao2023leapfrog} &0.40 &0.66 &6.92 &0.74 &1.35 &16.59 \\
MADiff~\cite{zhu2023madiff} &0.39 &0.64 &6.70 &0.66 &1.23 &15.27 \\
\midrule
\textbf{TacticGen}  &\textbf{0.33} &\textbf{0.58} &\textbf{5.64} &\textbf{0.53} &\textbf{1.01} &\textbf{11.72} \\
\bottomrule
\end{tabular}%
}
\vspace{-0.05in}
\end{table}

We observe that all methods experience slight performance regressions compared to the random split setting in the main experiments, likely due to the gradual tactical evolution over time, which causes a slight difference in the distribution between the training and testing datasets. However, TacticGen still outperforms all other methods in the temporal split setting across all metrics, demonstrating its robustness to temporal shifts and its ability to maintain high accuracy in dynamically evolving tactical scenarios. This underscores the effectiveness of TacticGen's architectural design in capturing complex interactions and adapting to changing game conditions, positioning it as a powerful tool for tactical prediction in football.

\section{Evaluation on Trajectory Prediction Benchmark}\label{sec:nba-results}
We evaluate the predictive performance of TacticGen on a public sport trajectory prediction benchmark with existing state-of-the-art methods.

\begin{table}[htbp]
\centering
\caption{{Comparison with baseline models on the NBA dataset.} We report minADE\textsubscript{20} and minFDE\textsubscript{20} (in meters). The \textbf{bolded} values indicate the best results, respectively.}\label{tab:nba-results}
\resizebox{0.5\textwidth}{!}{%
\begin{tabular}{lcccc}
\toprule
Method & 1.0\,s & 2.0\,s & 3.0\,s & Total (4.0\,s) \\ \midrule
Social-GAN~\cite{gupta2018socialgan}        & 0.41/0.62           & 0.81/1.32           & 1.19/1.94           & 1.59/2.41           \\
STGAT~\cite{huang2019stgat}                & 0.35/0.51           & 0.73/1.10           & 1.04/1.75           & 1.40/2.18           \\
Social-STGCNN~\cite{mohamed2020socialstgcnn}     & 0.34/0.48           & 0.71/0.94           & 1.09/1.77           & 1.53/2.26           \\
PECNet~\cite{mangalam2020pecnet}            & 0.40/0.71           & 0.83/1.61           & 1.27/2.44           & 1.69/2.95           \\
STAR~\cite{yu2020star}                      & 0.43/0.66           & 0.75/1.24           & 1.03/1.51           & 1.13/2.01           \\
Trajectron++~\cite{salzmann2020trajectron++}& 0.30/0.38           & 0.59/0.82           & 0.85/1.24           & 1.15/1.57           \\
MemoNet~\cite{xu2022memonet}                & 0.38/0.56           & 0.71/1.14           & 1.00/1.57           & 1.25/1.47           \\
NPSN~\cite{bae2022npsn}                     & 0.35/0.58           & 0.68/1.23           & 1.01/1.76           & 1.31/1.79           \\
GroupNet~\cite{xu2022groupnet}              & 0.26/0.34           & 0.49/0.70           & 0.73/1.02           & 0.96/1.30           \\
MID~\cite{gu2022stochastic}                 & 0.28/0.37           & 0.51/0.72           & 0.71/0.98           & 0.96/1.27           \\
LED~\cite{mao2023leapfrog}                  & {0.21}/\textbf{0.31} & \textbf{0.42}/{0.63} & {0.65}/{0.92}           & {0.89}/{1.24}           \\
Diffuser~\cite{janner2022planning}          & 0.46/0.55           & 0.72/0.89           & 0.96/1.16           & 1.20/1.41           \\ 
DiT~\cite{peebles2023scalable}              & 0.36/0.47           & 0.63/0.87           & 0.92/1.19           & 1.18/1.45           \\ 
MADiff~\cite{zhu2023madiff}                 & 0.25/0.33           & 0.46/0.72           & 0.71/1.06           & 0.97/1.36           \\  \midrule
\textbf{TacticGen}                  & \textbf{0.20}/\textbf{0.31}           & \textbf{0.42}/\textbf{0.62}           & \textbf{0.61}/\textbf{0.88} & \textbf{0.84}/\textbf{1.18} \\ \bottomrule
\end{tabular}%
}
\end{table}

Specifically, we use the NBA SportVU Dataset (NBA)\footnote{\url{https://github.com/linouk23/NBA-Player-Movements}}, which was collected by the NBA using the SportVU tracking system during the 2015 to 2016 season. We follow the same data processing and subset selection as in~\cite{mao2023leapfrog}, resulting in 40,000 trajectory sequences in total. Each sequence contains 30 frames sampled at 5 Hz (6 seconds), including the $(x,y)$ positions of 10 players and the ball. Following previous works~\cite{mao2023leapfrog,capellera2025unified}, we train our model to observe the first 2 seconds (10 frames) and predict the following 4 seconds (20 frames). To ensure consistency with prior
works~\cite{mao2023leapfrog}, we report marginal metrics, computed independently over agents and samples. It is worth noting that, because the dataset lacks event information, we omit the event encoder in TacticGen-P for a fair comparison. The results for minADE\textsubscript{20} and minFDE\textsubscript{20} are presented in Table~\ref{tab:nba-results}.

The results show that TacticGen remains highly competitive with state-of-the-art methods, outperforming all baselines in both minADE\textsubscript{20} and minFDE\textsubscript{20} when evaluated over the full 4.0\,s prediction horizon. These findings highlight the effectiveness of TacticGen’s multi-agent diffusion transformer backbone and context encoder in capturing complex player movement patterns, demonstrating strong potential for generalization beyond the football domain.
\section{More Experiments on Trajectory Prediction}\label{sec:more-exp-pred}

\begin{figure}[htbp]
  \centering

  \begin{minipage}[t]{0.16\textwidth}
    \centering
    \small \textbf{(a)} Ground Truth\\
    \includegraphics[width=\linewidth]{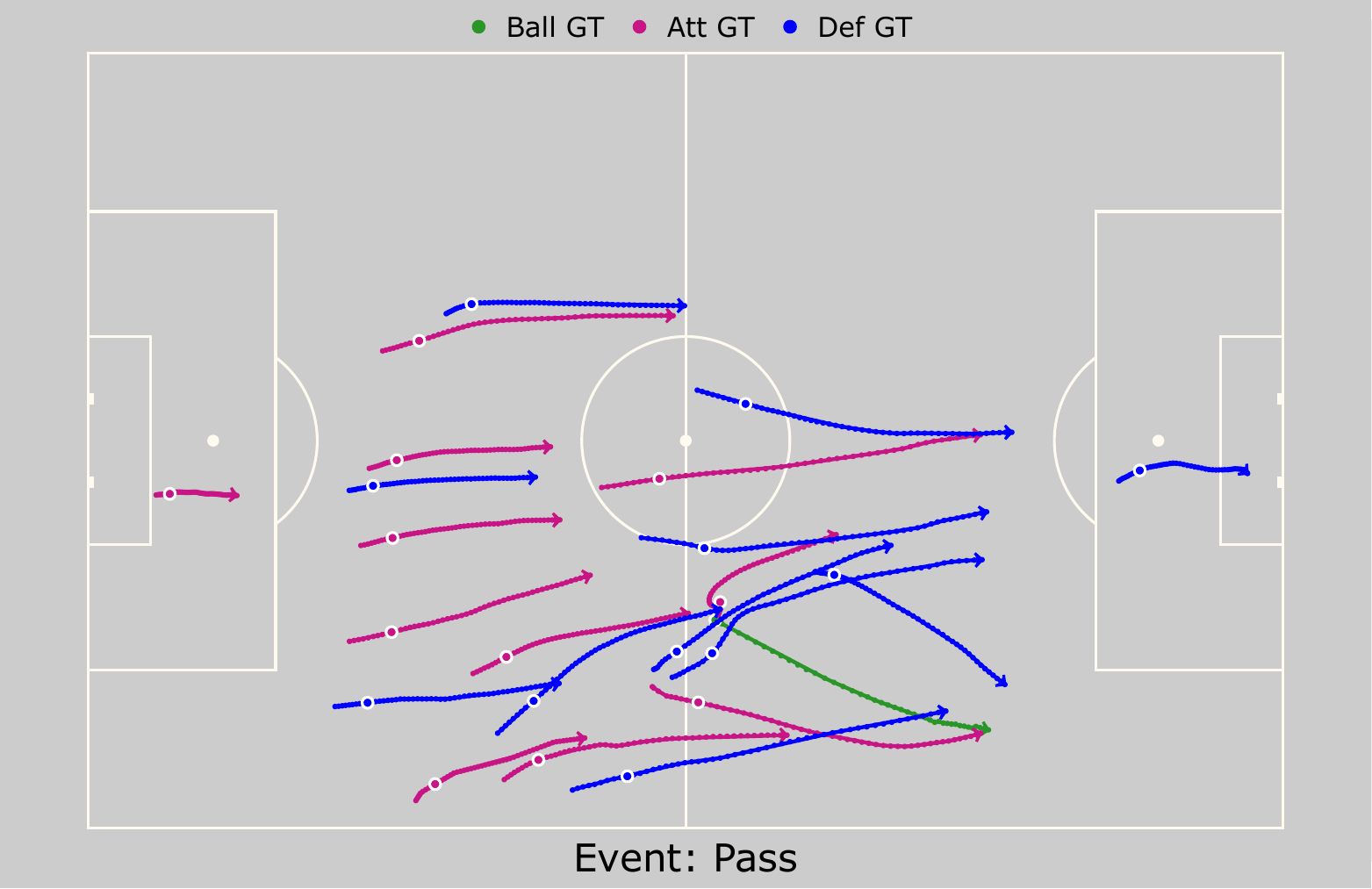}
  \end{minipage}\hfill
    \begin{minipage}[t]{0.16\textwidth}
    \centering
    \small \textbf{(b)} Diffuser\\
    \includegraphics[width=\linewidth]{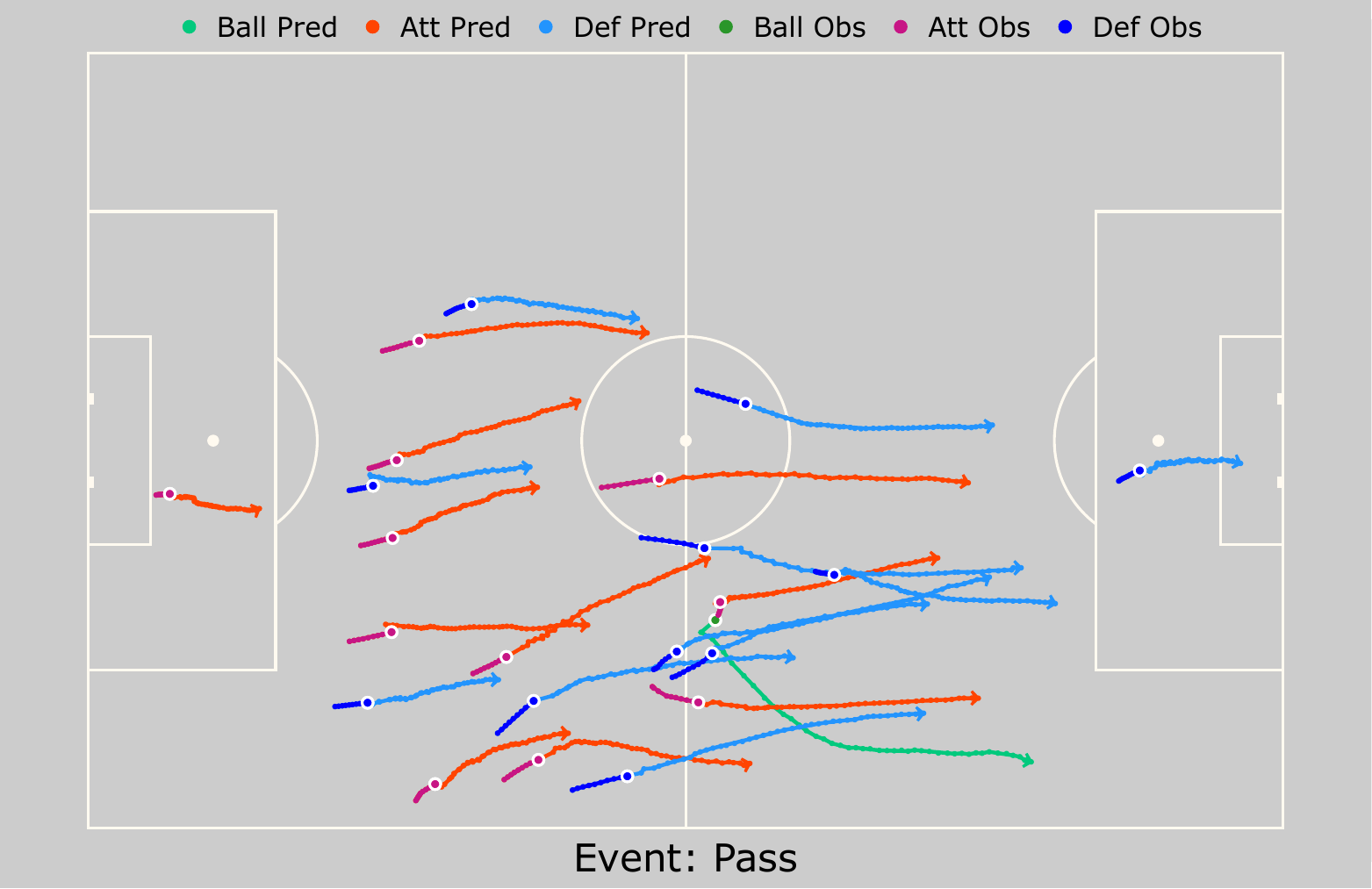}
  \end{minipage}\hfill
  \begin{minipage}[t]{0.16\textwidth}
    \centering
    \small \textbf{(c)} MID\\
    \includegraphics[width=\linewidth]{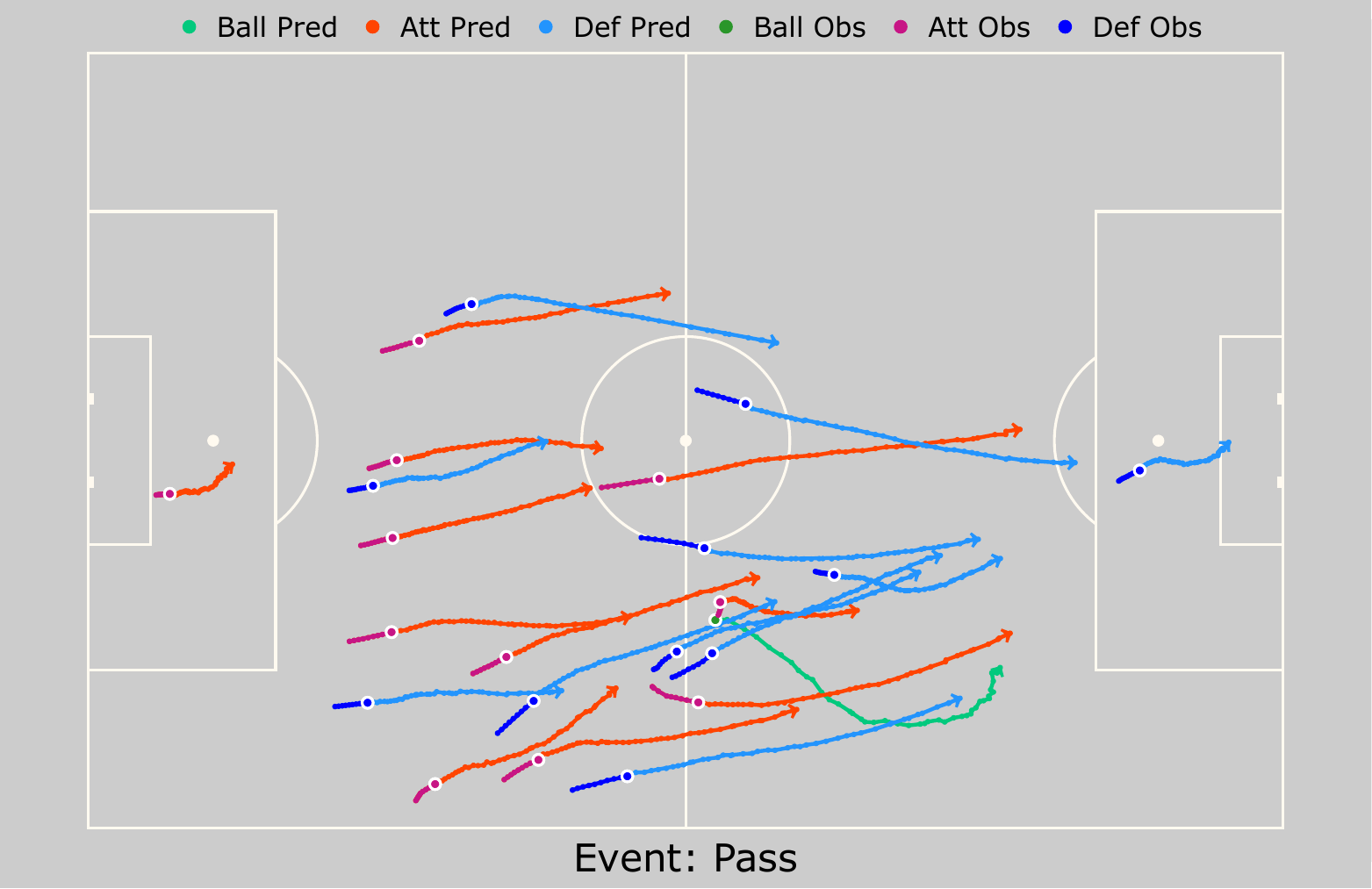}
  \end{minipage}

  \vspace{0.5em} 

  \begin{minipage}[t]{0.16\textwidth}
    \centering
    \small \textbf{(d)} MADiff\\
    \includegraphics[width=\linewidth]{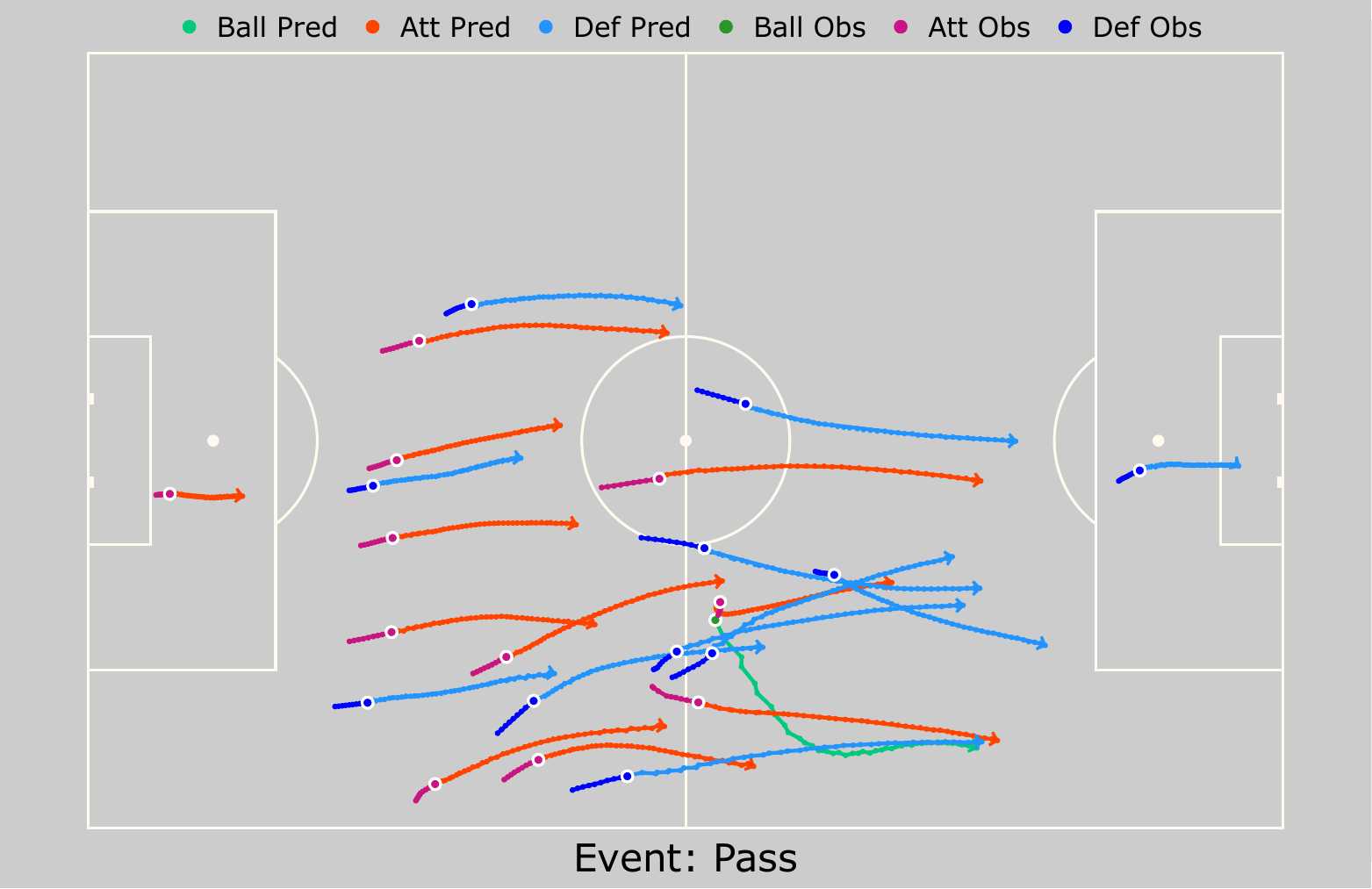}
  \end{minipage}\hfill
  \begin{minipage}[t]{0.16\textwidth}
    \centering
    \small \textbf{(e)} TacticGen-P\\
    \includegraphics[width=\linewidth]{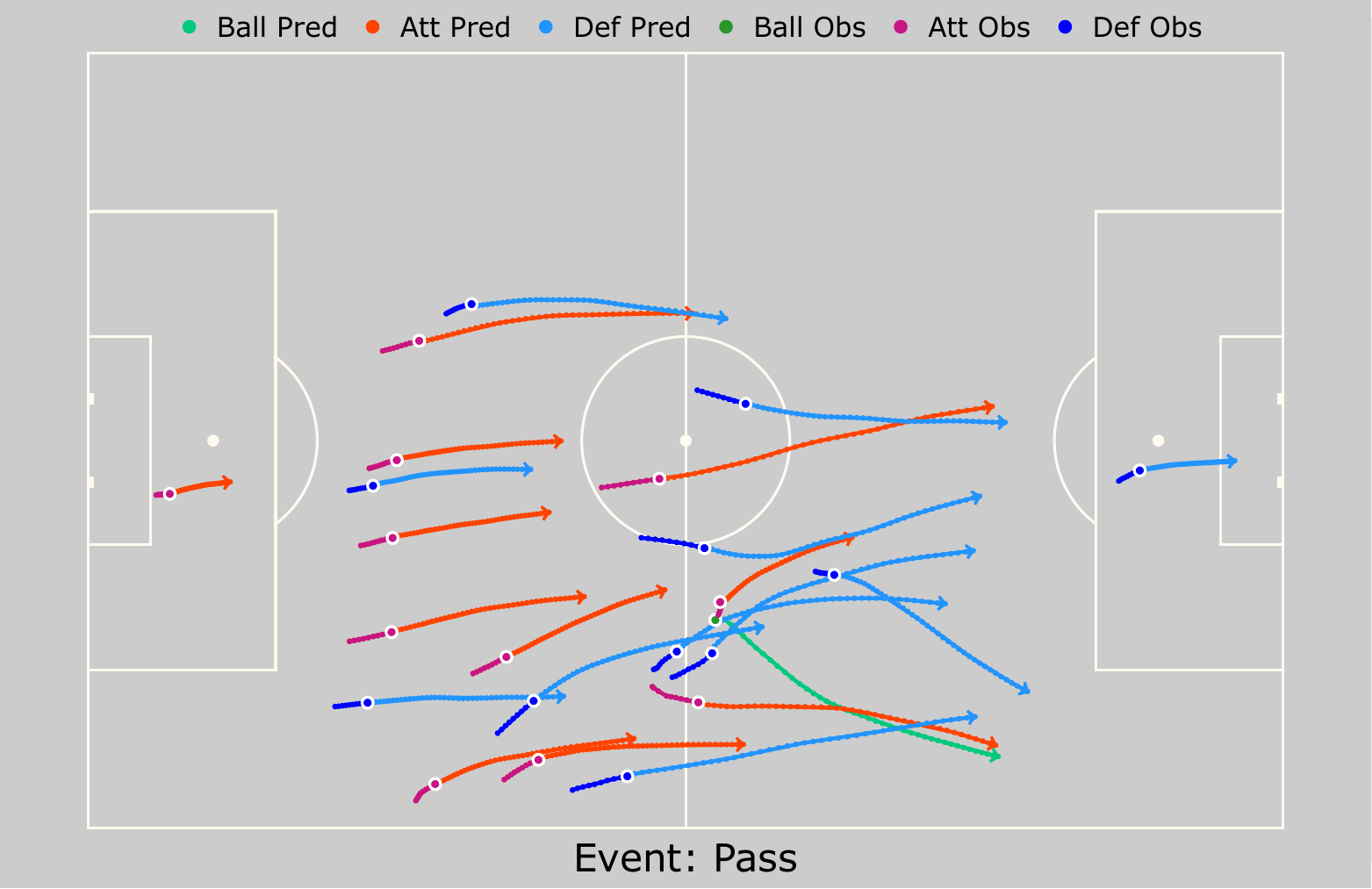}
  \end{minipage}\hfill
  \begin{minipage}[t]{0.16\textwidth}
    \centering
    \small \textbf{(f)} TacticGen-C\\
    \includegraphics[width=\linewidth]{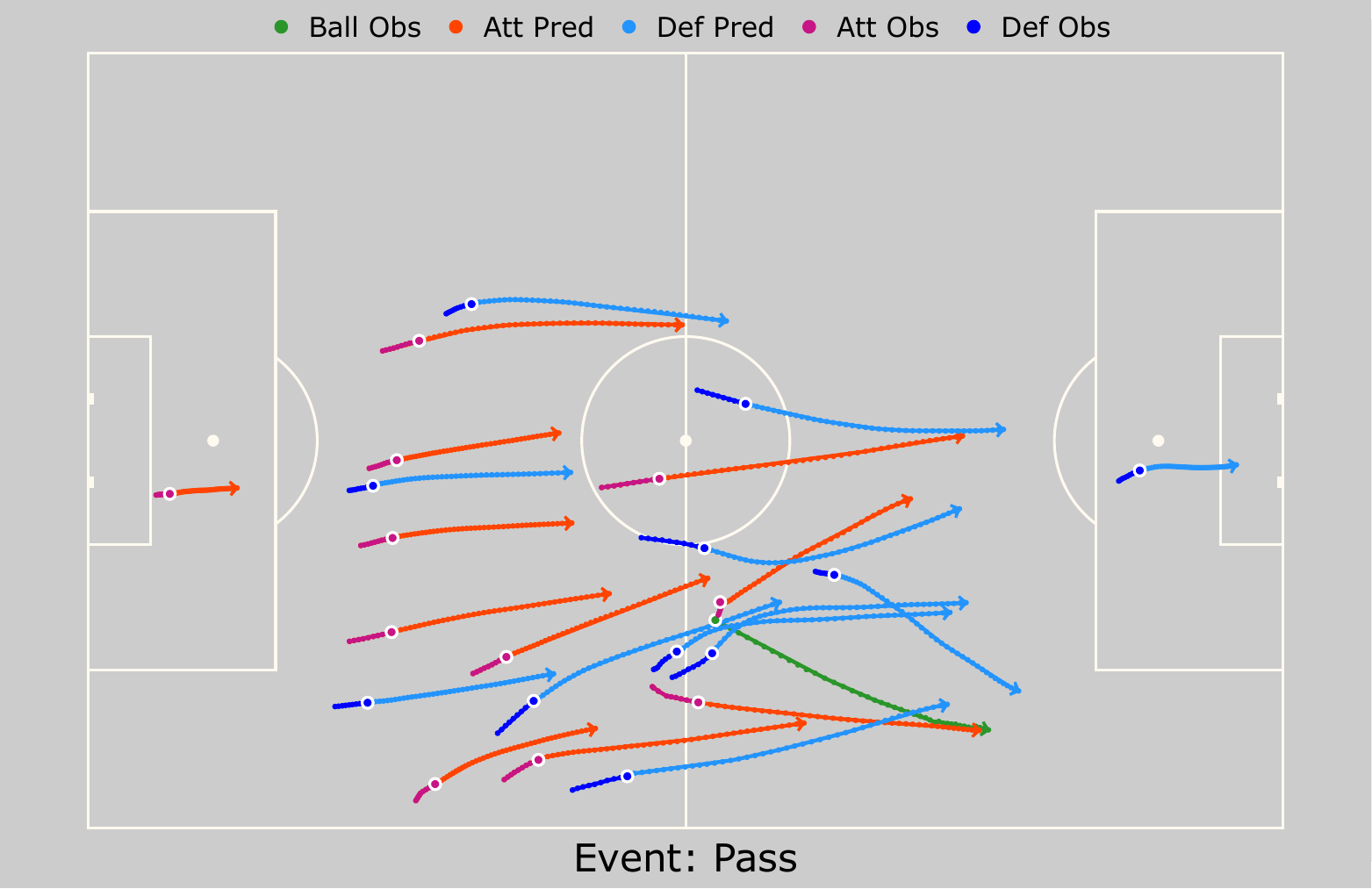}
  \end{minipage}

  \caption{{Ground truth trajectories for a pass event and the corresponding best-of-20 predictions generated by five methods.}}  \label{fig:pred-pass-52}
\end{figure}

\begin{figure}[htbp]
\vspace{-0.05in}
  \centering
  \resizebox{0.5\textwidth}{!}{%
    \begin{minipage}[t]{0.33\textwidth}
      \centering
      \small \textbf{(a)} TacticGen-P\\
      \includegraphics[width=\linewidth]{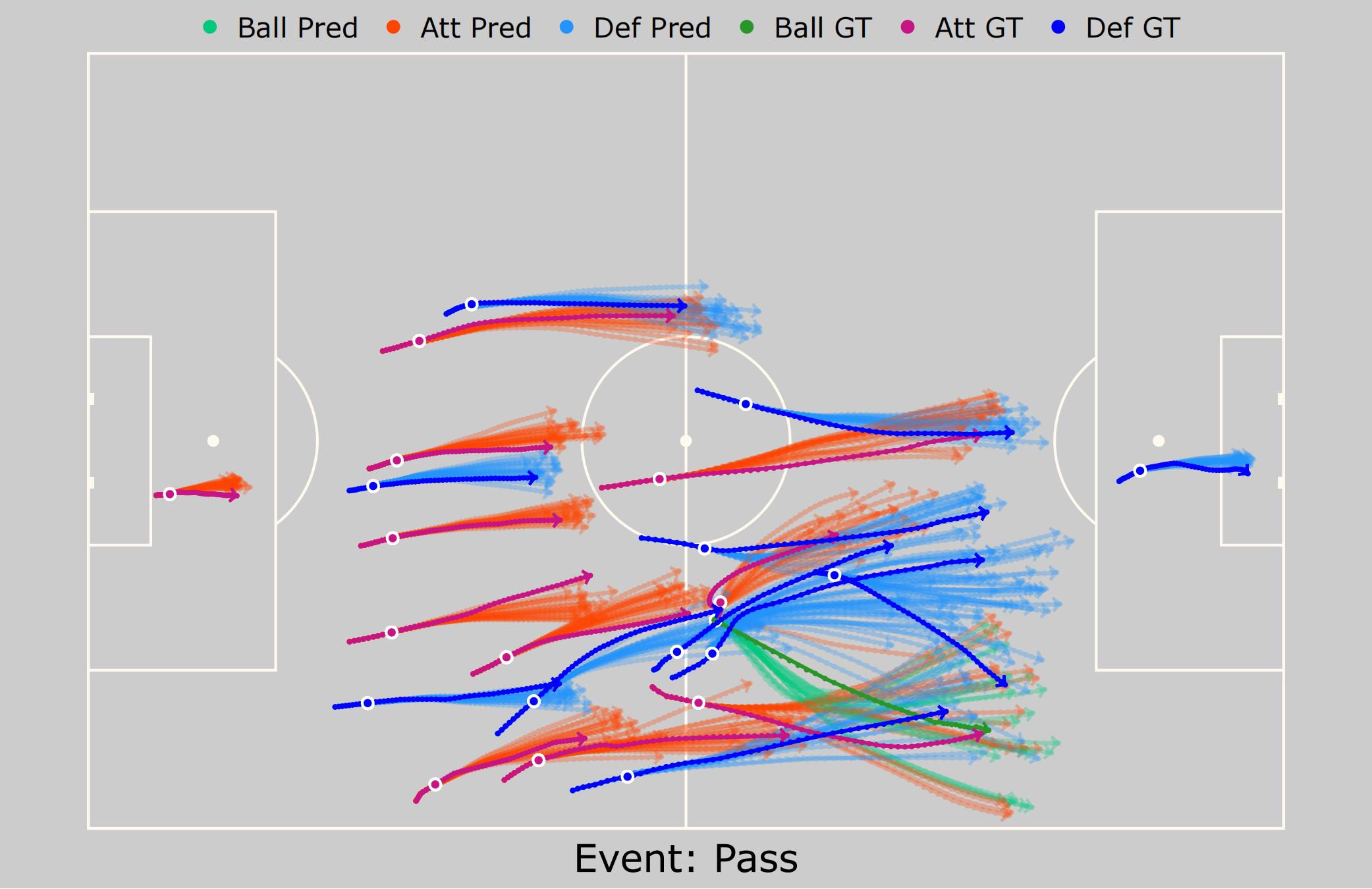}
    \end{minipage}
    \hspace{0.01\textwidth}
    \begin{minipage}[t]{0.33\textwidth}
      \centering
      \small \textbf{(b)} TacticGen-C\\
      \includegraphics[width=\linewidth]{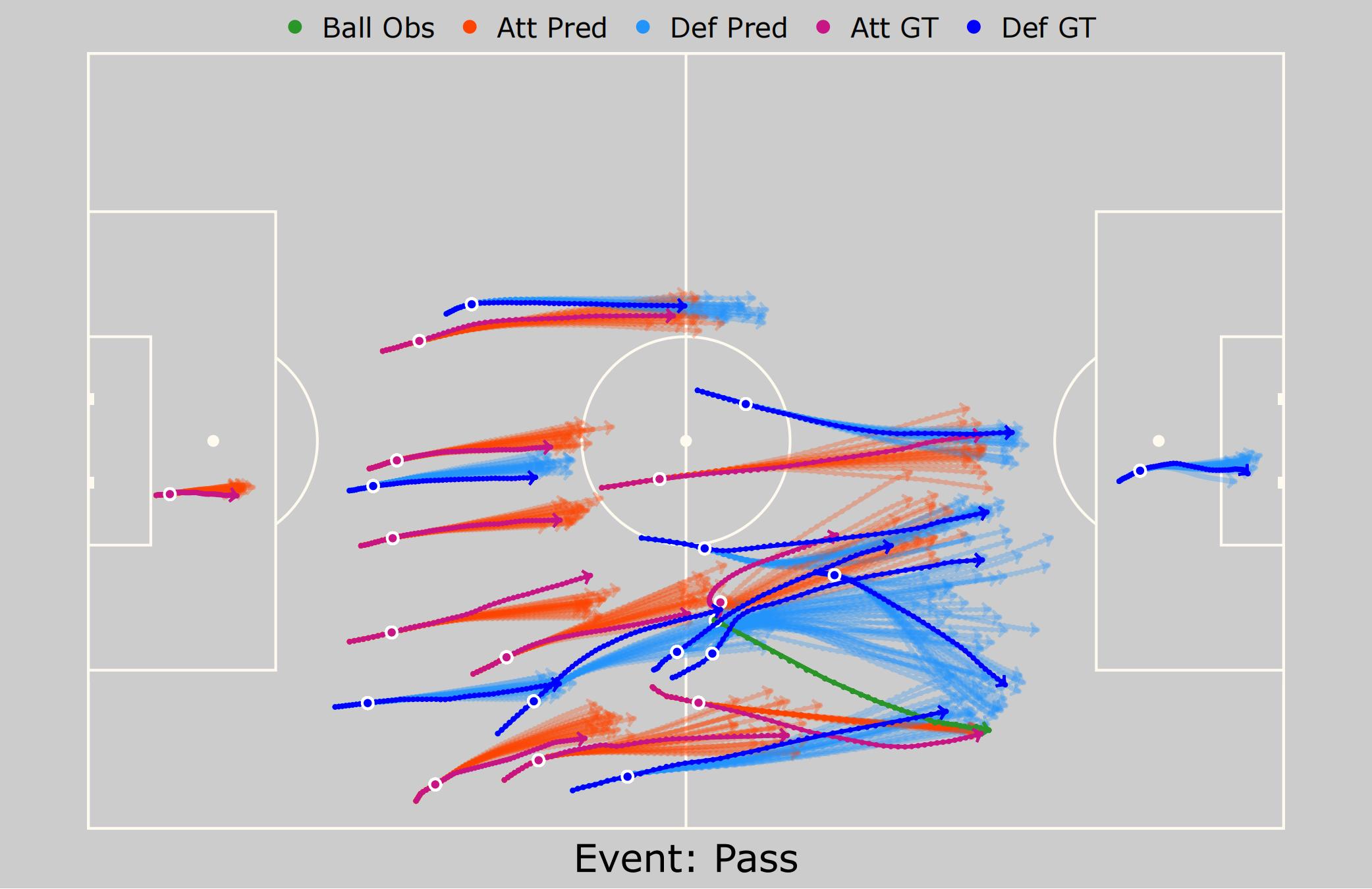}
    \end{minipage}%
  }
  \caption{{20 predicted trajectory samples by TacticGen variants for a pass event.}}
  \label{fig:footdiff-diversity-pass52}
  \vspace{-0.05in}
\end{figure}

Figure~\ref{fig:pred-pass-52} presents the best-of-20 trajectories generated by different models for a pass event, and Figure~\ref{fig:footdiff-diversity-pass52} illustrates the full set of 20 trajectories produced by TacticGen. We find that TacticGen consistently produces coherent and realistic trajectories that closely align with the ground truth, while also demonstrating strong generative capacity for diverse samples.

\begin{figure}[htbp]
  \centering

  \begin{minipage}[t]{0.16\textwidth}
    \centering
    \small \textbf{(a)} Ground Truth\\
    \includegraphics[width=\linewidth]{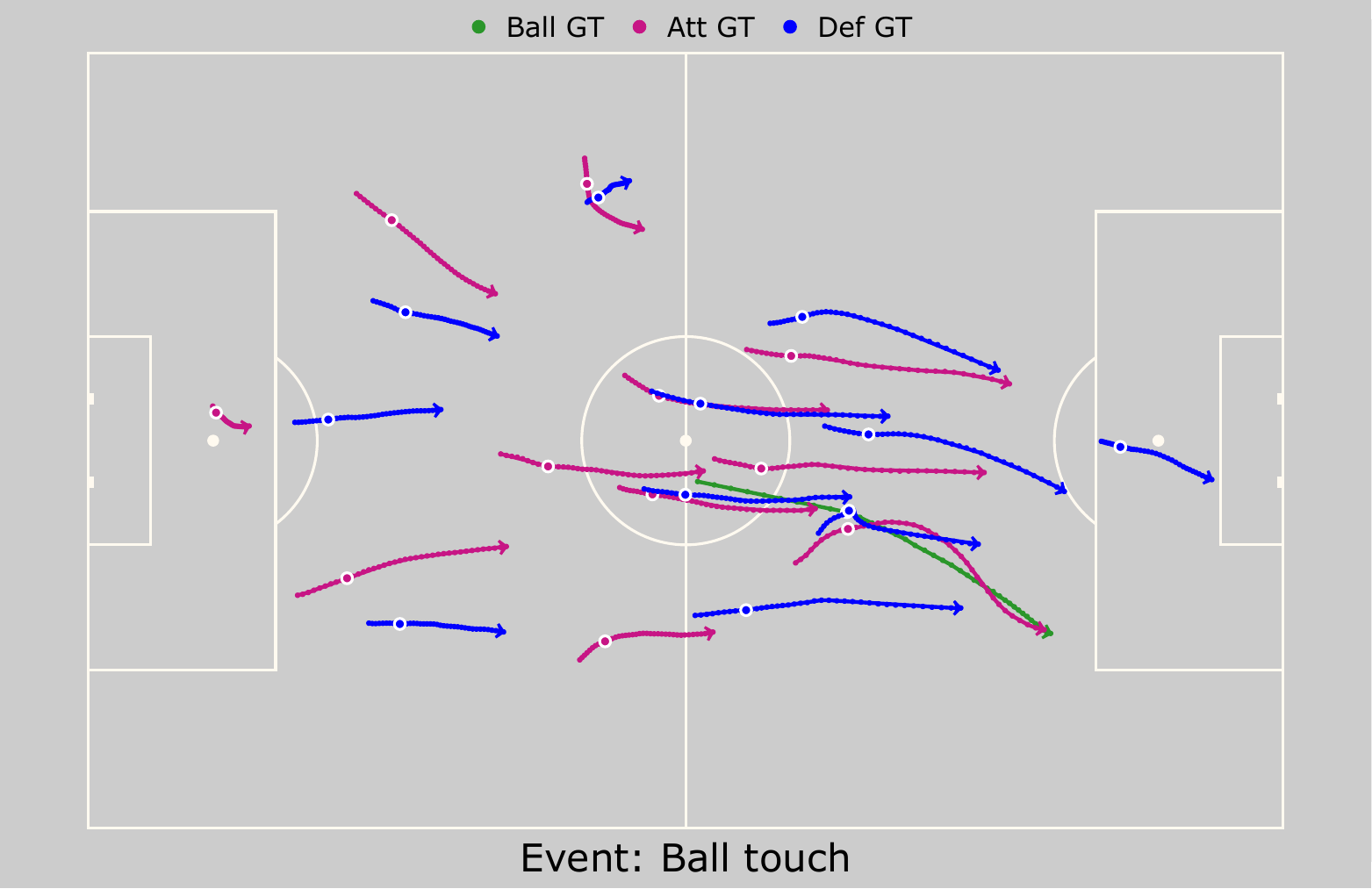}
  \end{minipage}\hfill
   \begin{minipage}[t]{0.16\textwidth}
    \centering
    \small \textbf{(b)} Diffuser\\
    \includegraphics[width=\linewidth]{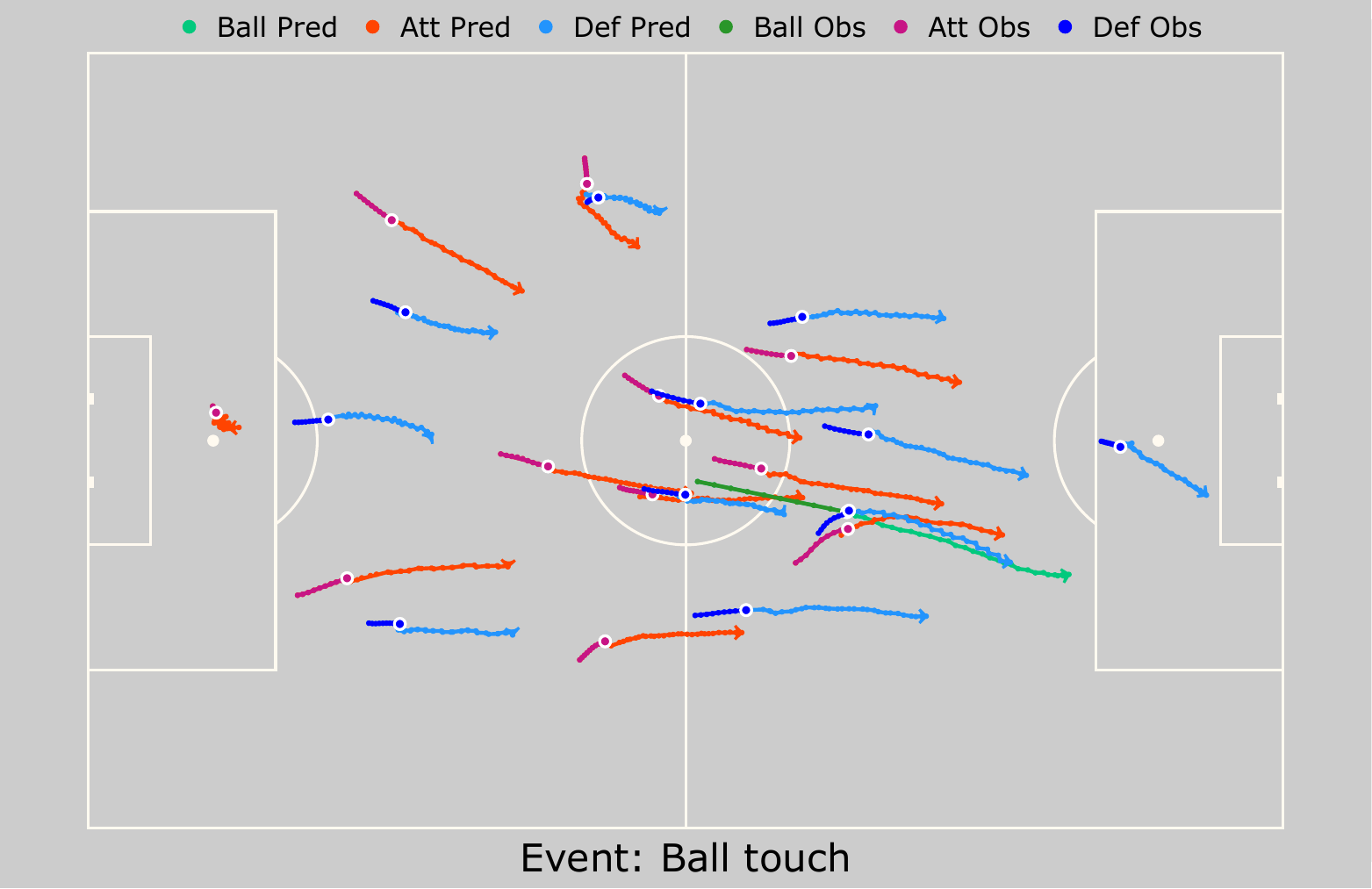}
  \end{minipage}\hfill
  \begin{minipage}[t]{0.16\textwidth}
    \centering
    \small \textbf{(c)} MID\\
    \includegraphics[width=\linewidth]{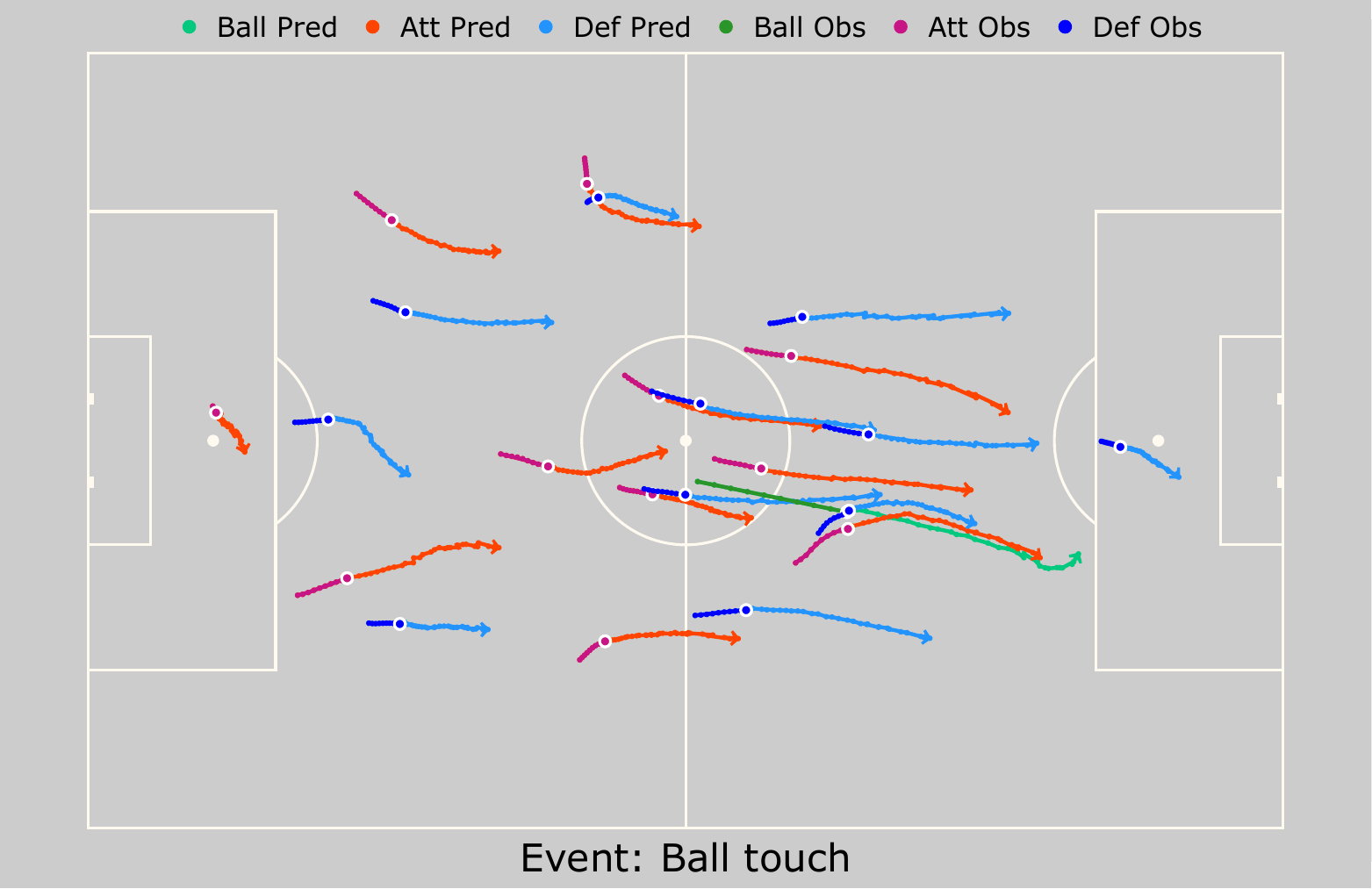}
  \end{minipage}

  \vspace{0.5em} 

  \begin{minipage}[t]{0.16\textwidth}
    \centering
    \small \textbf{(d)} MADiff\\
    \includegraphics[width=\linewidth]{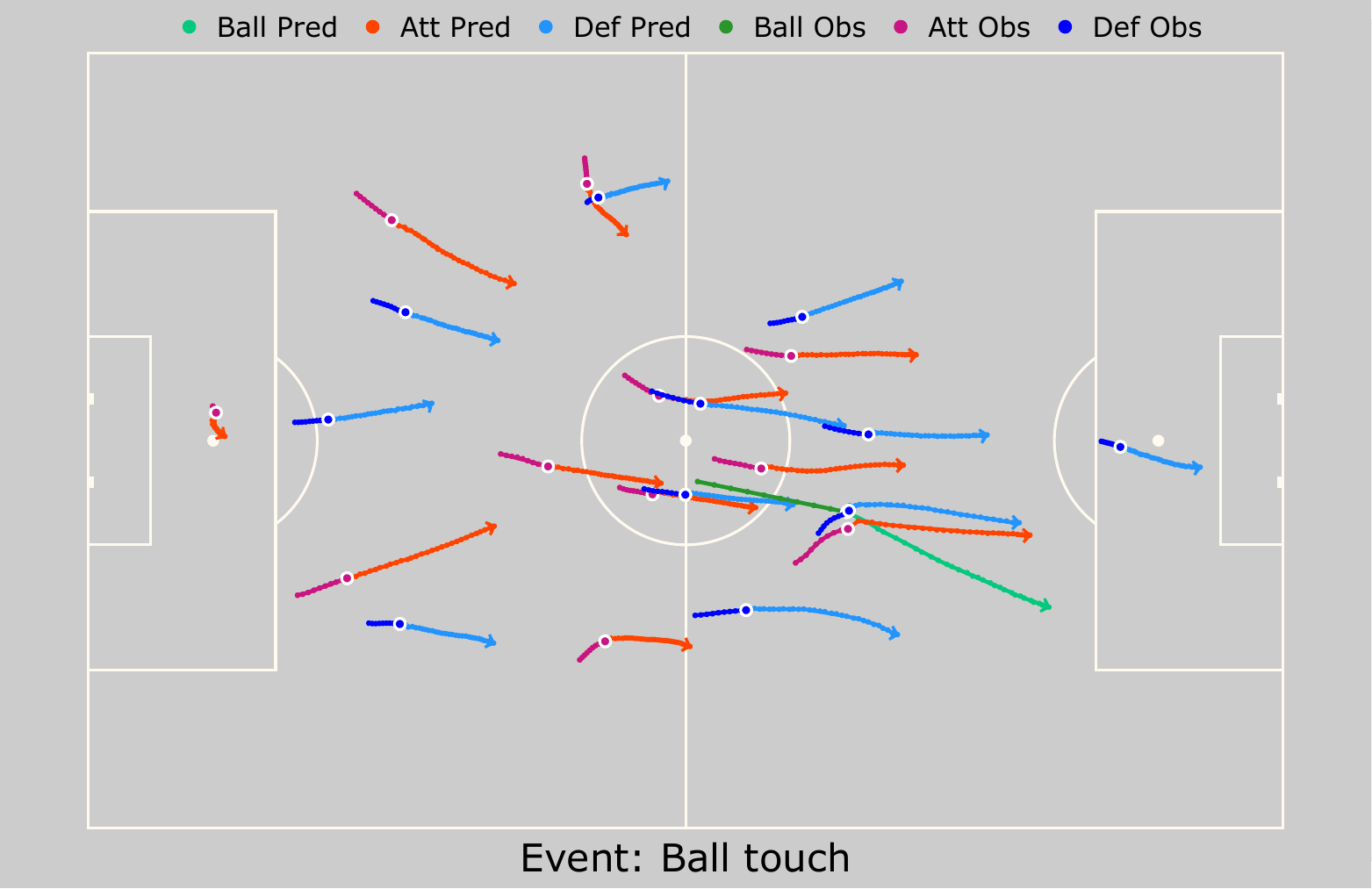}
  \end{minipage}\hfill
  \begin{minipage}[t]{0.16\textwidth}
    \centering
    \small \textbf{(e)} TacticGen-P\\
    \includegraphics[width=\linewidth]{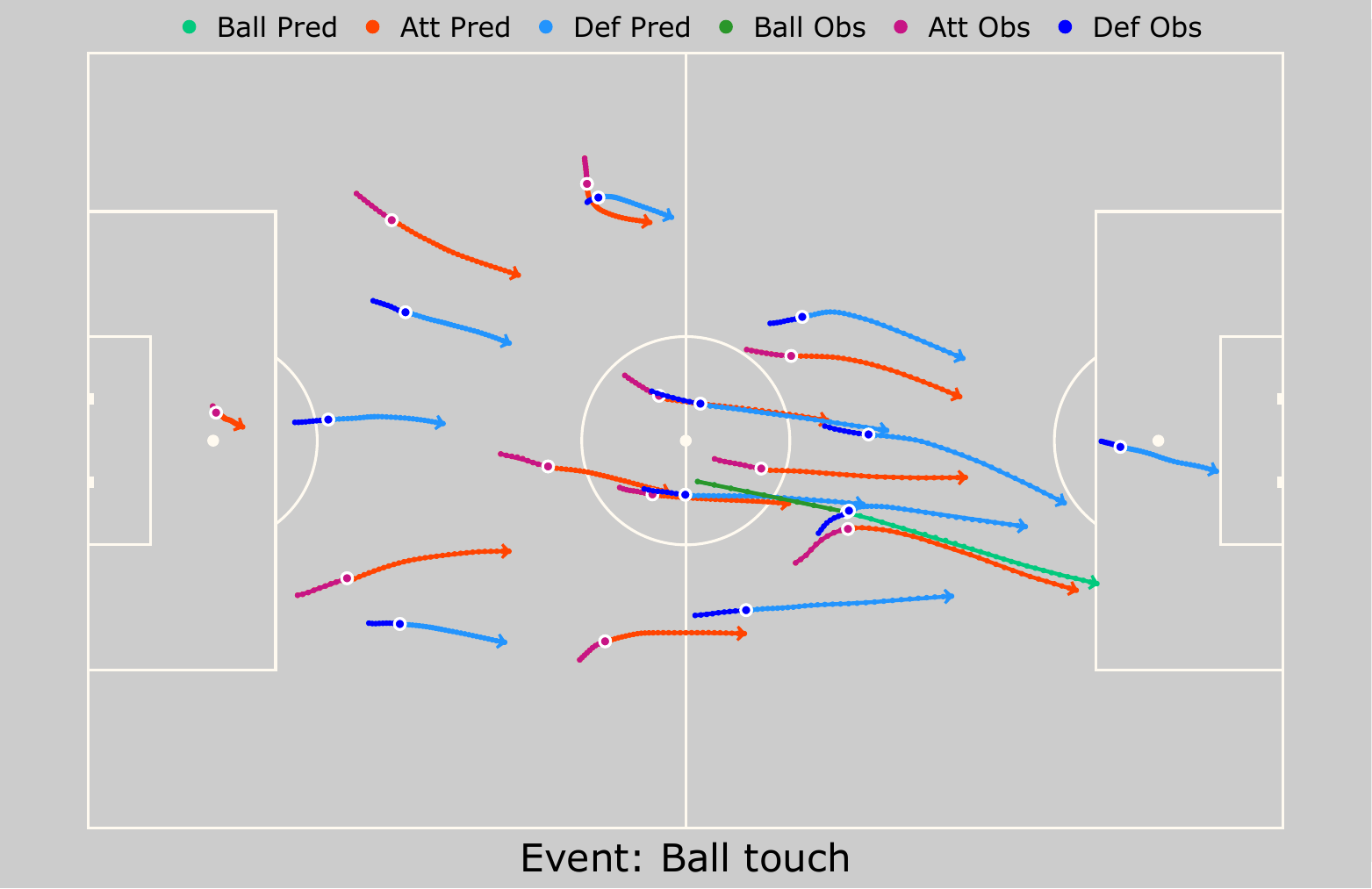}
  \end{minipage}\hfill
  \begin{minipage}[t]{0.16\textwidth}
    \centering
    \small \textbf{(f)} TacticGen-C\\
    \includegraphics[width=\linewidth]{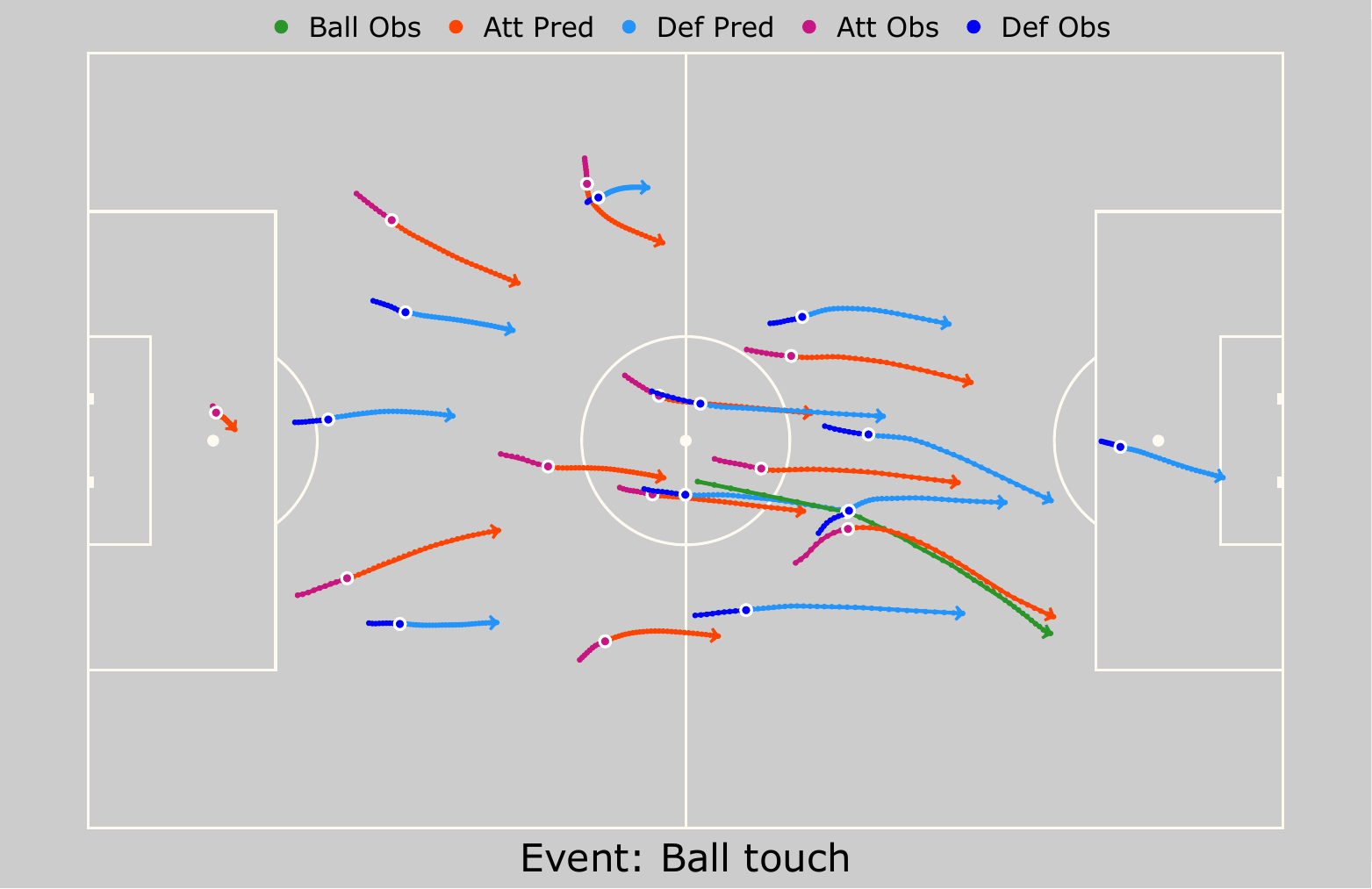}
  \end{minipage}

  \caption{{Ground truth trajectories for a ball touch event and the corresponding best-of-20 predictions generated by five methods.}}  \label{fig:pred-balltouch-39}
\end{figure}

\begin{figure}[htbp]
\vspace{-0.05in}
  \centering
  \resizebox{0.5\textwidth}{!}{%
    \begin{minipage}[t]{0.33\textwidth}
      \centering
      \small (a) TacticGen-P\\
      \includegraphics[width=\linewidth]{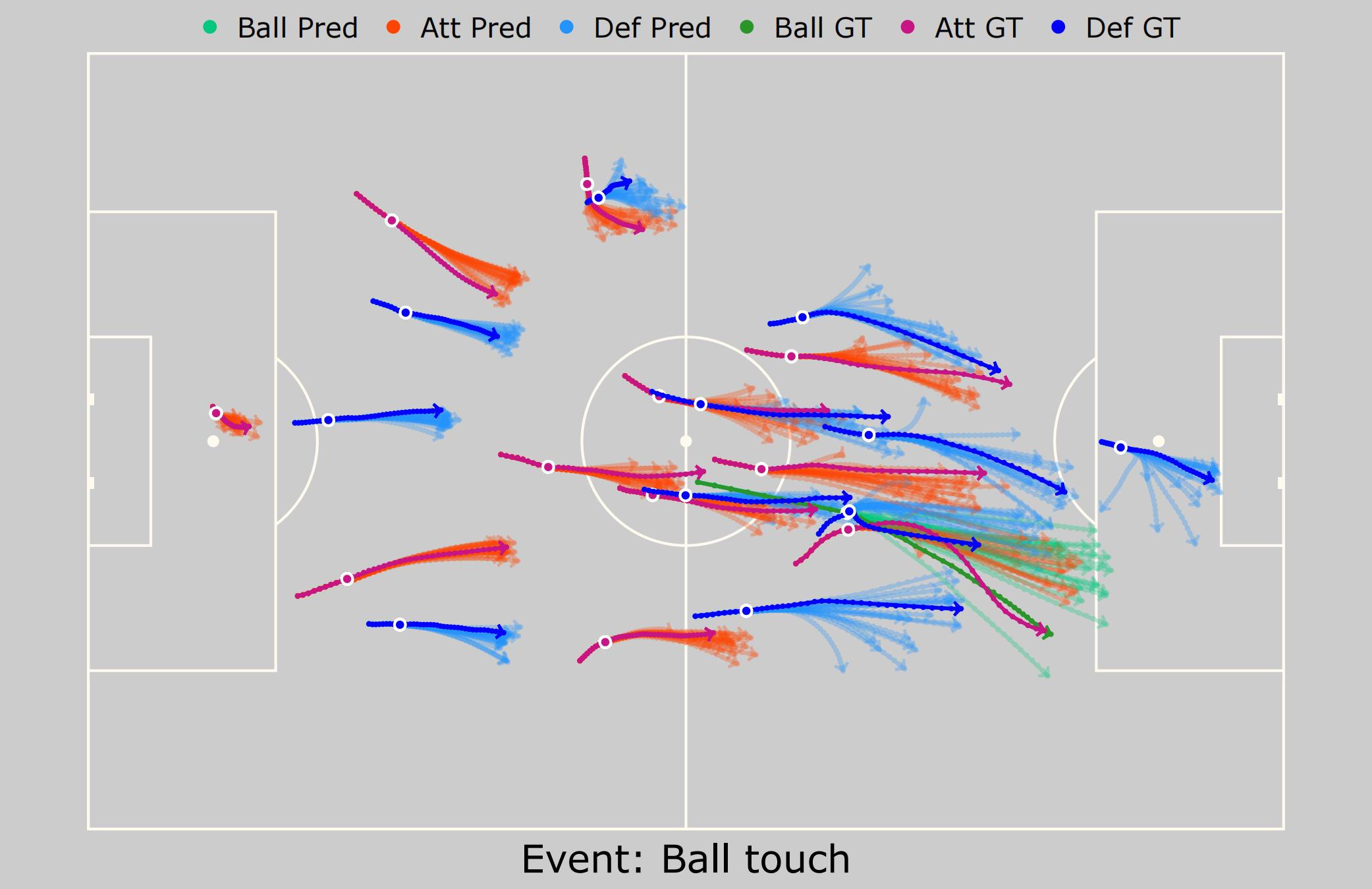}
    \end{minipage}
    \hspace{0.01\textwidth}
    \begin{minipage}[t]{0.33\textwidth}
      \centering
      \small (b) TacticGen-C\\
      \includegraphics[width=\linewidth]{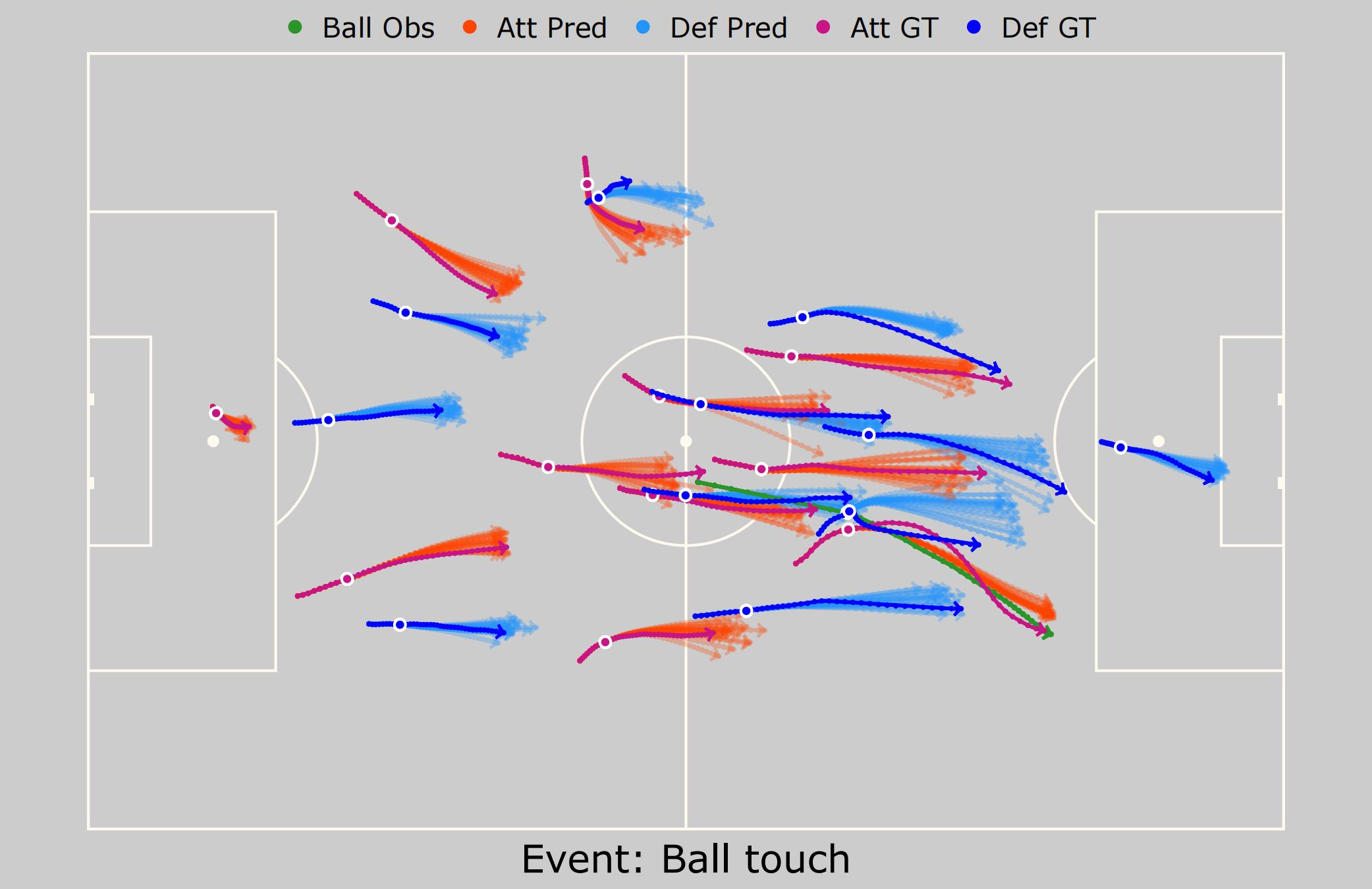}
    \end{minipage}%
  }
  \caption{{20 predicted trajectory samples by TacticGen variants for a ball touch event.}}
  \label{fig:footdiff-diversity-balltouch39}
  \vspace{-0.05in}
\end{figure}

Figure~\ref{fig:pred-balltouch-39} presents the best-of-20 trajectories generated by different models for a ball touch event, and Figure~\ref{fig:footdiff-diversity-balltouch39} illustrates the full set of 20 trajectories produced by TacticGen. We find that, for ball-touch events, TacticGen also generates realistic trajectories in which players exhibit greater attention to the ball compared to the baselines. Notably, TacticGen-C exhibits relatively limited diversity in this event, as conditioning on the complete ball trajectory makes player movements more deterministic in such a simple scenario.

\begin{figure}[htbp]
\vspace{-0.05in}
  \centering

  \begin{minipage}[t]{0.16\textwidth}
    \centering
    \small \textbf{(a)} Ground Truth\\
    \includegraphics[width=\linewidth]{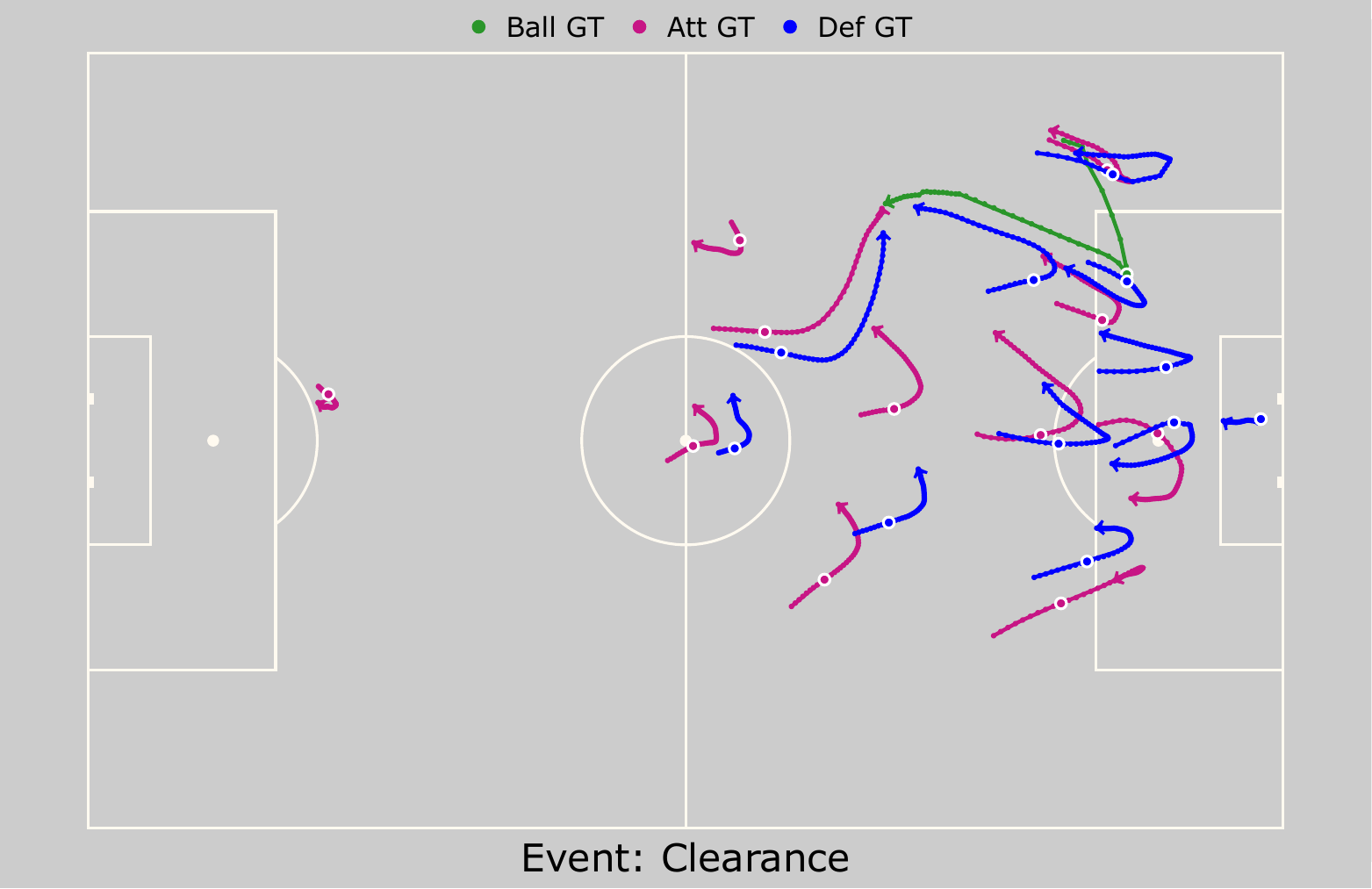}
  \end{minipage}\hfill
   \begin{minipage}[t]{0.16\textwidth}
    \centering
    \small \textbf{(b)} Diffuser\\
    \includegraphics[width=\linewidth]{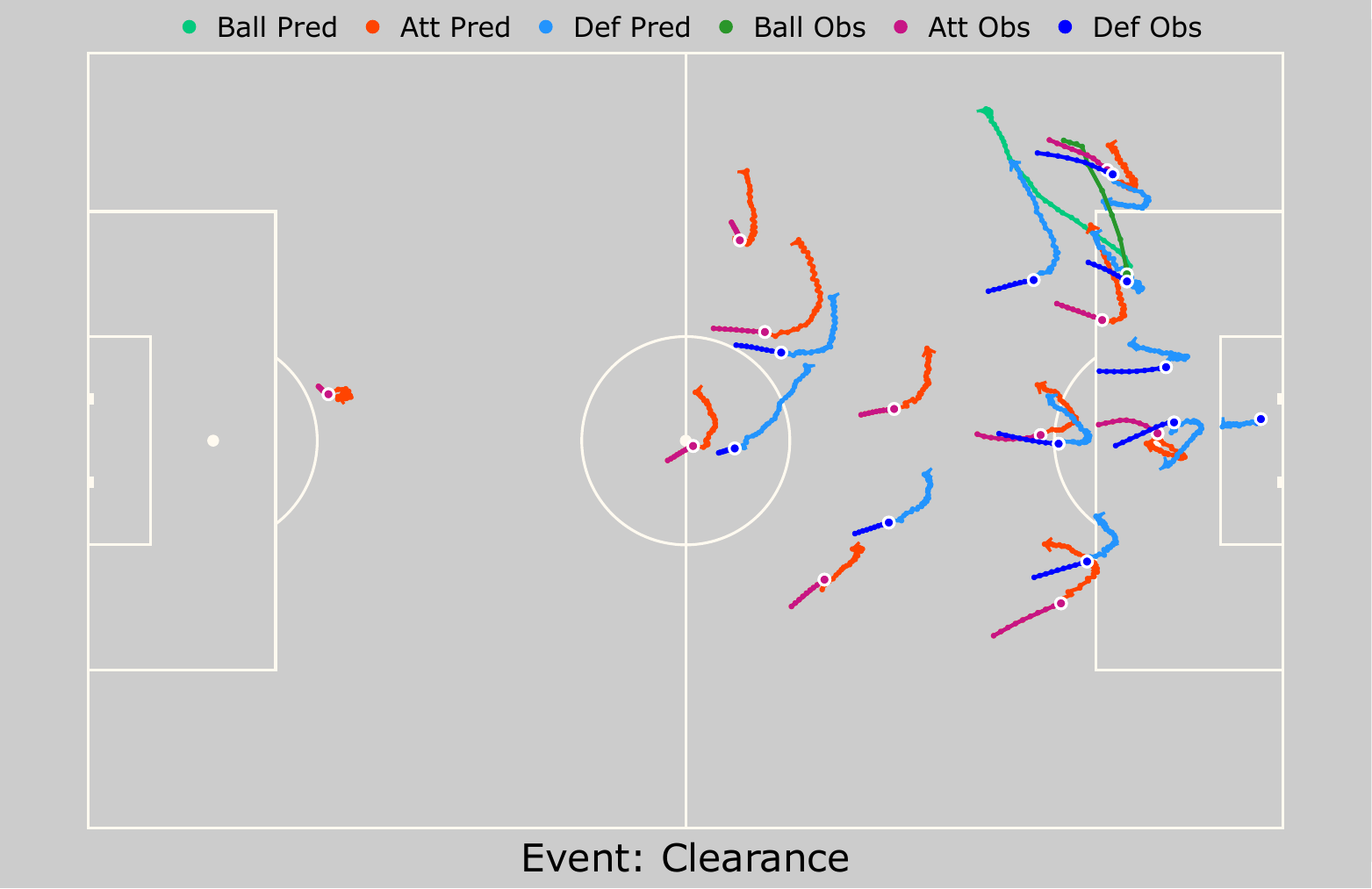}
  \end{minipage}\hfill
  \begin{minipage}[t]{0.16\textwidth}
    \centering
    \small \textbf{(c)} MID\\
    \includegraphics[width=\linewidth]{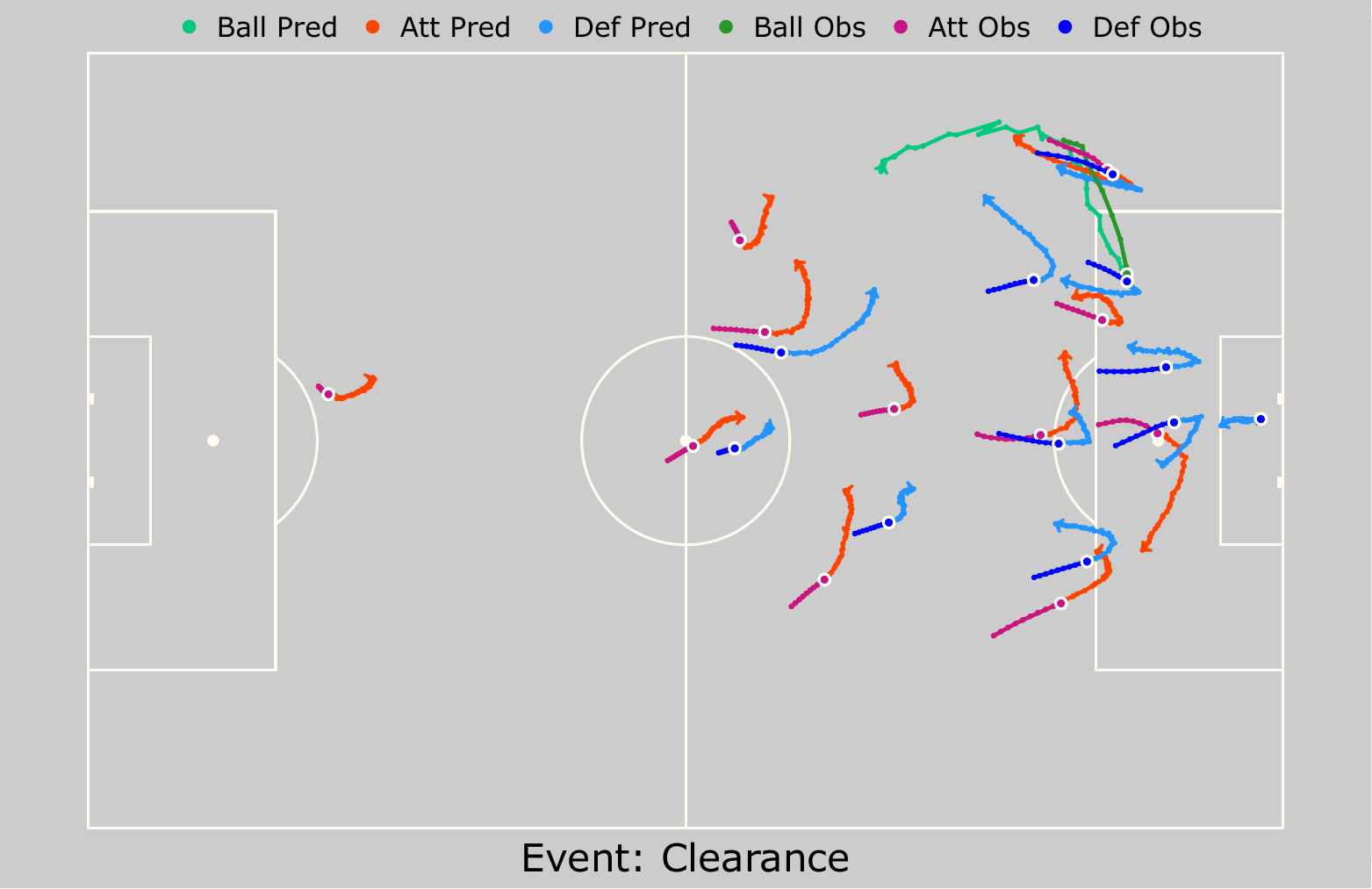}
  \end{minipage}

  \vspace{0.5em} 

  \begin{minipage}[t]{0.16\textwidth}
    \centering
    \small \textbf{(d)} MADiff\\
    \includegraphics[width=\linewidth]{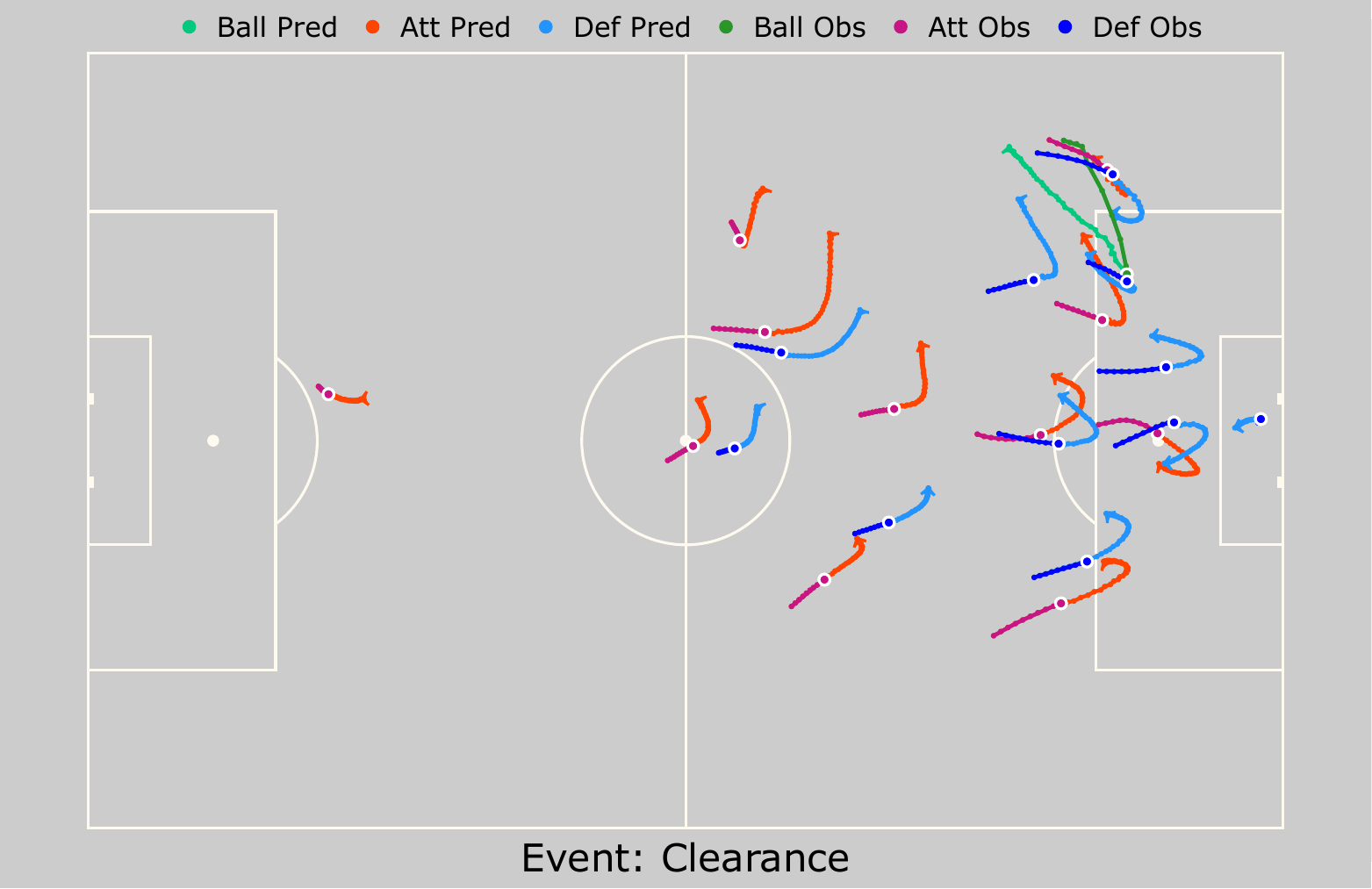}
  \end{minipage}\hfill
  \begin{minipage}[t]{0.16\textwidth}
    \centering
    \small \textbf{(e)} TacticGen-P\\
    \includegraphics[width=\linewidth]{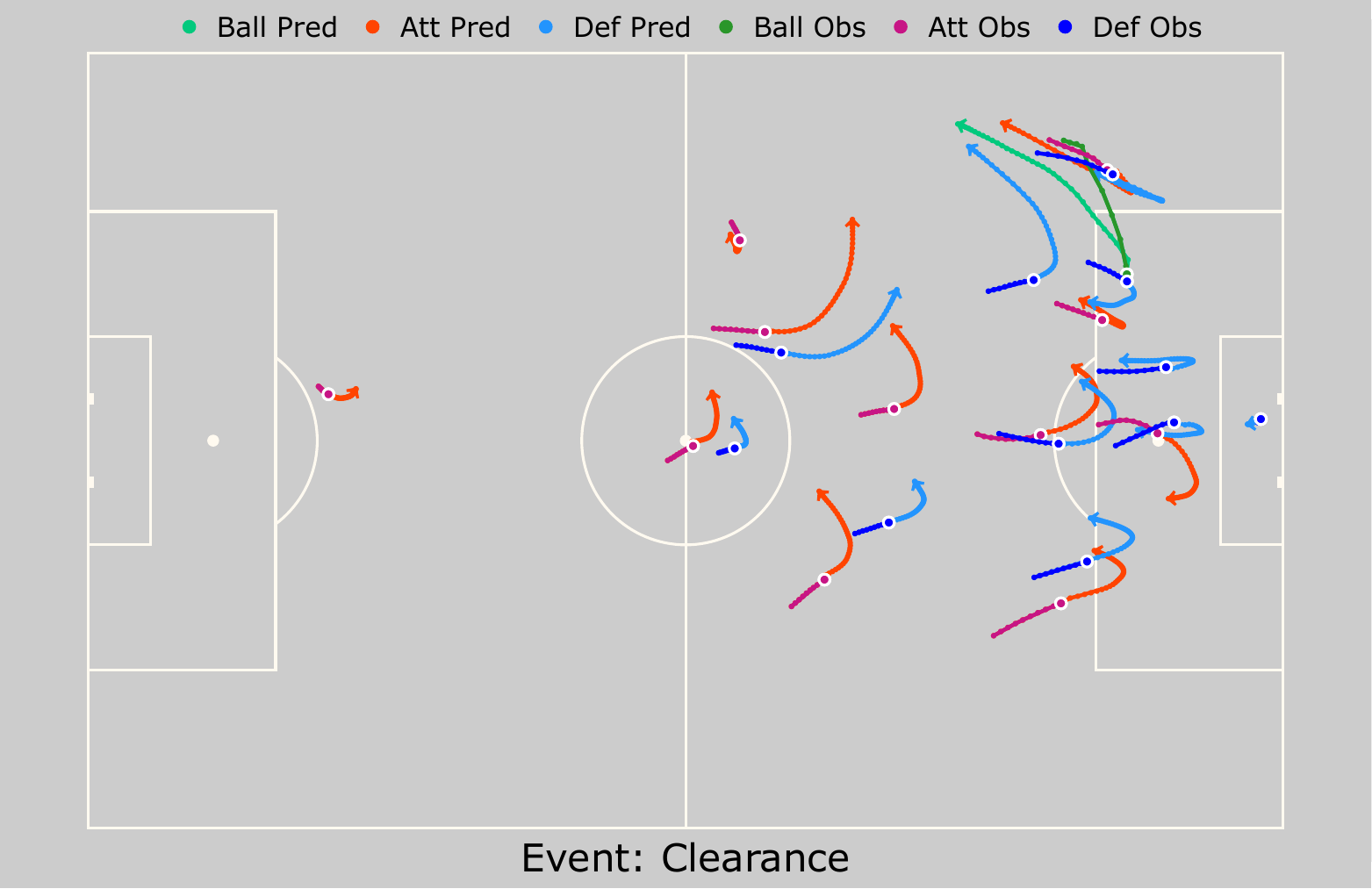}
  \end{minipage}\hfill
  \begin{minipage}[t]{0.16\textwidth}
    \centering
    \small \textbf{(f)} TacticGen-C\\
    \includegraphics[width=\linewidth]{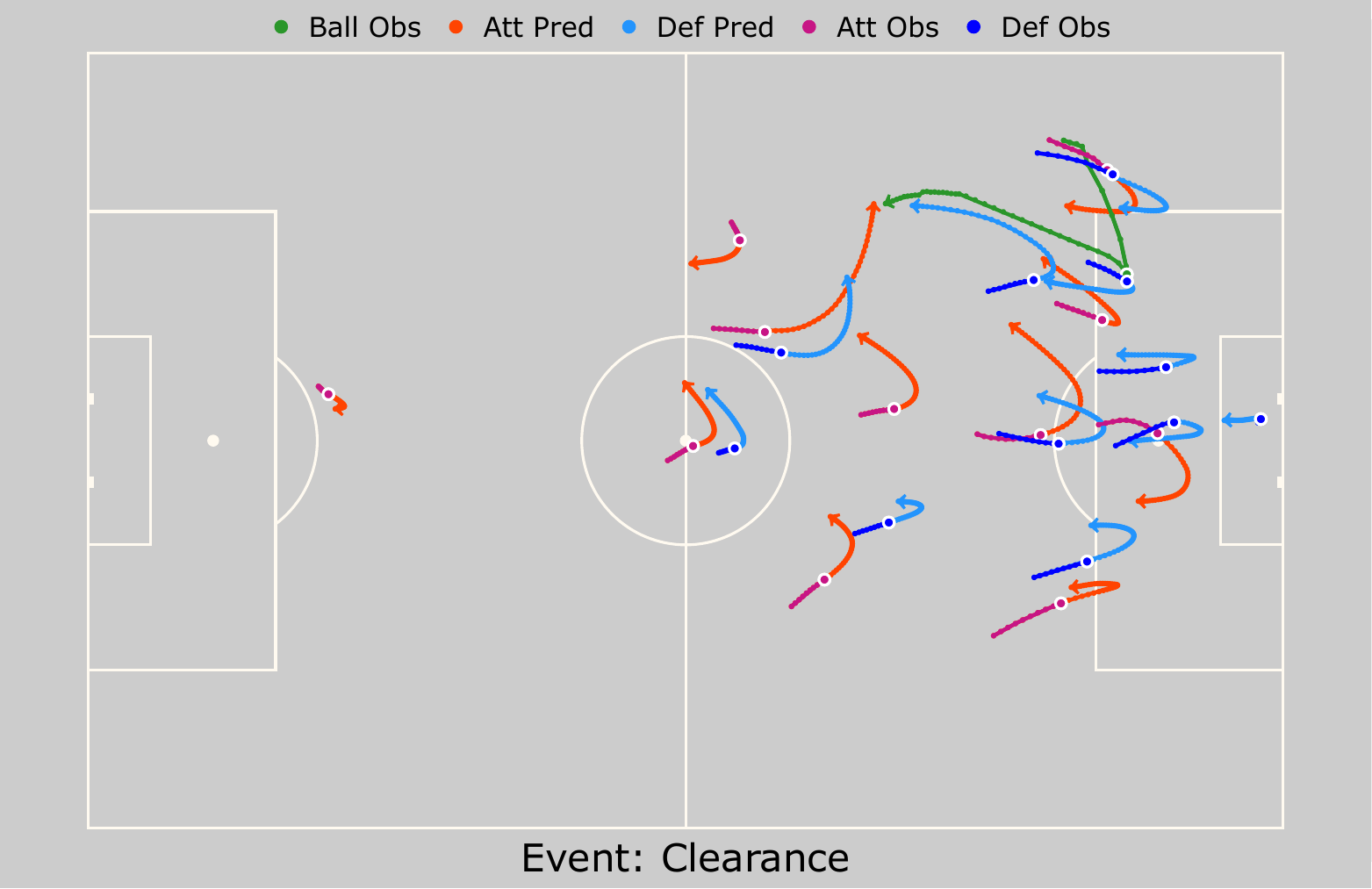}
  \end{minipage}

  \caption{{Ground truth trajectories for a clearance event and the corresponding best-of-20 predictions generated by five methods.}} \label{fig:pred-clearance}
\end{figure}

\begin{figure}[htbp]
\vspace{-0.2in}
  \centering
  \resizebox{0.5\textwidth}{!}{%
    \begin{minipage}[t]{0.33\textwidth}
      \centering
      \small \textbf{(a)} TacticGen-P\\
      \includegraphics[width=\linewidth]{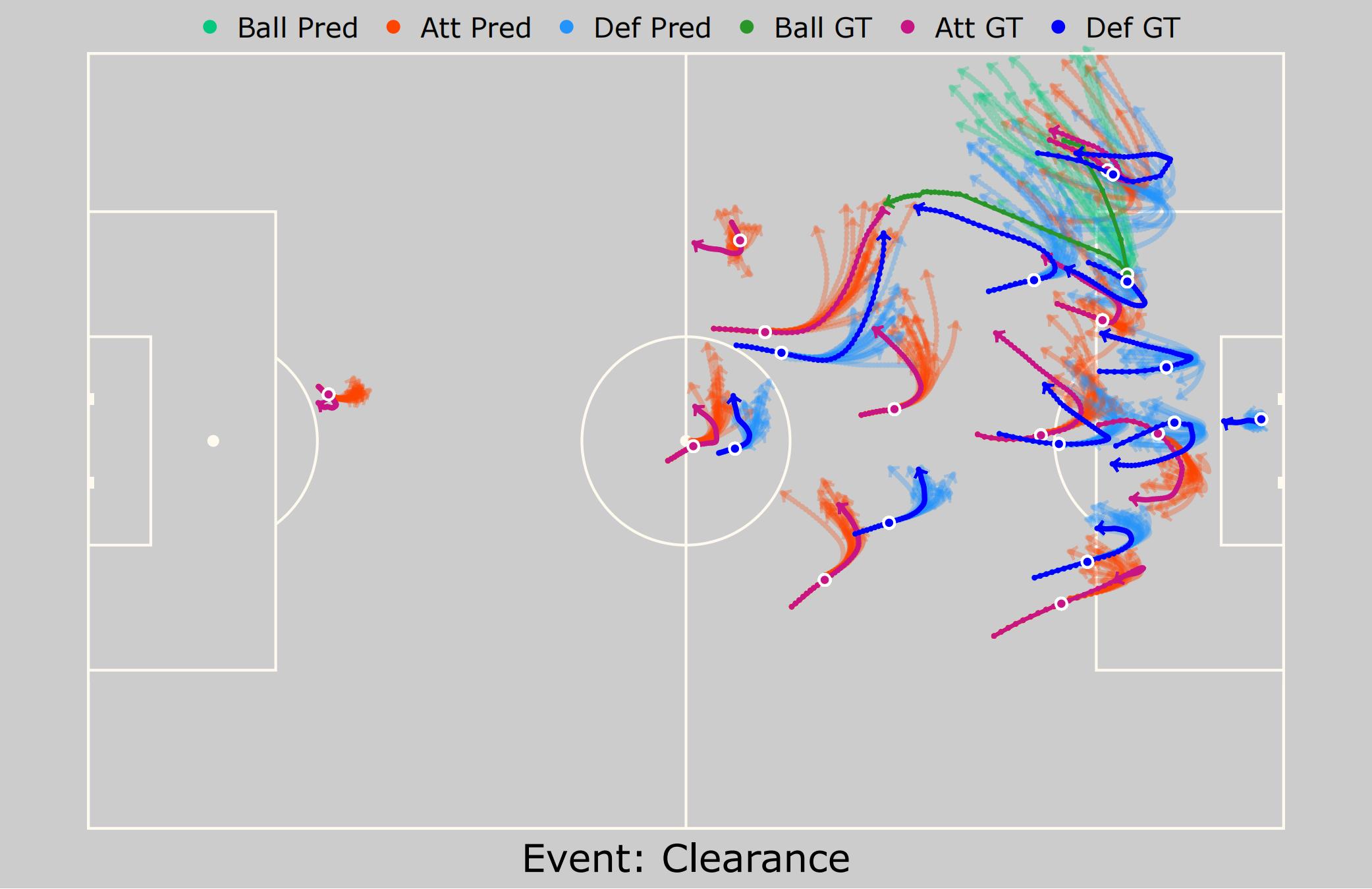}
    \end{minipage}
    \hspace{0.01\textwidth}
    \begin{minipage}[t]{0.33\textwidth}
      \centering
      \small \textbf{(b)} TacticGen-C\\
      \includegraphics[width=\linewidth]{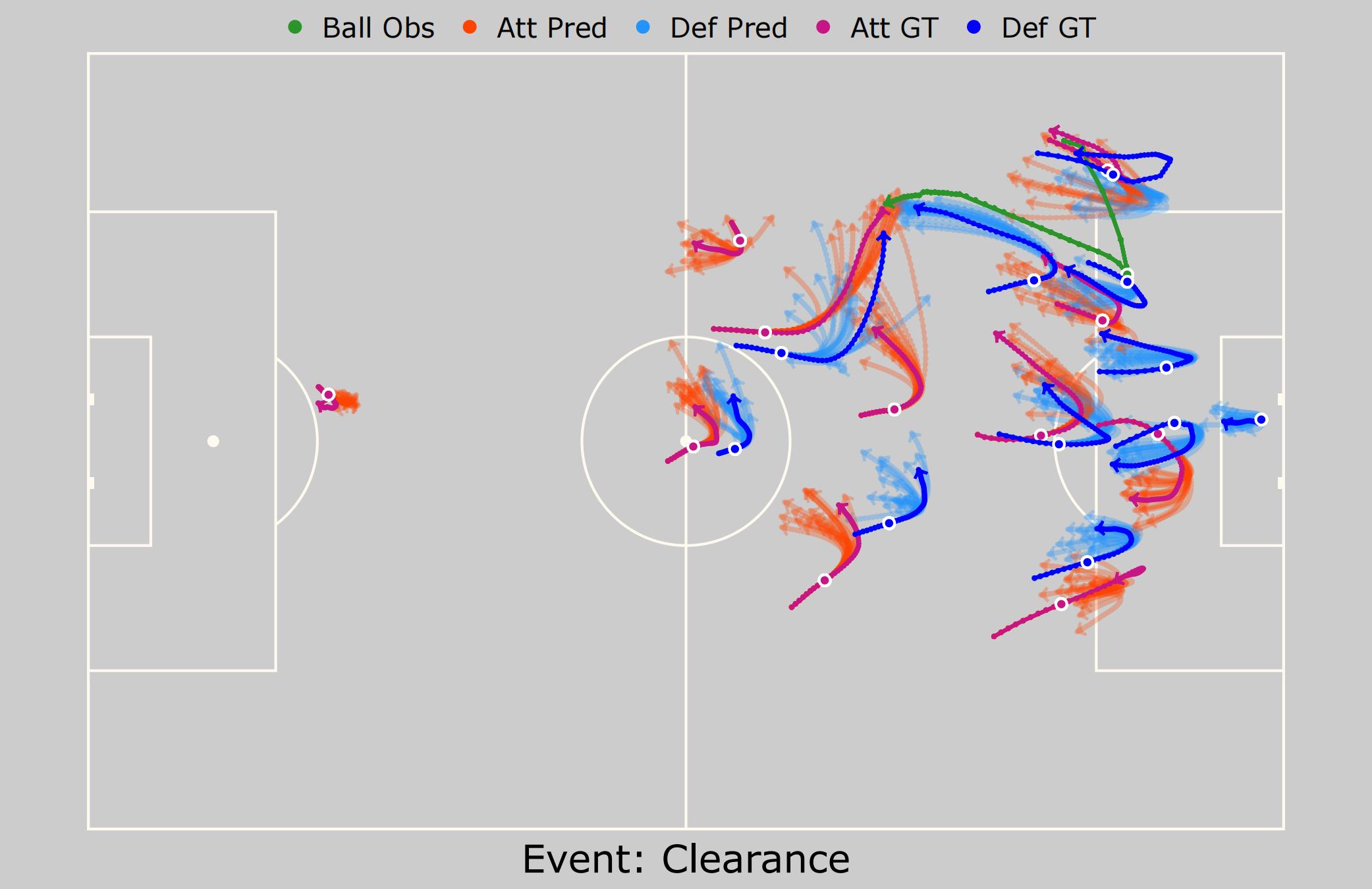}
    \end{minipage}%
  }
  \caption{{20 predicted trajectory samples by TacticGen variants for a clearance event.}}
  \label{fig:footdiff-diversity-clearance}
  \vspace{-0.05in}
\end{figure}

Figure~\ref{fig:pred-clearance} presents the best-of-20 trajectories generated by different models for a clearance event, and Figure~\ref{fig:footdiff-diversity-clearance} illustrates the full set of 20 trajectories produced by TacticGen. This event demonstrates TacticGen’s ability to generate more than just left-to-right movements: when the ball is played back, players still attend to it and react accordingly.

\section{More Experiments on Tactic Generation}\label{sec:more-exp-guidance}

\subsection{Results of Alternative Team Behaviors}\label{sec:replay-reactive-results}
\textbf{Replayed trajectories from original prediction.} We provide additional results where the unguided team follows the replayed data from the model’s original prediction. This setting is particularly valuable in scenarios where no real data are available for either team, so it is necessary to anticipate their trajectories. By first leveraging the model’s strong predictive capability, we can infer the teams’ movements and subsequently guide one side to achieve better tactics based on these anticipated trajectories. This approach enables the application of tactical adjustments even in partially observed or data-limited situations. Such a setup closely mirrors practical use cases in football analytics, where only incomplete tracking data are often available.

\begin{figure}[htbp]
  \centering

  \begin{minipage}[t]{0.16\textwidth}
    \centering
    \small \textbf{(a)} No Guidance
    \includegraphics[width=\linewidth]{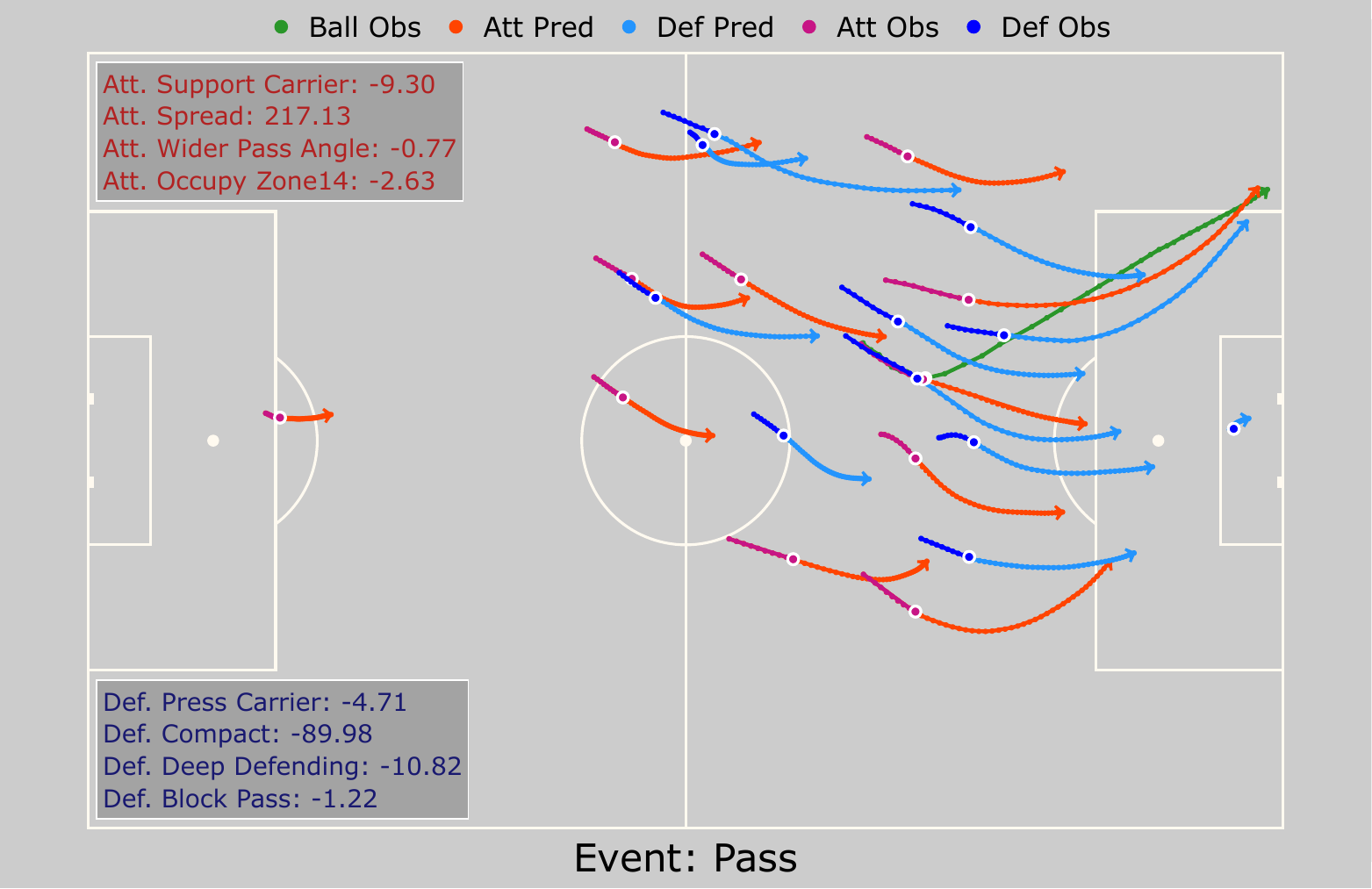}
  \end{minipage}\hfill
  \begin{minipage}[t]{0.16\textwidth}
    \centering
    \small \textbf{(b)} Att. Rule Guid.
    \includegraphics[width=\linewidth]{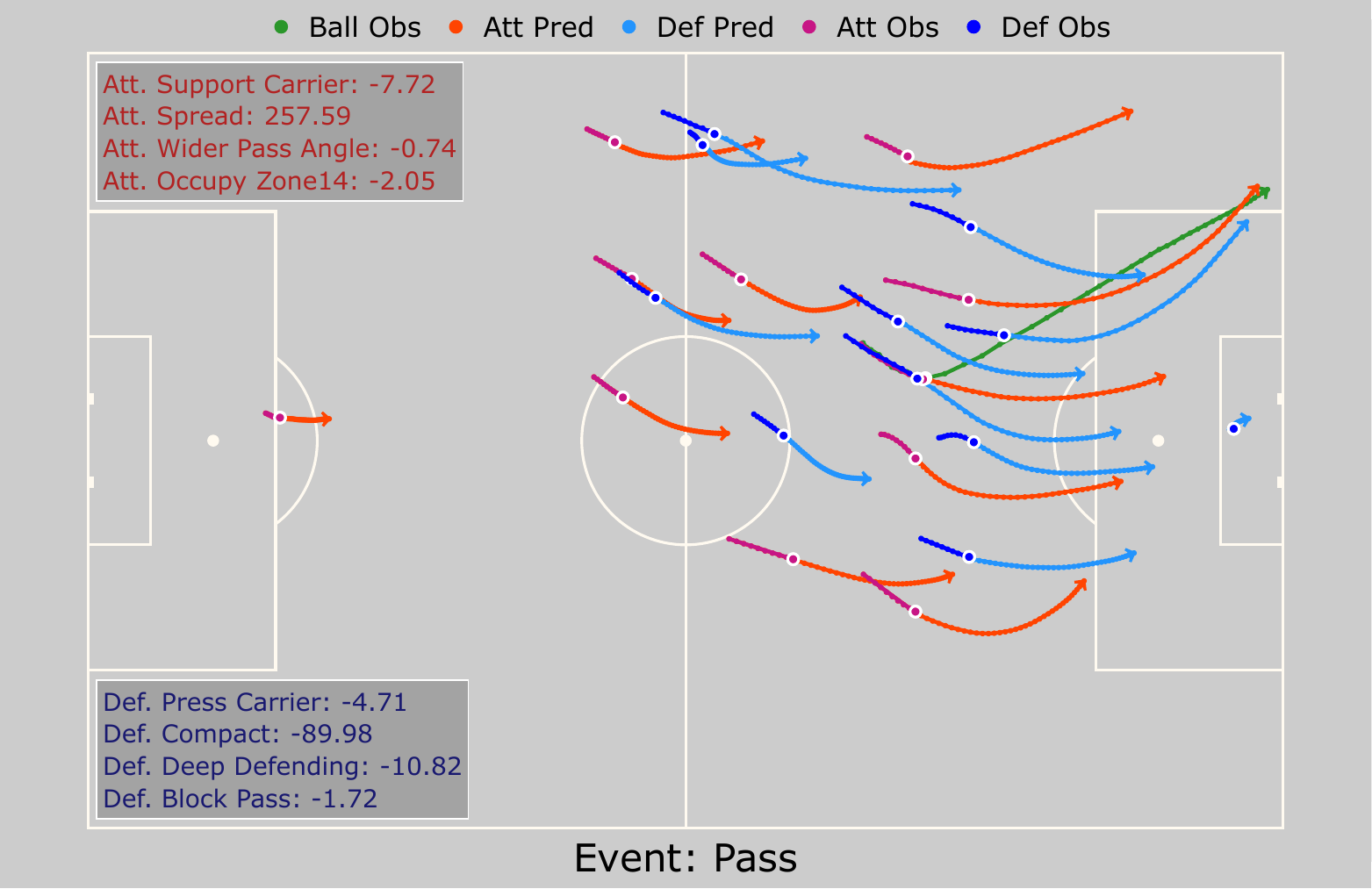}
  \end{minipage}\hfill
  \begin{minipage}[t]{0.16\textwidth}
    \centering
    \small \textbf{(c)} Def. Rule Guid.
    \includegraphics[width=\linewidth]{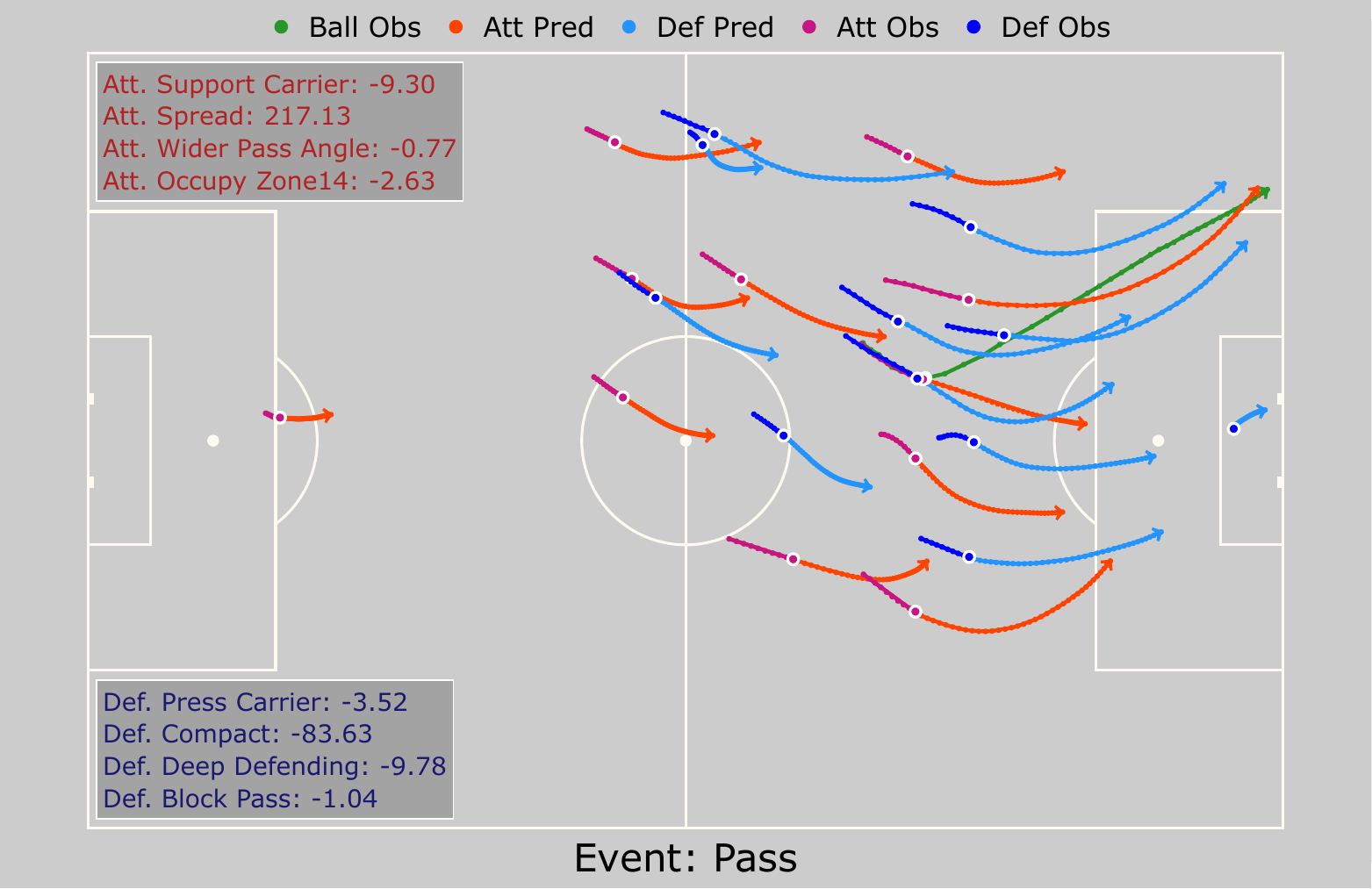}
  \end{minipage}

  \caption{{Trajectories generated by TacticGen for a pass event under rule-based guidance, where the unguided team follows the replayed data from the model's original prediction.}}\label{fig:fce-rule-guide-pass46-replaypred}
\end{figure}

\begin{figure}[htbp]
  \vspace{-0.05in}
  \centering

  \begin{minipage}[t]{0.16\textwidth}
    \centering
    \small \textbf{(a)} No Guidance\\
    \includegraphics[width=\linewidth]{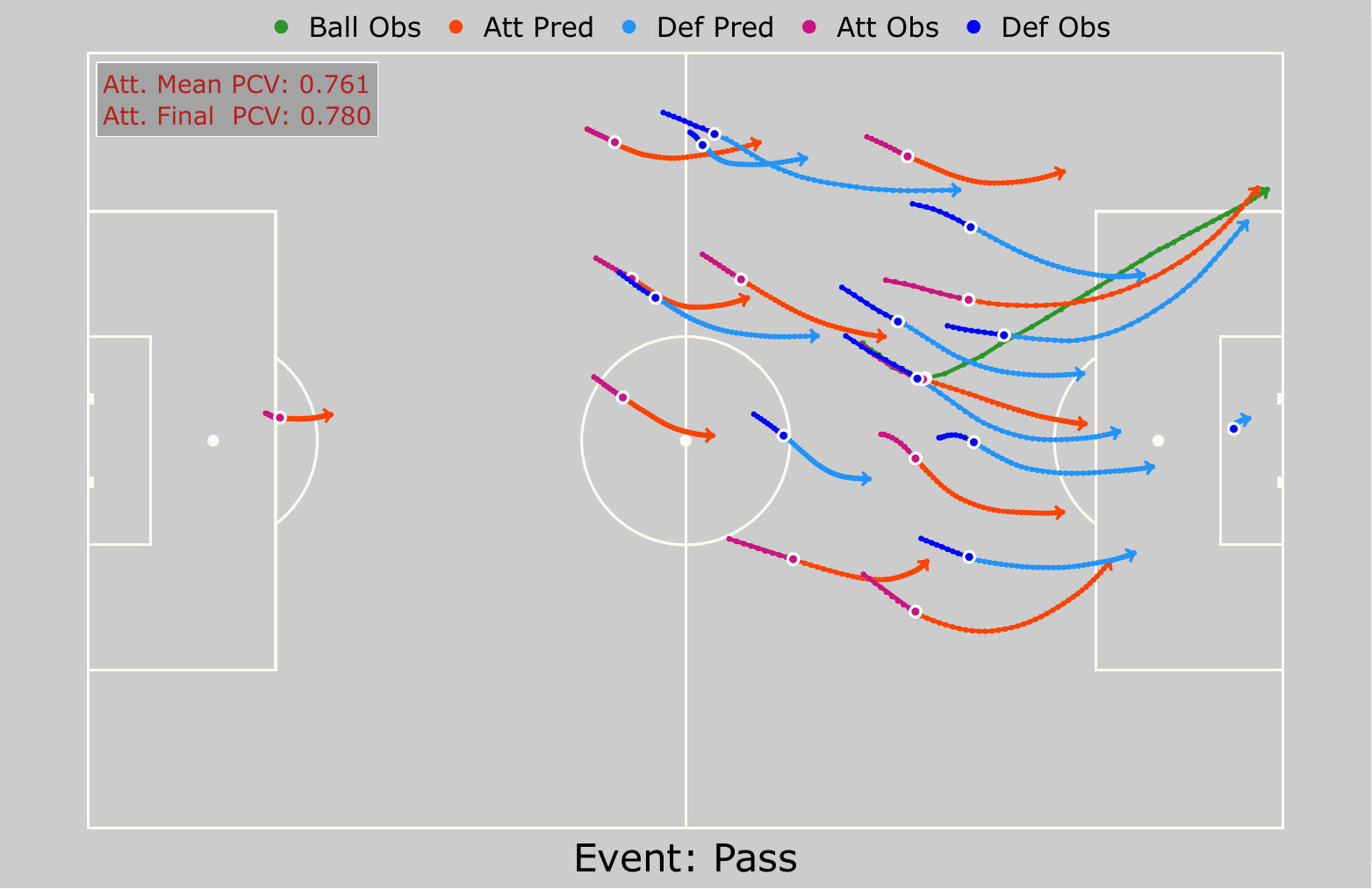}
  \end{minipage}\hfill
  \begin{minipage}[t]{0.16\textwidth}
    \centering
    \small \textbf{(b)} Att. High PCV\\
    \includegraphics[width=\linewidth]{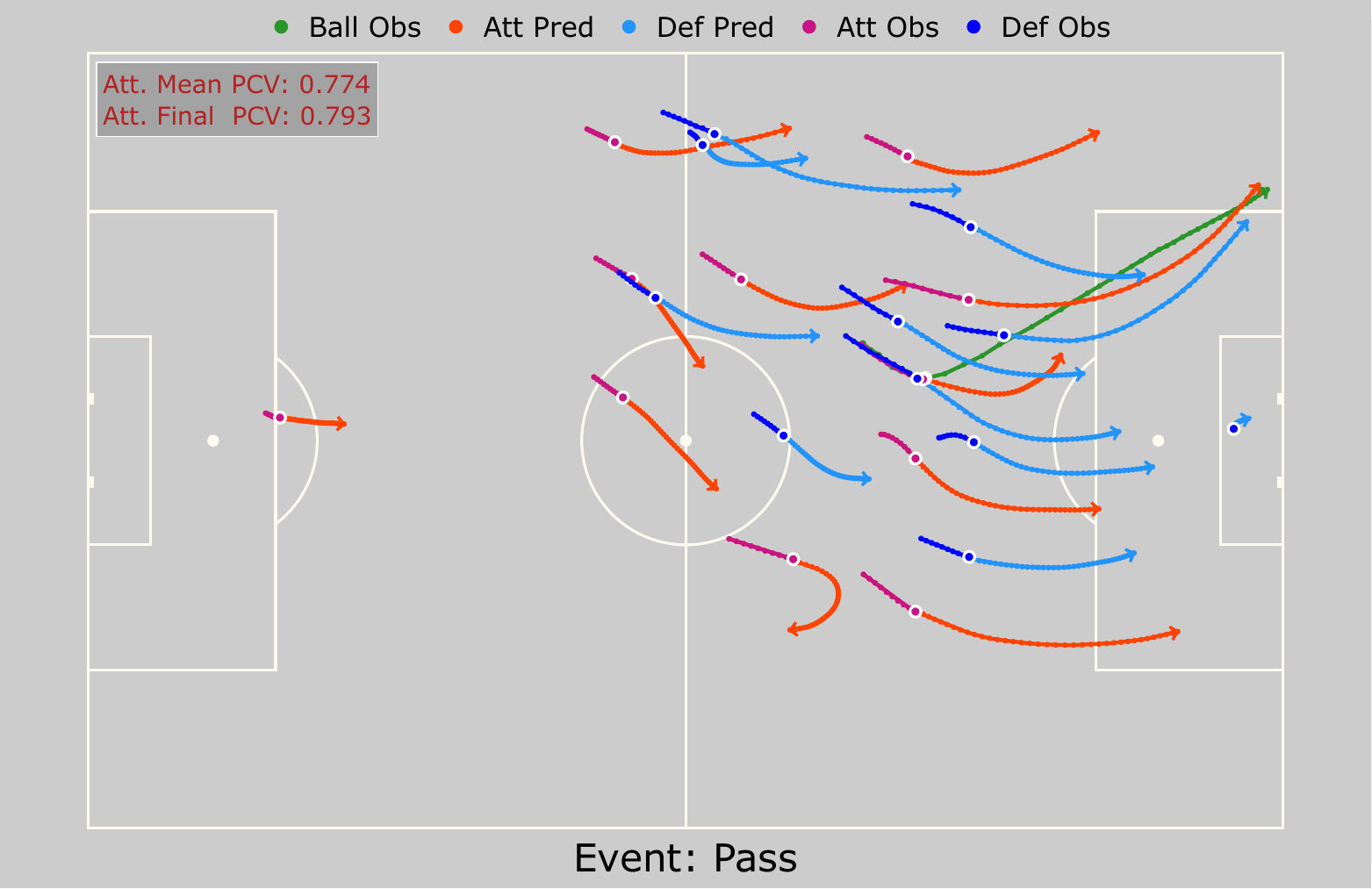}
  \end{minipage}\hfill
  \begin{minipage}[t]{0.16\textwidth}
    \centering
    \small \textbf{(c)} Def. High PCV\\
    \includegraphics[width=\linewidth]{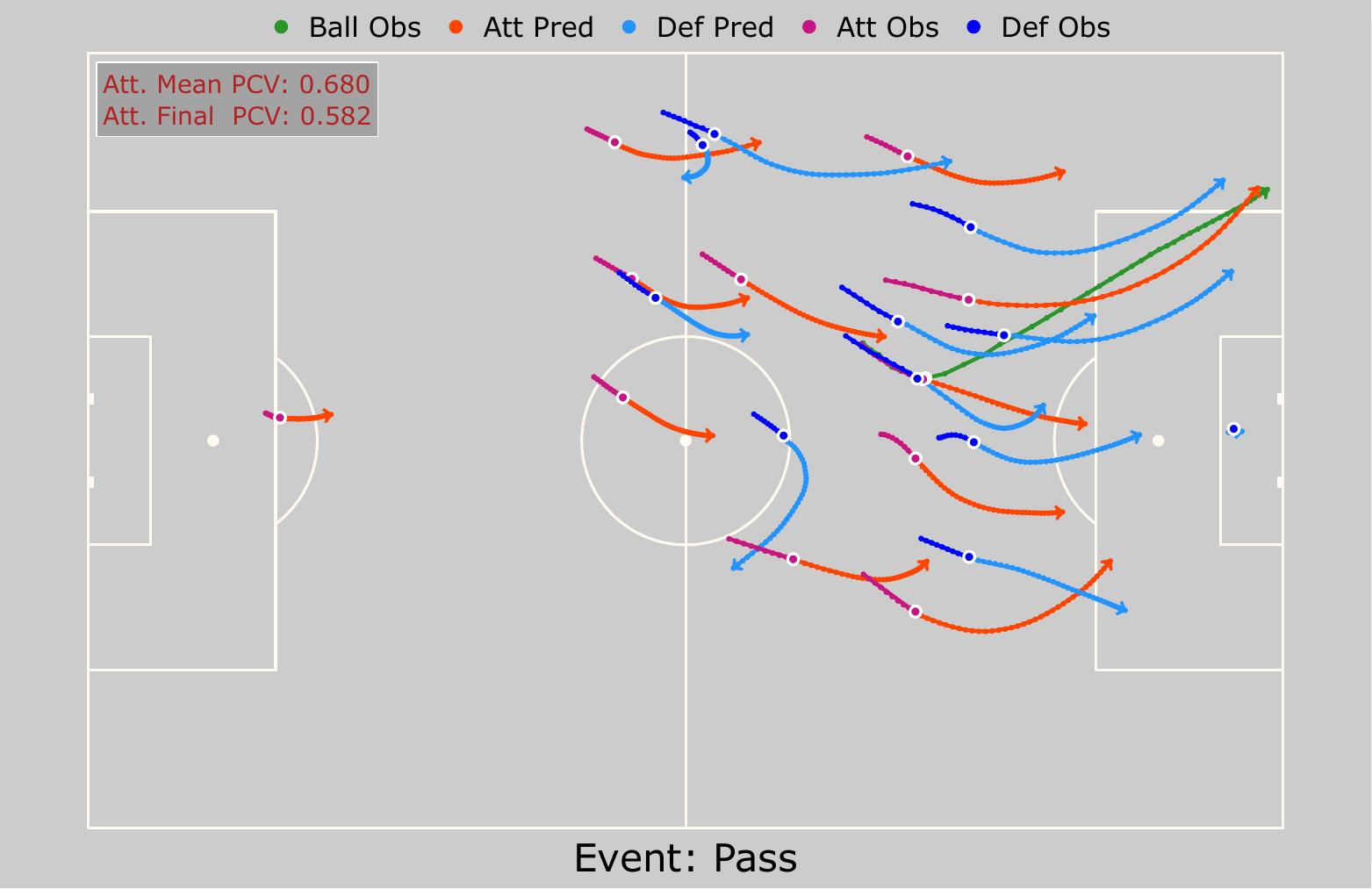}
  \end{minipage}

  \vspace{0.5em}

  \begin{minipage}[t]{0.16\textwidth}
    \centering
    \includegraphics[width=\linewidth]{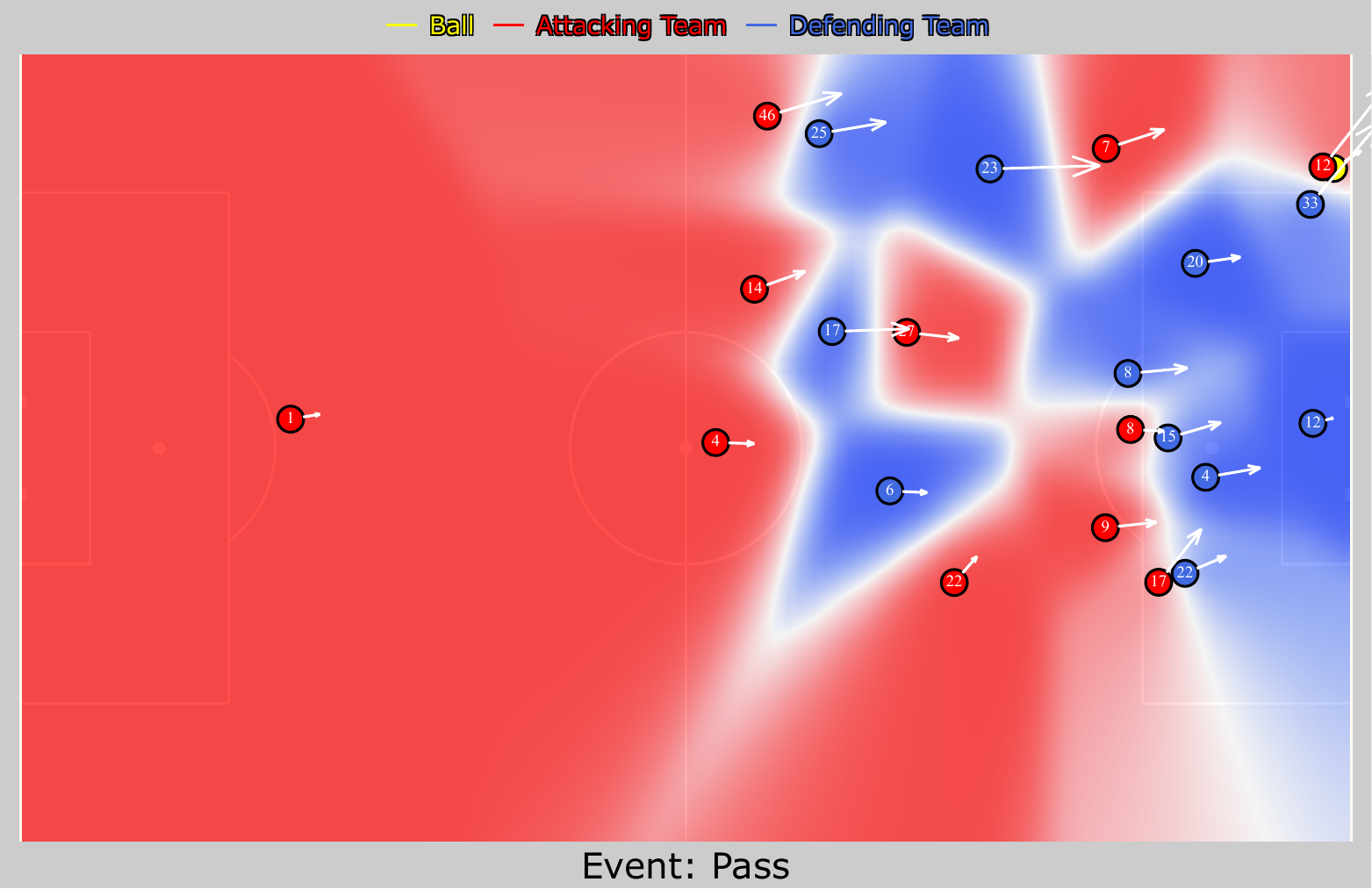}
  \end{minipage}\hfill
  \begin{minipage}[t]{0.16\textwidth}
    \centering
    \includegraphics[width=\linewidth]{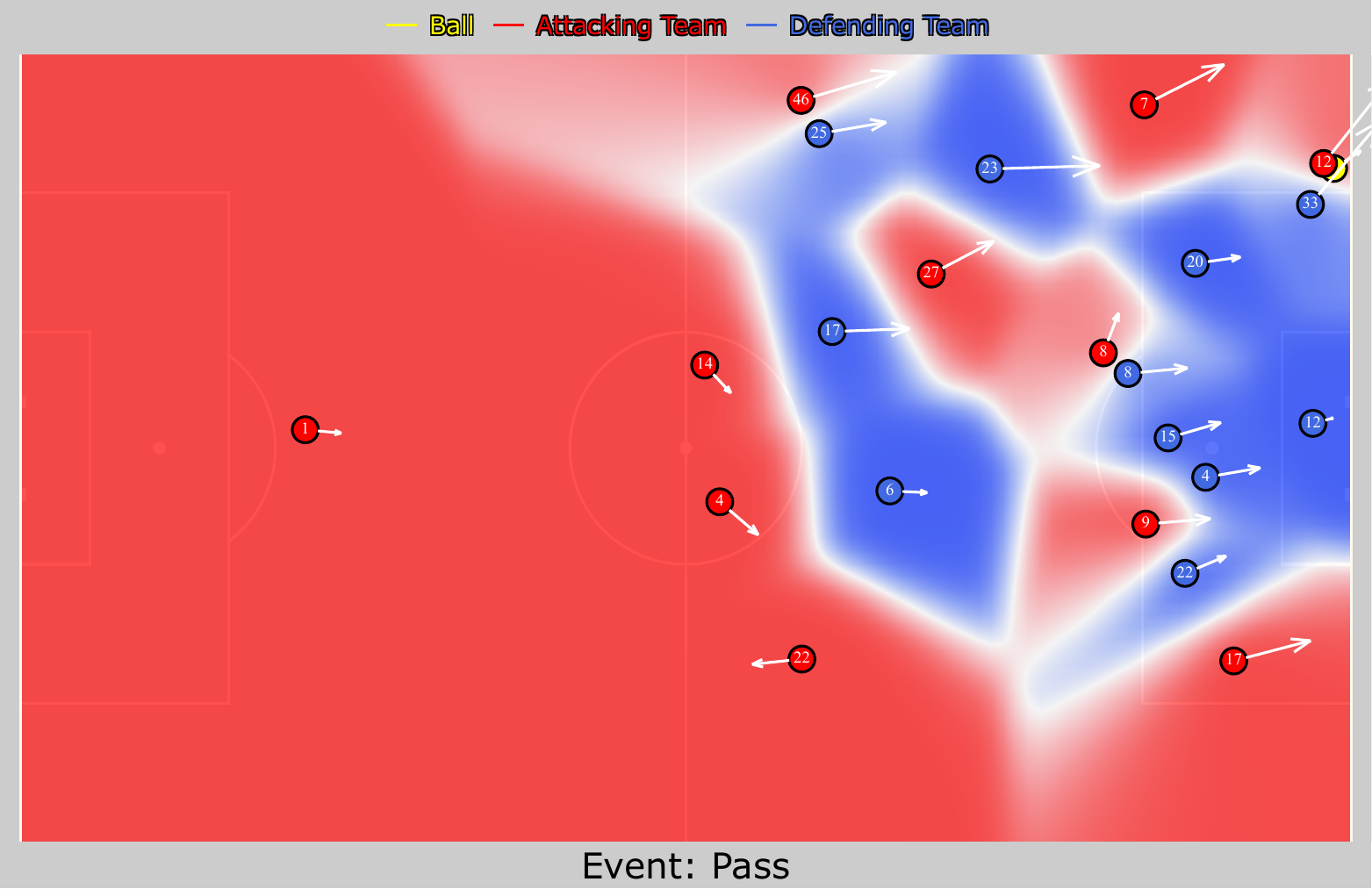}
  \end{minipage}\hfill
  \begin{minipage}[t]{0.16\textwidth}
    \centering
    \includegraphics[width=\linewidth]{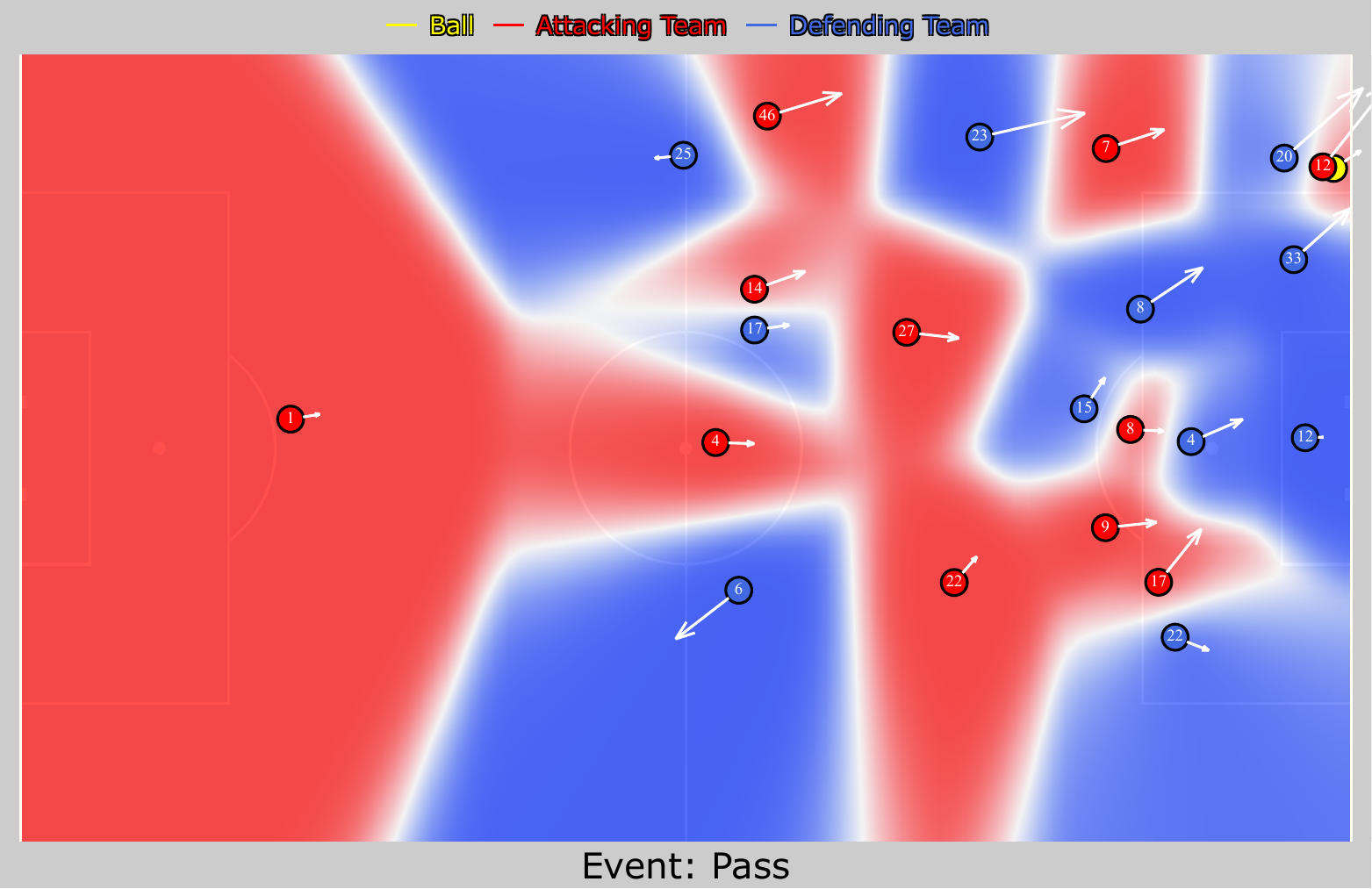}
  \end{minipage}

  \vspace{-0.1in}
  \caption{{Visualizations of trajectories (top) and pitch control values (PCV) at the final frame (bottom) generated by TacticGen for a pass event under pitch control guidance, where the unguided team follows the replayed data from the model's original prediction.}}\label{fig:fce-rule-guide-pcm-pass46-replaypred}
  \vspace{-0.1in}
\end{figure}

\begin{figure}[htbp]
  \vspace{-0.1in}
  \centering

  \begin{minipage}[t]{0.16\textwidth}
    \centering
    \includegraphics[width=\linewidth]{figures/prediction/Pass_986528_1816044_46/tactgen-c/Epoch100_GFalse_SBJ_L1.135732_Pred.pdf}
  \end{minipage}\hfill
  \begin{minipage}[t]{0.16\textwidth}
    \centering
    \includegraphics[width=\linewidth]{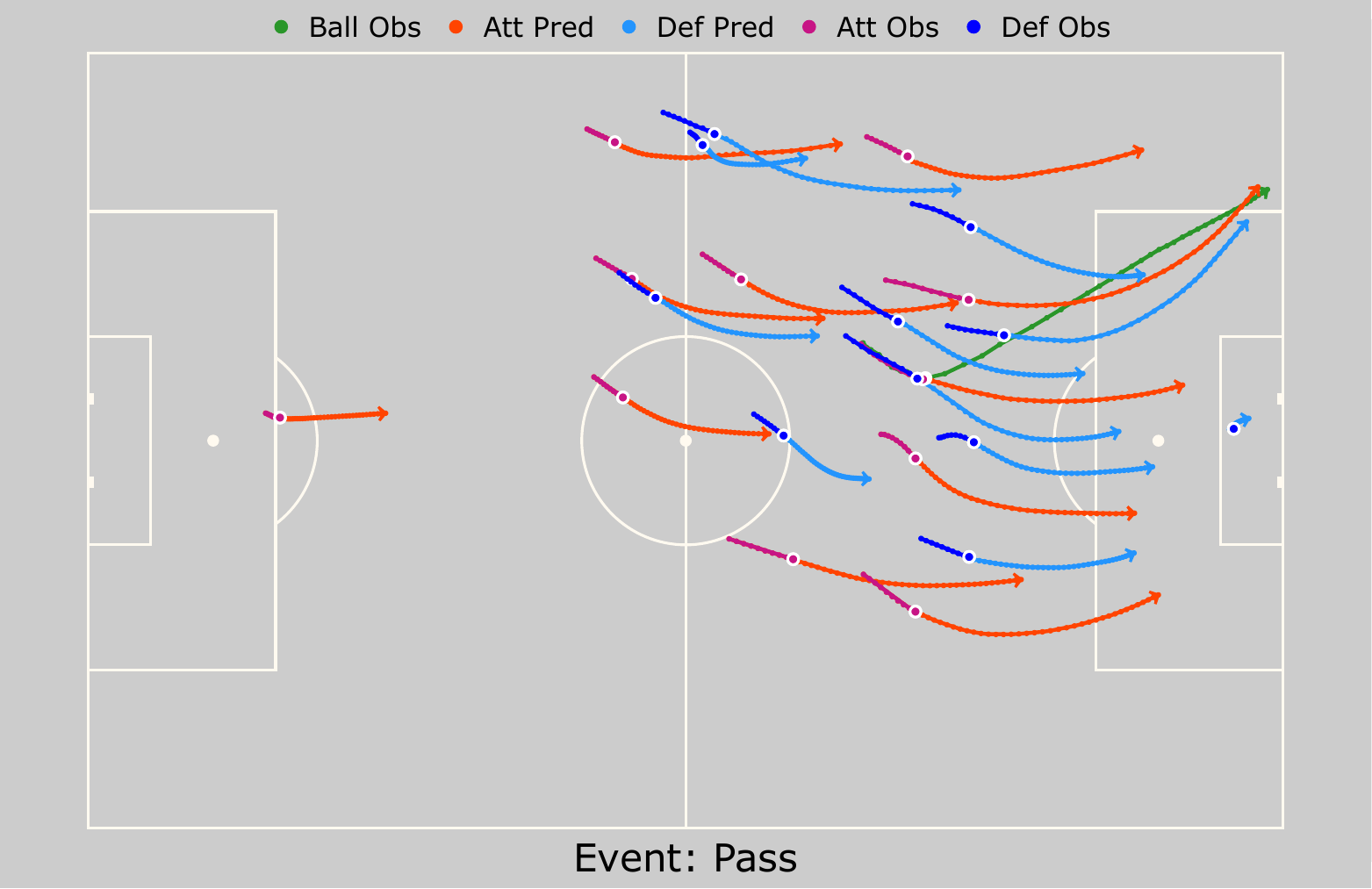}
  \end{minipage}\hfill
  \begin{minipage}[t]{0.16\textwidth}
    \centering
    \includegraphics[width=\linewidth]{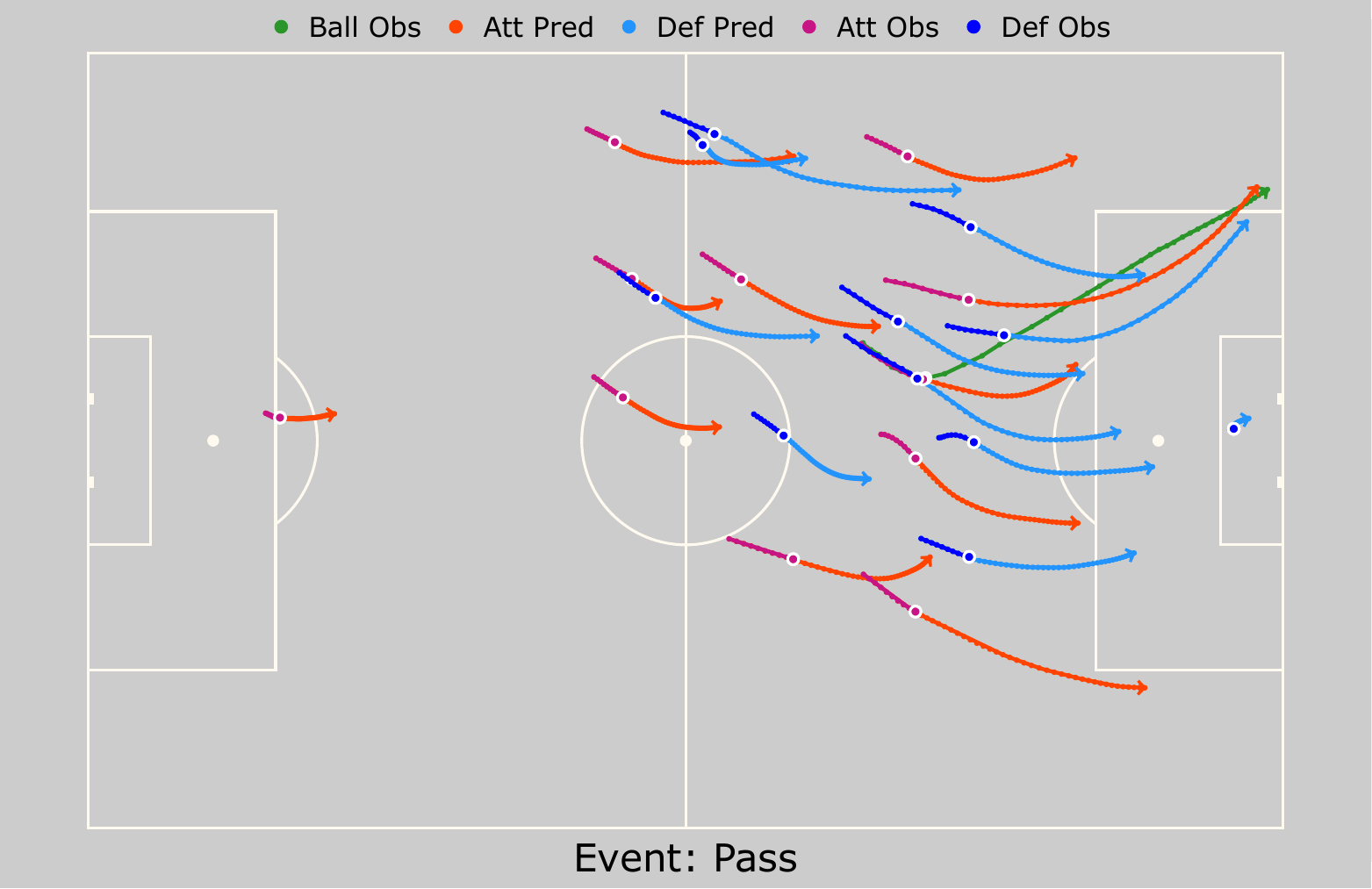}
  \end{minipage}

  \vspace{-0.1in}
  \caption{{Trajectories generated by TacticGen for a pass event under different guidance functions prompted by LLM, where the unguided team follows the replayed data from the model's original prediction.} \textbf{Left} Unguided generation. \textbf{Middle} Guided generation with the prompt, “Make the attacking team move forward more aggressively.” \textbf{Right} Guided generation with the prompt, “Make the right bottom player drift into the corner to stretch the defense and open up more space.”}
  \label{fig:fce-rule-guide-llm-pass46-replaypred}
\end{figure}

\begin{figure}[htbp]
  \centering

  \begin{minipage}[t]{0.158\textwidth}
    \centering
    \small \textbf{(a)} No Guidance\\
    \includegraphics[width=\linewidth]{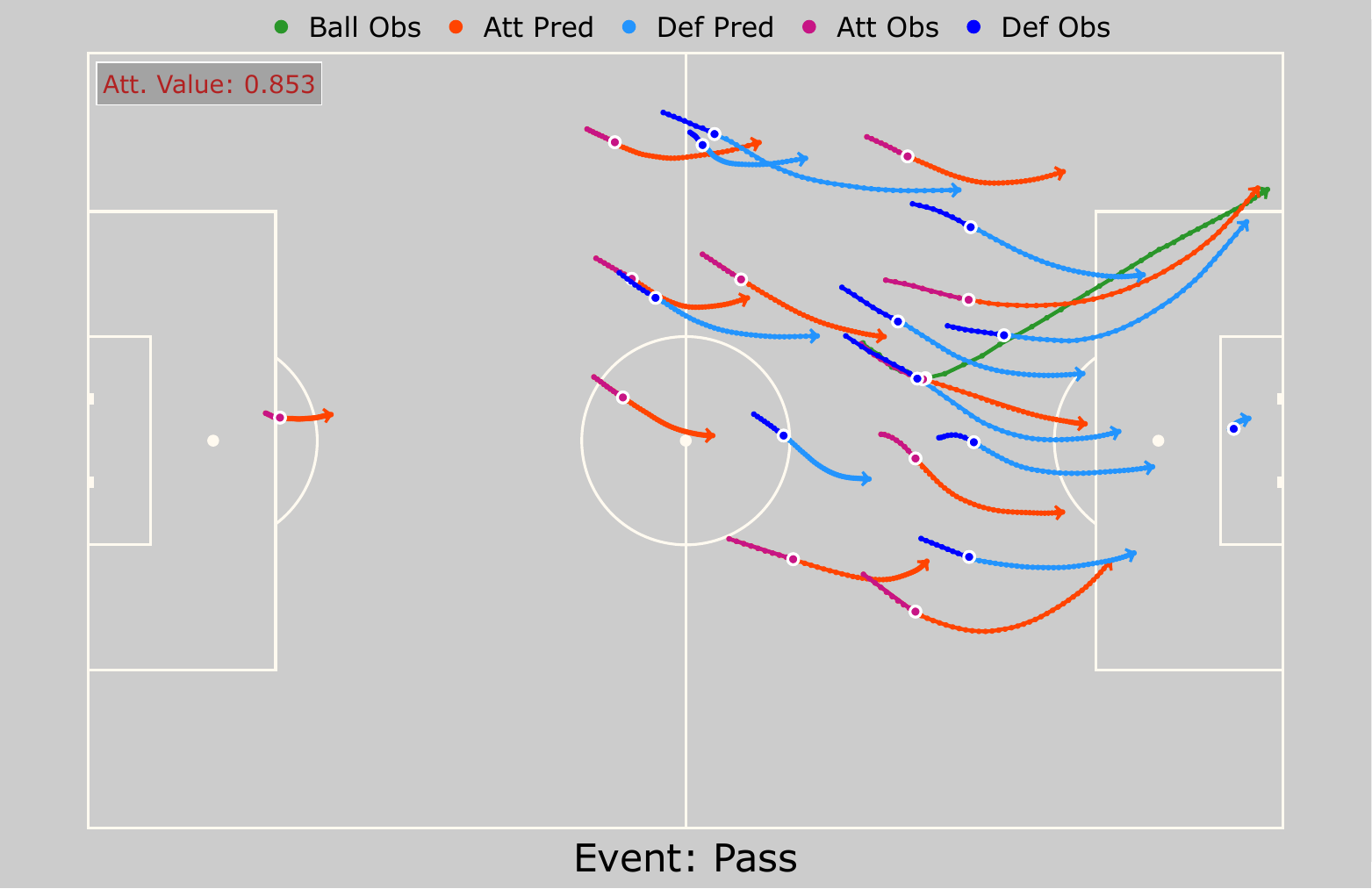}
  \end{minipage}\hfill
  \begin{minipage}[t]{0.158\textwidth}
    \centering
    \small \textbf{(b)} Att. High $V$\\
    \includegraphics[width=\linewidth]{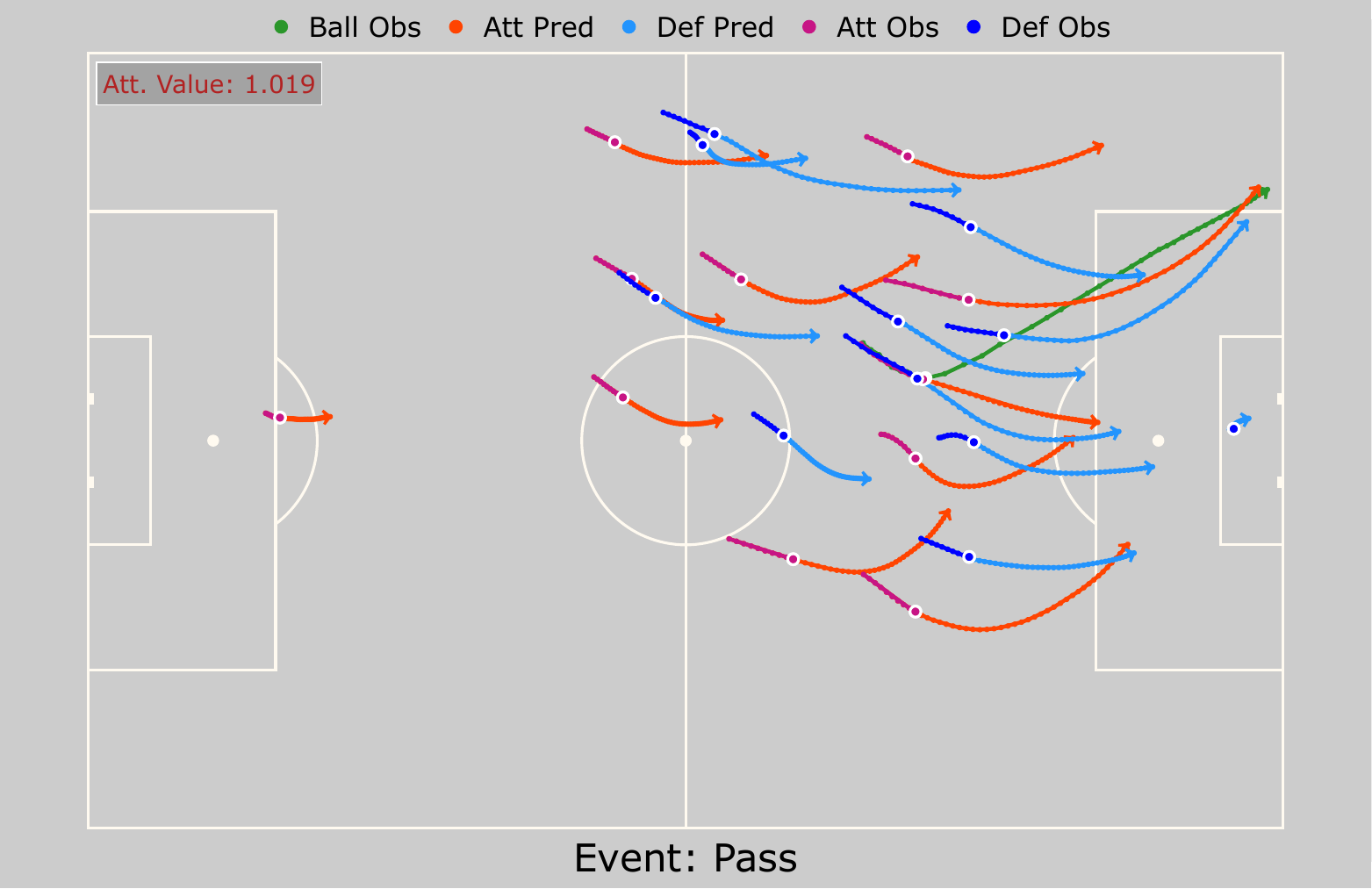}
  \end{minipage}\hfill
  \begin{minipage}[t]{0.158\textwidth}
    \centering
    \small \textbf{(c)} Def. High $V$\\
    \includegraphics[width=\linewidth]{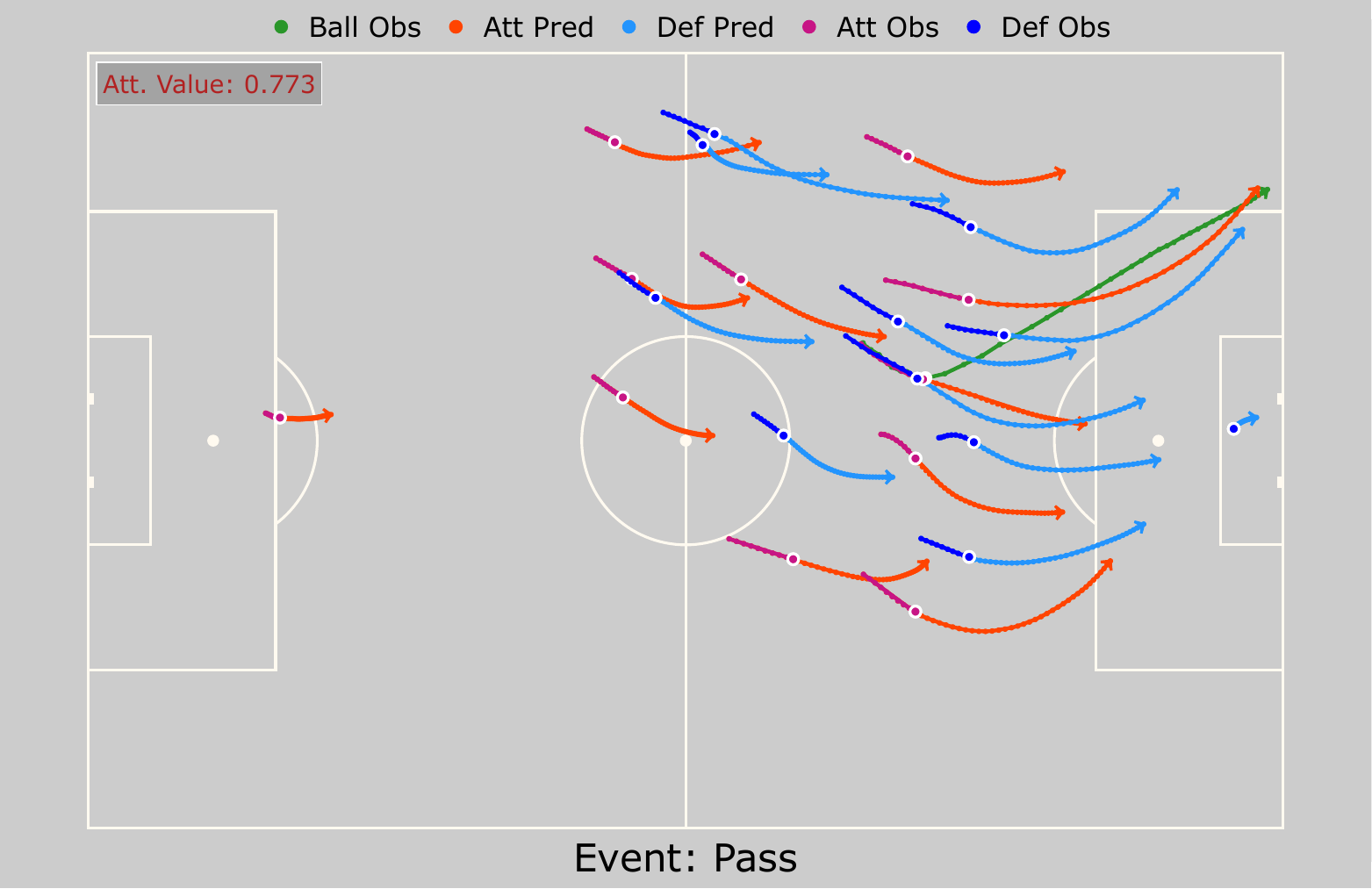}
  \end{minipage}
  \caption{{Visualizations of trajectories generated by TacticGen for a pass event under value guidance, where the unguided team follows the replayed data from the model's original prediction.}}
  \label{fig:fce-rule-guide-value-pass46-replaypred}
\end{figure}

Figure~\ref{fig:fce-rule-guide-pass46-replaypred} shows the model’s original prediction alongside the trajectories generated under rule-based guidance. Figure~\ref{fig:fce-rule-guide-pcm-pass46-replaypred} presents the corresponding results under pitch control value guidance, Figure~\ref{fig:fce-rule-guide-llm-pass46-replaypred} shows the results guided by LLM-generated functions, and Figure~\ref{fig:fce-rule-guide-value-pass46-replaypred} illustrates the results obtained with the trained value function. All configurations are kept identical to the main experiment in Section~\ref{sec:exp-guidance}, with the only difference being the unguided team’s behavior. In both attacking and defending cases, the guided team demonstrates improved behavior compared to the original prediction, underscoring TacticGen’s effectiveness and adaptability across different unguided team settings.

\textbf{Reactive generated trajectories.} We present results where the unguided team generates reactive trajectories. In such cases, the unguided team reacts naturally to the guided team’s behavior without receiving any external guidance signals. 

The key difference between this setup and the replayed-prediction setting is how the unguided team is handled. In the replayed-prediction case, the unguided team’s trajectories are fixed to the model’s initial predictions, ensuring controlled comparisons when guiding the other team. In contrast, in the reactive setting, the unguided team continues to evolve dynamically, responding to the guided team’s actions rather than following a predetermined path. While both approaches rely on the model’s predictive ability, the reactive setup more closely resembles real match conditions, where players continuously adjust to their opponents’ behavior. By allowing the unguided team to adapt reactively, we can better evaluate whether TacticGen generates behaviors that remain coherent and meaningful in the presence of uncontrolled opponents.

\begin{figure}[htbp]
  \vspace{-0.05in}
  \centering

  \begin{minipage}[t]{0.16\textwidth}
    \centering
    \small \textbf{(a)} No Guidance
    \includegraphics[width=\linewidth]{figures/guidance/Pass_986528_1816044_46/tactgen-c/rule/Epoch100_GFalse_SBJ_L1.135732_Pred.pdf}
  \end{minipage}\hfill
  \begin{minipage}[t]{0.16\textwidth}
    \centering
    \small \textbf{(b)} Att. Rule Guid.
    \includegraphics[width=\linewidth]{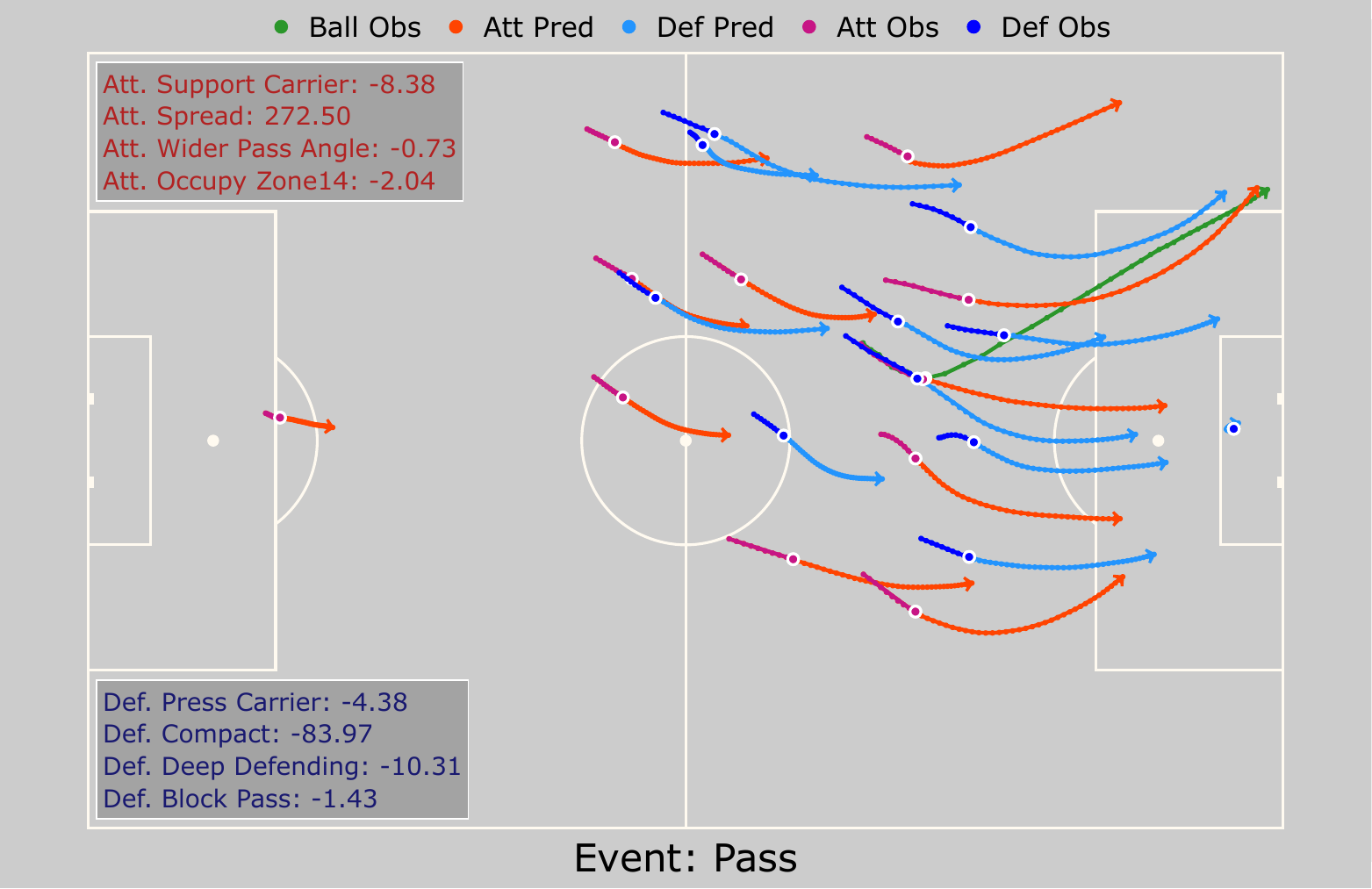}
  \end{minipage}\hfill
  \begin{minipage}[t]{0.16\textwidth}
    \centering
    \small \textbf{(c)} Def. Rule Guid.
    \includegraphics[width=\linewidth]{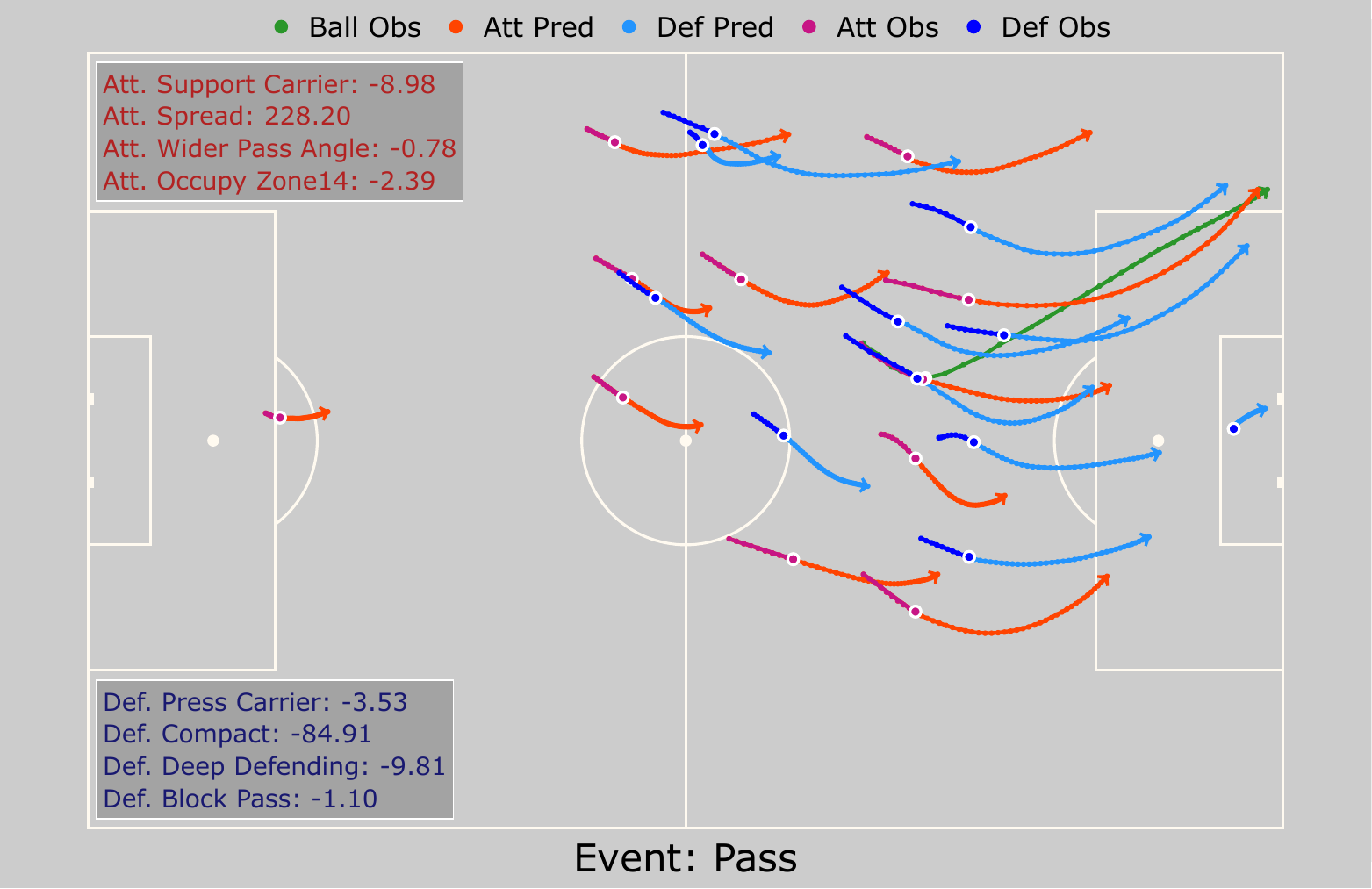}
  \end{minipage}

  \caption{{Trajectories generated by TacticGen for a pass event under rule-based guidance, where the unguided team generates reactive trajectories.}}\label{fig:fce-rule-guide-pass46-reactive}
  \vspace{-0.05in}
\end{figure}

\begin{figure}[htbp]
  \vspace{-0.05in}
  \centering

  \begin{minipage}[t]{0.16\textwidth}
    \centering
    \small \textbf{(a)} No Guidance\\
    \includegraphics[width=\linewidth]{figures/guidance/Pass_986528_1816044_46/tactgen-c/pcm/Epoch100_GFalse_SBJ_L1.135732_Pred.pdf}
  \end{minipage}\hfill
  \begin{minipage}[t]{0.16\textwidth}
    \centering
    \small \textbf{(b)} Att. High PCV\\
    \includegraphics[width=\linewidth]{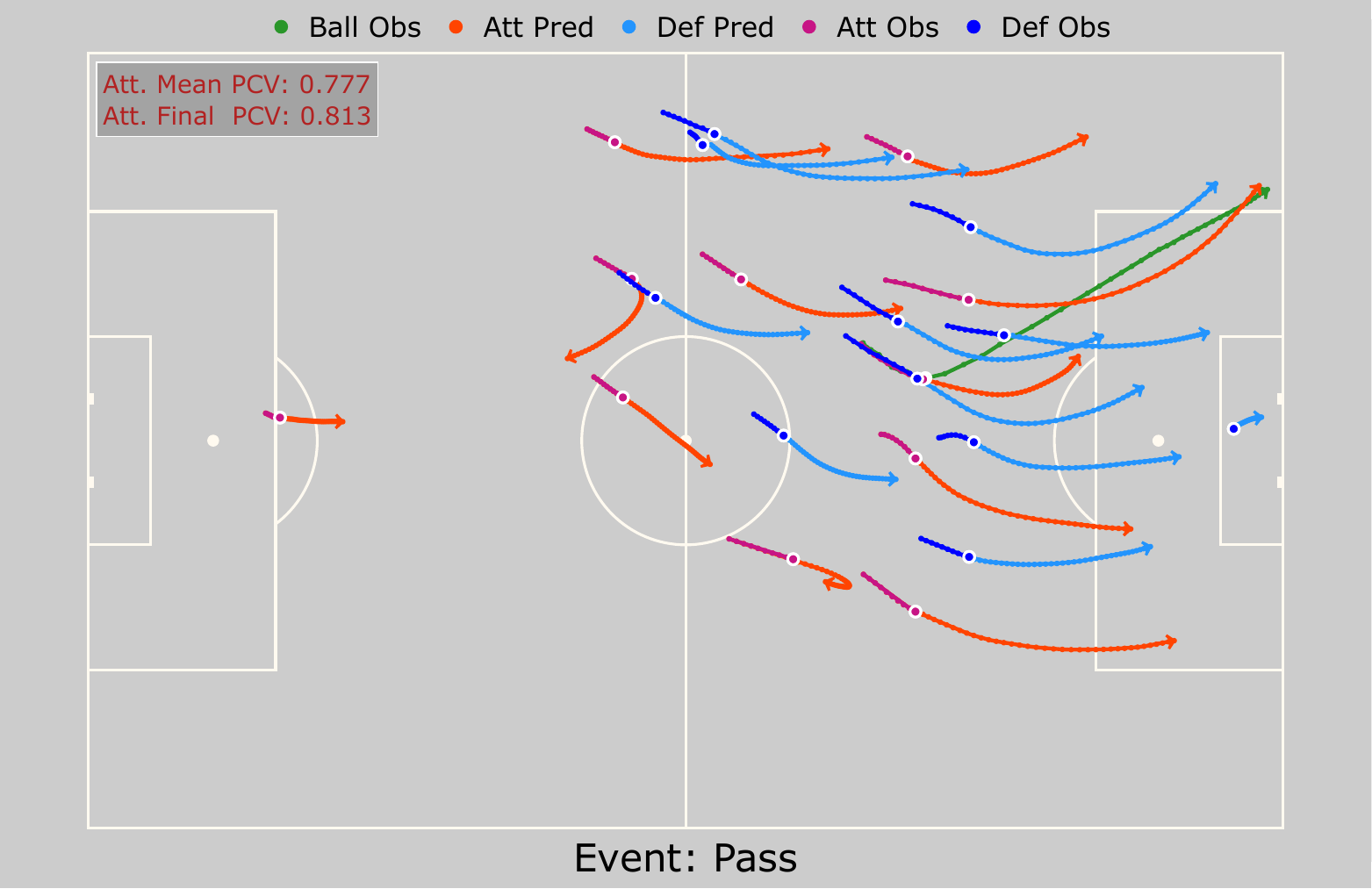}
  \end{minipage}\hfill
  \begin{minipage}[t]{0.16\textwidth}
    \centering
    \small (c) Def. High PCV\\
    \includegraphics[width=\linewidth]{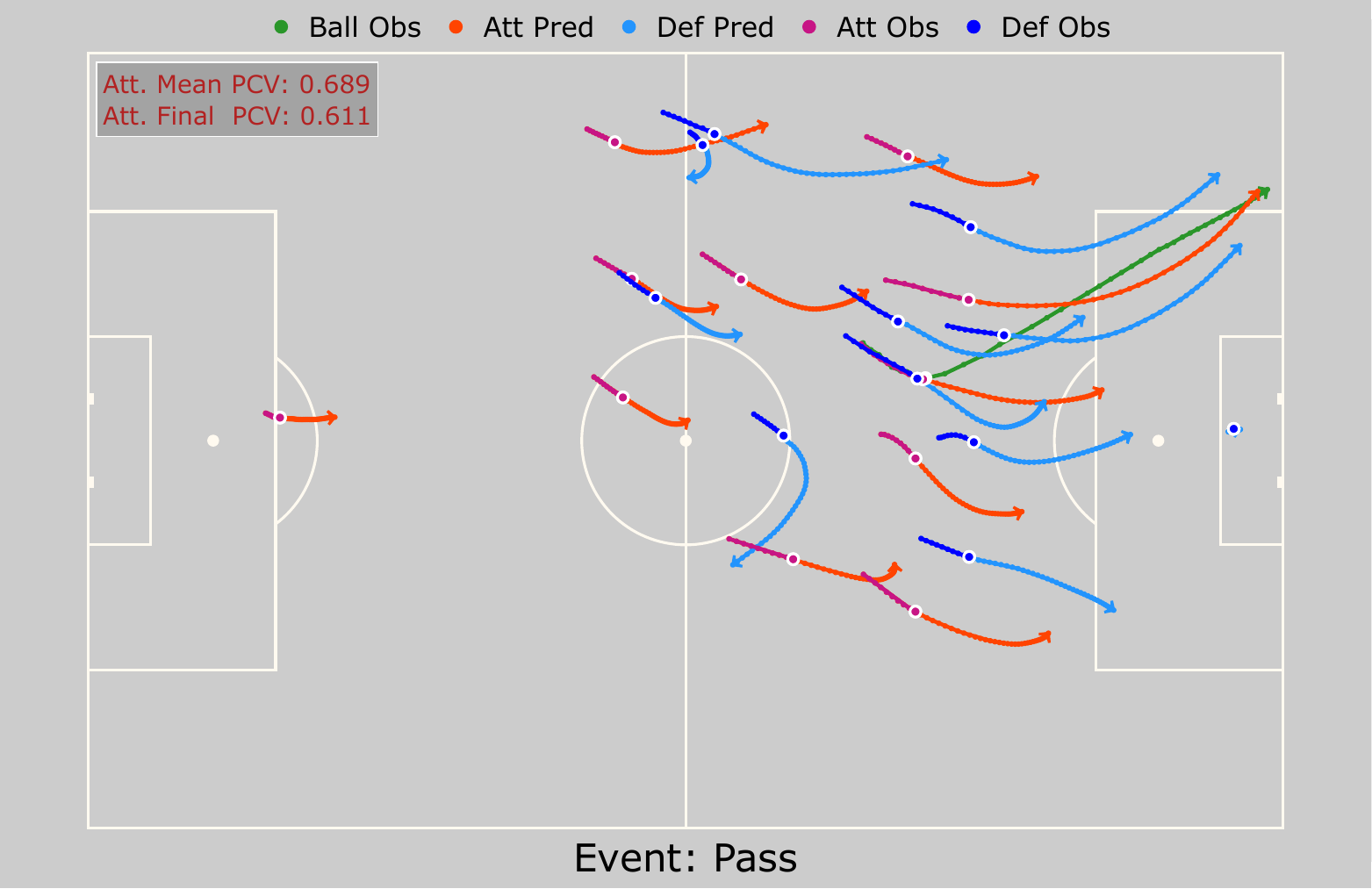}
  \end{minipage}

  \vspace{0.5em}

  \begin{minipage}[t]{0.16\textwidth}
    \centering
    \includegraphics[width=\linewidth]{figures/guidance/Pass_986528_1816044_46/tactgen-c/pcm/Epoch100_GFalse_SBJ_L1.135732_pcm.pdf}
  \end{minipage}\hfill
  \begin{minipage}[t]{0.16\textwidth}
    \centering
    \includegraphics[width=\linewidth]{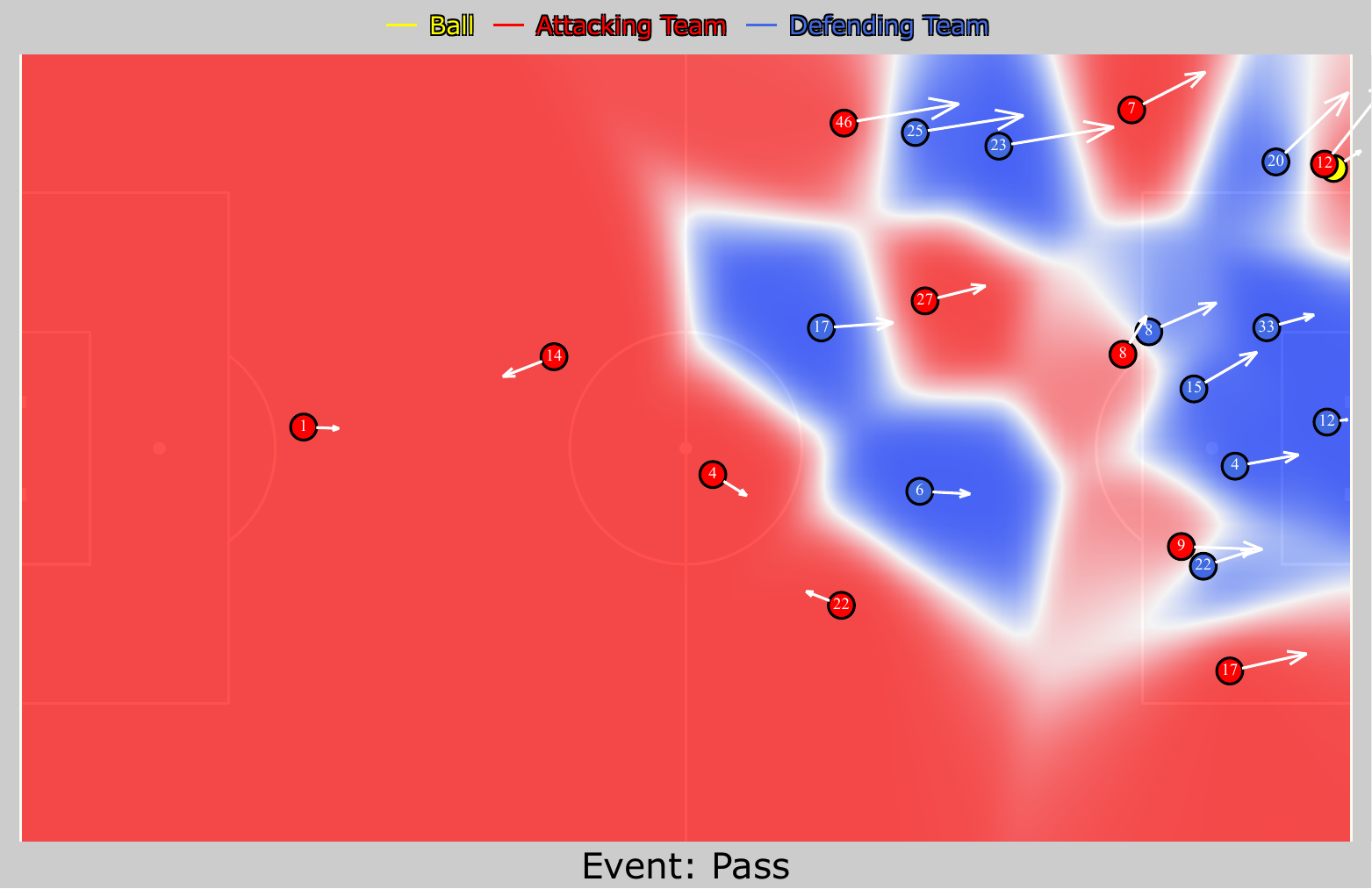}
  \end{minipage}\hfill
  \begin{minipage}[t]{0.16\textwidth}
    \centering
    \includegraphics[width=\linewidth]{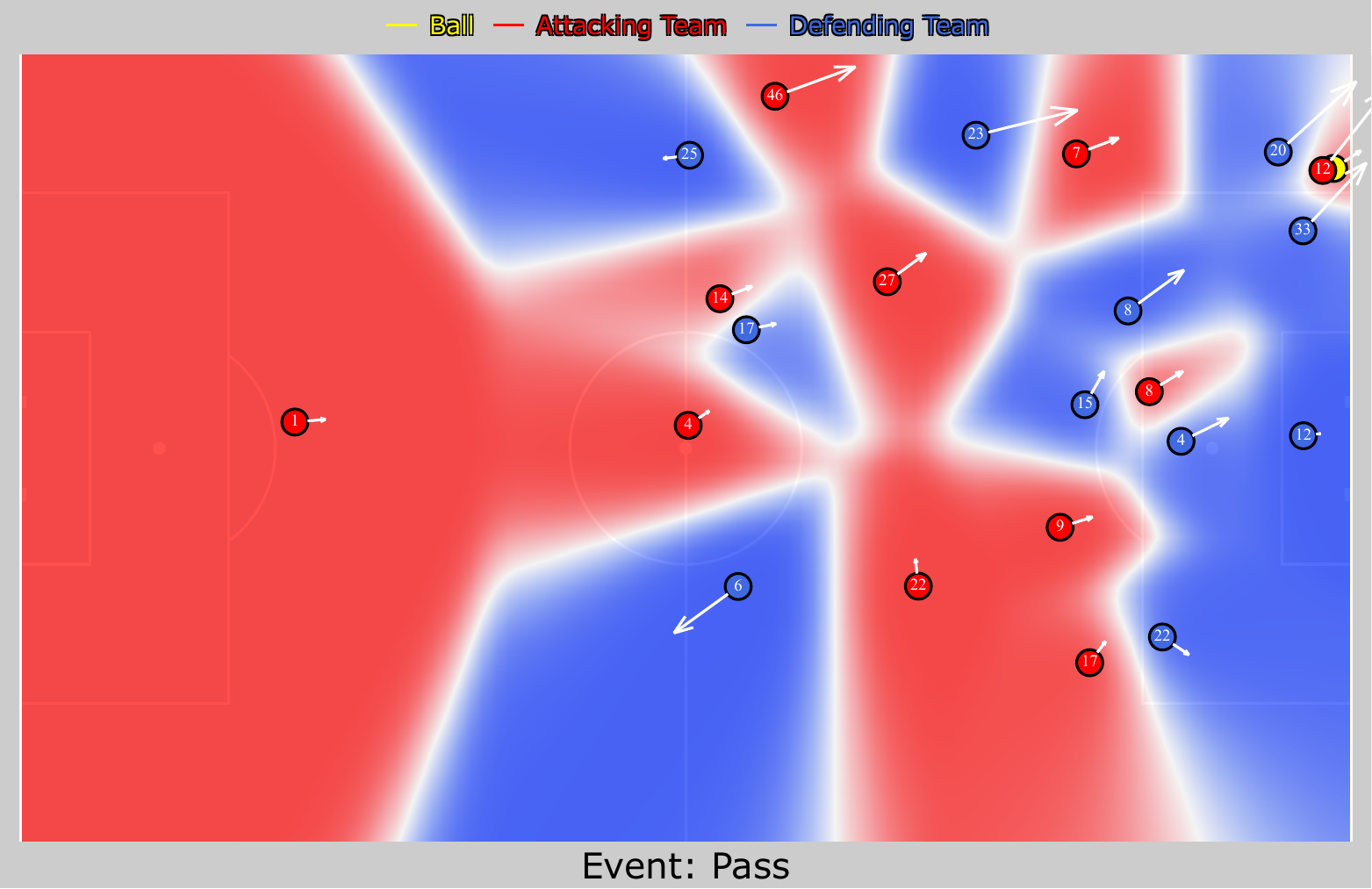}
  \end{minipage}

  \vspace{-0.1in}
  \caption{{Visualizations of trajectories (top) and pitch control values (PCV) at the final frame (bottom) generated by TacticGen for a pass event under pitch control guidance, where the unguided team generates reactive trajectories.}}\label{fig:fce-rule-guide-pcm-pass46-reactive}
\end{figure}

\begin{figure}[htbp]
  \vspace{-0.05in}
  \centering

  \begin{minipage}[t]{0.16\textwidth}
    \centering
    \includegraphics[width=\linewidth]{figures/prediction/Pass_986528_1816044_46/tactgen-c/Epoch100_GFalse_SBJ_L1.135732_Pred.pdf}
  \end{minipage}\hfill
  \begin{minipage}[t]{0.16\textwidth}
    \centering
    \includegraphics[width=\linewidth]{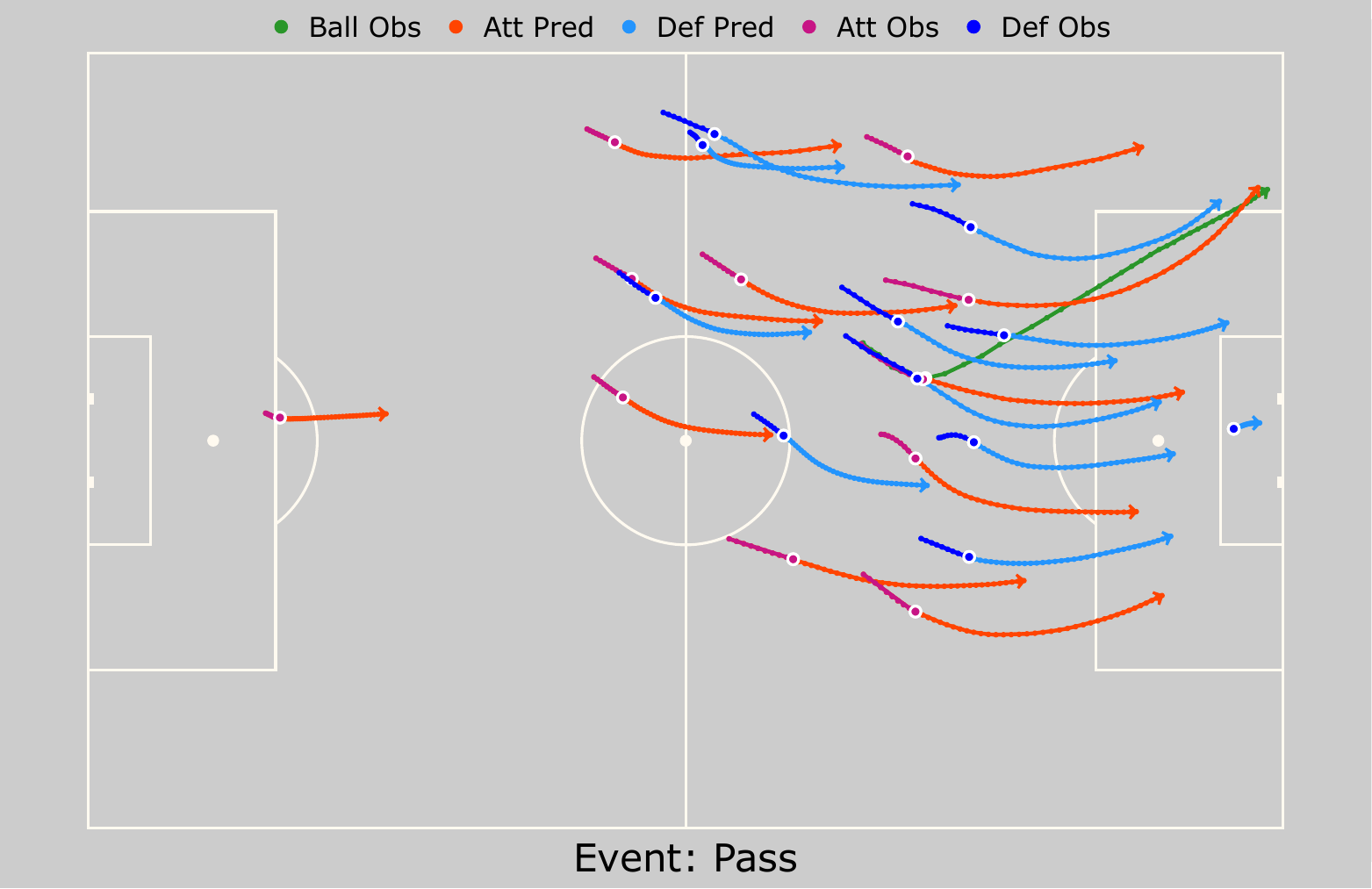}
  \end{minipage}\hfill
  \begin{minipage}[t]{0.16\textwidth}
    \centering
    \includegraphics[width=\linewidth]{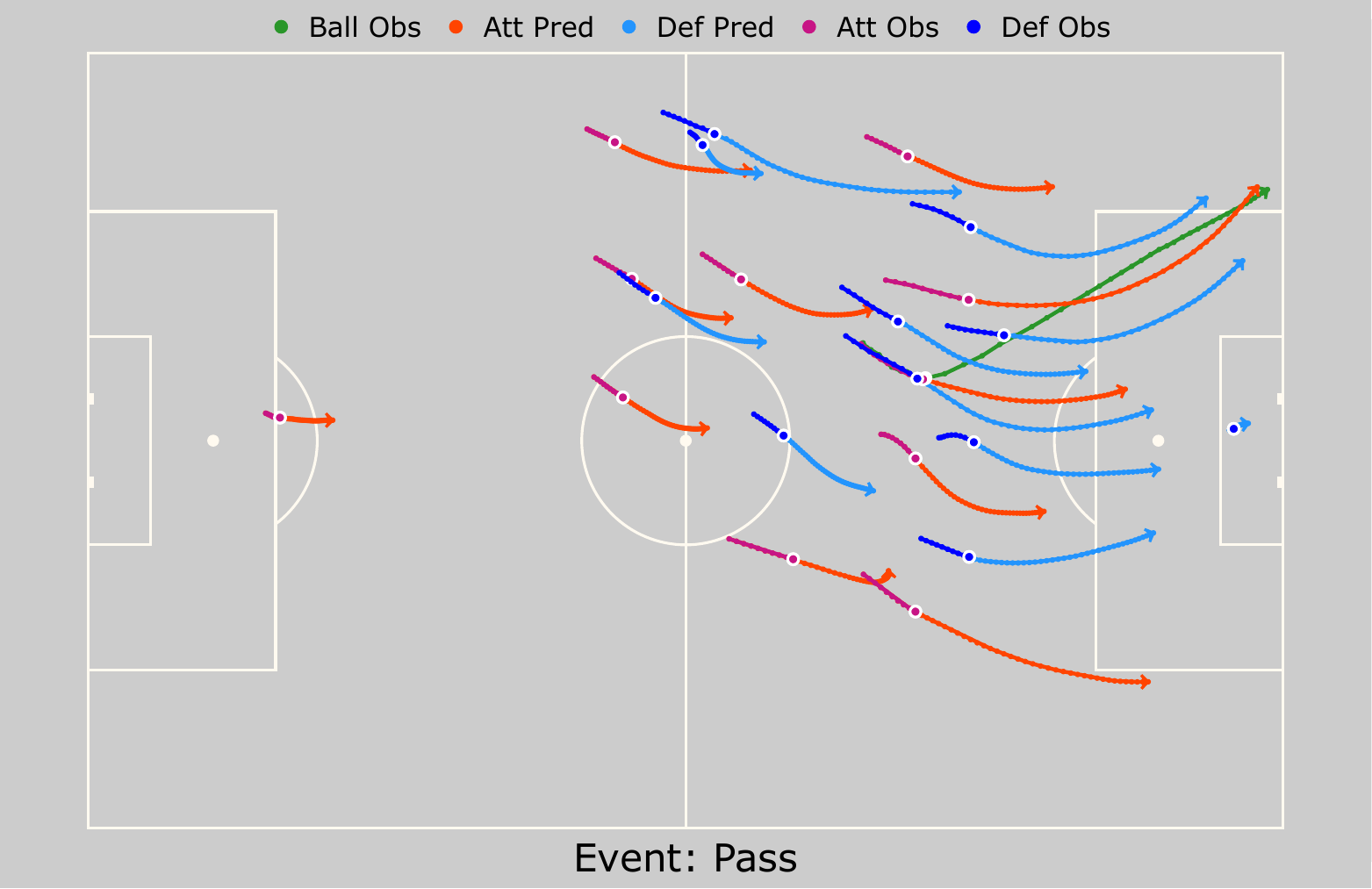}
  \end{minipage}

  \vspace{-0.1in}
  \caption{{Trajectories generated by TacticGen for a pass event under different guidance functions prompted by LLM, where the unguided team generates reactive trajectories.} \textbf{Left} Unguided generation. \textbf{Middle} Guided generation with the prompt, “Make the attacking team move forward more aggressively.” \textbf{Right} Guided generation with the prompt, “Make the right bottom player drift into the corner to stretch the defense and open up more space.”}
  \label{fig:fce-rule-guide-llm-pass46-reactive}
\end{figure}

\begin{figure}[htbp]
  \centering

  \begin{minipage}[t]{0.158\textwidth}
    \centering
    \small \textbf{(a)} No Guidance\\
    \includegraphics[width=\linewidth]{figures/guidance/Pass_986528_1816044_46/tactgen-c/value/Epoch100_GFalse_SBJ_L1.136_Pred.pdf}
  \end{minipage}\hfill
  \begin{minipage}[t]{0.158\textwidth}
    \centering
    \small \textbf{(b)} Att. High $V$\\
    \includegraphics[width=\linewidth]{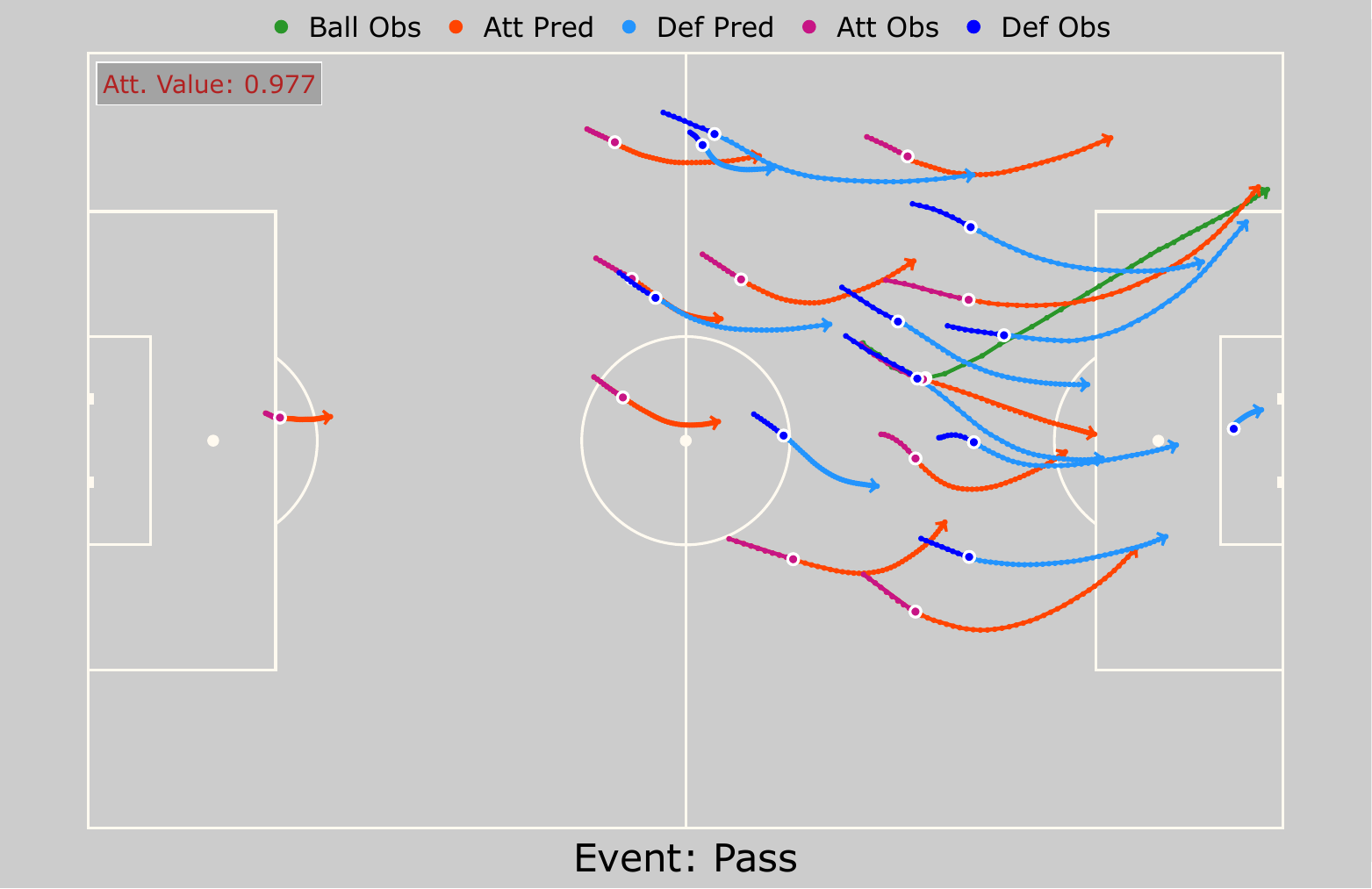}
  \end{minipage}\hfill
  \begin{minipage}[t]{0.158\textwidth}
    \centering
    \small \textbf{(c)} Def. High $V$\\
    \includegraphics[width=\linewidth]{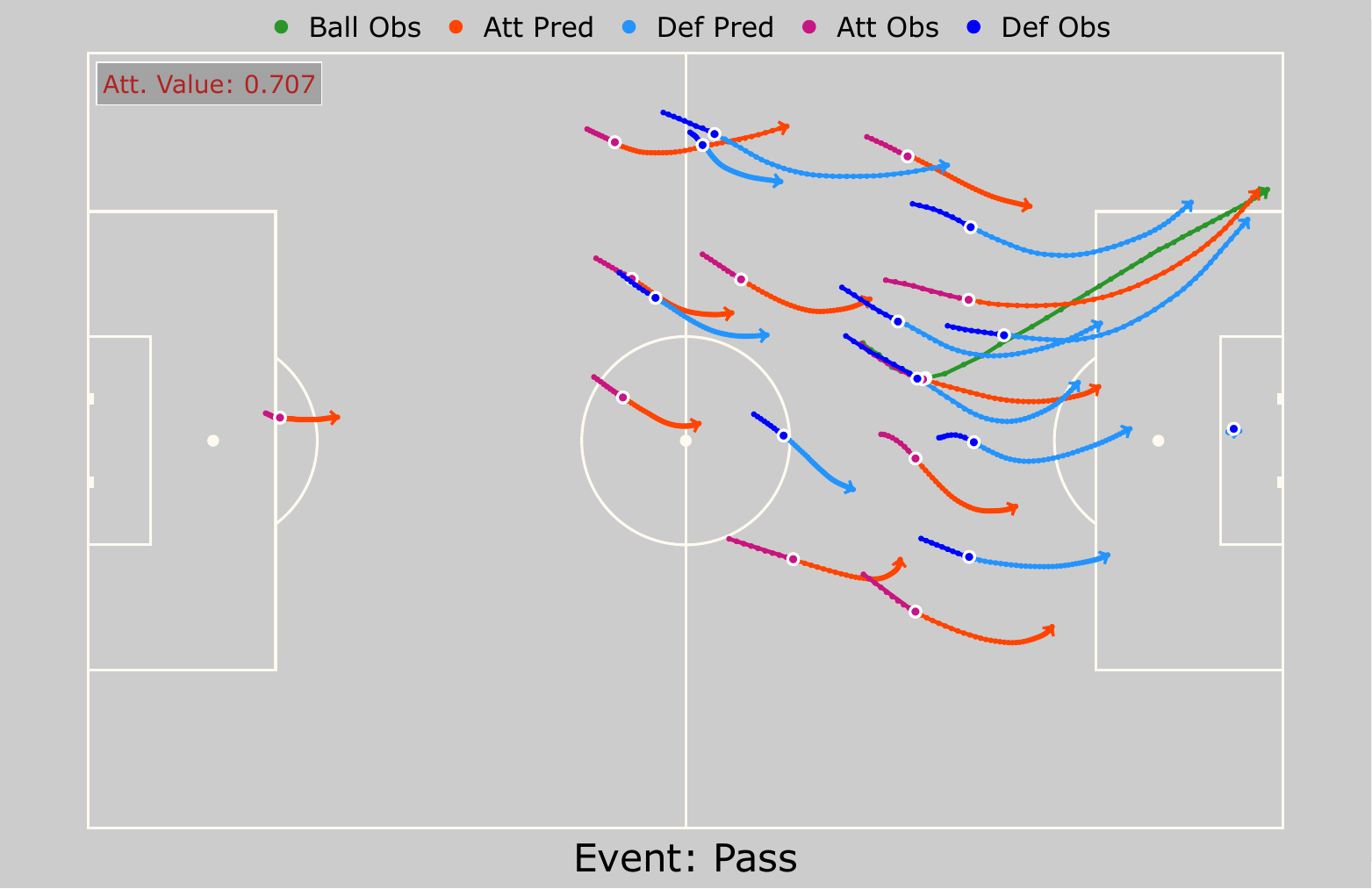}
  \end{minipage}
  \caption{{Visualizations of trajectories generated by TacticGen-C for a pass event under value guidance, where the unguided team generates reactive trajectories.}}
  \label{fig:fce-rule-guide-value-pass46-reactive}
\end{figure}

The results in Figures~\ref{fig:fce-rule-guide-pass46-reactive}, \ref{fig:fce-rule-guide-pcm-pass46-reactive}, \ref{fig:fce-rule-guide-llm-pass46-reactive}, and \ref{fig:fce-rule-guide-value-pass46-reactive} demonstrate TacticGen’s ability to generate football tactics even when the opposing team re-actively adjusts its movements in response to the guided team’s tactics. This highlights TacticGen’s effectiveness in handling dynamic, multi-agent interactions, where both sides continuously adapt to each other. Such a capability is particularly valuable for practical applications, as it more closely reflects real match conditions and allows practitioners to explore adjustments under realistic and adversarial settings.

\subsection{Results of More Events}\label{sec:more-event-result}

In addition to the event discussed in the main experiments, we present another representative pass event in this section, visualizing the generated trajectories under different guidance mechanisms. We keep the same experimental setting, with the unguided team following the ground-truth replayed data.

\begin{figure}[htbp]
\vspace{-0.1in}
  \centering

  \begin{minipage}[t]{0.16\textwidth}
    \centering
    \small \textbf{(a)} Ground Truth
    \includegraphics[width=\linewidth]{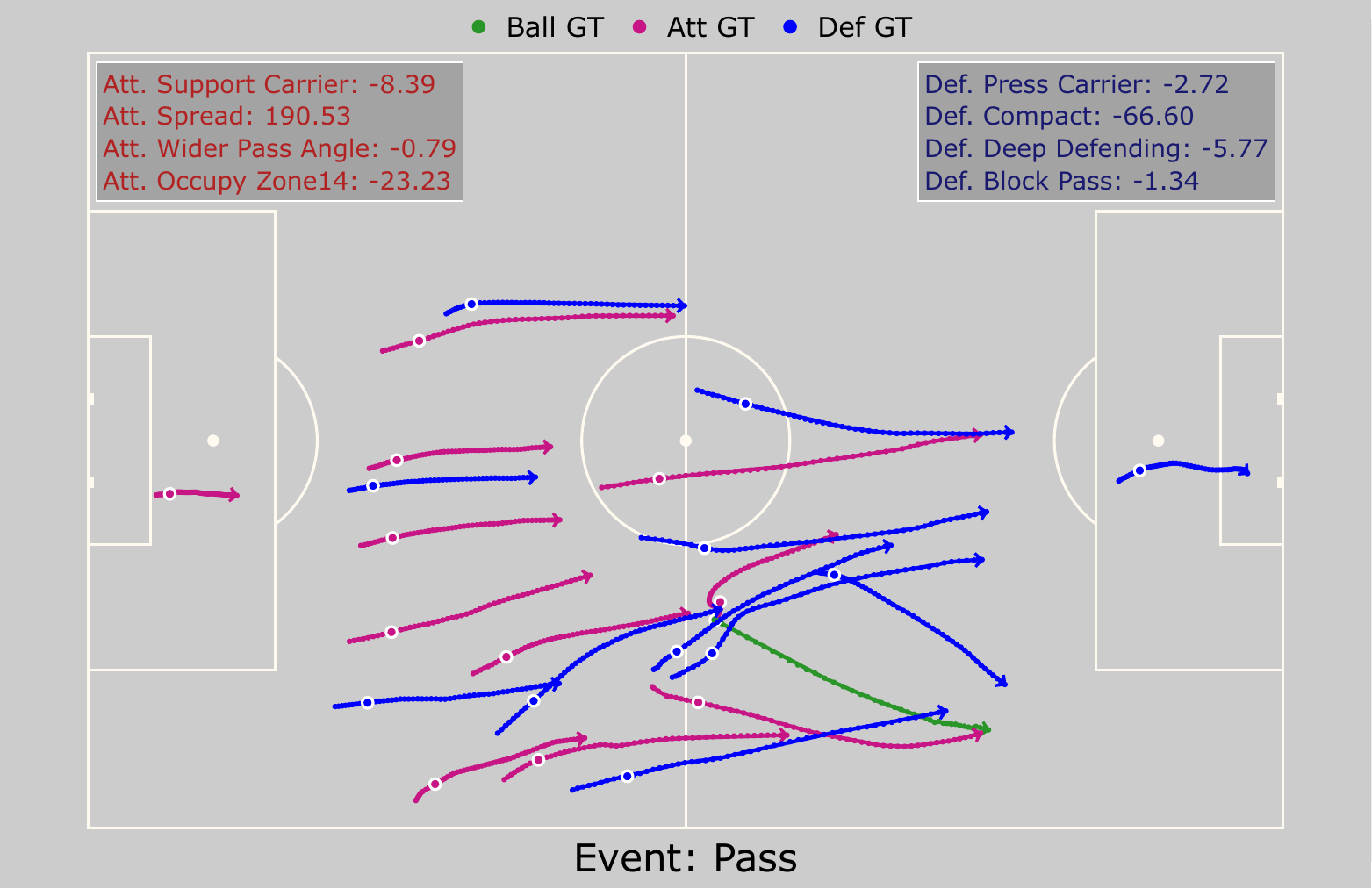}
  \end{minipage}\hfill
  \begin{minipage}[t]{0.16\textwidth}
    \centering
    \small \textbf{(b)} Att. Rule Guid.
    \includegraphics[width=\linewidth]{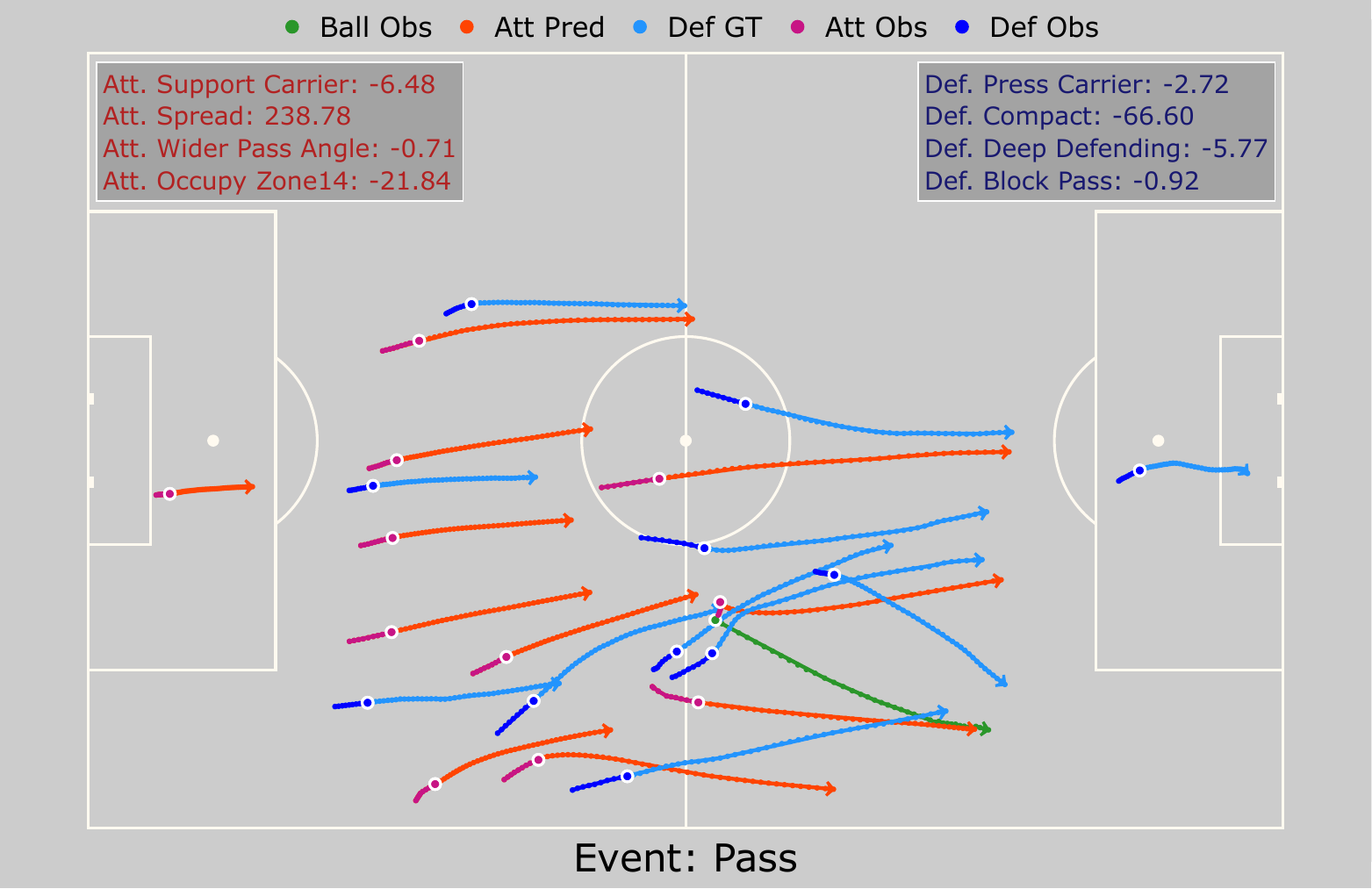}
  \end{minipage}\hfill
  \begin{minipage}[t]{0.16\textwidth}
    \centering
    \small \textbf{(c)} Def. Rule Guid.
    \includegraphics[width=\linewidth]{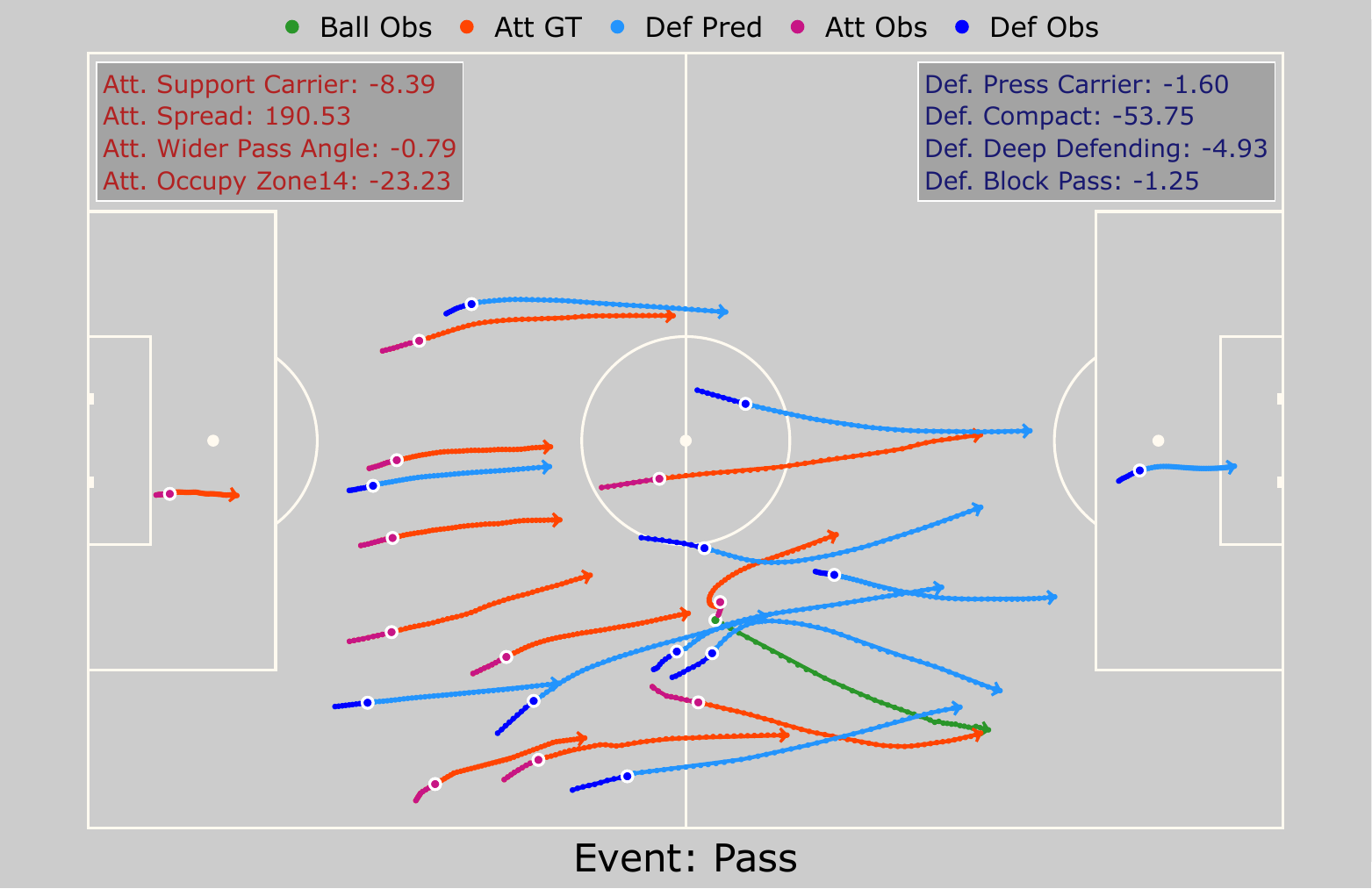}
  \end{minipage}

  \caption{{Trajectories generated by TacticGen for a pass event.} \textbf{(a)} Ground Truth. \textbf{(b)} Rule-based guidance for the attacking team. Notably, the ball kicker makes a rapid run toward the opponent’s goal to prepare for receiving the ball, compared to the ground truth one. \textbf{(c)} Rule-based guidance for the defending team. We find that the defending team adjusts by sending another defender to press the ball carrier compared to the ground truth one, while the original presser in the ground truth retreats rapidly to strengthen the defensive structure rather than pressing.}\label{fig:fce-rule-guide-pass52}
  \vspace{-0.1in}
\end{figure}

\begin{figure}[htbp]
\vspace{-0.2in}
  \centering

  \begin{minipage}[t]{0.16\textwidth}
    \centering
    \small \textbf{(a)} Ground Truth\\
    \includegraphics[width=\linewidth]{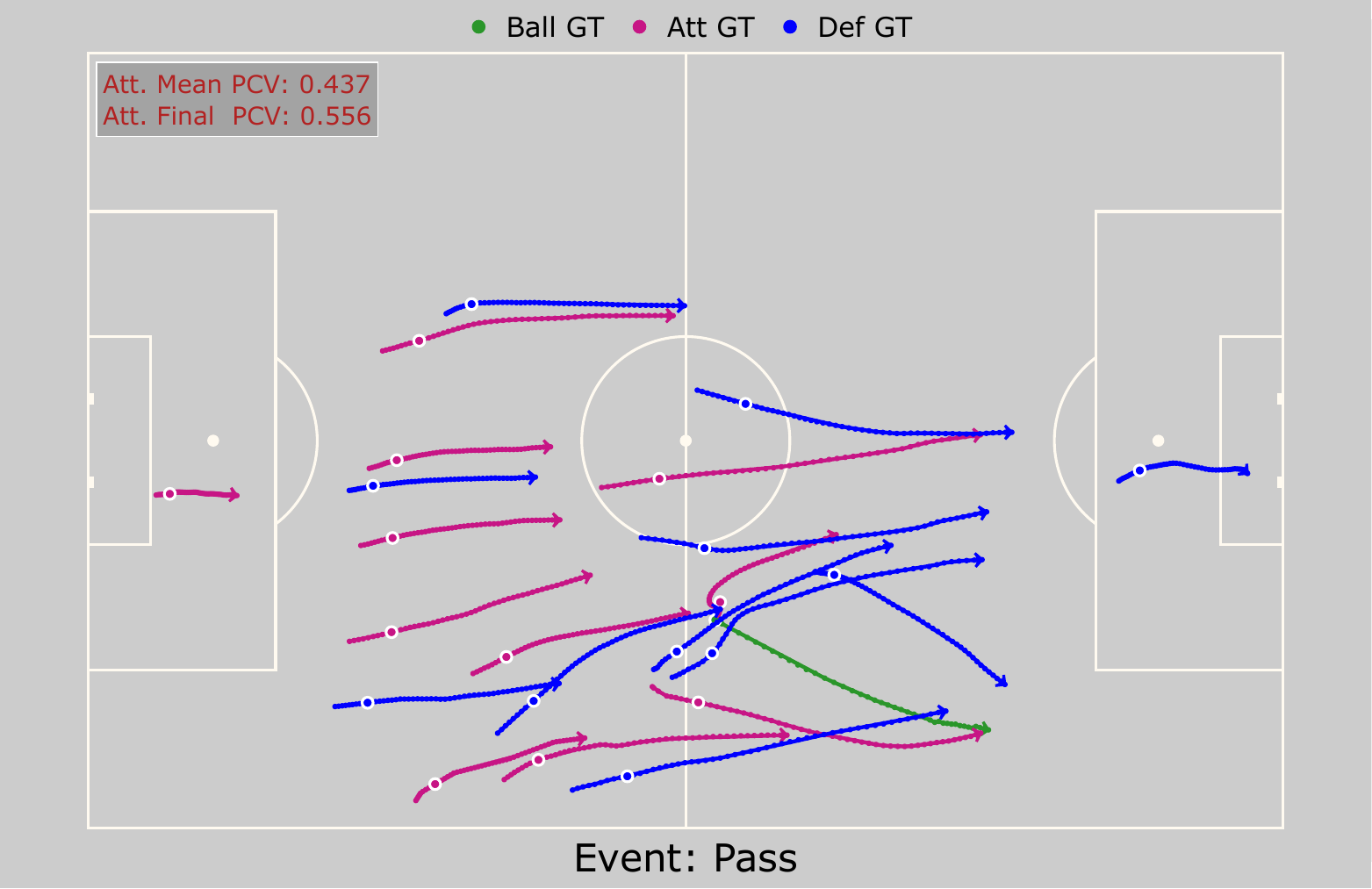}
  \end{minipage}\hfill
  \begin{minipage}[t]{0.16\textwidth}
    \centering
    \small \textbf{(b)} Att. High PCV\\
    \includegraphics[width=\linewidth]{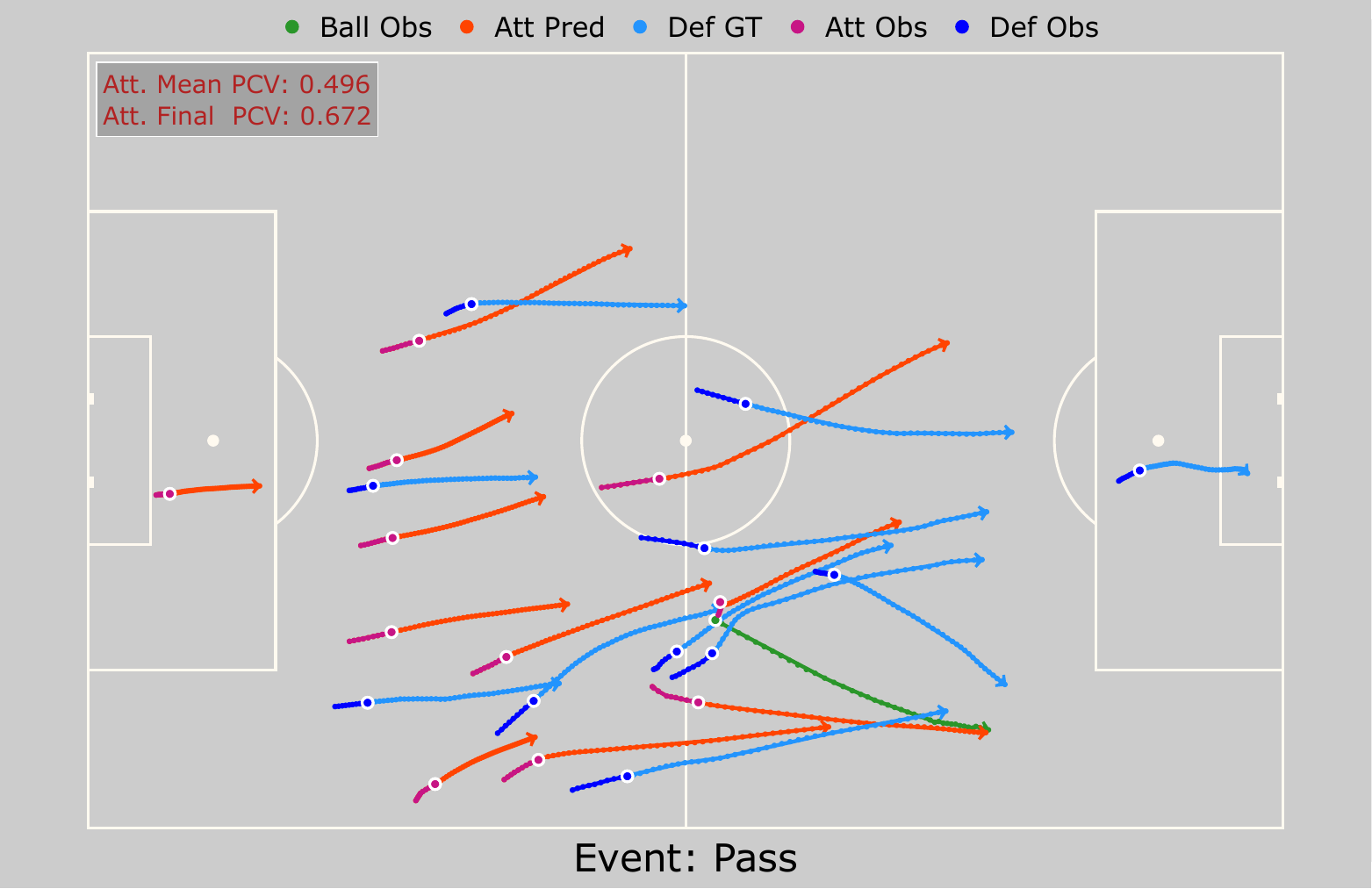}
  \end{minipage}\hfill
  \begin{minipage}[t]{0.16\textwidth}
    \centering
    \small \textbf{(c)} Def. High PCV\\
    \includegraphics[width=\linewidth]{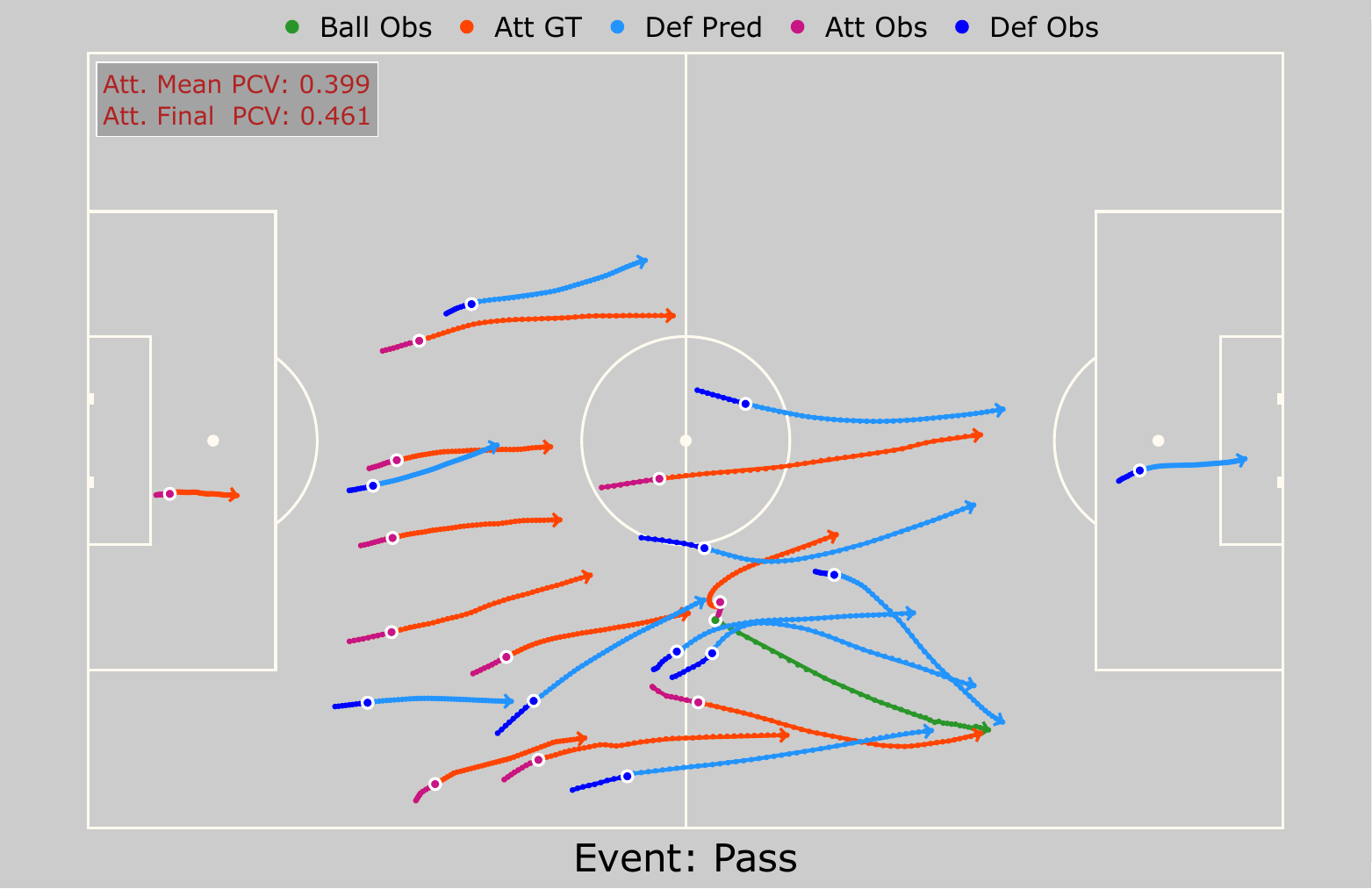}
  \end{minipage}

  \vspace{0.1em}

  \begin{minipage}[t]{0.16\textwidth}
    \centering
    \includegraphics[width=\linewidth]{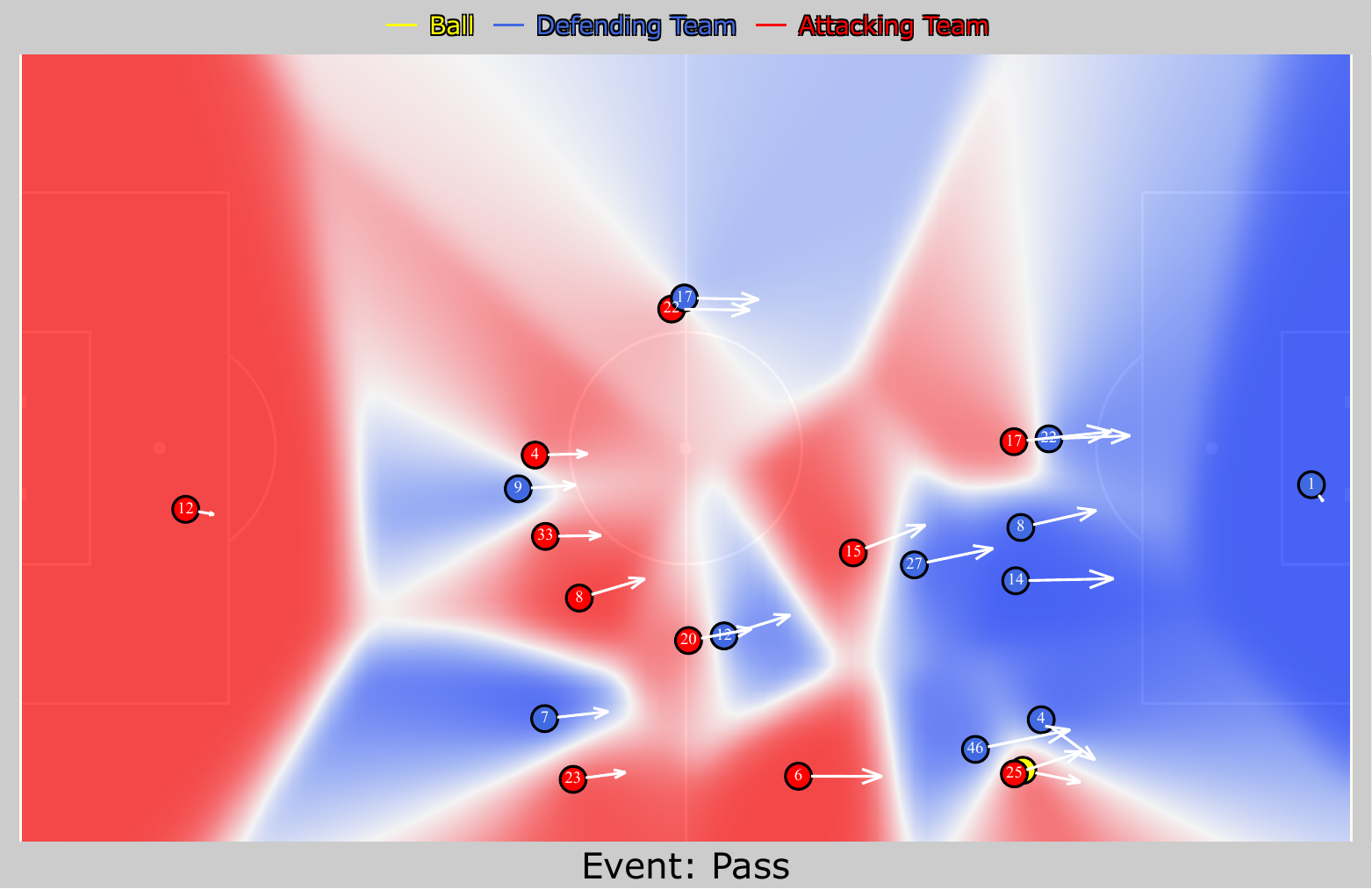}
  \end{minipage}\hfill
  \begin{minipage}[t]{0.16\textwidth}
    \centering
    \includegraphics[width=\linewidth]{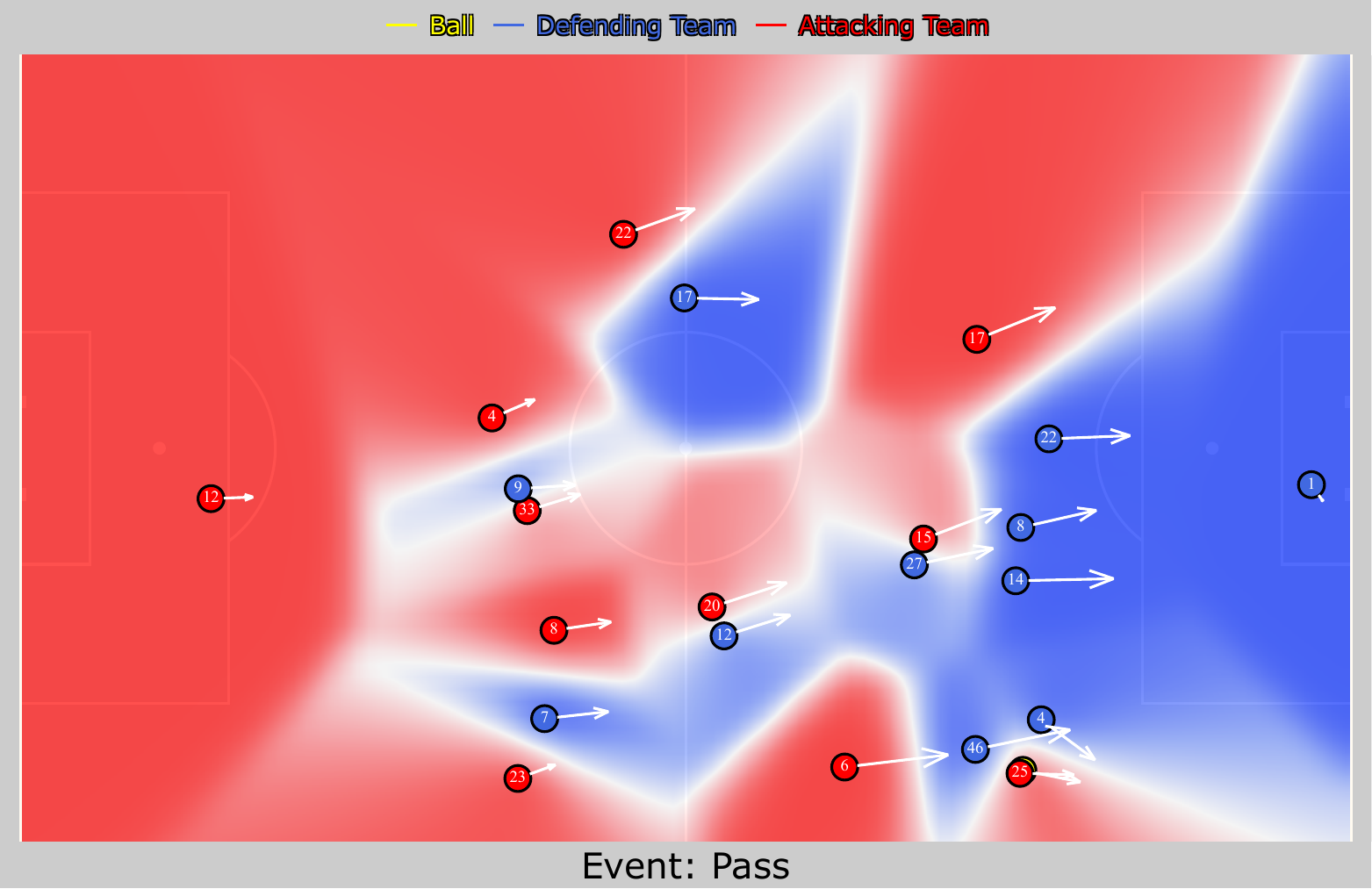}
  \end{minipage}\hfill
  \begin{minipage}[t]{0.16\textwidth}
    \centering
    \includegraphics[width=\linewidth]{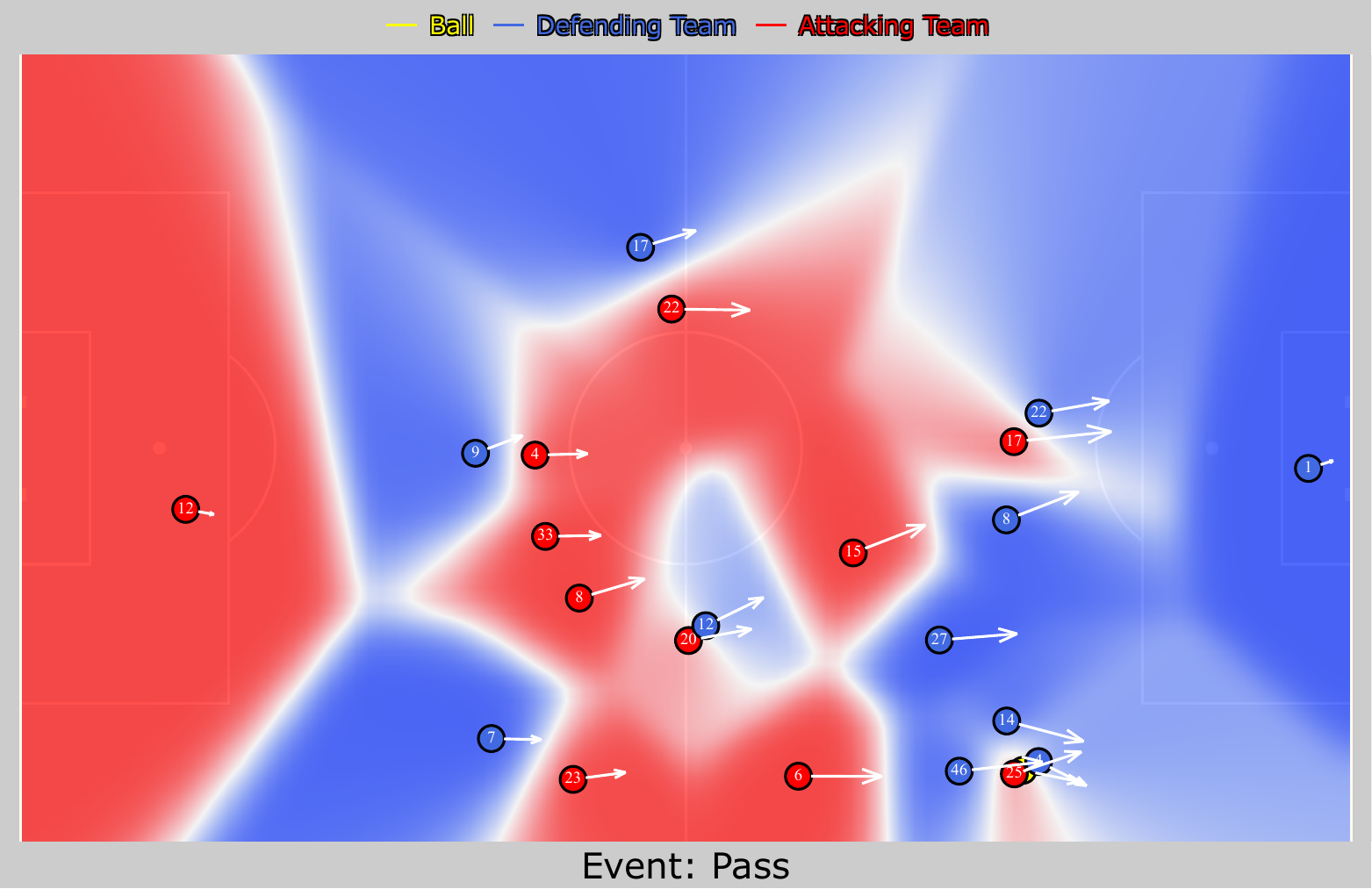}
  \end{minipage}

  \vspace{-0.1in}
  \caption{{Visualizations of trajectories (top) and pitch control values (PCV) at the final frame (bottom) generated by TacticGen for a pass event.} \textbf{(a)} Ground Truth. \textbf{(b)} Guided trajectories aimed at maximizing PCV for the attacking team. Notably, the upper-left and central attacking players run towards the upper-right to create more space. \textbf{(c)} Guided trajectories aimed at maximizing PCV for the defending team. Notably, the upper-left defender chooses to move towards the upper-right to create more space, and a defender is driven to indirectly press the ball carrier.}\label{fig:fce-rule-guide-pcm-pass52}
\end{figure}

\begin{figure}[htbp]
  \centering

  \begin{minipage}[t]{0.16\textwidth}
    \centering
    \includegraphics[width=\linewidth]{figures/prediction/Pass_986528_1843954_52/gt/GroundTruth.pdf}
  \end{minipage}\hfill
  \begin{minipage}[t]{0.16\textwidth}
    \centering
    \includegraphics[width=\linewidth]{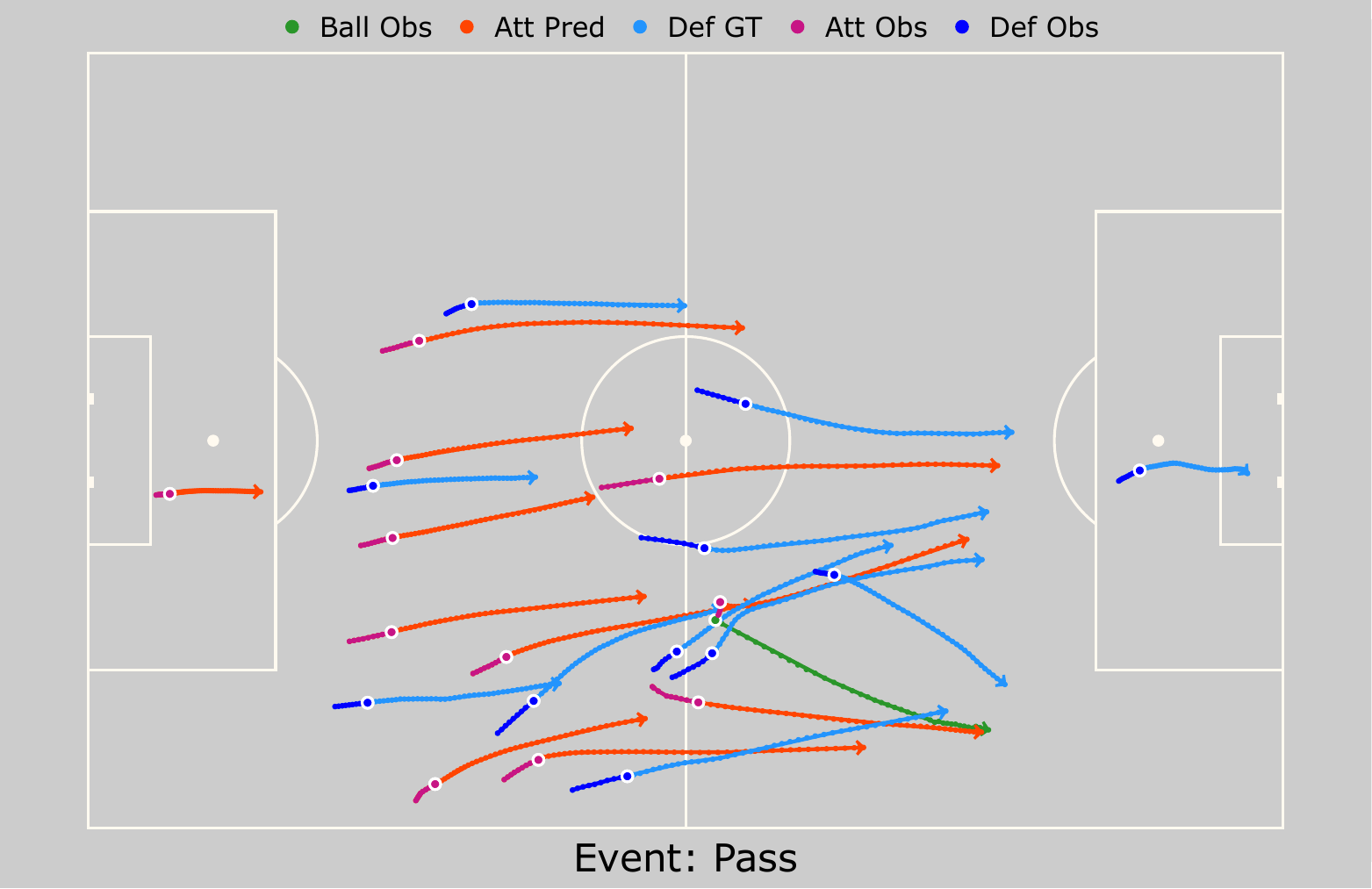}
  \end{minipage}\hfill
  \begin{minipage}[t]{0.16\textwidth}
    \centering
    \includegraphics[width=\linewidth]{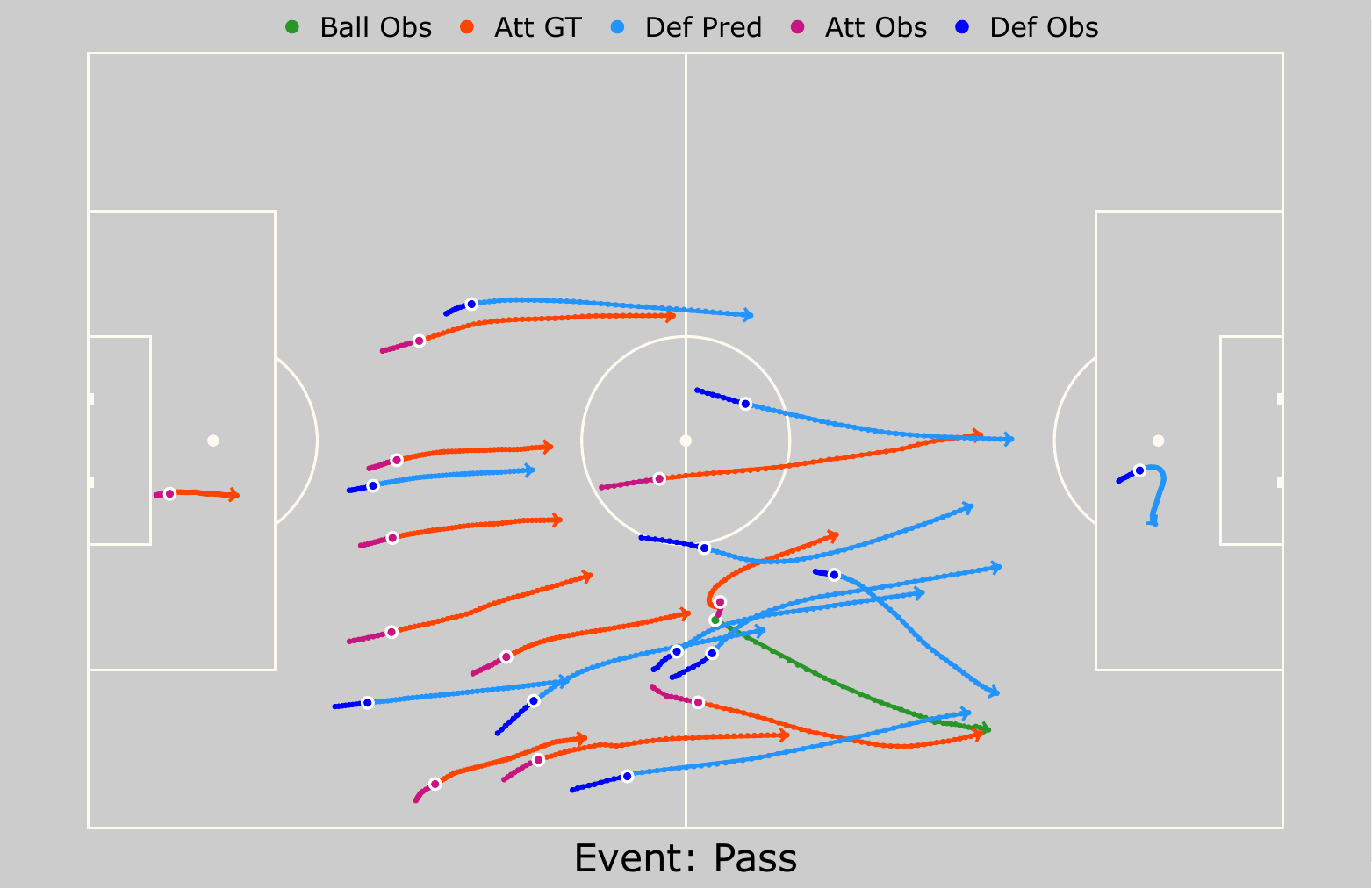}
  \end{minipage}

  \vspace{-0.1in}
  \caption{{Trajectories generated by TacticGen for a pass event under different guidance functions prompted by LLM.} \textbf{Left} Ground Truth. \textbf{Middle} Guided generation with the prompt, “Make the attacking team move forward more aggressively.” It is evident that the attacking players respond by increasing their speed and covering greater distances toward the defending goal, especially for the attacking players behind the midfield line. Note that although the players may appear to run faster, their movements do not tend to exceed speed limits, as TacticGen generates trajectories based on the learned movement patterns and prevents unrealistic behaviors. \textbf{Right} Guided generation with the prompt, “Make the defending goalkeeper on the right side move towards the ball, positioning to save a potential goal attempt.” Clearly, the defending goalkeeper adjusts the positioning towards the ball to better block a potential shot.}
  \label{fig:fce-rule-guide-llm-pass52}
\end{figure}

\begin{figure}[htbp]
  \centering

  \begin{minipage}[t]{0.158\textwidth}
    \centering
    \small \textbf{(a)} Ground Truth\\
    \includegraphics[width=\linewidth]{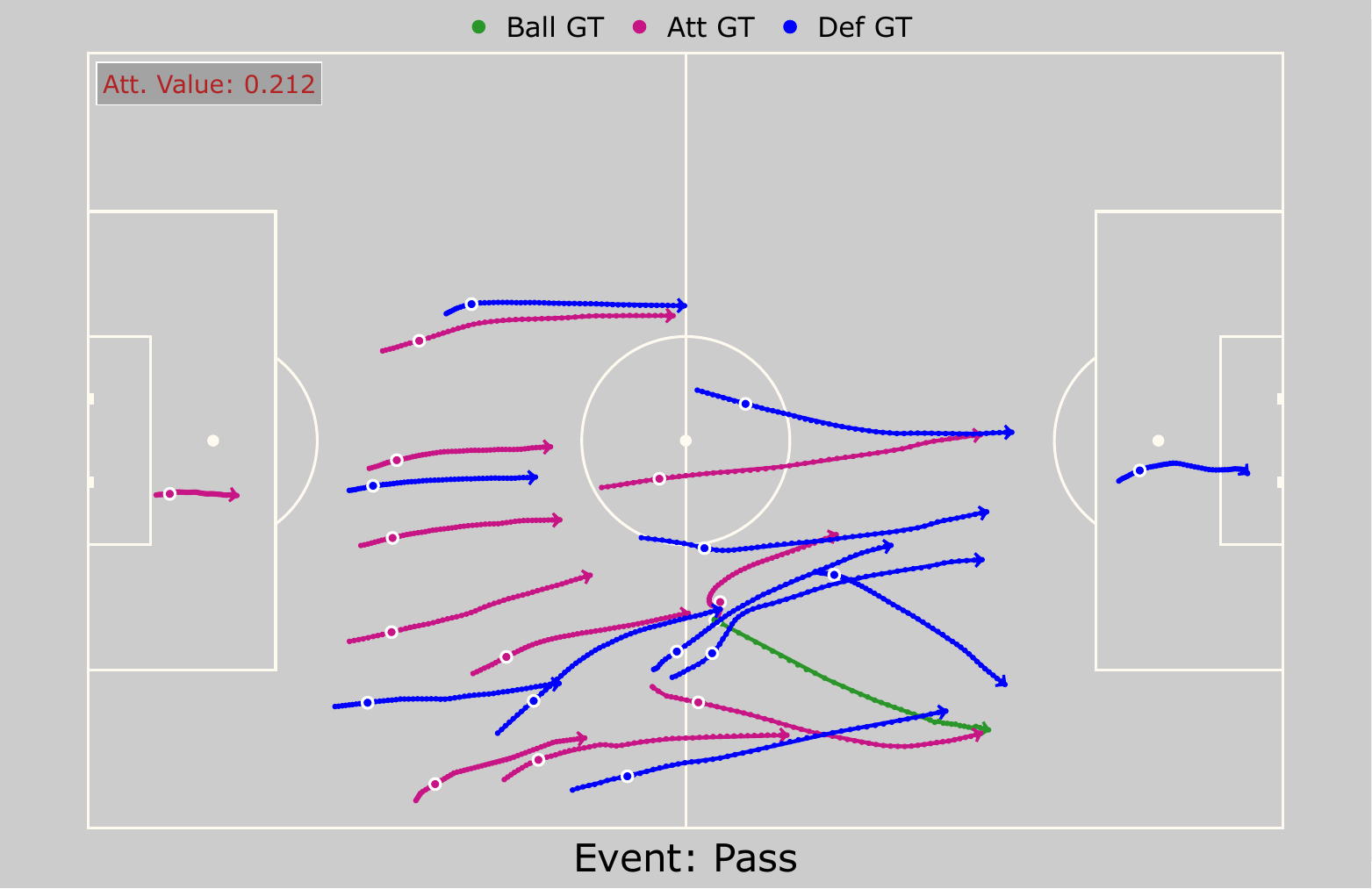}
  \end{minipage}\hfill
  \begin{minipage}[t]{0.158\textwidth}
    \centering
    \small \textbf{(b)} Att. High $V$\\
    \includegraphics[width=\linewidth]{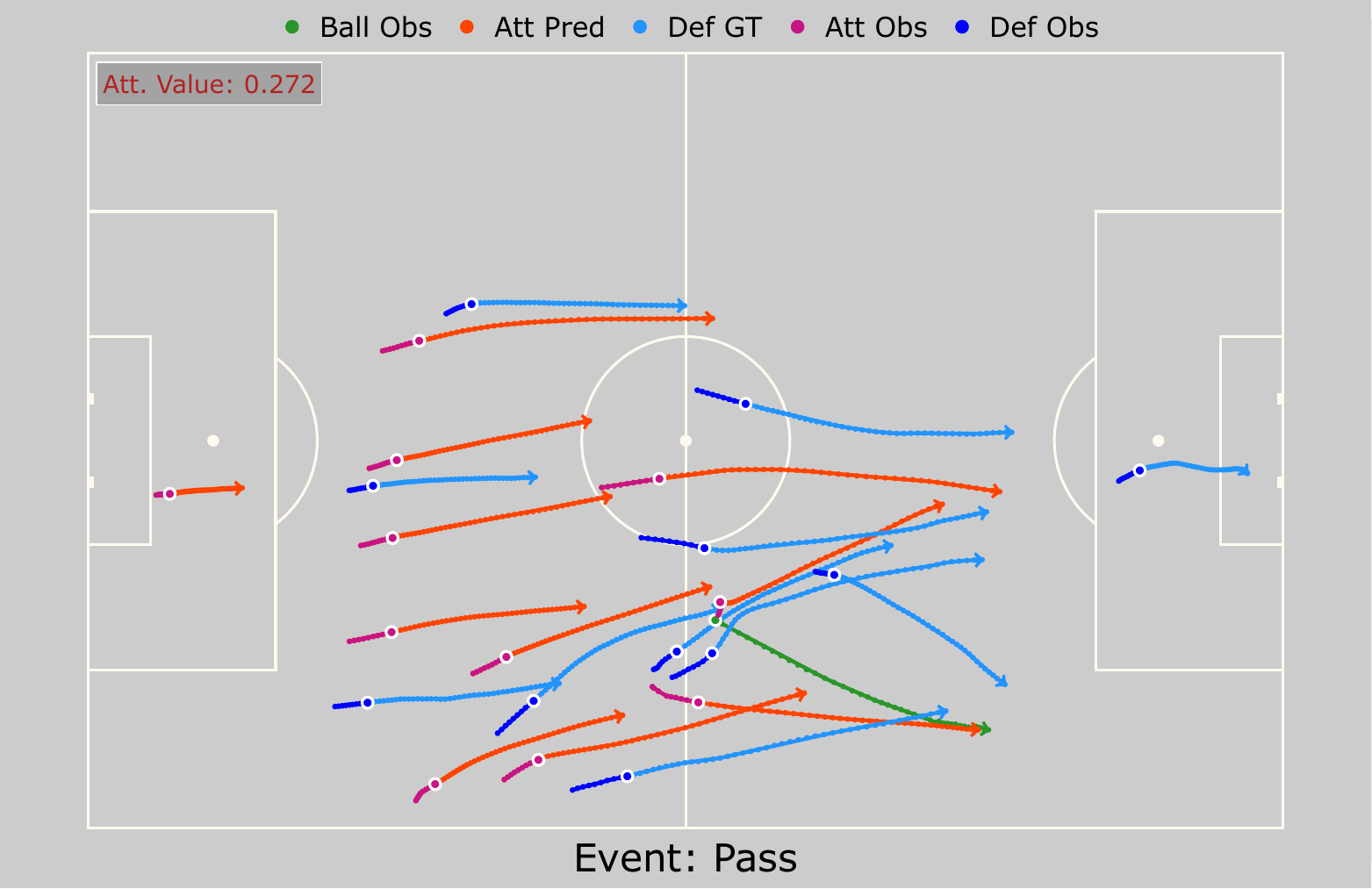}
  \end{minipage}\hfill
  \begin{minipage}[t]{0.158\textwidth}
    \centering
    \small \textbf{(c)} Def. High $V$\\
    \includegraphics[width=\linewidth]{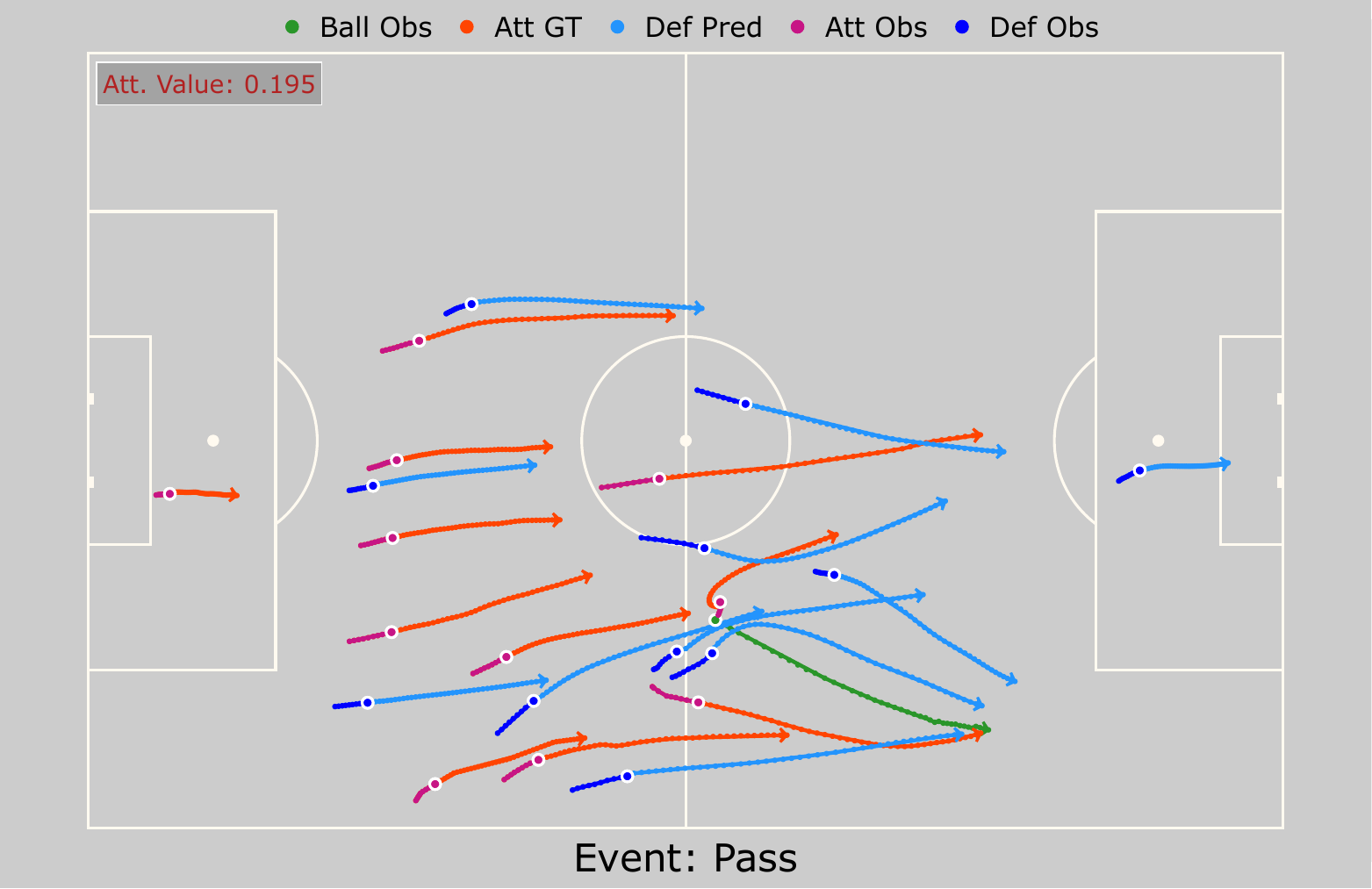}
  \end{minipage}
  \caption{{Visualizations of trajectories generated by TacticGen for a pass event.} \textbf{(a)} Ground Truth. \textbf{(b)} Guided trajectories aimed at maximizing the $V$ value for the attacking team. Notably, the attacking players increase their speed to push the team forward, bringing them closer to the defending area and potentially increasing the probability of scoring. \textbf{(c)} Guided trajectories aimed at maximizing the $V$ value for the defending team. Notably, a defender accelerates toward the ball carrier, creating a 3-vs-1 situation in an attempt to gain possession.}
  \label{fig:fce-rule-guide-value-pass52}
\end{figure}

The results in Figures~\ref{fig:fce-rule-guide-pass52}, \ref{fig:fce-rule-guide-pcm-pass52}, \ref{fig:fce-rule-guide-llm-pass52}, and \ref{fig:fce-rule-guide-value-pass52} demonstrate TacticGen’s ability to generate tactically coherent and meaningful behaviors in a scenario different from those in the main experiments.

\vspace{0.1in}

\section{Visualization Examples in Utility Case Study}\label{sec:vis-tactic}

In this section, we present two illustrative examples in which all five experts agreed that the generated trajectories demonstrated superior tactical effectiveness. In these examples, the guidance is applied exclusively to one team, while the unguided team follows the ground-truth movements.

\begin{figure}[htbp]
  \centering
  \begin{minipage}[t]{0.24\textwidth}
    \centering
    \small \textbf{(a)} Ground Truth\\
    \includegraphics[width=\linewidth]{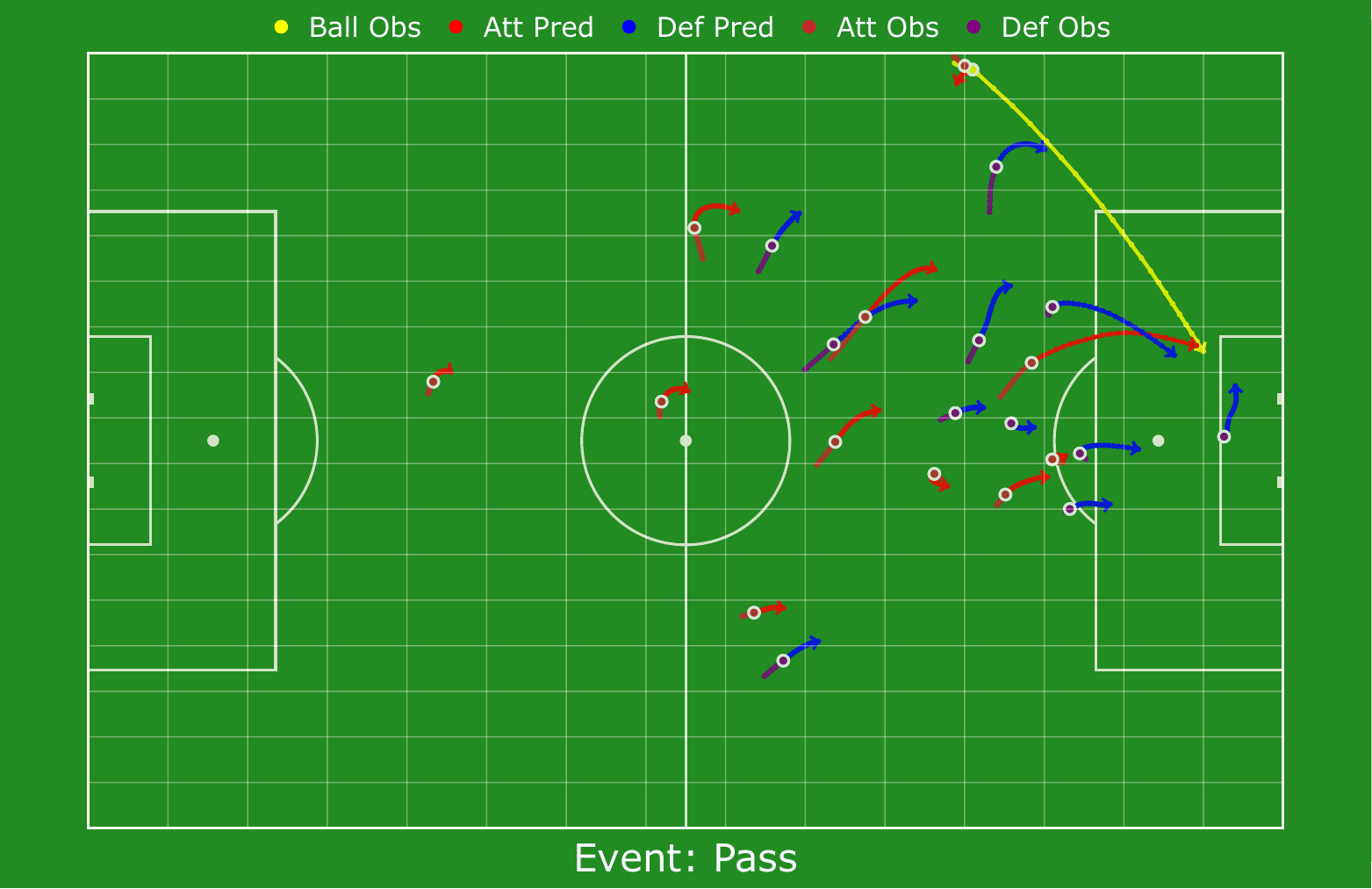}
  \end{minipage}\hfill
  \begin{minipage}[t]{0.24\textwidth}
    \centering
    \small \textbf{(b)} Generated\\
    \includegraphics[width=\linewidth]{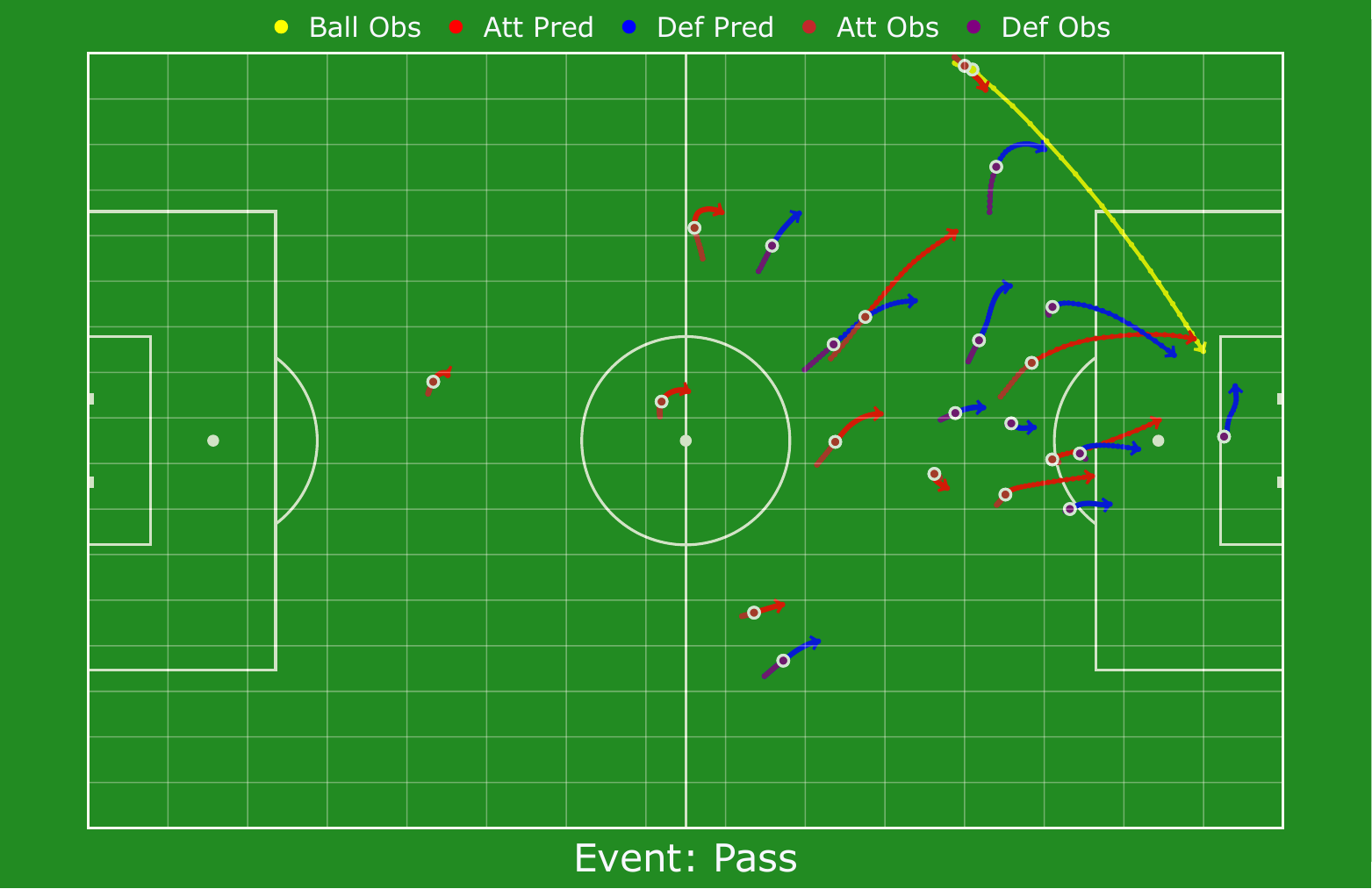}
  \end{minipage}

  \caption{{An illustrative example where the \textcolor{red}{attacking} team is guided.} Notably, the generated tactic positions an attacker to support the ball carrier, thereby creating alternatives beyond a direct shot. As one expert observed, “Support from teammates provides more options for the next action, preventing the goalkeeper from simply anticipating a shot on goal.”}
  \label{fig:vis-tactic-att}
\end{figure}

\begin{figure}[htbp]
  \centering
  \begin{minipage}[t]{0.24\textwidth}
    \centering
    \small \textbf{(a)} Ground Truth\\
    \includegraphics[width=\linewidth]{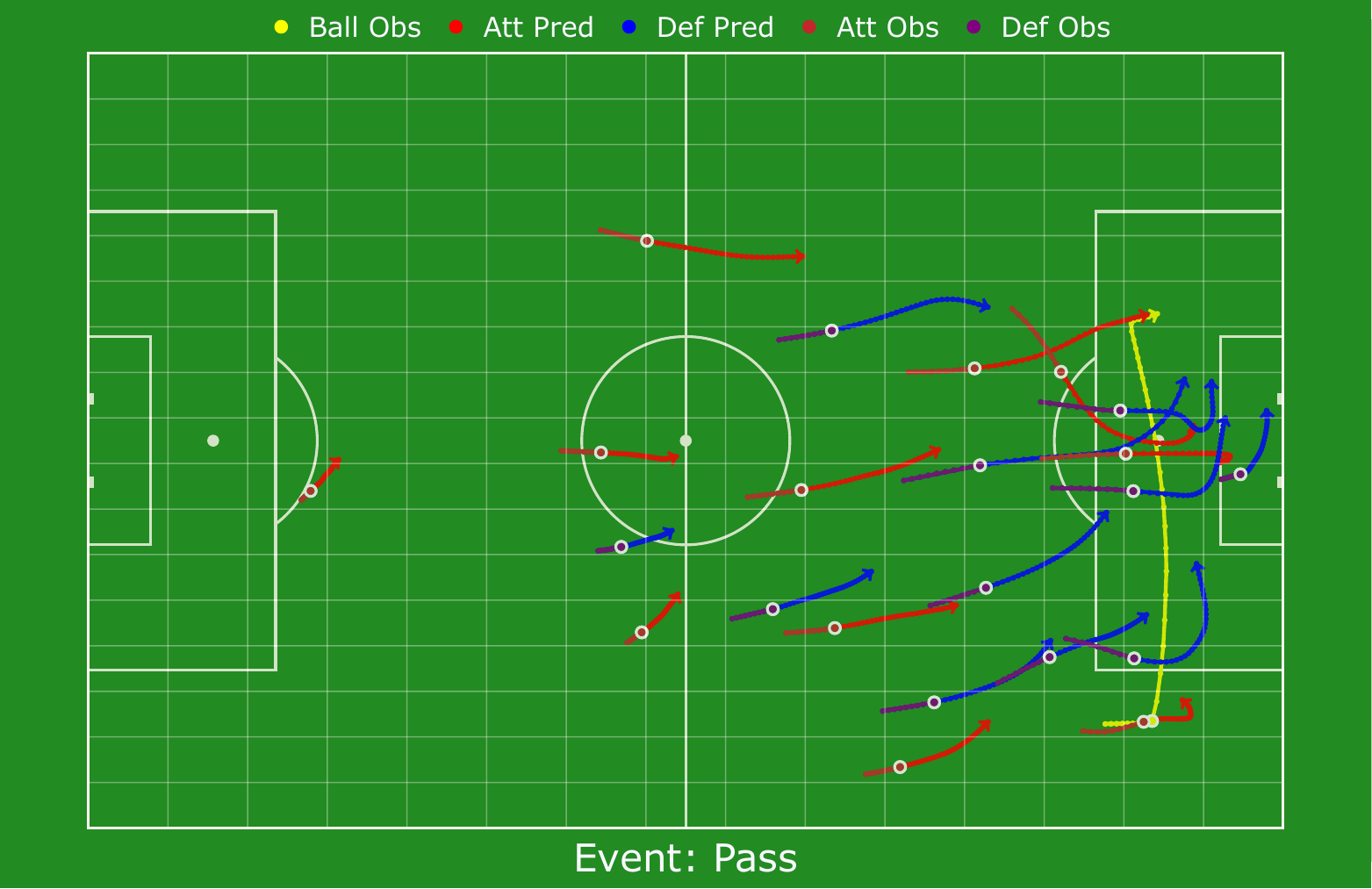}
  \end{minipage}\hfill
  \begin{minipage}[t]{0.24\textwidth}
    \centering
    \small \textbf{(b)} Generated\\
    \includegraphics[width=\linewidth]{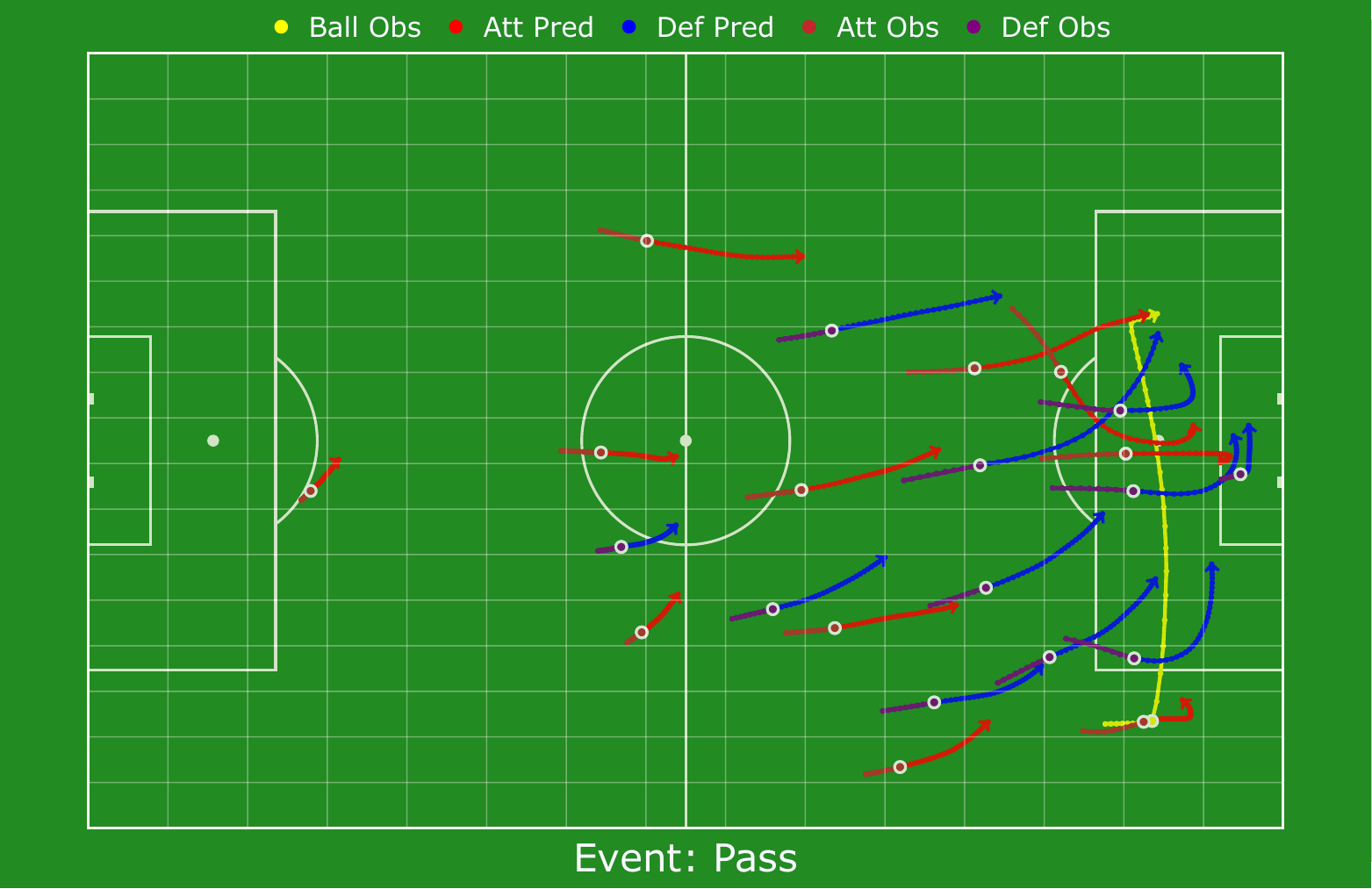}
  \end{minipage}

  \caption{\textbf{An illustrative example where the \textcolor{blue}{defending} team is guided.} Notably, in the generated tactic, the attacking ball carrier is more closely marked. As one expert remarked, “It is clearly better with defenders blocking the lonely attacker.”}
  \label{fig:vis-tactic-def}
\end{figure}

Figures~\ref{fig:vis-tactic-att} and~\ref{fig:vis-tactic-def} illustrate two examples of ground-truth and generated trajectories, where guidance is applied to the attacking team and the defending team, respectively. During the evaluations, experts were shown complete video clips of the trajectories. However, since videos cannot be included in the paper, we visualize the trajectory figure of each clip here.

\section{Implementation Details}\label{sec:imp-details}

\subsection{Sample Rule-based Guidance Functions}\label{sec:rule-example}
We provide some examples of rule-based guidance functions as follows.
\begin{itemize}
    \item \textbf{Applicable to both teams:}  
    \begin{itemize}
        \item \textbf{Pitch Control Value (PCV):} Quantifies the proportion of pitch space controlled by a team using a pitch control model with Gaussian smoothing to ensure differentiability.  

        \item \textbf{Ball Support:} Penalizes the top-$n$ players if their distance from the ball-carrier exceeds a support threshold.  

        \item \textbf{Spatial Spread / Compactness:} Encourages either dispersion (spread) or compactness by maximizing or minimizing the positional variance of the top-$n$ nearest players around the ball.  
    \end{itemize}  

    \item \textbf{Applicable to the attacking team:}  
    \begin{itemize}
        \item \textbf{Passing-Angle Spread:} Promotes wide angular separation of the top-$n$ nearest players around the ball by penalizing directional concentration.  

        \item \textbf{Zone-14 Presence:} Encourages occupation of Zone-14 by minimizing the per-frame distance of the closest player to the region, using a fully differentiable point-to-rectangle distance.  
    \end{itemize}  

    \item \textbf{Applicable to the defending team:}  
    \begin{itemize}
        \item \textbf{Deep Defending:} Encourages defenders to retreat and maintain positions deeper than the ball, prioritizing goal protection.

        \item \textbf{Pass-Lane Blocking:} Encourages defenders to occupy passing lanes between the ball and top-$n$ nearest opponents by penalizing the perpendicular distance of the closest defender to each lane.
    \end{itemize}  
\end{itemize}  

\subsection{Generating Guidance Functions via Prompting LLM}\label{sec:llm-func-detail}

We use the following prompts with GPT-5 to generate guidance functions.

\begin{promptbox}{Prompt for generating guidance functions}
\textbf{Background.} You are a helpful assistant. Your task is to propose practical guidance rules to steer a diffusion-based trajectory generator for tactical football trajectory generation. The diffusion model should guide team movements based on the given ball and team positions. You need to implement differentiable numerical guidance functions that can be directly applied in the diffusion process.
\vspace{0.1in}

\textbf{Function inputs.}
\vspace{-0.05in}
\begin{itemize}
  \item \texttt{ball\_pos}: tensor of shape \((\mathrm{batch}, \mathrm{length}, 1, 2)\) - ball positions over time.
  \item \texttt{team\_pos}: tensor of shape \((\mathrm{batch}, \mathrm{length}, 11, 2)\) - positions of 11 players (attacking or defending).
\end{itemize}

\textbf{Pitch information.}
\vspace{-0.05in}
\begin{itemize}
  \item The football pitch has dimensions of 105 x 68 meters.
  \item The bottom-left corner is (0, 0), and the top-right corner is (105, 68).
  \item The attacking direction is always from left to right.
\end{itemize}

\textbf{Examples.}

Example 1: \texttt{guide\_support} (encourage nearby players to support the ball carrier)
\begin{minted}{python}
import torch
import torch.nn.functional as F
def guide_support(ball_pos, team_pos):
    max_support_dist = 8.0
    top_k = 3

    # Distances from each player to the ball
    dist = (team_pos - ball_pos).norm(dim=-1)  # (B, H, N)

    # Top-k closest players (supporters)
    idx_closest = dist.topk(k=top_k, dim=-1, largest=False).indices
    mask = torch.zeros_like(dist, dtype=torch.float)
    mask.scatter_(-1, idx_closest, 1.0)  # one-hot mask

    # Penalty for being too far from the ball
    excess = F.relu(dist - max_support_dist)
    masked_penalty = excess * mask

    # Aggregate score (higher is better)
    k_frames = mask.sum(dim=(1, 2)) + 1e-6
    score = -masked_penalty.sum(dim=(1, 2)) / k_frames
    return score
\end{minted}

\vspace{0.05in}
Example 2: \texttt{guide\_compact} (keep team shape compact near the ball)
\begin{minted}{python}
import torch
def guide_compact(ball_pos, team_pos):
    top_k = 3
    N = team_pos.shape[2]
    k = min(max(top_k, 1), N)

    dists = (team_pos - ball_pos).norm(dim=-1)  # (B, H, N)
    sorted_idx = torch.argsort(dists, dim=-1)
    sel_idx = sorted_idx[..., :k]

    # Gather the k closest defenders
    sel = sel_idx.unsqueeze(-1).expand(-1, -1, -1, 2)
    close_pos = torch.gather(team_pos, 2, sel)  # (B, H, k, 2)

    # Compactness via positional variance
    var_x = close_pos[..., 0].var(dim=2, unbiased=False)  # (B, H)
    var_y = close_pos[..., 1].var(dim=2, unbiased=False)
    score = -(var_x + var_y).mean(dim=1)  # (B,)
    return score
\end{minted}

\vspace{0.05in}
Example 3: \texttt{guide\_zone14\_presence} (guide the nearest player to occupy Zone 14)
\begin{minted}{python}
import torch
import torch.nn.functional as F
def guide_zone14_presence(ball_pos, team_pos):
    x_min, x_max = 88.0, 100.0  # Zone 14 (x range)
    y_min, y_max = 22.0, 46.0   # Zone 14 (y range)

    x_pos, y_pos = team_pos[..., 0], team_pos[..., 1]  # (B, H, N)

    # L2 distance to the axis-aligned zone
    dx = F.relu(x_min - x_pos) + F.relu(x_pos - x_max)  # (B, H, N)
    dy = F.relu(y_min - y_pos) + F.relu(y_pos - y_max)
    dist = torch.sqrt(dx * dx + dy * dy + 1e-9)

    # Encourage at least one player close to the zone each frame
    frame_pen = dist.min(dim=-1).values  # (B, H)
    score = -frame_pen.mean(dim=1)       # (B,)
    return score
\end{minted}

\vspace{0.05in}
\textbf{Your task.} You are guiding the \emph{\{guided\_team\}}. Design a practical guidance function that achieves the following objectives:
\emph{\{your\_objective\}}. 
The function should accept \texttt{ball\_pos} and \texttt{team\_pos} and return a scalar score per batch (higher is better). Provide {Python code only}, with imports and inline comments, no extra text.

\vspace{0.05in}
\textbf{Function format.}
\begin{minted}{python}
def guidance_function(ball_pos, team_pos):  # do not rename the function
    # Your implementation here
    return score
\end{minted}
\end{promptbox}

By specifying \emph{\{guided\_team\}} and replacing \emph{\{your\_objective\}} with natural language descriptions of desired tactics, the LLM can automatically generate a corresponding function to guide the generation towards the specified objective. The functions generated in the main experiments in Section~\ref{sec:exp-llm} are provided below.

\begin{promptbox}{Prompt: Make the attacking team move forward more aggressively.}

\begin{minted}{python}
import torch
import torch.nn.functional as F
def guidance_function(ball_pos, team_pos):  # do not rename the function
    # team_pos: (B, H, N, 2) with x at index 0

    # Compute mean x-position of the attacking team per frame
    mean_team_x = team_pos[..., 0].mean(dim=-1)      # (B, H)

    # Compute frame-to-frame forward movement (delta x)
    delta_x = mean_team_x[:, 1:] - mean_team_x[:, :-1]  # (B, H-1)

    # Score is the average forward movement (higher means more aggressive)
    score = delta_x.mean(dim=1)  # (B,)

    return score
\end{minted}
\end{promptbox}

\begin{promptbox}{Prompt: Make the right bottom player drift into the corner to stretch the defense and open up more space.}

\begin{minted}{python}
import torch
import torch.nn.functional as F
def guidance_function(ball_pos, team_pos):  # do not rename the function
    """
    Encourage the right-bottom player to drift into the attacking corner (105,0)
    to stretch the defense.
    """
    # Extract x and y coordinates
    x = team_pos[..., 0]  # (B, H, N)
    y = team_pos[..., 1]  # (B, H, N)

    # Identify the right-bottom player by maximizing x - y
    scores = x - y                              # (B, H, N)
    idx = torch.argmax(scores, dim=-1)          # (B, H)

    # Gather the selected player's position
    idx_exp = idx.unsqueeze(-1).unsqueeze(-1).expand(-1, -1, 1, 2)  # (B, H, 1, 2)
    corner_player = torch.gather(team_pos, 2, idx_exp).squeeze(2)  # (B, H, 2)

    # Define the target corner position
    corner = team_pos.new_tensor([105.0, 0.0])   # (2,)

    # Compute distance to the corner
    diff = corner_player - corner               # (B, H, 2)
    dist = torch.sqrt((diff ** 2).sum(dim=-1) + 1e-6)  # (B, H)

    # Higher score for smaller distance (drifted into the corner)
    score = -dist.mean(dim=1)                   # (B,)
    return score
\end{minted}
\end{promptbox}

\subsection{Hyperparameters}\label{sec:hyperparameters}
The model configuration includes several critical hyperparameters. The batch size is 512, and the diffusion steps are 20. The learning rate is set to 3e-5, and the AdamW optimizer~\cite{loshchilov2017decoupled} is employed, with an EMA decay of 0.995. The model is configured with a horizon of 64. The chosen diffusion model is DDPM~\cite{ho2020denoising} with cosine Beta schedule, and the loss function used is of type L2. The model-related configurations are detailed in Table~\ref{tab:model-config-scale}. The full set of hyperparameters can be found in the supplementary code.

Regarding the number of diffusion steps $K$, we conduct a small ablation study by evaluating different values of $K$ while keeping all other parameters fixed. Table~\ref{tab:ablation-diffusion-step} reports the results. Overall, increasing $K$ from 10 to 20 leads to a clear improvement across all metrics, suggesting that a moderate number of diffusion steps helps the model better capture the underlying trajectory distribution. When $K$ increases further to 30, the performance remains comparable to $K=20$, indicating diminishing returns from additional diffusion steps. However, setting $K$ to a larger value such as 50 slightly degrades performance, possibly due to increased optimization difficulty during the diffusion process. Based on these observations, we adopt $K=20$ as the default configuration, which achieves the best overall balance between accuracy and efficiency.

\begin{table}[htbp]
\centering
\begin{minipage}{0.5\textwidth}
\centering
\caption{{Performance of different diffusion steps $K$.}}
\label{tab:ablation-diffusion-step}
\resizebox{1.0\textwidth}{!}{%
\begin{tabular}{@{}ccccccc@{}}
\toprule
\multirow{2}{*}{\bf Method} 
  & \multicolumn{3}{c}{\bf Marginal} 
  & \multicolumn{3}{c}{\bf Joint} \\
\cmidrule(lr){2-4} \cmidrule(lr){5-7}
  & ADE & FDE & MR (\%) & JADE & JFDE & JMR (\%) \\
\midrule
{$k=10$}  &{0.33} &{0.56} &{5.68} &{0.53} &{1.02} &{12.01} \\
{$k=20$}  &{0.29} &{0.52} &{4.73} &{0.45} &{0.92} &{10.66} \\
{$k=30$}  &{0.29} &{0.54} &{4.99} &{0.45} &{0.92} &{10.78} \\
{$k=50$}  &{0.31} &{0.54} &{5.09} &{0.47} &{0.98} &{11.34} \\
\bottomrule
\end{tabular}%
}
\end{minipage}
\vspace{-0.05in}
\end{table}

\subsection{Model Configurations in Scaling Experiments}\label{sec:scale-config-detail}

We report the model configurations used in the scaling experiments, including Small (S), Base (B), Large (L), XLarge (XL), and XXLarge (XXL), in Table~\ref{tab:model-config-scale}.

\begin{table}[htbp]
\centering
\caption{{TacticGen model configurations across different scales.}}\label{tab:model-config-scale}
\resizebox{0.5\textwidth}{!}{%
\begin{tabular}{@{}ccccc c@{}}
\toprule
\textbf{Model Size} & \textbf{Emb. Dim} & \textbf{Hid. Dim} & \textbf{Layers} & \textbf{Heads} & \textbf{Parameters (M)} \\
\midrule
S     & 32   & 192   & 2  & 4  & 1.74   \\
B     & 64   & 320   & 3  & 4  & 6.56   \\
L     & 128  & 512   & 6  & 8  & 30.83  \\
XL    & 192  & 768   & 12 & 12 & 132.87 \\
XXL   & 256  & 1024  & 16 & 16 & 311.50 \\
\bottomrule
\end{tabular}
}
\end{table}

\subsection{Case Study Details}\label{sec:case-study-detail}
We provide more details about the case study performed in this section.

\textbf{Expert Profiles.} The case study involved five domain experts with extensive experience in football analytics and practice:  
\begin{itemize}
\item \textbf{Three senior data scientists:} Experts in football tracking data, match video visualization and analysis, and tactical modeling. Each has over 10 years of experience collaborating with professional clubs and football analytics companies.  
\item \textbf{A former professional football player:} With more than 20 years of experience in professional football, followed by a career in player recruitment and team management.  
\item \textbf{A professor in sports analytics:} A faculty member at a leading university specializing in sports analysis, especially football studies, with decades of academic research and consulting experience in the field.  
\end{itemize}

Together, this group represents a balanced mix of quantitative analysts, practical practitioners, and academic researchers, ensuring that the evaluation captures multiple perspectives on both the realism and utility of generated trajectories.  

\textbf{Blinding Procedure.} To avoid potential bias, all case study evaluations were conducted under blinded conditions. Experts were not informed of the ground-truth labels, and the order of presentation for real and generated samples was randomized with a fixed random seed. This ensured that no participant could infer the authenticity of trajectories based on metadata or ordering.  

\textbf{Case Study on Realism.} In the first experiment, experts were asked to evaluate the \textit{realism} of trajectories. We randomly sampled 50 realistic trajectories and 50 generated by TacticGen, covering a diverse set of action types such as passes, corners, interceptions, take-ons, clearances, saves, and blocked passes.

Each video clip followed a standardized structure:  
\begin{itemize}
    \item \textbf{Event context (10 frames):} Real tracking data showing the ball and all players immediately before the on-ball event.  
    \item \textbf{Post-event segment ($N$ frames):} After the on-ball event, the video continued until the next action occurred, the tracking data ended, or a 64-frame limit was reached. The ball trajectory was always authentic; the player trajectories were either entirely real or entirely generated.
\end{itemize}

All clips were rendered at 5 frames per second from tracking data captured at 10 Hz, so that one second of video corresponded to 0.5 seconds of real play. Raters were asked to classify whether the post-event trajectories appeared \textit{real} or \textit{AI-generated}.

\textbf{Case Study on Utility.} In the second experiment, experts were asked to evaluate the \textit{utility} of generated trajectories. Here, 25 events were selected by an independent football expert specialized in data visualization (not involved in the rating process), who reviewed a pool of 50 randomly sampled match clips and identified scenarios of high tactical importance, such as those leading to scoring opportunities, critical defensive interventions, or decisive moments. Events covered a range of action types, including passes, ball touches, recoveries, clearances, attempts, etc. Importantly, the selector had no access to model outputs or evaluation tasks, ensuring unbiased sample selection.

Each pair contained both the ground-truth trajectories and the guided TacticGen-generated version, presented in random order. Experts were instructed to judge which of the two clips displayed superior tactical quality for the target team (attacking or defending).

\subsection{Computational Resources}
In this paper, we utilized a total of 8 NVIDIA A800 GPUs, each with 80 GB of memory. Training TacticGen on the full dataset takes roughly 80 hours under the base configuration on a single A800 GPU. Actual runtime may vary depending on model size and dataset scale.

\end{document}